\documentclass[12pt]{article}
\usepackage{graphicx} % Required for inserting images
\usepackage{subcaption}
\usepackage{amsmath,amssymb,xcolor,physics}
\usepackage{url}
\usepackage{ascmac}
\usepackage{caption}
\usepackage{comment}
\usepackage{eclbkbox}

\usepackage{pgfplots}
\pgfplotsset{compat=1.17}
\usetikzlibrary{matrix}
\usetikzlibrary{arrows.meta}
\usetikzlibrary{shapes.geometric,shapes.misc}
\pgfplotsset{
every axis x label/.style={at={(axis description cs:1,0)}, anchor=west, font = \Large},
every axis y label/.style={at={(axis description cs:0,1)}, anchor=south, font = \Large},
axis line style={-{Stealth[length=3mm,width=2mm]}},
}

\allowdisplaybreaks[4]

\usepackage[hidelinks]{hyperref}

\begin{document}

\begin{center}
{\large Deep Learning for Physicists: From Linear Regression to Reasoning Models}
\end{center}
\vspace{0.1cm}
\vspace{0.1cm}
\begin{center}

Masanori Hanada

\end{center}
\vspace{0.3cm}

\begin{center}
School of Mathematical Sciences, Queen Mary University of London\\
Mile End Road, London, E1 4NS, United Kingdom
\end{center}

\vspace{0.5cm}

\begin{center}
  {\bf Abstract}
\end{center}
These lecture notes are based on an introductory course on neural networks and deep learning that the author taught in the School of Mathematical Sciences at Queen Mary University of London in 2023/24 and 2024/25. The author, however, is a physicist, and it shows: wherever possible, ideas are explained with physical intuition rather than heavy formalism. Readers with a background in physics or mathematics who want a self-contained, introductory route into modern machine learning may therefore find these notes particularly friendly.

 The notes start from maximum likelihood, least squares, linear regression and (stochastic) gradient descent. They then cover neural networks, backpropagation, the vanishing gradient problem, batch and layer normalization, overfitting and regularization, adversarial examples, and the kernel method. The second half turns to natural language processing: word embeddings, recurrent networks and LSTMs, the attention mechanism and Transformers, and large language models, including scaling laws, the history of GPT, instruction tuning, reinforcement learning from human feedback, hallucination, alignment faking, chain-of-thought reasoning and test-time scaling. The emphasis on large language models reflects requests from the students.

The notes are aimed at third- and fourth-year students in the UK system. Because the course was also taken by students from other departments, very little is assumed: even basic material such as matrix multiplication is explained from scratch, so readers familiar with it can safely skip such parts. No originality is claimed; the aim is a coherent, intuitive path through the material. Developments after spring 2025 are not covered, but the basic ideas emphasized here are unlikely to go out of date soon.
 
 \newpage

\noindent
\textbf{These are lecture notes, not a research paper: they contain no new results, and I do not claim originality.}\\

\noindent
These notes are based on material I gathered from many sources: preprints on arXiv, textbooks in English and Japanese, seminars, and lectures at other universities. I learned a great deal from the recorded lectures of Stanford's Computer Science department, especially CS229, CS230 and CS224N. I watched these lectures, took my own notes, and then rewrote (or at least tried to rewrite) the material in my own words. Standard derivations and equations naturally coincide with those found in many other places. Figures adapted from published papers are credited to the original references.\\

\noindent
If anything here is original, it is only that I tried hard to understand the material and to explain it in plain language, or at least in a way that is easy for me (a physicist) to grasp. 
I used AI tools to polish the English, to redraw the figures, and to check the text for errors and unclear explanations. Some explanations, including a few full paragraphs, were added or expanded following their suggestions; I checked all of them and take full responsibility for the content.\\

\noindent
The reference list is far from complete. Many people have contributed to machine learning, neural networks and deep learning, and numerous important ideas have been reinvented more than once.\\

\noindent
Many important topics are not covered. In particular, convolutional neural networks, optimizers beyond stochastic gradient descent (such as Adam), the initialization of the weights, and generative models for images are not discussed.

\newpage
\tableofcontents
%%%%%%%%%%%%%%%%%%
%%%%%%%%%%%%%%%%%%
\part{Foundations of deep learning}
%%%%%%%%%%%%%%%%%%
%%%%%%%%%%%%%%%%%%
\section{Preliminaries}
%%%%%%%%%%%
\subsection{The task}
%%%%%%%%%%%
Deep learning is a specific kind of machine learning. In machine learning, we start with a dataset. Any sort of dataset is fine: animal images, YouTube videos, text scanned from books or Wikipedia, outcomes of scientific experiments, etc. The basic task is to find some patterns in the dataset. We want to do it without giving too many explicit instructions in a computer program. In other words, we want computers to learn by themselves as much as possible. Once the patterns are found, many applications are possible.

The quality of data is important, and the meaning of `good quality' or `bad quality' depends on the actual task. If the dataset consists of images of cats or dogs and the task is to classify cat images and dog images, resolution must be sufficiently high so that at least we human beings can distinguish them, but if the resolution is too high then it is a waste of computational resources.\footnote{Note that, when a high-resolution picture is given, we can always reduce the resolution.} The sweet spot would depend on whether you only have to classify zoomed-in images or more complicated ones like a puppy walking at Piccadilly Circus surrounded by hundreds of people. Furthermore, a good balance of dog and cat images is required.\footnote{For speech recognition, it is desirable to have all kinds of accents from all over the world. In the past, I noticed that automatic subtitle generation did not work well for my Japanese accent although it worked well for my colleague's Indian accent. It would mean either my accent was too bad or the dataset was biased.}

\begin{figure}[htbp]
\centering
\begin{tikzpicture}
\begin{axis}[
width=0.4\textwidth, height=0.3\textwidth,
xmin=0, xmax=7.6, ymin=0, ymax=6,
axis lines=left, xlabel={$x$}, ylabel={$y$},
xtick=\empty, ytick=\empty,
]
\addplot[only marks, mark=*, mark size=1.8pt] coordinates {
(0.5,1.3) (1.2,1.4) (2.0,2.3) (2.7,2.1) (3.5,3.1)
(4.1,2.8) (4.9,3.9) (5.6,3.6) (6.4,4.6) (7.1,4.4)};
\addplot[blue, thick, domain=0:7.6] {0.513*x+0.999};
\end{axis}
\end{tikzpicture}
\hspace{1em}
\begin{tikzpicture}
\begin{axis}[
width=0.4\textwidth, height=0.3\textwidth,
xmin=0, xmax=7.2, ymin=0, ymax=6.5,
axis lines=left, xlabel={$x$}, ylabel={$y$},
xtick=\empty, ytick=\empty,
]
\addplot[only marks, mark=*, mark size=1.8pt] coordinates {
(0.3,0.98) (0.9,2.95) (1.5,3.59) (2.1,2.99) (2.7,2.90) (3.3,2.41)
(3.9,2.46) (4.5,1.84) (5.1,2.13) (5.7,1.30) (6.3,3.74) (6.9,4.62)};
\addplot[red, thick, domain=0:7.2, samples=100]
{0.1242*x^3-1.2821*x^2+3.5907*x+0.3007};
\end{axis}
\end{tikzpicture}
\caption{Fitting data points by a straight line (left) and by a cubic function (right).}
\label{fig:fit_function}
\end{figure}
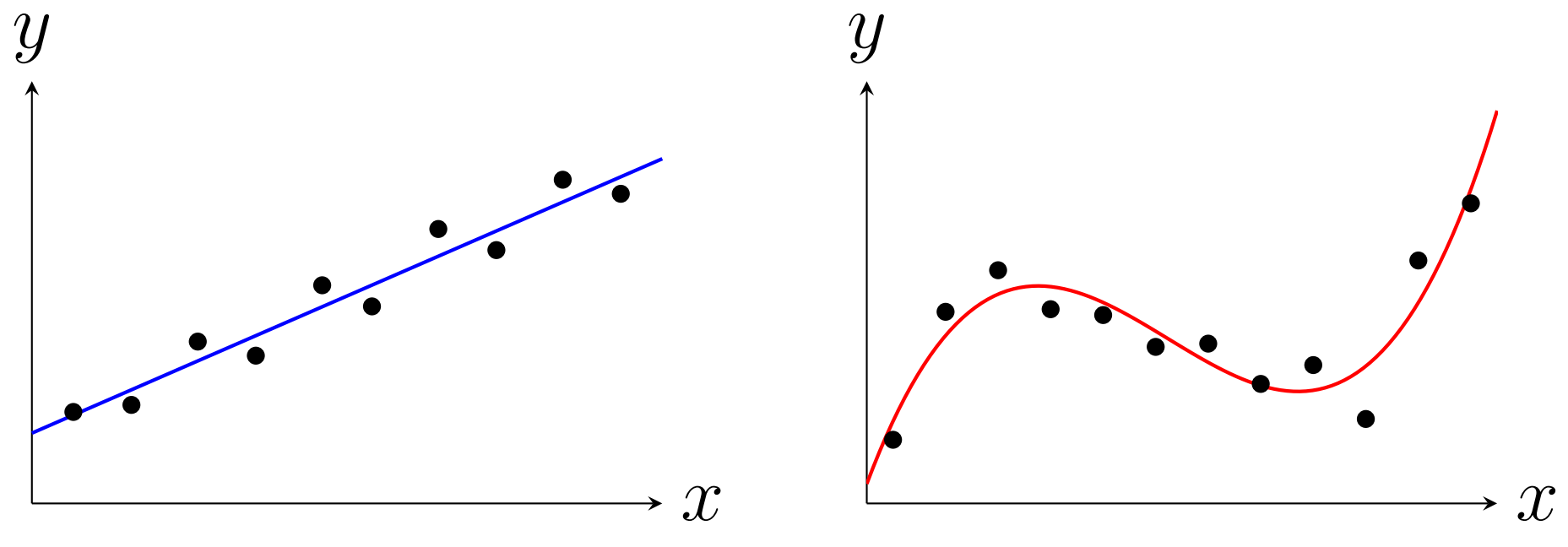

Basically, what people do is to build a map from the dataset to something. In the case of cat/dog image classification, we can convert images to numbers. For a $100 \times 100$-pixel color picture, we need three numbers (red, green, and blue) for each pixel, in total 30,000 numbers. We need to build a map from these 30,000 numbers describing the dataset to `something' that represents cat and dog. For a simple classification task, we can assign 0 to dog pictures and 1 to cat pictures,\footnote{It is more complicated if we allow the possibility of having both dog and cat in the same picture.} respectively, and hence our task is to find a good map from 30,000 numbers to $\{0,1\}$. In the case of speech recognition, the input data is audio data which consists of binary numbers. The output is a sequence of words. We can label all words in the dictionary by natural numbers $1,2,3,4, \cdots$, and then, the output is a sequence of natural numbers. Therefore, our task is to find a good map from binary numbers to sequences of natural numbers. We assume a certain ansatz (architecture) with some parameters $\vec{\theta}$ and determine `good' ansatz and parameters that fit the data well. Conceptually, there is no difference from the simple fits depicted in Fig.~\ref{fig:fit_function}.

Suppose images are labeled by 0 (dog) or 1 (cat). (In other words, we assume that the answer is given by someone.) To define the meaning of `good' or `bad' ansatz/parameters, we split the cat/dog dataset into a training dataset and a test dataset. The training dataset is used to determine the optimal parameters. Specifically, some cost function (equivalently, loss function) is defined and the parameters are tuned to minimize the cost function. The performance of the network is measured using the test dataset. It is important to separate the training set and the test set because we want to use the trained architecture for new data. If we use all data for training, the architecture could simply memorize all data points so that a perfect answer is given for the old data, but it does not necessarily guarantee that it works for new data.\footnote{We can try different architectures and choose the one that works best. A common mistake is to make this choice by looking at the performance on the test dataset. If we did this, our choice would be fitted to the test dataset, and the test dataset would no longer tell us how well the chosen architecture works for new data. For this purpose, a third set of data, called the validation dataset, is used; see Sec.~\ref{sec:overfit}.}

At the conceptual level, our to-do list is very simple! There are only five steps:

\begin{itemize}
  \item Decide the task. What is the input $x$ (images, etc.), what is the output $y$ (dog or cat, etc.), and what are the fit parameters $\vec{\theta}$?
  \item Get or create a dataset. This could be the hardest part.
  \item Training, i.e., determine the optimal parameters $\vec{\theta}$. What is the cost function, and how do you minimize it?
  \item Test it and make sure it works well.
  \item Use it.
\end{itemize}

Practically, each step is not easy and we must work hard. But it is all part of the fun.
%%%%%%%%%%%%%%%%%%%%%%%%%%%%%%%%%%%%%%%%%%%
\subsection{Maximum likelihood method}\label{sec:MLE}
%%%%%%%%%%%%%%%%%%%%%%%%%%%%%%%%%%%%%%%%%%%
The maximum likelihood method is an effective approach for selecting optimal fitting parameters.\footnote{This section is largely based on my book MCMC from Scratch: A Practical Introduction to Markov Chain Monte Carlo with So Matsuura~\cite{Hanada2022}.}$^,$\footnote{The maximum likelihood method was popularized by Ronald Fisher, though its history dates back to Gauss and Laplace.
}  If you don’t fully grasp this method, you won’t be able to understand machine learning. I recommend you carefully study the following, then close the notes and attempt to reproduce everything on your own. Repeat this process until it becomes second nature.

Let us throw a coin and count the number of heads and tails. In the ideal world, the most natural assumption is that the head and tail appear with the same probability every time the coin is tossed. Suppose, however, that the head appeared too many times, say 17 times out of 20 trials. There can be $2^{20}=1{,}048{,}576$ outcomes when the coin is tossed 20 times, among which only $\binom{20}{3}=1140$ outcomes are head 17 -- tail 3. Assuming a `fair' coin that returns head and tail with the same probability, there is only about a $0.1 \%$ chance of such an outcome. Maybe we should consider the possibility that somebody manipulated the coin. But we must be careful when we criticize someone; we need to have a quantitative argument.

So, we want to estimate the probability of heads $p$ at each toss from this outcome. Intuitively, $p=\frac{17}{20}$ sounds likely, but can we justify our intuition by defining the ``likelihood" quantitatively?

If the probability of a head is $p$ at each coin toss, the probability that the head appears 17 times out of 20 tosses is

\begin{align}
1140 p^{17}(1-p)^{3}\, .
\end{align}

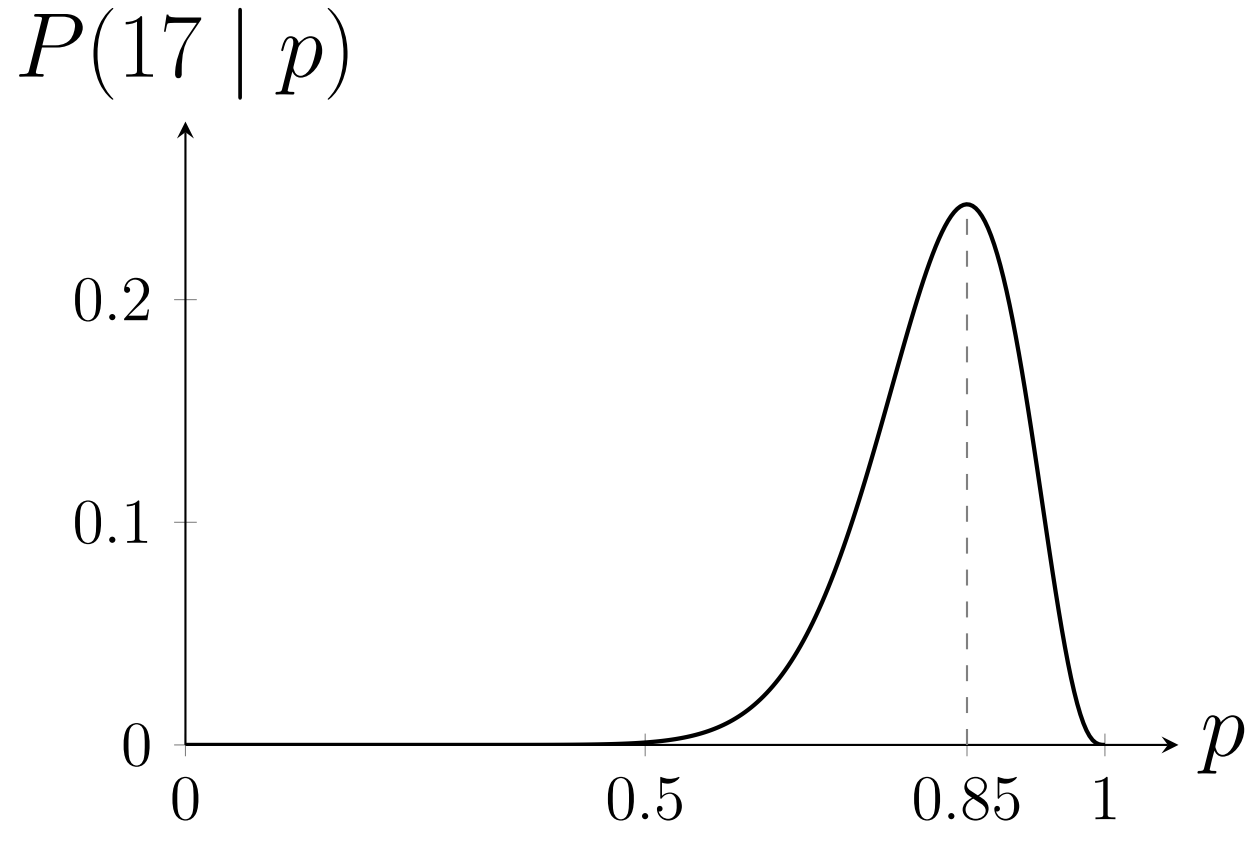
\begin{figure}[htbp]
\centering
\begin{tikzpicture}
\begin{axis}[
width=0.5\textwidth, height=0.35\textwidth,
domain=0:1, samples=300,
xmin=0, xmax=1.08, ymin=0, ymax=0.28,
axis lines=left, xlabel={$p$}, ylabel={$P(17\mid p)$},
xtick={0,0.5,0.85,1}, ytick={0,0.1,0.2},
yticklabel style={/pgf/number format/fixed},
scaled y ticks=false,
]
\addplot[black, thick] {1140*x^17*(1-x)^3};
\draw[dashed, gray] (axis cs:0.85,0) -- (axis cs:0.85,0.2428);
\end{axis}
\end{tikzpicture}
\caption{$P(17\mid p)=1140\, p^{17}(1-p)^{3}$, which is the probability of 17 heads and 3 tails when the probability of a head at each toss is $p$. The maximum is at $p=0.85$.}
\label{fig:coin_toss}
\end{figure}

We plot this function in Fig.~\ref{fig:coin_toss}. This function takes its maximum $(\simeq 0.2428)$ at $p=\frac{17}{20}$. Hence it seems to be reasonable, at least to some extent, to conclude that $p=\frac{17}{20}=0.85$. On the other hand, if $p=\frac{1}{2}$, such an event (head 17 -- tail 3) can happen only with probability 0.00109, which is not very likely. But we cannot immediately conclude $p=0.85$, because $1140 p^{17}(1-p)^{3}$ is large in a finite range, say between $p=0.7$ and $p=0.95$, and hence the values of $p$ around that range are not too unlikely.

Therefore, we adopt the value $1140 p^{17}(1-p)^{3}$, which is ``the probability that the observed outcome (head 17 -- tail 3) can happen when the probability of a head at each toss is $p$," as a quantitative index characterizing the likelihood that the probability of a head is actually $p$. We call this quantity the likelihood. Note that there is no way to `prove' that this quantity is the best. We use it because many people think it is natural. You can use something different if you like, but then you have to convince other people that what you use is legitimate.

More generally, the probability that the head appears $k$ times out of $n$ tosses is

\begin{align}
P(k \mid p)=\binom{n}{k} \cdot p^{k}(1-p)^{n-k}=\frac{n!}{k!(n-k)!} \cdot p^{k}(1-p)^{n-k}\, . 
\end{align}
To emphasize the assumption (``the probability of a head is $p$"), we used the notation $P(k \mid p)$. This quantity $P(k \mid p)$ is the likelihood, given the observed outcome that the head appeared $k$ times out of $n$ trials. In the maximum likelihood method, we infer that the value of $p$ that maximizes the likelihood is the most natural and hence the right value. Again, this is merely our choice and not a mathematical theorem.

In this way, the likelihood is defined as the probability that a certain event happens, when such an event is observed. The difference from the usual probability is that the quantity characterizing the event (in this case, $p$) is regarded as the variable. ``Probability" and ``likelihood" are the same function, but the interpretations are different:

\begin{itemize}
  \item If the probability of a head $p$ is given, then $P(k \mid p)$ is the ``probability" that the head appears $k$ times.
  \item If the outcome (the head appeared $k$ times) is given and we try to estimate the value of $p$ from the given outcome, $P(k \mid p)$ is the ``likelihood" that the value is actually $p$.
\end{itemize}
If you find it confusing, it would be better to use different notation, say $P(k \mid p)$ for the probability and $L(p \mid k)$ for the likelihood.\footnote{$P(p \mid k)$ and $P(k \mid p)$ are different conditional probabilities.
} In these lecture notes, we use only $P(k \mid p)$.

The same idea can be applied to other probability distributions. Consider the Gaussian distribution
\begin{align}
\rho(x \mid \mu, \sigma)=\frac{e^{-\frac{1}{2} \frac{(x-\mu)^{2}}{\sigma^{2}}}}{\sqrt{2 \pi \sigma^{2}}}\, . 
\end{align}
If we generate $n$ random numbers following this distribution, the probability that $x^{(1)}$, $x^{(2)}$, $\cdots$, $x^{(n)}$ are obtained is

\begin{align}
P\left(x^{(1)}, \cdots, x^{(n)} \mid \mu, \sigma\right) & =\rho\left(x^{(1)} \mid \mu, \sigma\right) \times \rho\left(x^{(2)} \mid \mu, \sigma\right) \times \cdots \times \rho\left(x^{(n)} \mid \mu, \sigma\right) \nonumber\\
& =\prod_{i=1}^{n} \rho\left(x^{(i)} \mid \mu, \sigma\right)\, .  
\end{align}

Suppose you have a black-box routine that creates Gaussian random numbers, but you do not know the values of $\mu$ and $\sigma$. You used this routine $n$ times and obtained $x^{(1)}, \cdots, x^{(n)}$. From such an observed outcome, you want to make a good guess at the values of $\mu$ and $\sigma$. Then, by interpreting $P\left(x^{(1)}, \cdots, x^{(n)} \mid \mu, \sigma\right)$ as a function of $\mu$ and $\sigma$, we can regard $(\mu, \sigma)$ that makes $P\left(x^{(1)}, \cdots, x^{(n)} \mid \mu, \sigma\right)$ larger as more likely. Therefore, $P\left(x^{(1)}, \cdots, x^{(n)} \mid \mu, \sigma\right)$ can be interpreted as the likelihood as well.

In traditional statistics, it is assumed that parameters such as $p$ for the coin toss and $\mu, \sigma$ for the Gaussian distribution are (even if you don't know the actual values, in principle) already uniquely determined. The values of those parameters are determined by using, e.g., the maximum likelihood method. In the Bayesian approach, on the other hand, the parameters themselves are treated as random variables: starting from a prior distribution, we use the observed data to obtain a posterior distribution, which tells us how strongly we should believe each value of the parameters. We will not use the full Bayesian approach in these notes. The only exception is Sec.~\ref{sec:overfit}, where a prior distribution leads to regularization; even there, we only look for the maximum of the posterior distribution (maximum a posteriori, or MAP, estimation), which is close in spirit to the maximum likelihood method.

So far, we have assumed that the functional forms of the likelihood function $P(k \mid p)$ or $P\left(x^{(1)}, \cdots, x^{(n)} \mid \mu, \sigma\right)$ are known. In actual applications, even if data like $k$ or $\left\{x^{(i)}\right\}$ are given, usually we do not know the form of the likelihood functions, and we have to find reasonable likelihood functions. 
%%%%%%%%%%%%%%%%%%%%%%%%%%%%%%%%%
\subsubsection{Least squares method}
%%%%%%%%%%%%%%%%%%%%%%%%%%%%%%%%%
The least squares method~\cite{Legendre1805,Gauss1809} is the simplest version of the maximum likelihood method.

Imagine a theory parametrized by $\vec{x}$. In physics problems, temperature $T$, pressure $p$, magnetic field $h$, etc., which you specify when you do experiments, correspond to $\vec{x}$. Suppose that a physical quantity (say the energy $E$) is measured at different points in the parameter space, $\vec{x}_{1}, \cdots, \vec{x}_{K}$, and the results $y_{1}, \cdots, y_{K}$ are obtained. We fit the data by using a function $y=f\left(\vec{x} ; \theta_{1}, \theta_{2}, \cdots\right)$. You can imagine something like $E=\theta_{1} T^{\theta_{2}}+\theta_{3} p^{\theta_4} h^{\theta_5}$. (This specific ansatz is nonsense, though.) The fitting parameters are $\theta_{1}$, $\theta_{2}$, $\cdots$. The least squares method gives us a guideline for the fit. In the least squares method, $\theta_{1}$, $\theta_{2}$, $\cdots$ are chosen such that the cost function (or loss function) defined by
\begin{align}
J(\vec{\theta}) 
\equiv 
\frac{1}{2} \sum_{k=1}^{K}\left(y_{k}-f\left(\vec{x}_{k} ; \vec{\theta}\right)\right)^{2}
\label{eq:least_square_J}
\end{align}
is minimized. The errors in the experiments often follow the Gaussian distribution. So, let us assume it here. It is natural to identify the width of the Gaussian distribution at each point $\vec{x}_k$ with the error bar associated with the measured value $y_k$, which we denote by $\sigma_k$. We can estimate them by using, e.g., the jackknife method. Then, when the fitting parameters $\theta_1,\theta_2,\cdots$ are fixed, the outcome would be $y_1,\cdots,y_K$ with the following probability:
\begin{align}
&
 P(y_1,\cdots,y_K|\vec{\theta})=
 \prod_{k=1}^K
 \rho_{\sigma_k,\vec{x}_k}(y_k|\vec{\theta})\, , \\
 &
 \rho_{\sigma_k,\vec{x}_k}(y_k|\vec{\theta})
 \equiv
 \frac{1}{\sqrt{2\pi}\sigma_k}
 \exp\left(
 -\frac{(y_k-f(\vec{x}_k;\vec{\theta}))^2}{2\sigma_k^2}
 \right)\, . 
\end{align}
 We interpret this $P$ as the likelihood function. In the maximum likelihood method, $\theta_1$, $\theta_2$, $\cdots$ are chosen such that the likelihood is maximized, or equivalently, such that
\begin{align}
J(\vec{\theta}) 
\equiv 
\frac{1}{2} \sum_{k=1}^{K}
\frac{\left(y_{k}-f\left(\vec{x}_{k} ; \vec{\theta}\right)\right)^2}{\sigma_k^2}
\label{eq:least_square_J_2}
\end{align}
is minimized.
If we do not know the variance $\sigma_k^2$, let us just assume $\sigma_1=\cdots=\sigma_K=\sigma$.  
Then, regardless of the value of $\sigma$, the minimization of \eqref{eq:least_square_J_2} reduces to the minimization of \eqref{eq:least_square_J}. In this sense, the least squares method is a version of the maximum likelihood method. The minimization of \eqref{eq:least_square_J_2} is called the weighted least squares method.

%%%%%%%%%%%%%%%%%%%%%%%%%
\subsection{Gradient descent}
%%%%%%%%%%%%%%%%%%%%%%%%%

Suppose that there are two parameters $\theta_1$ and $\theta_2$ which correspond to the $x$ and $y$ coordinates, and the cost function $J(\theta_1,\theta_2)$ is the gravitational potential. (We assume that the earth is flat.) Then, the minimization of the cost function is the search for the lowest point.
Suppose it is very dark and we cannot see anything, and for some reason, we have a ball.
Then, we can find the lowest point (more precisely, locally the lowest point) by putting a ball on the ground with zero velocity, letting it roll a little bit, stopping it, letting it roll a little bit, stopping it, ... (repeat). The force acting on the ball is $\vec{F}=-\frac{\partial J}{\partial\vec{\theta}}$. Therefore, the ball rolls in this direction. We let the ball move a little bit, so, roughly speaking, we let the ball move by $\Delta\vec{\theta}=-\epsilon\frac{\partial J}{\partial\vec{\theta}}$ (equivalently, $\Delta\theta_i=-\epsilon\frac{\partial J}{\partial\theta_i}$). The parameter $\epsilon$ is called the learning rate.
The same holds for a generic number of `coordinates' $\theta_1,\cdots,\theta_n$ and `potential' $J(\vec{\theta})$. To see
that the cost function does decrease regardless of the details, we only have to use the Taylor expansion:
\begin{align}
J(\vec{\theta}+\Delta\vec{\theta})
&=
J(\vec{\theta})
+
\Delta\vec{\theta}\cdot\frac{\partial J}{\partial\vec{\theta}}
+
O((\Delta\vec{\theta})^2)
\nonumber\\
&=
J(\vec{\theta})
-
\epsilon\left(\frac{\partial J}{\partial\vec{\theta}}\right)^2
+
O(\epsilon^2)\, . 
\end{align}
Therefore, if $\epsilon$ is sufficiently small and $O(\epsilon^2)$-terms are negligible, the cost function decreases. The gradient descent method~\cite{Cauchy1847} uses this property.
\begin{itembox}[c]{Gradient descent method}
\begin{itemize}
\item
Change $\theta_i$ a little bit as $\theta_i\to\theta'_i=\theta_i-\epsilon\cdot\frac{\partial J}{\partial\theta_i}$.
\item
Repeat it until $\frac{\partial J}{\partial\theta_i}$ becomes almost zero for all $i$.
\end{itemize}
\end{itembox}
\\

Note that we can find only a \textit{local minimum} by gradient descent. It may or may not be the global minimum.

%%%%%%%%%%%%%%%%%%%%%%%%%
\subsection{Linear regression}
%%%%%%%%%%%%%%%%%%%%%%%%%
Let us consider a simple fit ansatz, 
\begin{align}
y
=
f(\vec{x};\vec{\theta})
=
\theta_0
+
\sum_{i=1}^d\theta_ix_i
=
\sum_{i=0}^d\theta_ix_i
=
\vec{\theta}\cdot\vec{x}\, . 
\end{align}
Here, we took $x_0=1$. 
This ansatz is linear in the parameters $\theta_i$. It is not necessarily linear in $x$, because we can take $x_1=x$, $x_2=x^2$ and $f(x;\vec{\theta})=\theta_0+\theta_1x+\theta_2x^2$, for example. We can use a more complicated ansatz, such as $f(x,x';\vec{\theta})=\theta_0+\theta_1\tan(x+x')+\theta_2\cosh(\sin(x/x'))$. 
In this case, $x_1=\tan(x+x')$ and $x_2=\cosh(\sin(x/x'))$. 

This linear ansatz can be very powerful. For example, we can truncate a Taylor series at a sufficiently high order so that the convergence is good in the range of $x$ we are interested in. 

To find the optimal $\vec{\theta}$, let us minimize the square loss function, 
\begin{align}
J(\vec{\theta})
=
\frac{1}{2K}\sum_{k=1}^K
\left(
y^{(k)}
-
f(\vec{x}^{(k)};\vec{\theta})
\right)^2
=
\frac{1}{2K}\sum_{k=1}^K
\left(
y^{(k)}
-
\vec{\theta}\cdot\vec{x}^{(k)}
\right)^2\, . 
\end{align}
Then, by using $A_{ij}\equiv\overline{x_ix_j}\equiv\frac{1}{K}\sum_kx_i^{(k)}x_j^{(k)}$, $B_{i}\equiv\overline{x_iy}\equiv\frac{1}{K}\sum_kx_i^{(k)}y^{(k)}$, and 
$C\equiv\overline{y^2}\equiv\frac{1}{K}\sum_k(y^{(k)})^2$, the cost function can be written as 
\begin{align}
J(\vec{\theta})
=
\frac{1}{2}\vec{\theta}\cdot A\vec{\theta}
-
\vec{B}\cdot\vec{\theta}
+
\frac{C}{2}\, . 
\end{align}
If $A$ is positive definite, we can rewrite it as 
\begin{align}
J(\vec{\theta})
=
\frac{1}{2}\left(\vec{\theta}-A^{-1}\vec{B}\right)
\cdot A
\cdot
\left(\vec{\theta}-A^{-1}\vec{B}\right)
-
\frac{1}{2}\vec{B}\cdot A^{-1}\vec{B}
+
\frac{C}{2}\, . 
\end{align}
Hence, there is a unique minimum at $\vec{\theta}=A^{-1}\vec{B}$. There are no other local minima. 

%%%%%%%%%%%%%%%%%%%%%%%%%
\subsection{Stochastic gradient descent and mini-batch gradient descent}\label{sec:SGD}
%%%%%%%%%%%%%%%%%%%%%%%%%
In the original `vanilla' gradient descent (GD), all samples are used to define the cost function. 
To update the parameters following $\theta_i\to\theta_i-\epsilon\frac{\partial J}{\partial\theta_i}$, 
\begin{align}
\frac{\partial J(\vec{\theta})}{\partial\theta_i}
=
-\frac{1}{K}
\sum_{k:\ {\rm all\ samples}}
\left(
y^{(k)}
-
f(\vec{x}^{(k)};\vec{\theta})
\right)
\cdot
\frac{\partial f(\vec{x}^{(k)};\vec{\theta})}{\partial\theta_i}\, , 
\end{align}
is used. 
In the stochastic gradient descent method (SGD), the update process is changed to $\theta_i\to\theta_i-\epsilon\frac{\partial J_\Lambda}{\partial\theta_i}$, where
\begin{align}
\frac{\partial J_\Lambda(\vec{\theta})}{\partial\theta_i}
=
-\frac{1}{K_{\rm batch}}
\sum_{\ell\in\Lambda}
\left(
y^{(\ell)}
-
f(\vec{x}^{(\ell)};\vec{\theta})
\right)
\cdot
\frac{\partial f(\vec{x}^{(\ell)};\vec{\theta})}{\partial\theta_i}\, . 
\end{align}
$\Lambda$ is a randomly chosen mini-batch, which should be refreshed at every step.

When people say they use mini-batch gradient descent, usually it means that training is done over multiple epochs. In each epoch, all $K$ samples are split into $N_{\rm batch}$ batches of size $K_{\rm batch}$ ($K=N_{\rm batch}K_{\rm batch}$), and each batch is used once and only once.

Usually, `batch gradient descent' means the vanilla gradient descent explained above. This is because `the batch' (which is not a \textit{mini}-batch) could mean the entire dataset.

\begin{figure}[htbp]
\centering
\begin{tikzpicture}
\begin{axis}[
width=0.75\textwidth, height=0.45\textwidth,
domain=-2.8:2.8, samples=200,
xmin=-2.8, xmax=2.8, ymin=-2.5, ymax=4,
axis lines=left,
xlabel={$\theta$}, ylabel={$J$},
xtick=\empty, ytick=\empty,
legend style={at={(0.97,0.97)}, anchor=north east, draw=none, font=\small},
]
\addplot[black, very thick] {0.1*x^4-0.6*x^2-0.3*x};
\addlegendentry{$J$ (all samples)}
\addplot[blue, thick, dashed] {0.1*x^4-0.6*x^2-0.3*x+0.5*(1-tanh((x+1)/0.5))-0.25+0.1*x};
\addlegendentry{$J_{\Lambda_1}$ (mini-batch 1)}
\addplot[red, thick, dotted] {0.1*x^4-0.6*x^2-0.3*x+0.2*sin(deg(1.3*x+2.0))+0.35};
\addlegendentry{$J_{\Lambda_2}$ (mini-batch 2)}
% a ball trapped in the local minimum of J
\node[circle, fill=gray, inner sep=2.5pt, anchor=south] at (axis cs:-1.59,-0.40) {};
\end{axis}
\end{tikzpicture}
\caption{Schematic picture of the cost functions for all samples (solid) and for two different mini-batches (dashed and dotted). The local minimum of $J$ where the ball is trapped does not exist in $J_{\Lambda_1}$, so the parameters can escape from it when the mini-batch is changed. The location of the global minimum depends only weakly on the choice of mini-batch.}
\label{fig:minibatch_minima}
\end{figure}
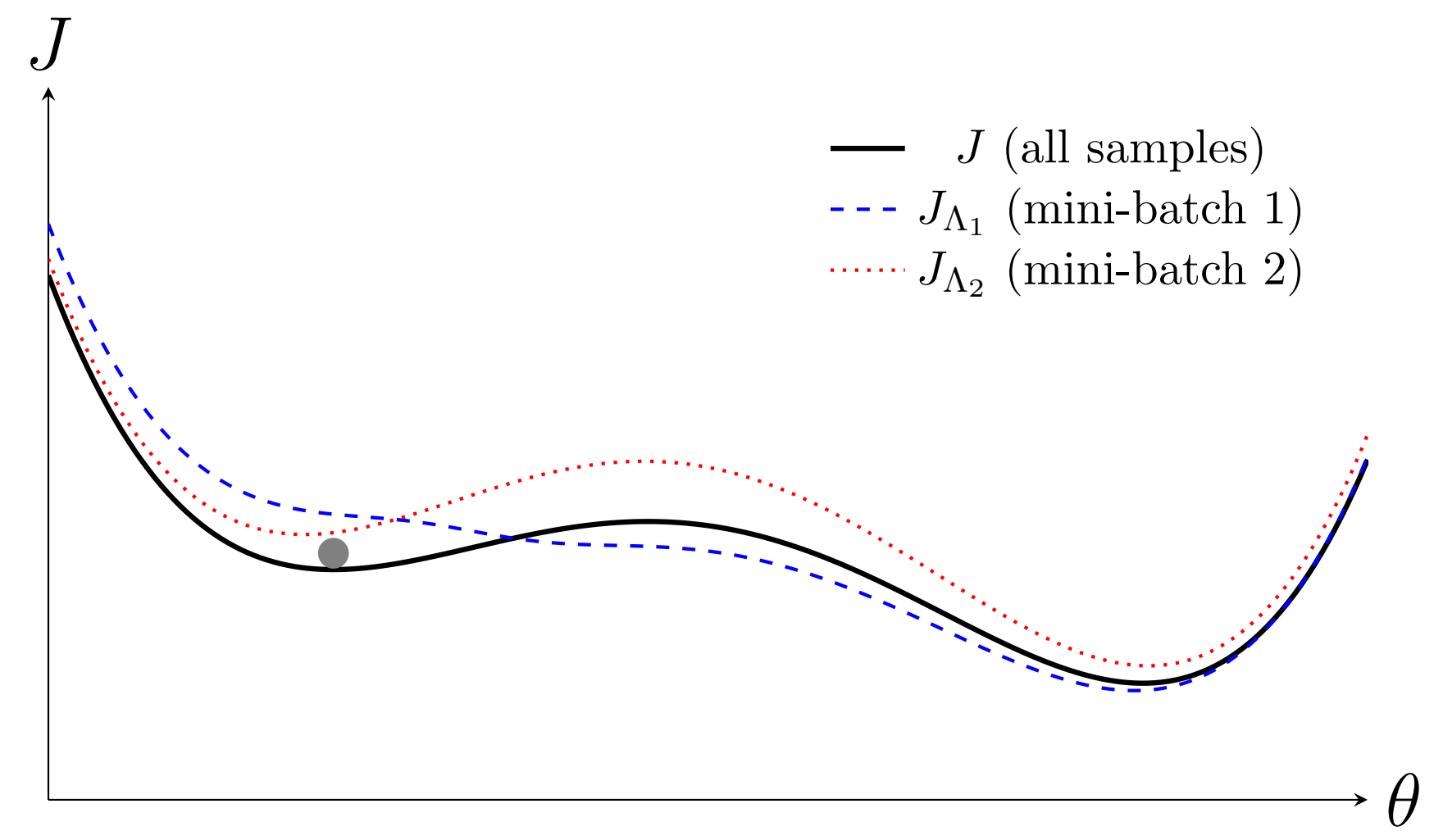

When the batch size is large, $\frac{\partial J_\Lambda}{\partial\theta_i}$ and $\frac{\partial J}{\partial\theta_i}$ should not be so different for a `healthy' dataset that is not dominated by a small number of outliers (the ML approach would work only for such `healthy' datasets), and we should get more
or less the same result whether we use vanilla GD or SGD.\footnote{In the limit of an infinitesimally small learning rate $\epsilon\to 0$, vanilla GD and SGD are equivalent.} Even when the batch size is small, it often works.
There are several (potential) advantages over vanilla GD:
\begin{itemize}
\item
The computational cost at each step is significantly lower.

\item
Because the locations of local minima depend on the choice of batch, it is less likely to be trapped in a local minimum. At least, we have a better chance of finding a local minimum which is robust under batch choice, which is likely to be a good one.
See Fig.~\ref{fig:minibatch_minima}. 

\item
It has an unexpected positive effect when applied to deep learning. Specifically, when the batch size is not too big, the model is less likely to fit accidental patterns in the training data that do not hold for new data. Such a failure is called \textit{overfitting}, and this effect of SGD is called implicit regularization; we will discuss them in Sec.~\ref{sec:overfit}. (This is similar to the second point above.)

\end{itemize}

In an image classification task, suppose that there were several pictures of Siberian Huskies with snow in the background. Then there might be a local minimum corresponding to a misleading pattern, ``snow  $\to$ dog". Such a local minimum disappears when those pictures are not in a batch. 

%%%%%%%%%%%%%%%%%%%%%%%%%
%%%%%%%%%%%%%%%%%%%%%%%%%
\section{Perceptrons and logic gates}
%%%%%%%%%%%%%%%%%%%%%%%%%
%%%%%%%%%%%%%%%%%%%%%%%%%
Roughly speaking, computer programs are made of logic gates, and neural networks are made of perceptrons. By seeing how they are related, we can understand some basic aspects of neural networks. Furthermore, we can easily see that anything that can be programmed on a computer can also be realized by using a neural network.

%%%%%%%%%%%%%%%%%%%%%%%%%
\subsection{Logic gates}
%%%%%%%%%%%%%%%%%%%%%%%%%
Computer programs are made of logic gates. If we have a computer that can execute basic logic gates, any complicated program can be executed~\cite{nakashima1936,shannon1938,piesch1939,stankovic2007}. We will consider four types of logic gates with two inputs and one output: the AND gate (Table~\ref{table:and-gate}), the NAND gate (Table~\ref{table:nand-gate}), the OR gate (Table~\ref{table:or-gate}), and the XOR gate (Table~\ref{table:xor-gate}). Each gate has two binary inputs $x_1$ and $x_2$ and one binary output $y$. Binary means 0 and 1. The common convention is that 0 is False and 1 is True. Let us also add one gate with one input and one output: the NOT gate (Table~\ref{table:not-gate}).

\begin{itemize}
\item
The NOT gate returns $y=1$ if $x=0$ and $y=0$ if $x=1$, i.e., the output $y$ is \textit{not} the same as the input $x$.
\item
The AND gate returns $y=1$ (True) for $x_1=1$ \textit{and} $x_2=1$, and $y=0$ (False) otherwise.
\item
NAND means \textit{Not} AND. It returns $y=0$ (False) for $x_1=x_2=1$ and $y=1$ (True) otherwise. The output is True when the inputs are \textit{not} ``True \textit{and} True".
We can also say this is a combination of the NOT gate and the AND gate.
\item
The OR gate returns $y=1$ (True) for $x_1=1$ \textit{or} $x_2=1$, and $y=0$ (False) otherwise.
\item
XOR means \textit{Exclusive} OR. When we say ``$x_1=1$ or $x_2=1$", usually ``$x_1=1$ \textit{and} $x_2 =1$" is allowed, but ``exclusive" means the latter case is excluded.
\end{itemize}

\begin{table}[htbp]
\centering
\begin{tabular}[htbp]{ |c|c||c|} 
\hline
 $x_1$ & $x_2$ & $y=\mathrm{AND}(x_1,x_2)$\\
\hline
\hline
 0 & 0 & 0 \\ 
\hline 
 1 & 0 & 0 \\ 
\hline 
 0 & 1 & 0 \\ 
\hline
 1 & 1 & 1 \\ 
\hline
\end{tabular}
\caption{AND gate. It returns $y=1$ if $x_1=1$ and $x_2=1$. Equivalently, the output $y$ is True when input $x_1$ is True and input $x_2$ is True.}\label{table:and-gate}
\end{table}

\begin{table}[htbp]
\centering
\begin{tabular}[htbp]{ |c|c||c|} 
\hline
 $x_1$ & $x_2$ & $y=\mathrm{NAND}(x_1,x_2)$\\
\hline
\hline
 0 & 0 & 1 \\ 
\hline 
 1 & 0 & 1 \\ 
\hline 
 0 & 1 & 1 \\ 
\hline
 1 & 1 & 0 \\ 
\hline
\end{tabular}
\caption{NAND gate (\textit{Not} AND gate). It returns $y=1$ if ``$x_1=1$ and $x_2=1$" does not hold.}\label{table:nand-gate}
\end{table}

\begin{table}[htbp]
\centering
\begin{tabular}[htbp]{ |c|c||c|} 
\hline
 $x_1$ & $x_2$ & $y=\mathrm{OR}(x_1,x_2)$\\
\hline
\hline
 0 & 0 & 0 \\ 
\hline 
 1 & 0 & 1 \\ 
\hline 
 0 & 1 & 1 \\ 
\hline
 1 & 1 & 1 \\ 
\hline
\end{tabular}
\caption{OR gate. It returns $y=1$ if $x_1=1$ \textit{or} $x_2=1$. Equivalently, the output $y$ is True when input $x_1$ is True or input $x_2$ is True.}\label{table:or-gate}
\end{table}

\begin{table}[htbp]
\centering
\begin{tabular}[htbp]{ |c|c||c|} 
\hline
 $x_1$ & $x_2$ & $y=\mathrm{XOR}(x_1,x_2)$\\
\hline
\hline
 0 & 0 & 0 \\ 
\hline 
 1 & 0 & 1 \\ 
\hline 
 0 & 1 & 1 \\ 
\hline
 1 & 1 & 0 \\ 
\hline
\end{tabular}
\caption{XOR gate, or \textit{Exclusive} OR gate. If we say ``$x_1=1$ or $x_2=1$", usually ``$x_1=1$ and $x_2=1$" is allowed, but ``exclusive" means the latter case is excluded.}\label{table:xor-gate}
\end{table}

\begin{table}[htbp]
\centering
\begin{tabular}[htbp]{ |c||c|} 
\hline
 $x$ & $y=\mathrm{NOT}(x)$\\
\hline
\hline 
 0 & 1 \\ 
\hline
 1 & 0 \\ 
\hline
\end{tabular}
\caption{NOT gate. It returns $y=1$ if $x=0$ and $y=0$ if $x=1$. In other words, the output $y$ is not the input $x$.}\label{table:not-gate}
\end{table}

Suppose we asked if people voted for the Conservative Party or the Labour Party in the last General Election. $(x_1,x_2)=(0,0)$ means they did not vote for these parties, and $(x_1,x_2)=(1,0)$ means they voted for the Conservatives and not for Labour. If $y=1$ or $0$, such a person did or did not exist, respectively. Unless someone votes twice, $(x_1,x_2)=(1,1)$ is impossible. Therefore, while $(x_1,x_2)=(0,0), (1,0), (0,1)$ could be true ($y=1$), $(x_1,x_2)=(1,1)$ should be false ($y=0$). This is the NAND gate. 

Let us classify British people based on their favorite football clubs. Suppose $x_1$ is whether a person is a Liverpool fan or not, $x_2$ is whether a person is an Everton fan or not, and $y$ is whether such a person can exist. Obviously, $(x_1,x_2)=(0,0)$ (i.e., neither a Liverpool fan nor an Everton fan) should return $y=1$ because many people support other teams. Furthermore, some people are not interested in football at all. $(x_1,x_2)=(1,1)$  and $y=1$ would mean there is a person who supports both Liverpool and Everton. There is a long debate about whether the Beatles support Everton or Liverpool. If you Google this subject, you can find a YouTube video in which Paul McCartney says he is a Liverpool and Everton fan, and he goes on like “I know it is not strictly allowed, but...”. From his comment, we can see that, although we would expect the NAND gate for this classification, this is not the case because Paul McCartney is an exception. If we repeat the same classification problem restricting the samples to people from Liverpool and removing Paul McCartney, XOR may be a good approximation.

We will see that a neural network can realize a combination of those gates, and hence, any complicated computer program can be realized by using a neural network.

As a simple example of a program, let us consider the addition of two non-negative integers. The simplest case is the addition of $x=0$ or 1 and $x'=0$ or 1. Let $x+x'\equiv(y_0,y_1)\equiv 2y_1+y_0$.
Then:
\begin{align}
\begin{array}{|c|c||c|c|}
\hline
 x & x' & y_0=\mathrm{XOR}(x,x') & y_1=\mathrm{AND}(x,x')\\
\hline
\hline 
 0 & 0 & 0 & 0\\ 
\hline 
 1 & 0 & 1 & 0\\ 
\hline 
 0 & 1 & 1 & 0\\ 
\hline
 1 & 1 & 0 & 1\\ 
\hline
\end{array}
\end{align}
We can see that $y_0 = \mathrm{XOR}(x,x')$ and $y_1 = \mathrm{AND}(x,x')$. 

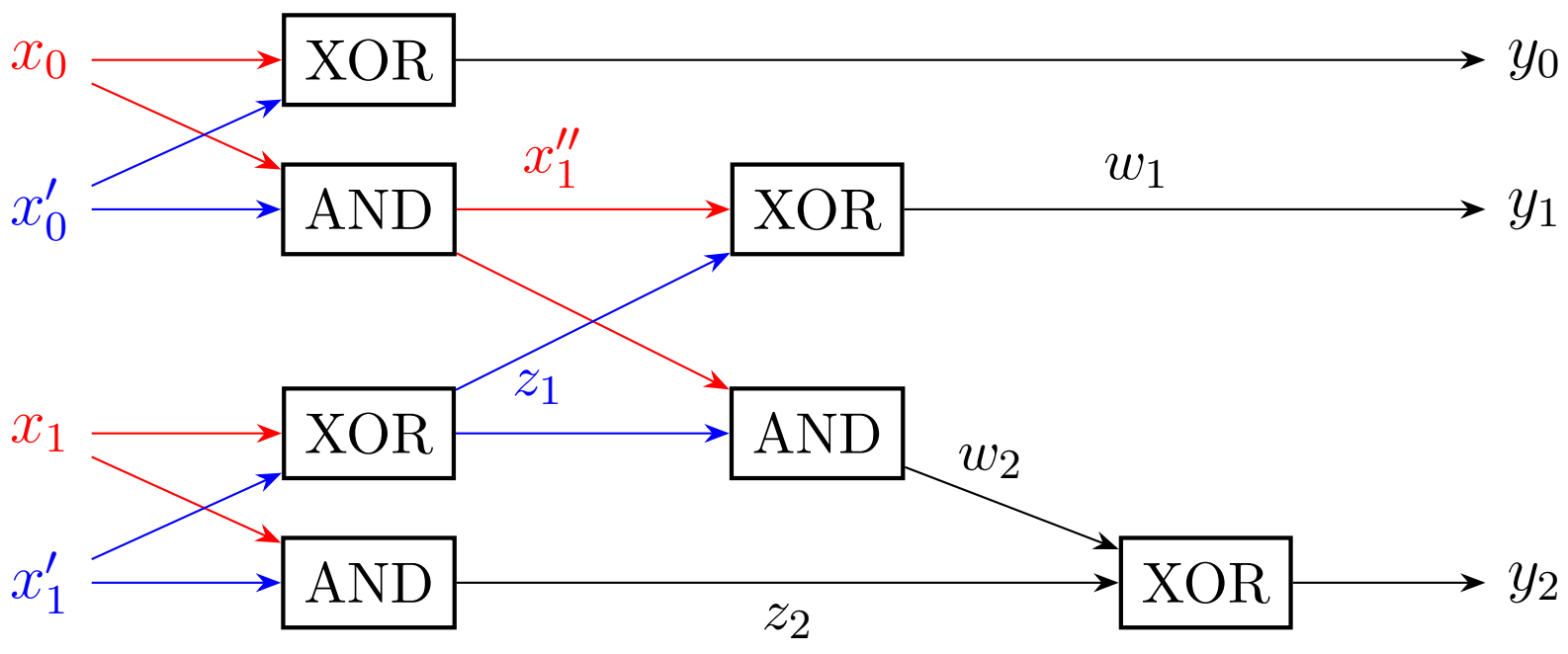
\begin{figure}[htbp]
\centering
\begin{tikzpicture}[>=Stealth, x=1cm, y=1cm,
gate/.style={draw, thick, minimum width=11mm, minimum height=6mm, font=\small}]
\node[red] (x0) at (0,4) {$x_0$};
\node[blue] (x0p) at (0,3) {$x'_0$};
\node[red] (x1) at (0,1.5) {$x_1$};
\node[blue] (x1p) at (0,0.5) {$x'_1$};
\node[gate] (g1) at (2.2,4) {XOR};
\node[gate] (g2) at (2.2,3) {AND};
\node[gate] (g3) at (2.2,1.5) {XOR};
\node[gate] (g4) at (2.2,0.5) {AND};
\node[gate] (g5) at (5.2,3) {XOR};
\node[gate] (g6) at (5.2,1.5) {AND};
\node[gate] (g7) at (7.8,0.5) {XOR};
\node (y0) at (10,4) {$y_0$};
\node (y1) at (10,3) {$y_1$};
\node (y2) at (10,0.5) {$y_2$};
\foreach \a/\b in {x0/g1,x0/g2,x1/g3,x1/g4}
\draw[->, red] (\a) -- (\b);
\foreach \a/\b in {x0p/g1,x0p/g2,x1p/g3,x1p/g4}
\draw[->, blue] (\a) -- (\b);
\draw[->] (g1) -- (y0);
\draw[->, red] (g2) -- node[pos=0.35,above,font=\small]{$x''_1$} (g5);
\draw[->, red] (g2) -- (g6);
\draw[->, blue] (g3) -- node[pos=0.3,below,font=\small]{$z_1$} (g5);
\draw[->, blue] (g3) -- (g6);
\draw[->] (g5) -- node[pos=0.4,above,font=\small]{$w_1$} (y1);
\draw[->] (g6) -- node[pos=0.4,above,font=\small]{$w_2$} (g7);
\draw[->] (g4) -- node[pos=0.5,below,font=\small]{$z_2$} (g7);
\draw[->] (g7) -- (y2);
\end{tikzpicture}
\caption{Addition of two-digit binary numbers in terms of logic gates.}
\label{fig:add-two-numbers}
\end{figure}

Next, let us consider the addition of two-digit numbers $(x_0,x_1)\equiv 2x_1+x_0$ and $(x'_0,x'_1)\equiv 2x'_1+x'_0$. Let the answer be $(y_0, y_1, y_2)\equiv 2^2y_2+2y_1+y_0$. Obviously, $y_0=\mathrm{XOR}(x_0, x'_0)$. 
To get $y_1$ and $y_2$, we need to add three numbers: $x_1$, $x'_1$ and $x''_1=\mathrm{AND}(x_0 ,x'_0)$. The gates we have accept only two inputs, and hence we cannot add three numbers at once. So, let us break the sum into three steps. Firstly, we compute $x_1+x'_1=(z_1,z_2)$, where $z_1=\mathrm{XOR}(x_1,x'_1)$ and $z_2=\mathrm{AND}(x_1,x'_1)$ as before. Then, we calculate $z_1+x''_1=(w_1,w_2)$ in the same way. Then, by construction, $y_1=w_1$ and $y_2=z_2+w_2$. The input-output table is as follows:
{\tiny
\begin{align}
\begin{array}{|c|c|c||c|c||c|c||c|c|}
\hline
 x_1 & x'_1 & x''_1 & z_1 =\mathrm{XOR}(x_1,x'_1)& z_2=\mathrm{AND}(x_1,x'_1) & w_1 =\mathrm{XOR}(z_1,x''_1)& w_2=\mathrm{AND}(z_1,x''_1) & y_1=w_1 & y_2 = z_2+w_2\\
\hline
\hline 
 0 & 0 & 0 & 0 & 0 & 0 & 0 & 0 & 0\\ 
\hline 
 1 & 0 & 0 & 1 & 0 & 1 & 0 & 1 & 0\\ 
\hline 
 0 & 1 & 0 & 1 & 0 & 1 & 0 & 1 & 0\\  
\hline 
 0 & 0 & 1 & 0 & 0 & 1 & 0 & 1 & 0\\ 
\hline
 1 & 1 & 0 & 0 & 1 & 0 & 0 & 0 & 1\\ 
 \hline
 1 & 0 & 1 & 1 & 0 & 0 & 1 & 0 & 1\\  
 \hline
 0 & 1 & 1 & 1 & 0 & 0 & 1 & 0 & 1\\ 
 \hline
 1 & 1 & 1 & 0 & 1 & 1 & 0 & 1 & 1\\  
\hline
\end{array}
\end{align}
}
As we can see from the table, $z_2$ and $w_2$ cannot be 1 simultaneously, and hence, $y_2=z_2+w_2=\mathrm{XOR}(z_2,w_2)=\mathrm{OR}(z_2, w_2)$. See Fig.~\ref{fig:add-two-numbers}.

%%%%%%%%%%%%%%%%%%%%%%%%%
\subsection{Single-layer perceptron and logic gates}
%%%%%%%%%%%%%%%%%%%%%%%%%
A perceptron~\cite{mcculloch1943,Rosenblatt1958} accepts two inputs $x_1,x_2\in\mathbb{R}$ and returns $y = 0$ or $y = 1$. Specifically, there are three parameters $w_1$, $w_2$ (weights) and $b$ (bias), and $y = 0$ when $w_1x_1 + w_2x_2 + b\le 0$, and $y = 1$ when $w_1x_1 + w_2x_2 + b > 0$.

\begin{figure}[htbp]
\centering
\begin{tikzpicture}[>=Stealth, unit/.style={circle, draw, thick, minimum size=13mm}]
\node[unit] (x0) at (0,1.8) {$x_0=1$};
\node[unit] (x1) at (0,0) {$x_1$};
\node[unit] (x2) at (0,-1.8) {$x_2$};
\node[unit] (y) at (4,0) {$y$};
\draw[->, thick] (x0) -- node[pos=0.45, above right]{$b$} (y);
\draw[->, thick] (x1) -- node[pos=0.45, above]{$w_1$} (y);
\draw[->, thick] (x2) -- node[pos=0.45, below right]{$w_2$} (y);
\end{tikzpicture}
\caption{A schematic picture of a perceptron.}
\label{fig:perceptron}
\end{figure}
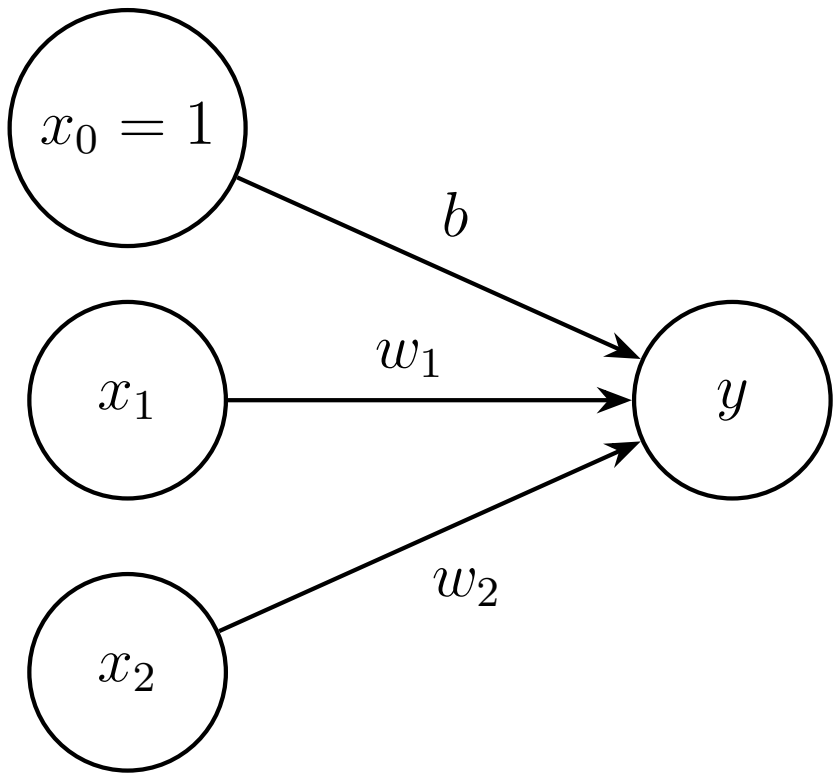

By using $\vec{x}=\begin{pmatrix} x_0 \\ x_1 \\ x_2\end{pmatrix}=\begin{pmatrix} 1 \\ x_1 \\ x_2\end{pmatrix}$, $\vec{w}=\begin{pmatrix} w_0 \\ w_1 \\ w_2\end{pmatrix}=\begin{pmatrix} b \\ w_1 \\ w_2\end{pmatrix}$, and the step function
\begin{align}
\Theta(z)
=
\left\{
\begin{array}{ccc}
0 & & (z\le 0)\\
1 & & (z>0)
\end{array}
\right.\, , 
\end{align}
we can write the above relation as
\begin{align}
y
=
\Theta(\vec{w}\cdot\vec{x})\, . 
\end{align}
Pictures like Fig.~\ref{fig:perceptron} are used to visualize the perceptron. Such pictures are useful when we consider neural networks. 

\begin{figure}[htbp]
\centering
\begin{tikzpicture}[scale=2.2]
\fill[gray!25] (-0.45,0.95) -- (0.95,-0.45) -- (-0.45,-0.45) -- cycle;
\draw[-{Stealth}] (-0.45,0) -- (1.5,0) node[right]{$x_1$};
\draw[-{Stealth}] (0,-0.45) -- (0,1.5) node[above]{$x_2$};
\draw[thick] (-0.45,0.95) -- (0.95,-0.45)
node[pos=0.95, below right, font=\small]{$x_1+x_2=0.5$};
\fill (-0.035,0.965) rectangle ++(0.07,0.07); \node[left, font=\small] at (-0.06,1) {1};
\fill (0.965,0.965) rectangle ++(0.07,0.07); \node[right, font=\small] at (1.06,1) {1};
\fill (0.965,-0.035) rectangle ++(0.07,0.07); \node[above, font=\small] at (1,0.05) {1};
\fill (0,0) circle (0.045); \node[below left, font=\small] at (0,0) {0};
\end{tikzpicture}
\hspace{2em}
\begin{tikzpicture}[scale=2.2]
\draw[-{Stealth}] (-0.45,0) -- (1.5,0) node[right]{$x_1$};
\draw[-{Stealth}] (0,-0.45) -- (0,1.5) node[above]{$x_2$};
\draw[thick] plot[smooth, tension=0.7] coordinates
{(-0.45,0.45) (0.25,0.4) (0.65,0.0) (1.0,-0.3) (1.32,-0.05)
(1.3,0.4) (0.68,0.75) (0.52,1.45)};
\fill (-0.035,0.965) rectangle ++(0.07,0.07); \node[left, font=\small] at (-0.06,1) {1};
\fill (0.965,-0.035) rectangle ++(0.07,0.07); \node[above, font=\small] at (1,0.05) {1};
\fill (1,1) circle (0.045); \node[right, font=\small] at (1.06,1) {0};
\fill (0,0) circle (0.045); \node[below left, font=\small] at (0,0) {0};
\end{tikzpicture}
\caption{Schematic pictures of perceptrons for the OR gate (left) and the XOR gate (right).}\label{fig:OR-and-XOR}
\end{figure}
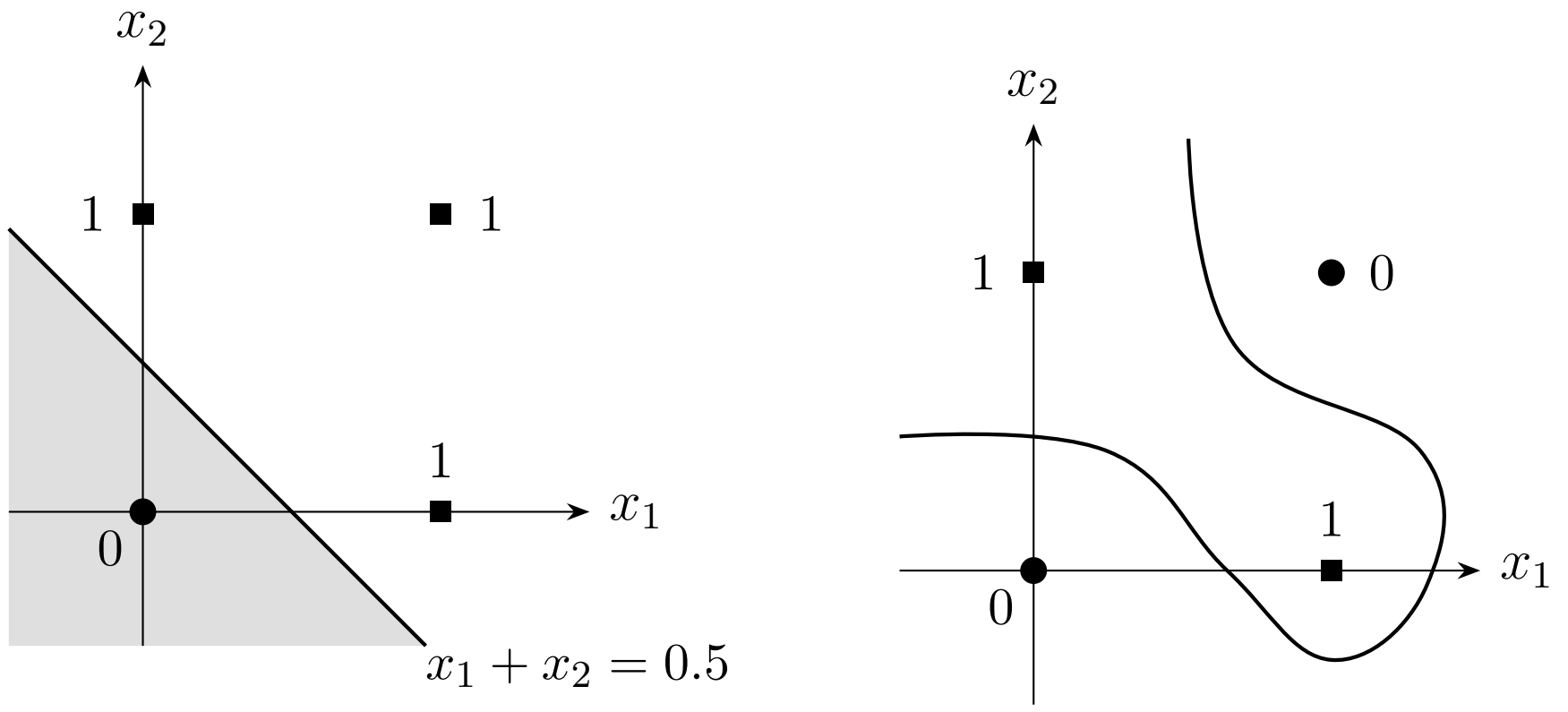

A nice feature of the perceptron is that we can make AND, NAND, and OR gates. For example, we can choose 
$\vec{w}=\begin{pmatrix} -1.5 \\ 1 \\ 1\end{pmatrix}$ for AND,
$\vec{w}=\begin{pmatrix} 1.5 \\ -1 \\ -1\end{pmatrix}$ for NAND, and 
$\vec{w}=\begin{pmatrix} -0.5 \\ 1 \\ 1\end{pmatrix}$
for OR, respectively. However, the XOR gate cannot be realized by a perceptron. To understand it, we can draw pictures like Fig.~\ref{fig:OR-and-XOR}. 
The left panel represents the OR gate. A straight line $\vec{w}\cdot\vec{x}=0$, which is $x_1+x_2=0.5$ in this case, splits $\mathbb{R}^2$ into two parts
-- $\vec{w}\cdot\vec{x}\le 0$ and $\vec{w}\cdot\vec{x}>0$ -- that correspond to $y=0$ and $y=1$. However, as we can see in the right panel, the XOR gate cannot be realized by drawing a straight line. We can still realize the XOR gate by drawing a complicated curve, but it requires some nonlinearity.

%%%%%%%%%%%%%%%%%%%%%%%%%
\subsection{Logic circuits from perceptrons}
%%%%%%%%%%%%%%%%%%%%%%%%%

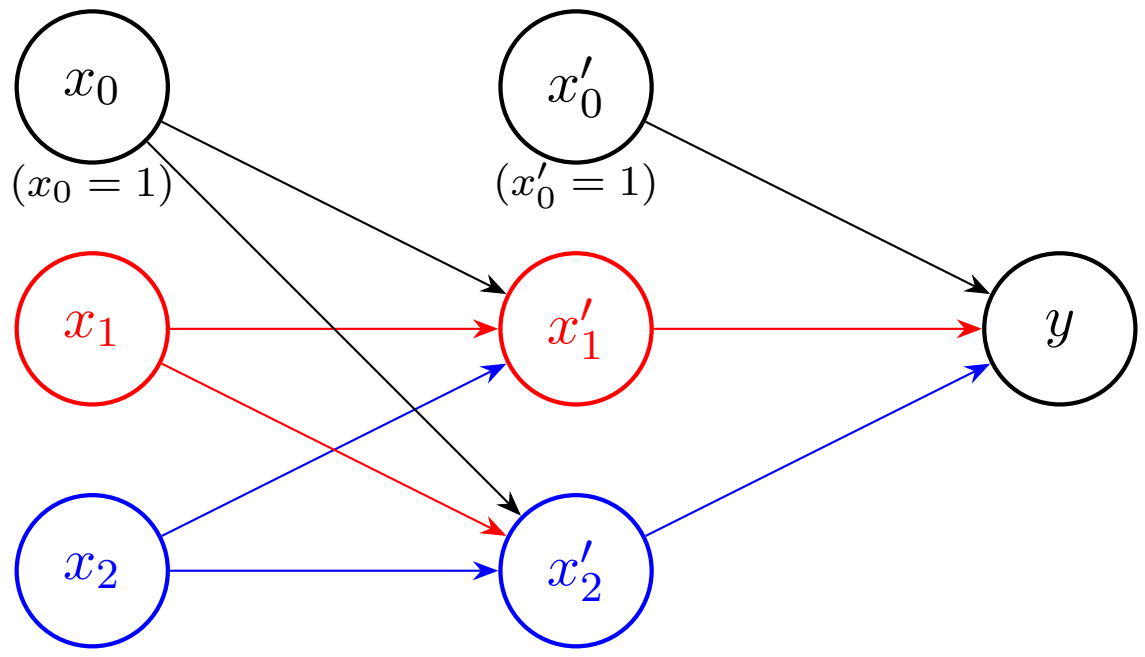
\begin{figure}[htbp]
\centering
\begin{tikzpicture}[>=Stealth,
  unit/.style={circle, draw, thick, minimum size=10mm, font=\small}]
\node[unit] (x0) at (0,2.4) {$x_0$};
\node[font=\scriptsize] at (0,1.75) {$(x_0=1)$};
\node[unit, red]  (x1) at (0,0.8)  {$x_1$};
\node[unit, blue] (x2) at (0,-0.8) {$x_2$};
\node[unit] (p0) at (3.2,2.4) {$x'_0$};
\node[font=\scriptsize] at (3.2,1.75) {$(x'_0=1)$};
\node[unit, red]  (p1) at (3.2,0.8)  {$x'_1$};
\node[unit, blue] (p2) at (3.2,-0.8) {$x'_2$};
\node[unit] (y) at (6.4,0.8) {$y$};
\foreach \t in {p1,p2} {
  \draw[->] (x0) -- (\t);
  \draw[->, red]  (x1) -- (\t);
  \draw[->, blue] (x2) -- (\t);
}
\draw[->] (p0) -- (y);
\draw[->, red]  (p1) -- (y);
\draw[->, blue] (p2) -- (y);
\end{tikzpicture}
\caption{XOR gate from two layers of perceptrons. Three perceptrons are used.}
\label{fig:XOR-as-circuit}
\end{figure}

The XOR gate can easily be realized by combining two layers of perceptrons. Let $x'_1$ and $x'_2$ be $x'_1 = \mathrm{NAND}(x_1,x_2)$ and $x'_2 = \mathrm{OR}(x_1,x_2)$, respectively. Then, $y = \mathrm{AND}(x'_1,x'_2)$ is the XOR gate. Equivalently, $\mathrm{XOR}(x_1, x_2) = \mathrm{AND}(\mathrm{NAND}(x_1, x_2), \mathrm{OR}(x_1, x_2))$. 
See Table~\ref{table:XOR-as-circuit} and Fig.~\ref{fig:XOR-as-circuit}.

\begin{table}[htbp]
\centering
\begin{tabular}[htbp]{|c|c||c|c||c|}
\hline
 $x_1$ & $x_2$ & $x'_1 =\mathrm{NAND}(x_1,x_2)$ & $x'_2=\mathrm{OR}(x_1,x_2)$ & $y =\mathrm{AND}(x'_1,x'_2)$\\
\hline
\hline 
 0 & 0 & 1 & 0 & 0\\
 \hline 
 1 & 0 & 1 & 1 & 1\\
 \hline 
 0 & 1 & 1 & 1 & 1\\
 \hline 
 1 & 1 & 0 & 1 & 0\\
\hline
\end{tabular}
\caption{XOR gate made of NAND, OR, and AND gates.}\label{table:XOR-as-circuit}
\end{table}

Any logic circuit can be built by using the logic gates shown above. Therefore, any logic circuit can be realized as a network of perceptrons.

Note that, in principle, any logic circuit can be realized by combining NAND gates. It is a good exercise to make AND and OR from NAND. The first step is to make NOT, which flips 0 and 1.

%%%%%%%%%%%%%%%%%%%%%%%%%
\subsection{Perceptrons with many inputs}
%%%%%%%%%%%%%%%%%%%%%%%%%
We can also have perceptrons with more than two inputs $x_1,x_2,\cdots,x_n$. We can write it as $y=\Theta(\vec{w}\cdot\vec{x})$, where
$\vec{x}=\begin{pmatrix} x_0 \\ x_1 \\ \vdots\\ x_n\end{pmatrix}=\begin{pmatrix} 1 \\ x_1 \\ \vdots \\ x_n\end{pmatrix}$, $\vec{w}=\begin{pmatrix} w_0 \\ w_1\\ \vdots \\ w_n\end{pmatrix}=\begin{pmatrix} b \\ w_1\\ \vdots \\ w_n\end{pmatrix}$.  

%%%%%%%%%%%%%%%%%%%%%%%%%
\subsection{Importance of nonlinearity in activation function}
%%%%%%%%%%%%%%%%%%%%%%%%%
The step function $\Theta$ is a simple example of an activation function. 
An important feature of an activation function is its nonlinearity. In fact, if we do not introduce nonlinearity via an activation function, there is no point in adding many layers. As an example, let us consider a two-layer network like Fig.~\ref{fig:XOR-as-circuit}. 
If we do not use the activation function and take $x'_1 = \vec{w}\cdot\vec{x}$, $x'_2 = \vec{w}'\cdot\vec{x}$, and $y = \vec{w}''\cdot\vec{x}'$, then
\begin{align}
y
&=
b''
+
w''_1(\vec{w}\cdot\vec{x})
+
w''_2(\vec{w}'\cdot\vec{x})
\nonumber\\
&=
(w''_1b+w''_2b'+b'')
+
(w''_1w_1+w''_2w'_1)x_1
+
(w''_1w_2+w''_2w'_2)x_2
\nonumber\\
&=
\vec{\tilde{w}}\cdot\vec{x}\, , 
\end{align}
where
\begin{align}
\vec{\tilde{w}}
=
\left(
\begin{array}{c}
w''_1b+w''_2b'+b''\\
w''_1w_1+w''_2w'_1\\
w''_1w_2+w''_2w'_2
\end{array}
\right)\, . 
\end{align}
This is just another single-layer perceptron.
%%%%%%%%%%%%
%%%%%%%%%%%%
\section{Neural networks}
%%%%%%%%%%%%
%%%%%%%%%%%%

%%%%%%%%%%%%
\subsection{Neural network}
%%%%%%%%%%%%

Let us define a neural network. Actually, I am not really sure if there is a consensus. Some people may call a multi-layer network of perceptrons a neural network. Probably, more people assume certain activation functions which are not the step function $\Theta(x)$. (We will see several examples of activation functions in Sec.~\ref{sec:activation_fnc}.)

A neuron (or simply `node') receives signals from outside or from other neurons. Each neuron is similar to a multi-input perceptron, except that the activation function $f$ is (usually) not a step function. Although we can imagine various ways to combine the inputs $\vec{x}$, we use $\vec{w}\cdot\vec{x}$ with a weight vector $\vec{w}$. It is converted to the output $f(\vec{w}\cdot\vec{x})$ and passed to other neurons. Historically, neural networks were inspired by biological neural networks. If you want to distinguish them clearly from biological neural networks, you can say artificial neural networks.

A not-so-nice feature of $\Theta(x)$ is that the derivative is zero at $x\neq 0$. This makes it hard to tune the parameters by using gradient descent. As an example, suppose that we have inputs $\vec{x}^{(i)}$ $(i=1,2,\cdots)$, say pictures of gorillas or chimpanzees, and labels $y^{(i)} = 0$ or $1$ (say 0 for gorilla and 1 for chimpanzee). Our task would be to tune parameters in a given network so that a cost function is minimized. However, when we vary the parameters $\vec{\theta}$, the cost function does not change at all, or otherwise it jumps abruptly.

For example, we could choose a single-layer perceptron and take the cost function as $J(\vec{w})=\frac{1}{2}\sum_k\left(\Theta(\vec{w}\cdot\vec{x}^{(k)})-y^{(k)}\right)^2$. Then, the derivative with respect to $\vec{w}$ is zero except for $\vec{w}\cdot\vec{x}^{(k)}=0$, and at $\vec{w}\cdot\vec{x}^{(k)}=0$ the derivative is not well defined.\footnote{
Because our convention is $\Theta(z)=0$ for $z\le 0$, we can define the left derivative of $\Theta(z)$ at $z=0$ as $\lim_{\Delta z\to +0}\frac{\Theta(0)-\Theta(-\Delta z)}{\Delta z}=0$. 
But it does not resolve the trouble here.
}
Therefore, the gradient descent method does not work.

\begin{figure}[htbp]
\centering
\begin{tikzpicture}
\begin{axis}[
width=0.55\textwidth, height=0.35\textwidth,
domain=-8:8, samples=200,
xmin=-8.5, xmax=9, ymin=-0.1, ymax=1.25,
axis lines=middle, xlabel={$x$}, ylabel={$\sigma(x)$},
ytick={0,0.5,1}, yticklabel style={/pgf/number format/fixed},
]
\addplot[blue, thick] {1/(1+exp(-x))};
\draw[dashed, gray] (axis cs:-8.5,1) -- (axis cs:8,1);
\end{axis}
\end{tikzpicture}
\caption{Sigmoid function $\sigma(x)=\frac{1}{1+e^{-x}}$.}
\label{fig:sigmoid}
\end{figure}
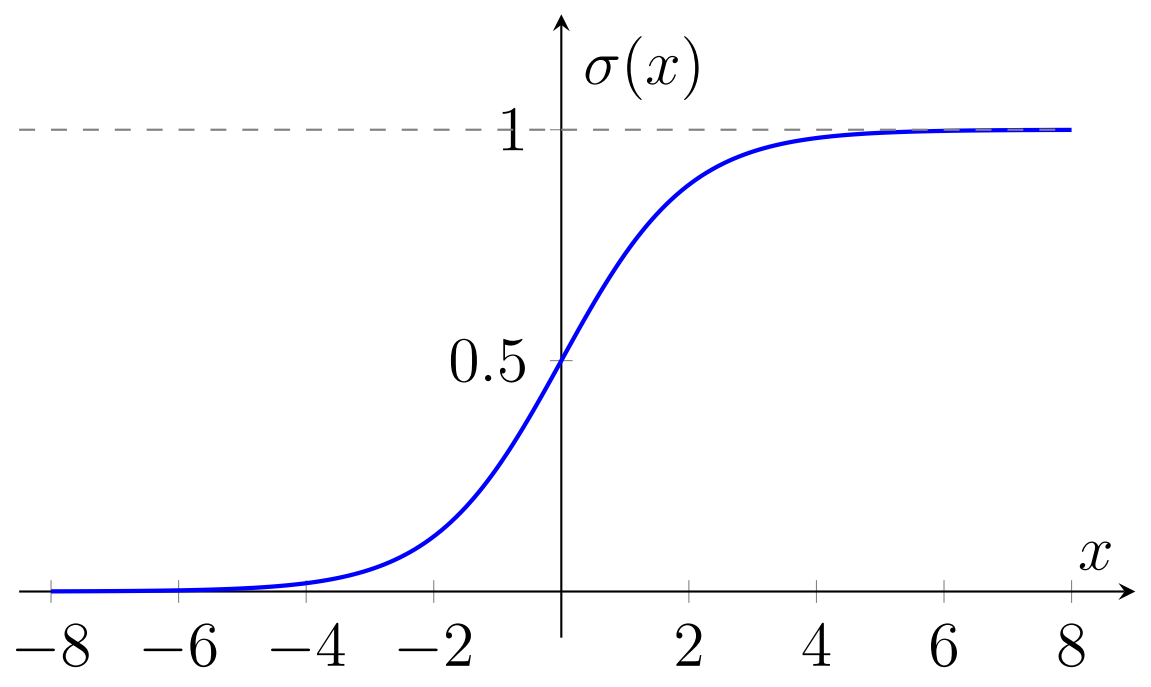

A simple cure is to replace $\Theta(x)$ with something similar but differentiable everywhere and not constant. The sigmoid function is such an example. It is defined by $\sigma(x)=\frac{1}{1+e^{-x}}$; see Fig.~\ref{fig:sigmoid}. Then, the cost function $J(\vec{w})=\frac{1}{2}\sum_k\left(\sigma(\vec{w}\cdot\vec{x}^{(k)})-y^{(k)}\right)^2$ is differentiable. The derivative is
\begin{align}
\frac{\partial J(\vec{w})}{\partial\vec{w}}
=
\sum_k\left(\sigma(\vec{w}\cdot\vec{x}^{(k)})-y^{(k)}\right)\cdot\sigma'(\vec{w}\cdot\vec{x}^{(k)})\cdot\vec{x}^{(k)}\, , 
\end{align}
where
\begin{align}
\sigma'(z)
=
\frac{1}{(e^{z/2}+e^{-z/2})^2}\, . 
\label{sigmoid-derivative}
\end{align}

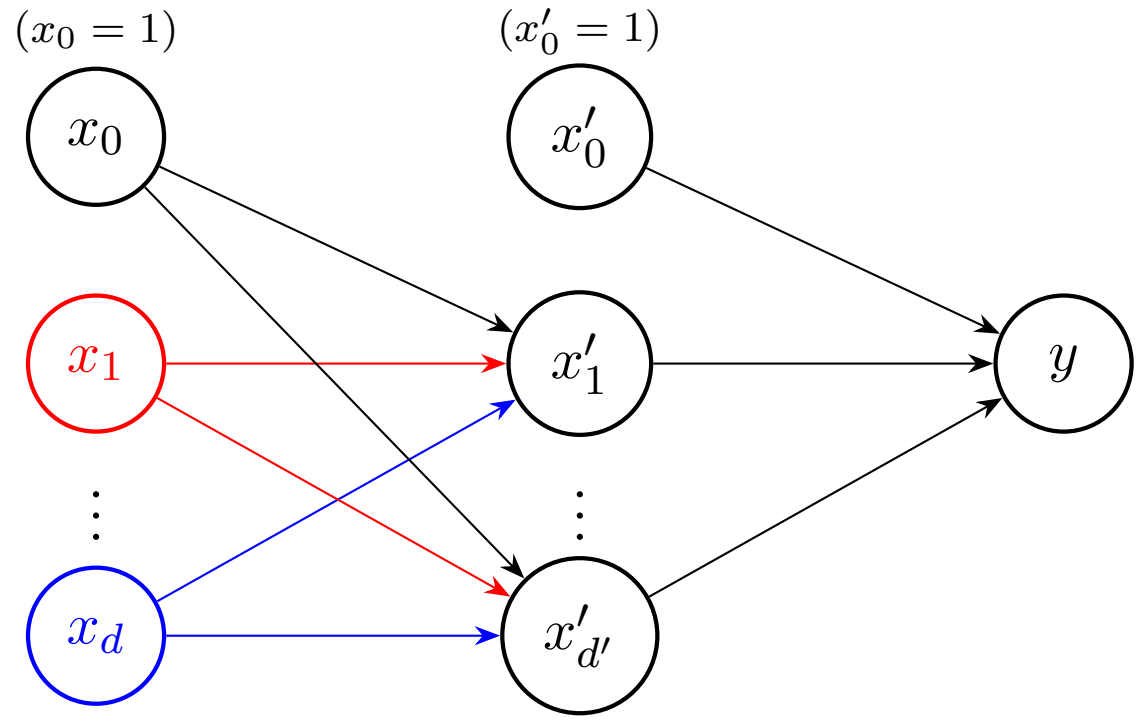
\begin{figure}[htbp]
\centering
\begin{tikzpicture}[>=Stealth,
  unit/.style={circle, draw, thick, minimum size=9mm, font=\small}]
\node[unit] (x0) at (0,3)  {$x_0$};
\node[font=\scriptsize] at (0,3.7) {$(x_0=1)$};
\node[unit, red]  (x1) at (0,1.5) {$x_1$};
\node[font=\small] at (0,0.6) {$\vdots$};
\node[unit, blue] (xd) at (0,-0.3) {$x_d$};
\node[unit] (p0) at (3.2,3)  {$x'_0$};
\node[font=\scriptsize] at (3.2,3.7) {$(x'_0=1)$};
\node[unit] (p1) at (3.2,1.5) {$x'_1$};
\node[font=\small] at (3.2,0.6) {$\vdots$};
\node[unit] (pd) at (3.2,-0.3) {$x'_{d'}$};
\node[unit] (y) at (6.4,1.5) {$y$};
\foreach \t in {p1,pd} {
  \draw[->] (x0) -- (\t);
  \draw[->, red]  (x1) -- (\t);
  \draw[->, blue] (xd) -- (\t);
}
\foreach \s in {p0,p1,pd} \draw[->] (\s) -- (y);
\end{tikzpicture}
\caption{A simple two-layer neural network. $x'_i = \sigma(\vec{w}_i\cdot\vec{x})$ ($i = 1, 2,\cdots, d'$) and $y = \sigma(\vec{w}'\cdot\vec{x}')$.}
\label{fig:2-layer-NN}
\end{figure}

As a slightly more complex example, let us consider $x'_i = \sigma(\vec{w}_i\cdot\vec{x})$ ($i = 1, 2,\cdots, d'$), $y = \sigma(\vec{w}'\cdot\vec{x}')$. Then, by combining these, we define $y = h_{\vec{\theta}}(\vec{x})$. Here, we used $\vec{\theta}$ to denote all parameters ($\vec{w}_1,\cdots,\vec{w}_{d'}$ and $\vec{w}'$). See Fig.~\ref{fig:2-layer-NN}. Let us use a cost function
\begin{align}
J(\vec{\theta})
=
\frac{1}{2}\sum_k\left(
h_{\vec{\theta}}(\vec{x}^{(k)})
-
y^{(k)}
\right)^2\, . 
\end{align}

To avoid confusion, let us show all indices explicitly. By using a matrix notation $w_{ij}$ for the components of $\vec{w}_i$, namely $\vec{w}_i=\begin{pmatrix}w_{i0}\\ w_{i1}\\ \vdots\\ w_{id}\end{pmatrix}=\begin{pmatrix}b_{i}\\ w_{i1}\\ \vdots\\ w_{id}\end{pmatrix}$, we have
\begin{align}
x'_i
=
\sigma\left(
\sum_j w_{ij}x_j
\right)
\end{align}
and
\begin{align}
h_{\vec{\theta}}(\vec{x})
=
\sigma\left(
\sum_i w'_ix'_i
\right)
=
\sigma\left(
\sum_i w'_i
\sigma\left(
\sum_j w_{ij}x_j
\right)
\right)\, . 
\end{align}
Therefore, 
\begin{align}
\frac{\partial h_{\vec{\theta}}(\vec{x})}{\partial w'_i}
&=
\left.
\frac{\mathrm{d}\sigma(z)}{\mathrm{d}z}
\right|_{z=\vec{w}'\cdot\vec{x}'}
\times
\frac{\partial}{\partial w'_i}
\left(
\sum_kw'_kx'_k
\right)
\nonumber\\
&=
\sigma'(\vec{w}'\cdot\vec{x}')\times x'_i\, , 
\end{align}
and
\begin{align}
\frac{\partial h_{\vec{\theta}}(\vec{x})}{\partial w_{ij}}
&=
\sigma'(\vec{w}'\cdot\vec{x}')\times 
\frac{\partial}{\partial w_{ij}}
\left(
\sum_kw'_k
\sigma\left(
\sum_lw_{kl}x_l
\right)
\right)
\nonumber\\
&=
\sigma'(\vec{w}'\cdot\vec{x}')\times 
\sum_kw'_k
\cdot
\frac{\partial}{\partial w_{ij}}
\sigma\left(
\sum_lw_{kl}x_l
\right)
\nonumber\\
&=
\sigma'(\vec{w}'\cdot\vec{x}')\times 
\sum_kw'_k
\cdot
\sigma'(\vec{w}_k\cdot\vec{x})
\cdot
\frac{\partial}{\partial w_{ij}}
\left(
\sum_lw_{kl}x_l
\right)
\nonumber\\
&=
\sigma'(\vec{w}'\cdot\vec{x}')\times 
\sum_kw'_k
\cdot
\sigma'(\vec{w}_k\cdot\vec{x})
\cdot
\sum_l\delta_{ik}\delta_{jl}x_l
\nonumber\\
&=
\sigma'(\vec{w}'\cdot\vec{x}')\times 
w'_i
\cdot
\sigma'(\vec{w}_i\cdot\vec{x})
\cdot
x_j\, .
\end{align}
See \eqref{sigmoid-derivative} for an explicit form of $\sigma'(z)$. By using these expressions, we can write
\begin{align}
\frac{\partial J(\vec{\theta})}{\partial\vec{\theta}}
=
\sum_k\left(
h_{\vec{\theta}}(\vec{x}^{(k)})
-
y^{(k)}
\right)
\cdot
\frac{\partial h_{\vec{\theta}}(\vec{x}^{(k)})}{\partial\vec{\theta}}
\end{align}
explicitly in terms of $\vec{\theta}$. 
%%%%%%%%%%%%
\subsection{Deep neural network}
\begin{figure}[htbp]
\centering
\begin{tikzpicture}[
  unit/.style={circle, draw, thick, minimum size=13mm, inner sep=0pt, font=\small},
  arr/.style={-{Triangle[length=5mm,width=6mm]}, line width=3mm, gray}]
\foreach \x/\l/\dm in {0/0/d_0, 3.2/1/d_1, 8.4/M/d_M} {
  \node[unit] at (\x,3.6)  {$x_0^{(\l)}$};
  \node[unit] at (\x,1.8)  {$x_1^{(\l)}$};
  \node       at (\x,0.6)  {$\vdots$};
  \node[unit] at (\x,-0.6) {$x_{\dm}^{(\l)}$};
}
\node[unit] at (11.6,2.7) {$y_1$};
\node       at (11.6,1.6) {$\vdots$};
\node[unit] at (11.6,0.3) {$y_n$};
\node[font=\large] at (0,4.8)    {input};
\node[font=\large] at (11.6,4.8) {output};
\draw[arr] (1.0,1.5) -- (2.2,1.5);
\draw[arr] (4.2,1.5) -- (5.1,1.5);
\node at (5.8,1.5) {$\cdots$};
\draw[arr] (6.5,1.5) -- (7.4,1.5);
\draw[arr] (9.4,1.5) -- (10.6,1.5);
\end{tikzpicture}
\caption{A simple multi-layer neural network with $M$ hidden layers.}
\label{fig:DNN-cartoon}
\end{figure}
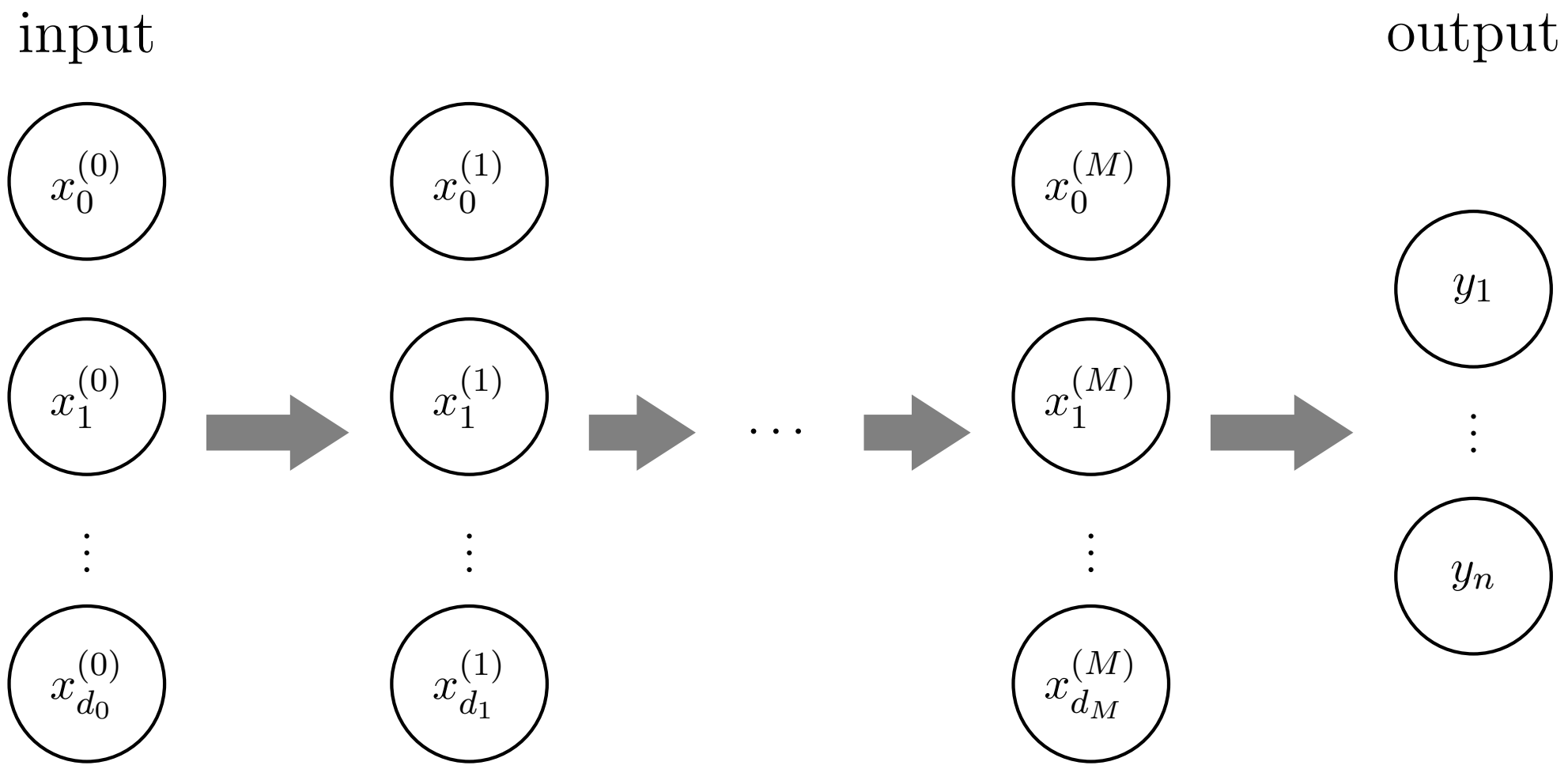

%%%%%%%%%%%%
In principle, any continuous function can be approximated arbitrarily well by a two-layer neural network with the sigmoid activation function~\cite{Cybenko89}. (See some comments below.) However, this statement only guarantees that good weights exist. It does not tell us how many nodes are needed, which can be enormous, or whether gradient descent can find such weights. A better strategy is to use simpler blocks to build complicated stuff: each layer does something simple, and later layers combine the outputs of the earlier ones. We can take such a strategy by making a network deeper. See Fig.~\ref{fig:DNN-cartoon} for a schematic picture of a multi-layer neural network with $M$ hidden layers (or intermediate layers).

That a network is \textit{deep} means the number of hidden layers $M$ is large. But how large? The minimum would be $M=2$. Some people may say a much larger value, which is the same as or larger than the depth they use. In these notes, we use $M\ge 2$ as the definition of \textit{deep}.

%%%%%%%%%%%%
\subsubsection{From sigmoid to arbitrary functions}
%%%%%%%%%%%%
Let us see how a two-layer neural network of the form shown in Fig.~\ref{fig:2-layer-NN} with the sigmoid activation function at the hidden layer can describe arbitrary continuous functions. ($y$ is taken to be a linear combination of $x'$s.) The point is to write a function you want to approximate by using a simple basis (e.g., sine and cosine) and show that the basis functions can be expressed by using such an architecture.

\begin{figure}[htbp]
\centering
\begin{tikzpicture}
\begin{axis}[
width=0.6\textwidth, height=0.38\textwidth,
declare function={f(\x)=3.2*exp(-((\x-3.2)^2)/4.5)*(1+0.25*sin(deg(1.4*\x)));},
xmin=0, xmax=6.8, ymin=0, ymax=4.2,
axis lines=left, xlabel={$x$}, ylabel={$y$},
xtick=\empty, ytick=\empty,
]
\addplot[ybar interval, fill=gray!20, draw=gray, domain=0.2:6.2, samples=13] {f(x+0.25)};
\addplot[red, thick, smooth, domain=0.2:6.2, samples=200] {f(x)};
\end{axis}
\end{tikzpicture}
\caption{We can use step functions to approximate any function. Each strip is a difference of two step functions, $\Theta(x-a)-\Theta(x-b)$, multiplied by a real number.}
\label{fig:approximation-by-step-fnc}
\end{figure}
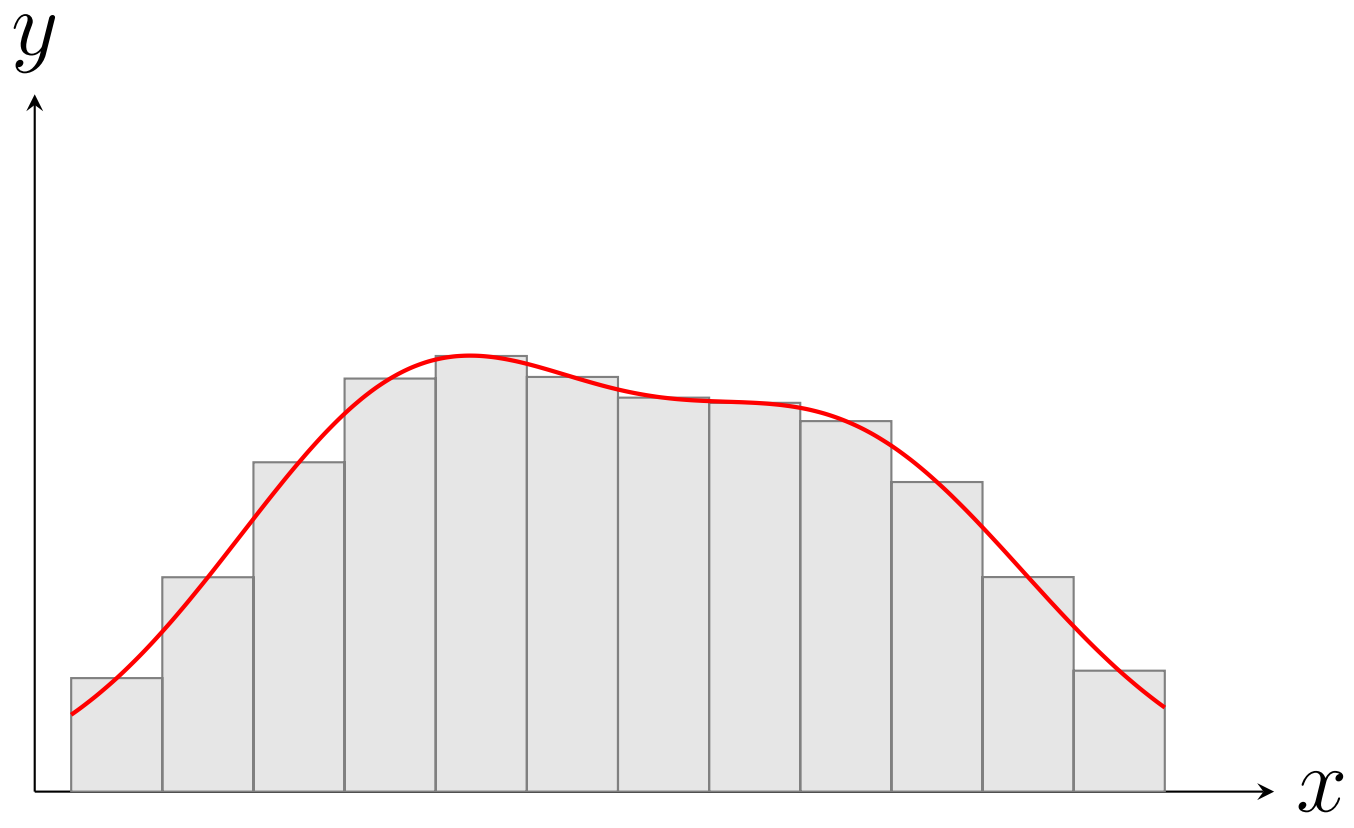

For univariate functions $f(x)$, we can easily find an explicit construction. 
Firstly, we can easily make a step function $\Theta(x-a)$ by taking $\vec{w}=\begin{pmatrix}-ca\\ c\end{pmatrix}$ and sending $c$ to $\infty$:
$\Theta(x-a) = \lim_{c\to\infty} \sigma(c(x-a))$ for $x\neq a$. (At $x=a$, the limit is $1/2$ rather than $\Theta(0)=0$, but a single point does not matter for the approximation below.) These step functions are obtained as $x'$ at the second layer of Fig.~\ref{fig:2-layer-NN}. Next, by using two step functions, we can define the following function for any $a < b$:
\begin{align}
\Theta(x-a)
-
\Theta(x-b)
=
\left\{
\begin{array}{ccl}
0 & &(x\le a)\\
1 & &(a< x \le b)\\
0 & &(x> b)
\end{array}
\right.\, .
\end{align}
By multiplying these functions by appropriate real numbers and taking a sum, any continuous function can be approximated; see Fig.~\ref{fig:approximation-by-step-fnc}. Therefore, any continuous function $y(x)$ can be obtained by taking a linear combination of $x'$, and hence, we can approximate any continuous function by using a two-layer neural network with the sigmoid activation function.

For any number of variables, the existence of a good approximation is mathematically proven, although the explicit construction was not given. A formal statement for the case of $d$ variables is as follows:

For any continuous function $f : [0,1]^d\to\mathbb{R}$ and $\epsilon>0$, there exist $m$ and parameters $\vec{w}_j$ and $w'_j$ ($j = 1,2,\cdots,m$) and $c$, such that
\begin{align}
\sup_{\vec{x}\in [0,1]^d}
\left|
f(\vec{x})
-
\sum_{j=1}^m w'_j \sigma(\vec{w}_j\cdot\vec{x})
-
c
\right|
<
\epsilon\, . 
\end{align}

This theorem does not say anything about the concrete value of $m$ or the values of the parameters. Therefore, it is hard to say if such a two-layer neural network is an efficient architecture or not.

When a neural network is used to approximate a function, the basis for the expansion is not fixed. Namely, it is not fixed to step functions or sines and cosines; rather, the parameters $\vec{w}_j$, $w'_j$ can change depending on the target function. Because of this feature, the number of functions needed for a good approximation can be smaller than in the case of a fixed basis. Functions that can efficiently approximate target functions are \textit{good features}. For the classification of animal images, for example, good features would be features that capture the visual regularities that matter for telling one animal from another. In practice, these range from low-level patterns such as edges, to mid-level ones such as fur textures and characteristic shapes of ears or snouts, and even to more abstract high-level motifs like ``striped pattern'' or ``round face.'' A network that learns such features can represent the essential structure of many different animals with far fewer units than a model forced to rely on a fixed basis like sines or step functions.
%%%%%%%%%%%%
\subsection{Activation functions}\label{sec:activation_fnc}
%%%%%%%%%%%%

%%%%%%%%%%%%
\subsubsection{Sigmoid}
%%%%%%%%%%%%
The sigmoid function $\sigma(x)$ is defined by
\begin{align}
\sigma(x) 
= 
\frac{1}{1+e^{-x}}\, .  
\end{align}
The sigmoid function is essentially the same as tanh:
\begin{align}
\frac{1}{1+e^{-x}}
=
\frac{1}{2}
\left(
\frac{1+e^{-x}}{1+e^{-x}}
+
\frac{1-e^{-x}}{1+e^{-x}}
\right)
=
\frac{1}{2}
\left(
1
+
\frac{e^{x/2}-e^{-x/2}}{e^{x/2}+e^{-x/2}}
\right)
=
\frac{1}{2}
\left(
1
+
\tanh\left(
\frac{x}{2}
\right)
\right)\, . 
\end{align}
See Fig.~\ref{fig:sigmoid}. 
Using Python, sigmoid can be coded and plotted as follows:
\begin{breakbox}
\begin{verbatim}
import numpy as np
import matplotlib.pyplot as plt
########################
### sigmoid function ###
########################
def sigmoid(x):
        return(1/(1+np.exp(-x)))
##########################
# print sigmoid function #
##########################
x = np.linspace(-8,8,1000)
y = sigmoid(x)
z = 0.5*(np.tanh(x/2)+1)
plt.plot(x,y)
plt.plot(x,z)
plt.grid()
plt.xlabel("$x$")
plt.show()
\end{verbatim}
\end{breakbox}

%%%%%%%%%%%%
\subsubsection{ReLU (Rectified Linear Unit) and friends}
%%%%%%%%%%%%
The ReLU function $h(x)$ is defined by~\cite{Fukushima69}
\begin{align}
h(x)
=
\left\{
\begin{array}{ccl}
0 & & (x\le 0)\\
x & & (x> 0)
\end{array}
\right.\, .
\end{align}
Although the derivative is not well defined at $x=0$,\footnote{We could use the left or right derivative, but still, the derivative jumps at $x = 0$.} the chance that exactly $x = 0$ is hit is zero, and hence we do not have to worry. ReLU is popular because it is useful for avoiding the vanishing gradient problem. (See Sec.~\ref{sec:vanishing_gradient}.)

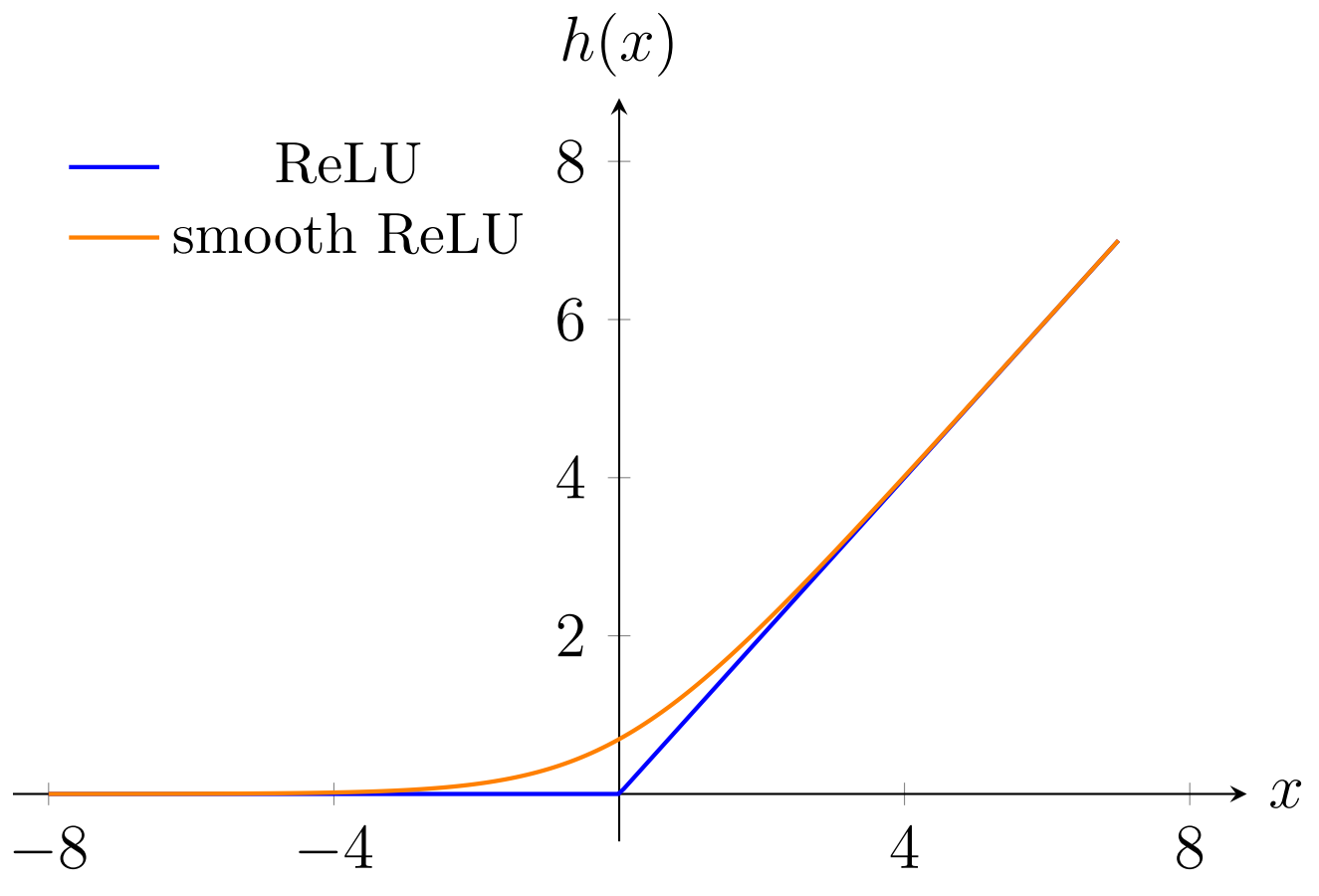
\begin{figure}[htbp]
\centering
\begin{tikzpicture}
\begin{axis}[
width=0.6\textwidth, height=0.4\textwidth,
domain=-8:7, samples=200,
xmin=-8.5, xmax=8.8, ymin=-0.6, ymax=8.8,
axis lines=middle,
xlabel={$x$}, ylabel={$h(x)$},
xlabel style={at={(ticklabel* cs:1)}, anchor=west},
ylabel style={at={(ticklabel* cs:1)}, anchor=south},
xtick={-8,-4,4,8}, ytick={2,4,6,8},
legend style={at={(0.03,0.97)}, anchor=north west, draw=none, font=\small},
]
\addplot[blue, thick] {max(0,x)};
\addlegendentry{ReLU}
\addplot[orange, thick] {ln(1+exp(x))};
\addlegendentry{smooth ReLU}
\end{axis}
\end{tikzpicture}
\caption{ReLU (blue) and smooth ReLU (orange).}
\label{fig:ReLU}
\end{figure}

There are many variants of ReLU. The smooth ReLU (also called softplus) is defined by
\begin{align}
h_{\rm smooth}(x)
=
\log(1+e^{x})\, . 
\end{align}
See Fig.~\ref{fig:ReLU} to see how ReLU and smooth ReLU are related. 
The leaky ReLU is defined by
\begin{align}
h_{{\rm leaky},\alpha}(x)
=
\left\{
\begin{array}{ccl}
\alpha x & & (x\le 0)\\
x & & (x>0)
\end{array}
\right.
\end{align}
for a small positive number $\alpha$, say $\alpha=0.1$. The smooth ReLU and leaky ReLU have
a nonzero derivative even for $x \le 0$.

Implementation in Python is straightforward. We also include the step function $\Theta(x)$ for later use:
\begin{breakbox}
\begin{verbatim}
########################
### ReLU function ###
########################
def ReLU(x):
        return np.maximum(0,x)
###################
### Smooth ReLU ###
###################
def smooth_ReLU(x):
        return np.log(1+np.exp(x))
###################
### Leaky ReLU ###
###################
def Leaky_ReLU(x,alpha):
        return np.maximum(0,x)-np.maximum(0,-x)*alpha
#####################
### step function ###
#####################
def step_function(x):
         if x <= 0:
	         return 0
         else:
	         return 1   
\end{verbatim}
\end{breakbox}        
        
%%%%%%%%%%%%
\subsubsection{Perceptron with arbitrary activation function}
%%%%%%%%%%%%

Now we have several activation functions. Let us define a perceptron with an arbitrary activation function:
\begin{breakbox}
\begin{verbatim}
#################################################################
### Multi-input perceptron with arbitrary activation function ###
#################################################################
# w and x are vectors. Use np.array like x = np.array([1,2,3])
# Number of element of x and w must be the same.
# x0 = 1
# f is the activation function
def perceptron(w,x,f):
        return f(sum(x*w))
\end{verbatim}
\end{breakbox}    
We can put all activation functions in \verb|activation_functions.py| so that we can call it as a module. (The line \verb|import numpy as np| is needed at the top of this file as well.) Then, we can use \verb|perceptron| like this:
\begin{breakbox}
\begin{verbatim}
import numpy as np
import activation_functions as act
w = np.array([2,3,4,5])
x = np.array([1,4,2,3])
print(sum(x*w))
print(perceptron(w,x,act.ReLU))
print(perceptron(w,x,act.step_function))
w = np.array([2,3,4,5])
x = np.array([1,1,-1,-1])
print(sum(x*w))
print(perceptron(w,x,act.ReLU))
print(perceptron(w,x,act.step_function))
\end{verbatim}
\end{breakbox}  
        
%%%%%%%%%%%%
\subsection{Softmax function}
%%%%%%%%%%%%
Softmax is often used as the activation function for the output layer. In the last step of Fig.~\ref{fig:DNN-cartoon}, suppose that we obtained $a_j = \vec{w}_j\cdot\vec{x}^{(M)}$ and then we want to convert $a$’s to $y$’s. The softmax function maps $a_1,\cdots,a_n$ to $y_1,\cdots,y_n$ as
\begin{align}
y_j
=
\textrm{softmax}(a_j)
\equiv
\frac{e^{a_j}}{\sum_{i=1}^ne^{a_i}}\,  .
\end{align}
We also use a vector notation $\textrm{softmax}(\vec{a})$, 
\begin{align}
\textrm{softmax}(\vec{a})
=
\left(
\begin{array}{c}
\textrm{softmax}(a_1)\\
\textrm{softmax}(a_2)\\
\vdots\\
\textrm{softmax}(a_n)
\end{array}
\right)\, . 
\end{align}
Note that, unlike the activation functions we have seen before, all $a_i$’s are used to obtain each $y_j$.

By construction, all $y$’s are positive,
\begin{align}
y_j>0\, , 
\end{align}
and the sum of $y$’s is 1:
\begin{align}
\sum_{j=1}^ny_j =1\, . 
\end{align}
Therefore, $y$’s can be regarded as probabilities. This is good for classification problems. For example, if we want to classify animal pictures, we can take $y_1 = y_{\rm dog}$, $y_2 = y_{\rm cat}$, etc.

When we implement softmax in our program, we should use a trivial identity,
\begin{align}
\textrm{softmax}(\vec{a}) = \textrm{softmax}(\vec{a}-c)
\end{align}
that holds for any constant $c$. By taking $c$ to be the largest element of $\vec{a}$, we can avoid overflow, which can happen when the exponential of a large number is taken. It is straightforward to implement it using Python:
\begin{breakbox}
\begin{verbatim}
def softmax(x):
        max_value = np.max(x)
        return np.exp(x-max_value)/np.sum(np.exp(x-max_value))
\end{verbatim}
\end{breakbox}          
%%%%%%%%%%%%
%%%%%%%%%%%%
\section{Building simple neural networks}
%%%%%%%%%%%%
%%%%%%%%%%%%

%%%%%%%%%%%%
%%%%%%%%%%%%
\subsection{Some linear algebra}
%%%%%%%%%%%%
%%%%%%%%%%%%
This section is very elementary. Readers who are comfortable with vectors and matrices can safely skip it, although it may still be useful to check the notation used in these notes.

Let us remind ourselves of some basic linear algebra and write the deep neural network shown in Fig.~\ref{fig:DNN-cartoon} in a systematic manner. Note that we do not really use powerful theorems from linear algebra; we use it to make the notation and implementation simpler. We do not use anything more than vectors and matrices, and their products. If you are not familiar with these rules, you should repeat the same computation a hundred times until it becomes muscle memory.

An $N$-component vector consists of $N$ numbers. In these notes, we assume all entries are real numbers. The set of all $N$-component vectors is expressed by $\mathbb{R}^N$. There are column vectors $\begin{pmatrix}v_1\\ v_2\\ \vdots\\ v_N\end{pmatrix}$ and row vectors $(v_1, v_2, \cdots , v_N)$. Mathematically, they are the same set of the same numbers. However, when we consider a product of matrices and vectors, the distinction between column vectors and row vectors is important. Unless otherwise stated, we use the standard symbol $\vec{v}$ for column vectors. To denote a row vector, we use the transpose, $\vec{v}^{\rm T}$. 

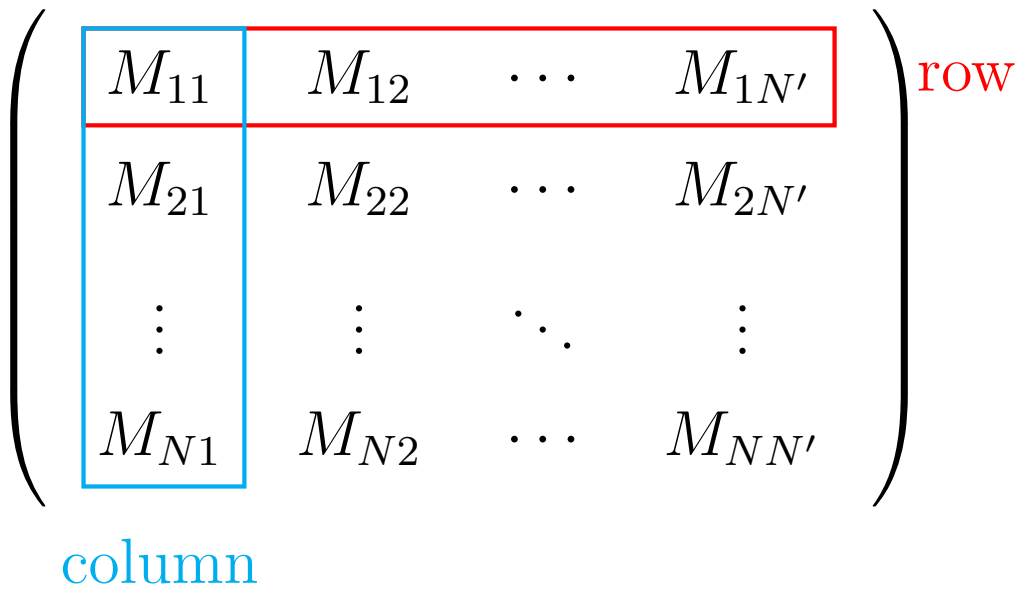
\begin{figure}[htbp]
\centering
\begin{tikzpicture}
\matrix (M) [matrix of math nodes, left delimiter=(, right delimiter=),
row sep=3pt, column sep=6pt, nodes={minimum width=9mm, minimum height=6mm}] {
M_{11} & M_{12} & \cdots & M_{1N'} \\
M_{21} & M_{22} & \cdots & M_{2N'} \\
\vdots & \vdots & \ddots & \vdots \\
M_{N1} & M_{N2} & \cdots & M_{NN'} \\
};
\draw[red, thick] (M-1-1.north west) rectangle (M-1-4.south east);
\node[red, right, xshift=4mm] at (M-1-4.east) {row};
\draw[cyan, thick] (M-1-1.north west) rectangle (M-4-1.south east);
\node[cyan, below, yshift=-2mm] at (M-4-1.south) {column};
\end{tikzpicture}
\caption{An $N\times N'$ matrix $M\in\mathbb{R}^{N\times N'}$ consists of $N$ rows and $N'$ columns.}
\label{fig:matrix}
\end{figure}

An $N \times N'$ matrix consists of $N$ rows and $N'$ columns. A row is horizontal, and a column is
vertical. See Fig.~\ref{fig:matrix}. In these notes, we assume all entries are real numbers. The set of all $N\times N'$ real matrices is sometimes expressed by $\mathbb{R}^{N\times N'}$. Note that $\mathbb{R}^{2\times 3}$, $\mathbb{R}^{3\times 2}$ and $\mathbb{R}^{6}$ are the same as vector spaces, but we use them to mean $2\times 3$ matrices, $3\times 2$ matrices, and 6-component vectors. An $N$-component column vector is an $N\times 1$ matrix, and an $N'$-component row vector is a $1\times N'$ matrix. Therefore, $\mathbb{R}^{N\times 1}$ means the set of $N$-component column vectors, while $\mathbb{R}^{1\times N}$ means the set of $N$-component row vectors. If we do not care about the difference between column vectors and row vectors, we simply use $\mathbb{R}^{N}$.

The transpose of an $N \times N'$ matrix $M$, denoted by $M^{\rm T}$, is an $N' \times N$ matrix that satisfies $(M^{\rm T})_{ij} = M_{ji}$.

The product of an $N \times N'$ matrix $M$ and an $N' \times N''$ matrix $M'$ is an $N \times N''$ matrix defined by
\begin{align}
(MM')_{ij}
=
\sum_{k=1}^{N'} M_{ik}M'_{kj}\, . 
\end{align}
Multiplication of a vector and a matrix is defined in the same way, regarding vectors as matrices with only one row or column:
\begin{align}
(M\vec{v})_i
=
\sum_j M_{ij}v_j\, , 
\qquad
(\vec{v}^{\rm T}M)_j
=
\sum_i v_iM_{ij}\, . 
\end{align}
$\vec{v}^{\rm T}\vec{w}$ is the same as the inner product $\vec{v}\cdot\vec{w}$:
\begin{align}
\vec{v}^{\rm T}\vec{w}
=
\sum_i v_iw_i
=
\vec{v}\cdot\vec{w}\, . 
\end{align}
However, $\vec{v}\vec{w}^{\rm T}$ is different. This is a matrix:
\begin{align}
(\vec{v}\vec{w}^{\rm T})_{ij}
=
v_iw_j\, . 
\end{align}
If you write two matrices $M$ and $M'$ as
$\begin{pmatrix}\vec{v}^{\rm T}_1\\ \vec{v}^{\rm T}_2\\ \vdots\\ \vec{v}^{\rm T}_{N}\end{pmatrix}$ and $(\vec{w}_1, \vec{w}_2, \cdots , \vec{w}_{N''})$, where $\vec{v}_i$ and $\vec{w}_j$ are $N'$-component vectors, then
\begin{align}
(MM')_{ij}
=
\vec{v}_i^{\rm T}\vec{w}_j
= 
\vec{v}_i\cdot\vec{w}_j\, . 
\end{align}
%%%%%%%%%%%%
%%%%%%%%%%%%
\subsection{Building neural networks}
%%%%%%%%%%%%
%%%%%%%%%%%%
We will not care about the efficiency of training for now. We just build a network for a given set of parameters.

The neural network depicted in Fig.~\ref{fig:DNN-cartoon} is defined as follows. There are $M+1$ layers labeled by $0, 1,\cdots, M$ and the output layer. 
The 0-th layer is the input. The $i$-th layer is associated with $d_i$ real numbers $x_1^{(i)},x_2^{(i)},\cdots,x_{d_i}^{(i)}$.\footnote{Please do not confuse the label of layers ${}^{(i)}$ with the label of samples; hopefully, the difference is clear from the context. We also use $\vec{x}_i$ for the same vector. Written in components, $x_j^{(i)}$ and $x_{i,j}$ mean the same number.}
To make the notation simple, let us write $\vec{y}$ as $\vec{x}^{(M+1)}$. We use $d_{M+1}$ to mean the number of nodes in the output layer, $n$.

For a given input $(x_1^{(0)},x_2^{(0)},\cdots,x_{d_0}^{(0)})$, $x_j^{(i)}$’s are determined by using
\begin{align}
x_j^{(i+1)}
=
f(a_j^{(i+1)})\, , 
\qquad
a_j^{(i+1)}
=
\sum_{k=1}^{d_i}
w_{jk}^{(i)}x_k^{(i)}+b_j^{(i)}\, . 
\end{align}
Here, $f$ is the activation function. (In principle, we can use different activation functions for different nodes.)

We introduce $x_0^{(i)} \equiv 1$ and $w_{j0}^{(i)} \equiv b_j^{(i)}$, and
\begin{align}
\vec{x}^{(i)}
\equiv
\begin{pmatrix}x_0^{(i)}\\ x_1^{(i)}\\ \vdots\\ x_{d_i}^{(i)}\end{pmatrix}
\equiv
\begin{pmatrix}1\\ x_1^{(i)}\\ \vdots\\ x_{d_i}^{(i)}\end{pmatrix}\, , 
\qquad
\vec{w}_j^{(i)}
\equiv
\begin{pmatrix}w_{j0}^{(i)}\\ w_{j1}^{(i)}\\ \vdots\\ w_{jd_i}^{(i)}\end{pmatrix}
\equiv
\begin{pmatrix}b_{j}^{(i)}\\ w_{j1}^{(i)}\\ \vdots\\ w_{jd_i}^{(i)}\end{pmatrix}\,  , 
\end{align}
so that
\begin{align}
a_j^{(i+1)}
=
\vec{w}_j^{(i)}\cdot \vec{x}^{(i)}
=
(\vec{w}_j^{(i)})^{\rm T} \vec{x}^{(i)}\, . 
\label{eq:a=Wx_1}
\end{align}
By introducing a $d_{i+1}\times(d_i + 1)$ matrix $W^{(i)}$ defined by 
\begin{align}
W^{(i)}
=
\begin{pmatrix}(\vec{w}_1^{(i)})^{\rm T}\\ (\vec{w}_2^{(i)})^{\rm T}\\ \vdots\\ (\vec{w}_{d_{i+1}}^{(i)})^{\rm T}\end{pmatrix}\, , 
\end{align}
we can write \eqref{eq:a=Wx_1} as 
\begin{align}
\vec{a}^{(i+1)}
=
W^{(i)}\vec{x}^{(i)}\, . 
\label{eq:a=Wx_2}
\end{align}
Applying the activation function, we obtain
\begin{align}
\vec{x}^{(i+1)}
=
f(\vec{a}^{(i+1)})
=
f(W^{(i)}\vec{x}^{(i)})\, . 
\end{align}
Therefore, we only need to repeat the multiplications of $W$ and applications of the activation function:
\begin{align}
\vec{x}^{(0)}
\xrightarrow[]{W^{(0)}}
\vec{a}^{(1)}
\xrightarrow[]{f}
\vec{x}^{(1)}
\xrightarrow[]{W^{(1)}}
\vec{a}^{(2)}
\xrightarrow[]{f}
\cdots
\vec{x}^{(M)}
\xrightarrow[]{W^{(M)}}
\vec{a}^{(M+1)}
\xrightarrow[]{f}
\vec{x}^{(M+1)}=\vec{y}\, .
\end{align}

%%%%%%%%%%%%
%%%%%%%%%%%%
\subsubsection*{Implementation via Python}
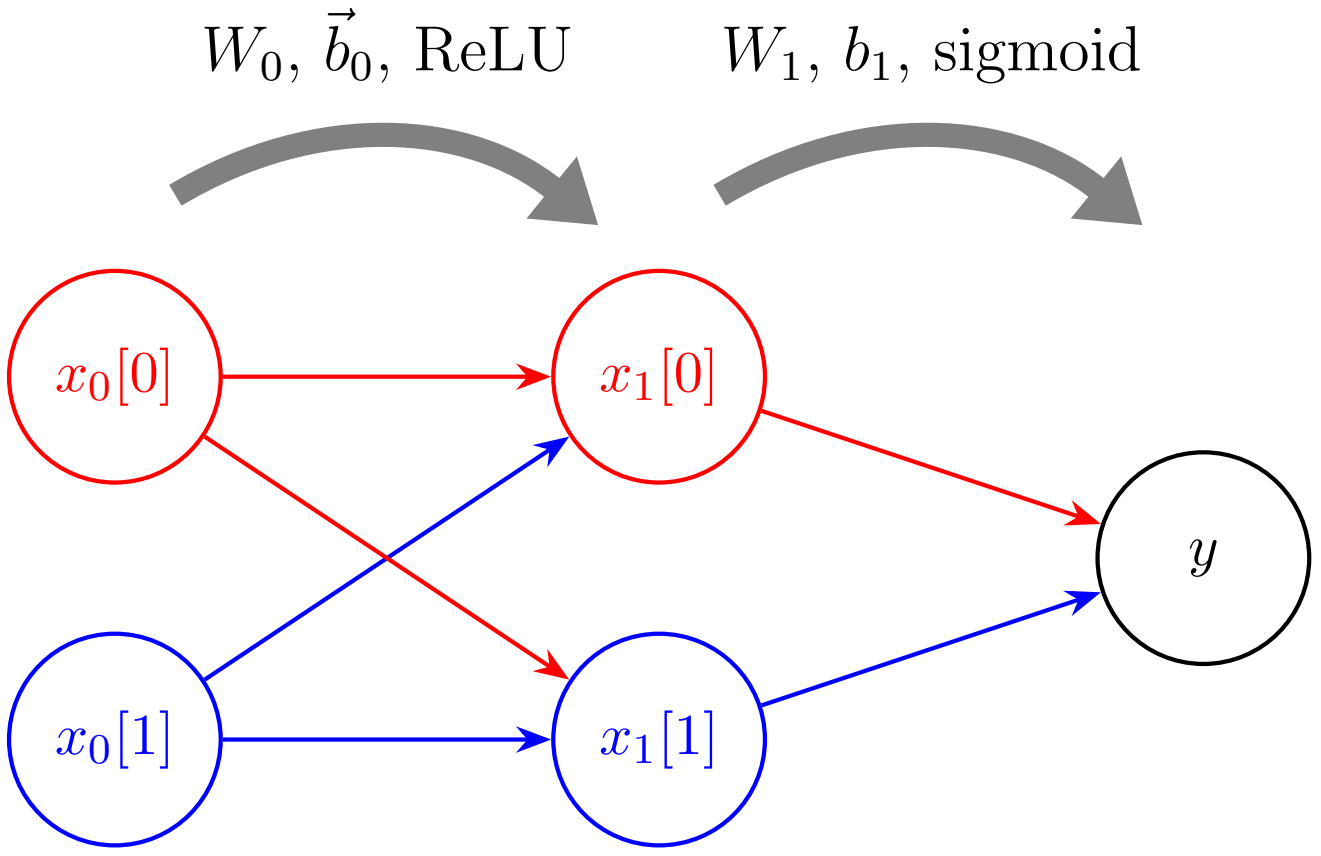
\begin{figure}[htbp]
\centering
\begin{tikzpicture}[>=Stealth,
  unit/.style={circle, draw, thick, minimum size=14mm, inner sep=0pt, font=\small}]
\node[unit, red]  (a) at (0,1.2)    {$x_0[0]$};
\node[unit, blue] (b) at (0,-1.2)   {$x_0[1]$};
\node[unit, red]  (c) at (3.6,1.2)  {$x_1[0]$};
\node[unit, blue] (d) at (3.6,-1.2) {$x_1[1]$};
\node[unit] (y) at (7.2,0) {$y$};
\foreach \t in {c,d} {
  \draw[->, thick, red]  (a) -- (\t);
  \draw[->, thick, blue] (b) -- (\t);
}
\draw[->, thick, red]  (c) -- (y);
\draw[->, thick, blue] (d) -- (y);
\draw[-{Triangle[length=4mm,width=5mm]}, line width=1.6mm, gray]
  (0.4,2.4) to[bend left=35] (3.2,2.2);
\node[above] at (1.8,3.0) {$W_0$, $\vec{b}_0$, ReLU};
\draw[-{Triangle[length=4mm,width=5mm]}, line width=1.6mm, gray]
  (4.0,2.4) to[bend left=35] (6.8,2.2);
\node[above] at (5.4,3.0) {$W_1$, $b_1$, sigmoid};
\end{tikzpicture}
\caption{A two-layer network that we use to reproduce the XOR gate.
Note that $x_\ell[0]$ and $x_\ell[1]$ correspond to $x^{(\ell)}_1$ and $x^{(\ell)}_2$ in the main text;
we used different notation to mimic the naming convention in the sample code in Python.}
\label{fig:XOR-as-NN}
\end{figure}

%%%%%%%%%%%%
%%%%%%%%%%%%
Let us train a very simple neural network by using a dataset consisting of four data points. We use the architecture shown in Fig.~\ref{fig:XOR-as-circuit} and create the XOR gate. (This is a silly but instructive example.) For convenience, we show the schematic picture again in Fig.~\ref{fig:XOR-as-NN}, slightly changing the notation.

The input is a two-component vector $\vec{x}_0$. In Python, the two components are \verb|x0[0]| and \verb|x0[1]|. As parameters, we use a $2\times 2$ weight matrix $W_0$, a two-component bias vector $\vec{b}_0$, a $1\times 2$ weight matrix $W_1$, and a one-component bias factor $b_1$. Specifically, we define the neural network as follows:
\begin{align}
&
\vec{x}_1
=
\textrm{ReLU}(\vec{a}_1)\, , 
\qquad
\vec{a}_1
=
W_0\vec{x}_0
+
\vec{b}_0\, , 
\nonumber\\
&
y
=
\textrm{Sigmoid}(a_2)\, , 
\qquad
a_2
=
W_1\vec{x}_1
+
b_1\, . 
\end{align}
We want to choose parameters so that $y_{\rm answer}=0,1,1,0$ follow from $\vec{x}_0=\begin{pmatrix}0\\ 0\end{pmatrix}$, 
$\begin{pmatrix}1\\ 0\end{pmatrix}$, 
$\begin{pmatrix}0\\ 1\end{pmatrix}$, and 
$\begin{pmatrix}1\\ 1\end{pmatrix}$. 
Therefore, we choose the cost function as
\begin{align}
J(W_0,W_1,\vec{b}_0,b_1)
=
\frac{1}{2}
\sum_{\vec{x}_0=(0,0),(1,0),(0,1),(1,1)}
\left(
f(\vec{x}_0; W_0,W_1,\vec{b}_0,b_1)
-
y_{\rm answer}(\vec{x}_0)
\right)^2\, . 
\end{align}
For training, we take the most naive and inefficient way: calculate all derivatives by using a naive numerical differentiation method. Later, we will improve our architecture by using the backpropagation method.

Let’s see some sample code. We skip the implementation of ReLU and sigmoid because we have already seen them before. The two-layer network can be defined in a straightforward manner:

\begin{breakbox}
\begin{verbatim}
##########################
# 2-layer neural network #
##########################
def Neural_Network(x0,w0,w1,b0,b1):
        a1 = np.dot(w0,x0) + b0
        x1 = ReLU(a1)
        a2 = np.dot(w1,x1) + b1
        # a2 has only 1 component
        # use a2[0] instead of a2, so that y becomes a scalar
        #  instead of 1-component ndarray.
        y = sigmoid(a2[0])
        return y
\end{verbatim}
\end{breakbox}   
Note that we do not have to specify the dimensions of vectors and matrices. In fact, we do not even have to specify the dimension of the output $y$, although, in this code, we specified that $y$ is a scalar.

Next, we define the cost function. Because we have only four data points, let’s write all possibilities explicitly:
\begin{breakbox}
\begin{verbatim}
###################
## Cost function ##
###################
def cost_function(w0,w1,b0,b1):
        cost=0.0
        x0 = np.array([0,0])
        y_answer = 0
        cost += 0.5*(Neural_Network(x0,w0,w1,b0,b1)-y_answer)**2.0
        x0 = np.array([1,0])
        y_answer = 1
        cost += 0.5*(Neural_Network(x0,w0,w1,b0,b1)-y_answer)**2.0
        x0 = np.array([0,1])
        y_answer = 1
        cost += 0.5*(Neural_Network(x0,w0,w1,b0,b1)-y_answer)**2.0
        x0 = np.array([1,1])
        y_answer = 0
        cost += 0.5*(Neural_Network(x0,w0,w1,b0,b1)-y_answer)**2.0
        
        return cost
\end{verbatim}
\end{breakbox}   

The derivative can be calculated by shifting each parameter by a small amount (i.e., vanilla numerical differentiation). Note that such a method is inefficient when the neural network is large. But we don’t care about efficiency for now. We can code the numerical differentiation straightforwardly:
\begin{breakbox}
\begin{verbatim}
###############################
## Derivative w.r.t. w and b ##
###############################
def cost_function_derivatives(w0,w1,b0,b1,h):
        dw0 = np.zeros((2,2))
        dw1 = np.zeros((1,2))
        db0 = np.zeros((2))
        db1 = np.zeros((1))
        temp0 = cost_function(w0,w1,b0,b1)
        for i in range(2):
                b0[i] += h
                temp = cost_function(w0,w1,b0,b1) - temp0
                b0[i] -= h
                db0[i] += temp/h
        b1[0] += h
        temp = cost_function(w0,w1,b0,b1) - temp0
        b1[0] -= h
        db1[0] = temp/h
        for i in range(2):
                for j in range(2):
                        w0[i,j] += h
                        temp = cost_function(w0,w1,b0,b1) - temp0
                        w0[i,j] -= h
                        dw0[i,j] = temp/h
                w1[0,i] += h
                temp = cost_function(w0,w1,b0,b1) - temp0
                w1[0,i] -= h
                dw1[0,i] = temp/h
        return dw0,dw1,db0,db1
\end{verbatim}
\end{breakbox}
Gradient descent is also straightforward:
\begin{breakbox}
\begin{verbatim}
######################
## Gradient descent ##
######################
def gradient_descent(w0,w1,b0,b1,num_step,lr):
        h=0.0001 # step size for taking derivative numerically
        for istep in range(num_step):
                dw0,dw1,db0,db1 = cost_function_derivatives(w0,w1,b0,b1,h)
                b0 = b0 - lr * db0
                b1 = b1 - lr * db1
                w0 = w0 - lr * dw0
                w1 = w1 - lr * dw1
                print(istep,cost_function(w0,w1,b0,b1))
        return w0,w1,b0,b1
\end{verbatim}
\end{breakbox}
Now we have all the routines needed: the neural network, the cost function, the derivative of the cost function, and gradient descent. The training can be conducted by calling only the routine for gradient descent. Let us take the initial values of the parameters to be Gaussian random numbers, and take 100K steps with learning rate 0.05:
\begin{breakbox}
\begin{verbatim}
w0 = np.random.randn(2,2)
w1 = np.random.randn(1,2)
b0 = np.random.randn(2)
b1 = np.random.randn(1)
num_step=100000
lr=0.05
w0,w1,b0,b1 = gradient_descent(w0,w1,b0,b1,num_step,lr)
\end{verbatim}
\end{breakbox}
If you actually try it, you will see that the cost function does not always converge to zero. We can get a good result once in 3 or 4 times. 
%%%%%%%%%%%%
%%%%%%%%%%%%
\subsection{Cross-entropy loss}\label{sec:cross_entropy}
%%%%%%%%%%%%
%%%%%%%%%%%%

%%%%%%%%%%%%
%%%%%%%%%%%%
\subsubsection{Cross-entropy loss}
%%%%%%%%%%%%
%%%%%%%%%%%%
Suppose we know the target distribution $p(\vec{x})$ for $\vec{x}\in\mathbb{R}^d$, and the distribution we could realize as an output of a neural network is $q(\vec{x})$. Then, the cross-entropy is given by
\begin{align}
H(p,q)
=
-\int \mathrm{d}\vec{x}\, p(\vec{x})\log q(\vec{x})\, . 
\end{align}
This is minimized when $q(\vec{x})=p(\vec{x})$. To see it, let us change $q(\vec{x})$ a little bit, as 
$q(\vec{x})\to q'(\vec{x}) =q(\vec{x})+\Delta q(\vec{x})$. 
Then, 
\begin{align}
\Delta H(p, q)
=
-\int \mathrm{d}\vec{x}\, p(\vec{x})\Delta\log q(\vec{x})
=
-\int \mathrm{d}\vec{x}\, p(\vec{x})\cdot\frac{\Delta q(\vec{x})}{q(\vec{x})}\, . 
\end{align}
If $q(\vec{x})=p(\vec{x})$, then
\begin{align}
-\int \mathrm{d}\vec{x}\, p(\vec{x})\cdot\frac{\Delta q(\vec{x})}{q(\vec{x})}
=
-\int \mathrm{d}\vec{x}\, \Delta q(\vec{x})
=
0\, . 
\end{align}
Here, we used $\int\mathrm{d}\vec{x}\, q(\vec{x})=\int\mathrm{d}\vec{x}\, q'(\vec{x})=1$. Therefore, $q(\vec{x})=p(\vec{x})$ extremizes the cross-entropy loss, at least. To see that this is actually a minimum, let us look at the higher-order terms in the expansion around $q(\vec{x})=p(\vec{x})$. We take $q'(\vec{x})=p(\vec{x})+\Delta q(\vec{x})$, and expand $H$ in terms of $\Delta q$. Because
\begin{align}
\log q'
=
\log p + \log\left(1+\frac{\Delta q}{p}\right)
=
\log p 
+
\frac{\Delta q}{p}
-
\frac{1}{2}\left(\frac{\Delta q}{p}\right)^2
+
\cdots\, , 
\end{align}
we have 
\begin{align}
H(p,q'=p+\Delta q)
&=
\int \mathrm{d}\vec{x}
\left(
-p
\log p 
-
\Delta q
+
\frac{(\Delta q)^2}{2p}
+
\cdots
\right)
\nonumber\\
&=
\int \mathrm{d}\vec{x}
\left(
-p
\log p 
+
\frac{(\Delta q)^2}{2p}
+
\cdots
\right)\, . 
\end{align}
Therefore, a small change from $q=p$ increases the cross-entropy. 

Strictly speaking, we only proved that $q=p$ is at least a local minimum. To show that this is the global minimum, we can use Jensen's inequality
\begin{align}
f\left(
\int\mathrm{d}\vec{x}\, p(\vec{x})g(\vec{x})
\right)
\ge 
\int\mathrm{d}\vec{x}\, p(\vec{x})f\left(g(\vec{x})\right)
\end{align}
that holds for any function $f$ which is concave (such as $\log$).
Equivalently, by using $\mathbb{E}_p$ to mean the expectation value with respect to the probability distribution $p$, 
\begin{align}
f\left(
\mathbb{E}_p[g(\vec{x})]
\right)
\ge 
\mathbb{E}_p[f\left(g(\vec{x})\right)]\, . 
\end{align}
We take $g=q/p$ and $f=\log$. Then, 
\begin{align}
\log
\left(
\int\mathrm{d}\vec{x}\, p(\vec{x})\cdot\frac{q(\vec{x})}{p(\vec{x})}
\right)
=
\log
\left(
\int\mathrm{d}\vec{x}\, q(\vec{x})
\right)
=
\log 1
=0\, , 
\end{align}
while 
\begin{align}
\int\mathrm{d}\vec{x}\, p(\vec{x})\log(q(\vec{x})/p(\vec{x}))
=
\int\mathrm{d}\vec{x}\, p(\vec{x})\log q(\vec{x})
-
\int\mathrm{d}\vec{x}\, p(\vec{x})\log p(\vec{x})\, ,
\end{align}
and hence, we obtain
\begin{align}
0
\ge 
\int\mathrm{d}\vec{x}\, p(\vec{x})\log q(\vec{x})
-
\int\mathrm{d}\vec{x}\, p(\vec{x})\log p(\vec{x})\, . 
\end{align}
This means $H(p,q)\ge H(p,p)$, i.e., $H(p,q)$ is minimized at $q=p$.

The difference $H(p,q)-H(p,p)=\int\mathrm{d}\vec{x}\, p(\vec{x})\log\frac{p(\vec{x})}{q(\vec{x})}\ge 0$ is called the \textit{Kullback--Leibler (KL) divergence}. It is often used as a measure of the difference between two probability distributions.
%%%%%%%%%%%%
%%%%%%%%%%%%
\subsubsection{Cross-entropy loss for supervised learning}
%%%%%%%%%%%%
%%%%%%%%%%%%
Let us see how the cross-entropy loss can be used for supervised learning. Specifically, we consider the classification of pictures of animals. We have inputs $\vec{x}^{(k)}$ ($k = 1, 2, \cdots, K$), and the label $\vec{y}^{(k)}= (y_{\rm cat}^{(k)}, y_{\rm dog}^{(k)} , \cdots )$ gives the probability that the picture is of a cat, a dog, etc. (Usually, one of the elements of $\vec{y}^{(k)}$ is 1 and the others are 0 (i.e., $\vec{y}^{(k)}$ is a one-hot vector), but we can imagine a more generic situation in which a picture could be either a dog or a fox, and hence $y_{\rm dog}^{(k)}=y_{\rm fox}^{(k)}=0.5$.) Equivalently, the label is the conditional probability $y_{\rm cat}^{(k)} = p({\rm cat}|\vec{x}^{(k)})$, $y_{\rm dog}^{(k)} = p({\rm dog}|\vec{x}^{(k)})$, etc.

Let $q_{\vec{\theta}}({\rm animal}|\vec{x})$ be the output of the neural network. We tune $\vec{\theta}$ so that $q_{\vec{\theta}}$ becomes closer to $p$. Therefore, we can use the cross-entropy loss defined by
\begin{align}
J(\vec{\theta})
&=
-
\sum_{k=1}^K
\sum_{{\rm animal}\, =\, {\rm cat},{\rm dog},\cdots}
p({\rm animal}|\vec{x}^{(k)})
\log q_{\vec{\theta}}({\rm animal}|\vec{x}^{(k)})
\nonumber\\
&=
-
\sum_{k=1}^K
\sum_{{\rm animal}\, =\, {\rm cat},{\rm dog},\cdots}
y_{\rm animal}^{(k)}
\log q_{\vec{\theta}}({\rm animal}|\vec{x}^{(k)})
\nonumber\\
&\equiv
-
\sum_{k=1}^K
\vec{y}^{(k)}\cdot
\log \vec{q}_{\vec{\theta}}(\vec{x}^{(k)})\, . 
\end{align}
Here, 
\begin{align}
\vec{q}_{\vec{\theta}}(\vec{x}^{(k)})
\equiv
\left(
\begin{array}{c}
q_{\vec{\theta}}({\rm cat}|\vec{x}^{(k)})\\
q_{\vec{\theta}}({\rm dog}|\vec{x}^{(k)})\\
\vdots
\end{array}
\right)\, ,
\qquad
\log\vec{q}_{\vec{\theta}}(\vec{x}^{(k)})
\equiv
\left(
\begin{array}{c}
\log q_{\vec{\theta}}({\rm cat}|\vec{x}^{(k)})\\
\log q_{\vec{\theta}}({\rm dog}|\vec{x}^{(k)})\\
\vdots
\end{array}
\right)\, .
\end{align}

Let us consider the classification of only cats and dogs, and suppose that all labels are one-hot vectors, i.e., $(y_{\rm cat}^{(k)},y_{\rm dog}^{(k)})=(1, 0)$ or $(0, 1)$. Because $0\le q({\rm animal}|\vec{x}^{(k)})\le 1$ and $\log q_{\vec{\theta}}({\rm animal}|\vec{x}^{(k)})\le 0$, the minimum value of the cost function is zero, which is realized if $q_{\vec{\theta}}({\rm animal}|\vec{x}^{(k)})=y_{\rm animal}^{(k)}$ for both ${\rm animal}={\rm cat}$ and ${\rm animal}={\rm dog}$.

Note that the cross-entropy loss function can be used only when the output of the neural network is a probability distribution. On the other hand, the square loss function works for generic targets. 
%%%%%%%%%%%%
%%%%%%%%%%%%
\subsubsection*{Avoiding $\log 0$}
%%%%%%%%%%%%
%%%%%%%%%%%%
In principle, $q_{\vec{\theta}}({\rm animal}|\vec{x}^{(k)})$ can be zero. But then $\log q_{\vec{\theta}}({\rm animal}|\vec{x}^{(k)})=\log 0$, which is not well defined. To avoid it, we can use $\log(q_{\vec{\theta}}+\epsilon)$ instead of $\log q_{\vec{\theta}}({\rm animal}|\vec{x}^{(k)})$, where $\epsilon$ is a small positive number.

%%%%%%%%%%%%
%%%%%%%%%%%%
\subsubsection{Neural network with cross-entropy loss}
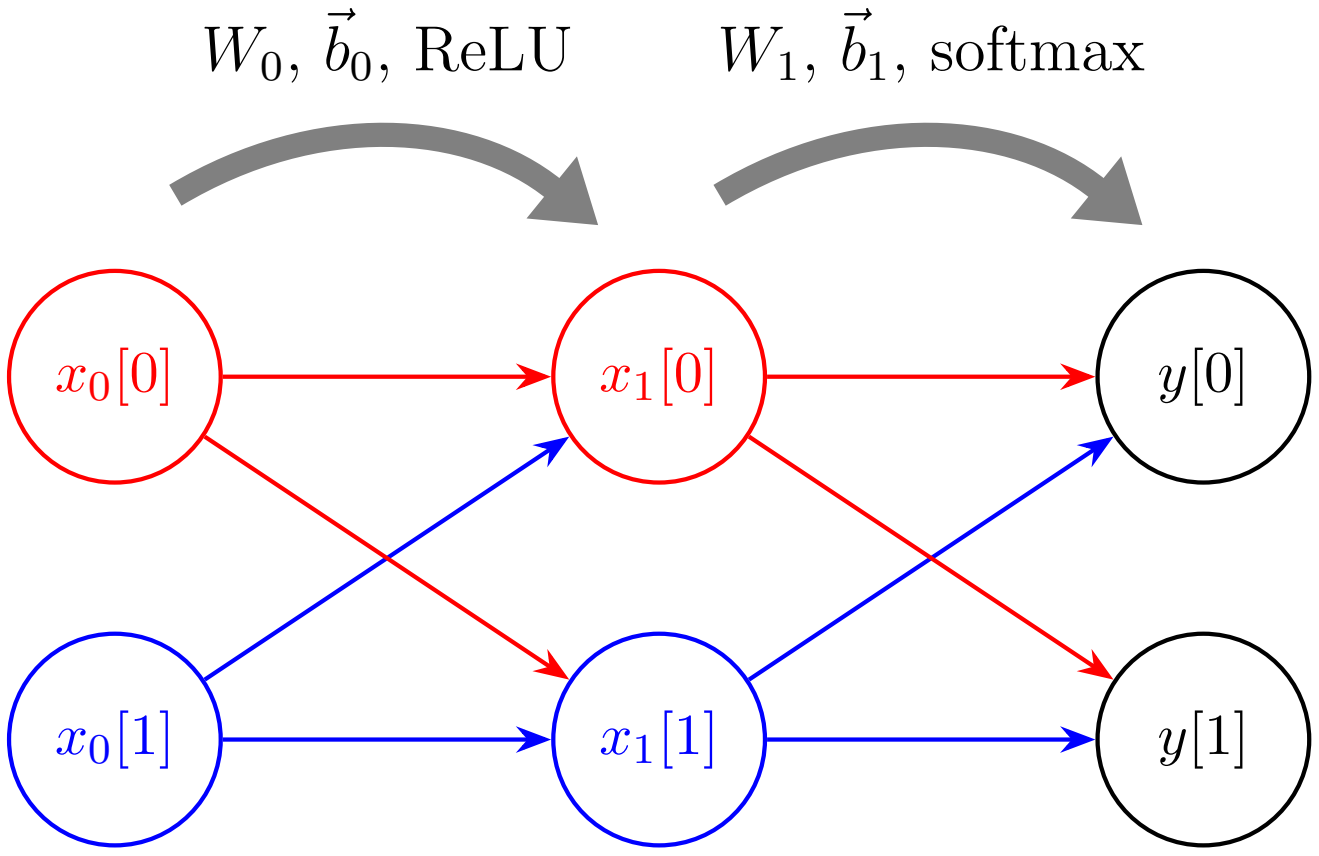
\begin{figure}[htbp]
\centering
\begin{tikzpicture}[>=Stealth,
  unit/.style={circle, draw, thick, minimum size=14mm, inner sep=0pt, font=\small}]
\node[unit, red]  (a) at (0,1.2)    {$x_0[0]$};
\node[unit, blue] (b) at (0,-1.2)   {$x_0[1]$};
\node[unit, red]  (c) at (3.6,1.2)  {$x_1[0]$};
\node[unit, blue] (d) at (3.6,-1.2) {$x_1[1]$};
\node[unit] (e) at (7.2,1.2)  {$y[0]$};
\node[unit] (f) at (7.2,-1.2) {$y[1]$};
\foreach \t in {c,d} {
  \draw[->, thick, red]  (a) -- (\t);
  \draw[->, thick, blue] (b) -- (\t);
}
\foreach \t in {e,f} {
  \draw[->, thick, red]  (c) -- (\t);
  \draw[->, thick, blue] (d) -- (\t);
}
\draw[-{Triangle[length=4mm,width=5mm]}, line width=1.6mm, gray]
  (0.4,2.4) to[bend left=35] (3.2,2.2);
\node[above] at (1.8,3.0) {$W_0$, $\vec{b}_0$, ReLU};
\draw[-{Triangle[length=4mm,width=5mm]}, line width=1.6mm, gray]
  (4.0,2.4) to[bend left=35] (6.8,2.2);
\node[above] at (5.4,3.0) {$W_1$, $\vec{b}_1$, softmax};
\end{tikzpicture}
\caption{A two-layer network that we use to reproduce the XOR gate, by using softmax for the output and the cross-entropy loss as the cost function.}
\label{fig:XOR-as-NN-softmax}
\end{figure}

%%%%%%%%%%%%
%%%%%%%%%%%%
Let us build the XOR gate by using the cross-entropy loss. Now we need two outputs; previously we had $y_{\rm answer} = 0$ and 1. We imagine that $y_{\rm answer} = 0$ is a dog and $y_{\rm answer} = 1$ is a cat, and assign $\vec{y}_{\rm answer} = \begin{pmatrix} 1 \\ 0\end{pmatrix}$ and $\vec{y}_{\rm answer} = \begin{pmatrix} 0 \\ 1\end{pmatrix}$ to them. Hence, we change the
architecture to the one shown in Fig.~\ref{fig:XOR-as-NN-softmax}. Written explicitly, we have:
\begin{align}
&
\vec{x}_1
=
\textrm{ReLU}(\vec{a}_1)\, , 
\qquad
\vec{a}_1
=
W_0\vec{x}_0
+
\vec{b}_0\, , 
\nonumber\\
&
\vec{y}
=
\textrm{Softmax}(\vec{a}_2)\, , 
\qquad
\vec{a}_2
=
W_1\vec{x}_1
+
\vec{b}_1\, . 
\end{align}

The cost function is
\begin{align}
J(W_0,W_1,\vec{b}_0,\vec{b}_1)
=
-
\sum_{\vec{x}_0=(0,0),(1,0),(0,1),(1,1)}
\vec{y}_{\rm answer}(\vec{x}_0)
\cdot
\log \vec{y}(\vec{x}_0; W_0,W_1,\vec{b}_0,\vec{b}_1)\, . 
\end{align}
The neural network is modified as
\begin{breakbox}
\begin{verbatim}
##########################
# 2-layer neural network #
##########################
def Neural_Network(x0,w0,w1,b0,b1):
        a1 = np.dot(w0,x0) + b0
        x1 = ReLU(a1)
        a2 = np.dot(w1,x1) + b1
        y = softmax(a2)
        
        return y
\end{verbatim}
\end{breakbox}
and the cost function is changed as follows:
\begin{breakbox}
\begin{verbatim}
###################
## Cost function ##
###################
def cost_function(w0,w1,b0,b1):
        cost=0.0
        x0 = np.array([0,0])
        #y_answer = 0
       	y_answer=np.array([1.0,0.0])
        cost -= np.sum(np.log(Neural_Network(x0,w0,w1,b0,b1))*y_answer)
        
       	x0 = np.array([1,0])
        #y_answer = 1
        y_answer=np.array([0.0,1.0])
        cost -= np.sum(np.log(Neural_Network(x0,w0,w1,b0,b1))*y_answer)

        x0 = np.array([0,1])
        #y_answer = 1
        y_answer=np.array([0.0,1.0])
        cost -= np.sum(np.log(Neural_Network(x0,w0,w1,b0,b1))*y_answer)

        x0 = np.array([1,1])
        #y_answer = 0
        y_answer=np.array([1.0,0.0])
        cost -= np.sum(np.log(Neural_Network(x0,w0,w1,b0,b1))*y_answer)

        return cost
\end{verbatim}
\end{breakbox}
Other parts of the code must be rewritten slightly, to accommodate the change in the size of $W_1$ and $\vec{b}_1$.

%%%%%%%%%%%%
%%%%%%%%%%%%
\section{Training deep networks}\label{sec:training}
%%%%%%%%%%%%
%%%%%%%%%%%%
So far, we have built neural networks and trained a very small one in a naive way. To make larger networks work in practice, we need to answer three questions. (i) How can we compute the gradient efficiently, and how can we keep it from vanishing as the network becomes deep? (ii) How can we make sure that a network that fits the training data also works for unseen data? (iii) How should we measure the performance, and how can a trained network fail? This section and the next two address these questions in turn.

%%%%%%%%%%%%
%%%%%%%%%%%%
\subsection{Backpropagation}
%%%%%%%%%%%%
%%%%%%%%%%%%

%%%%%%%%%%%%
%%%%%%%%%%%%
\subsubsection*{Chain rule}
%%%%%%%%%%%%
%%%%%%%%%%%%
You can understand backpropagation if you know the chain rule. You might have forgotten it, or you might be a bit too shy to say you don’t really understand the chain rule, so let’s recall what the chain rule was. Let $z=g(y)$, $y=f(x)$, and $z=h(x)\equiv g(f(x))$. Often, we simply write it as $z(x)=z(y(x))$ etc., if there is no risk of confusion. The derivative of $h(x)$ with respect to $x$ is obtained by using the chain rule,
\begin{align}
\frac{\mathrm{d}h}{\mathrm{d}x}
=
\left.
\frac{\mathrm{d}g}{\mathrm{d}y}
\right|_{y=f(x)}
\cdot
\frac{\mathrm{d}f}{\mathrm{d}x}\, . 
\end{align}
Equivalently,
\begin{align}
\frac{\mathrm{d}z}{\mathrm{d}x}
=
\frac{\mathrm{d}z}{\mathrm{d}y}
\cdot
\frac{\mathrm{d}y}{\mathrm{d}x}\, . 
\end{align}
This relation holds almost by definition. Roughly speaking, $\Delta z = \Delta (g(y)) =  \frac{\mathrm{d}g}{\mathrm{d}y}\Delta y + O((\Delta y)^2)$ is the definition of the derivative $\frac{\mathrm{d}g}{\mathrm{d}y}$, $\Delta y = \Delta (f(x)) =  \frac{\mathrm{d}f}{\mathrm{d}x}\Delta x + O((\Delta x)^2)$ is the definition of $\frac{\mathrm{d}f}{\mathrm{d}x}$, and $\Delta z = \Delta (h(x)) =  \frac{\mathrm{d}h}{\mathrm{d}x}\Delta x + O((\Delta x)^2)$ is the definition of the derivative $\frac{\mathrm{d}h}{\mathrm{d}x}$. 

If there are multiple variables and $z_i=g_i(y_1,y_2,\cdots)$, $y_j=f_j(x_1,x_2,\cdots)$, and $z_i=h_i(x_1,x_2,\cdots)\equiv g_i(f_1(x_1,x_2,\cdots),f_2(x_1,x_2,\cdots),\cdots)$, then
\begin{align}
\frac{\partial h_i}{\partial x_k}
=
\sum_j
\left.
\frac{\partial g_i}{\partial y_j}
\right|_{y_j=f_j(x_1,x_2,\cdots)}
\cdot
\frac{\partial f_j}{\partial x_k}\, . 
\end{align}
Equivalently,
\begin{align}
\frac{\partial z_i}{\partial x_k}
=
\sum_j
\frac{\partial z_i}{\partial y_j}
\cdot
\frac{\partial y_j}{\partial x_k}\, . 
\end{align}
Again, this is almost by definition, because $\Delta z_i = \Delta(g_i(y_1, y_2,\cdots))=\sum_j\frac{\partial z_i}{\partial y_j}\Delta y_j + O((\Delta y)^2)$
is the definition of the partial derivative $\frac{\partial z_i}{\partial y_j}$, etc.
%%%%%%%%%%%%
%%%%%%%%%%%%
\subsubsection*{Backpropagation}
%%%%%%%%%%%%
%%%%%%%%%%%%

In the neural network depicted in Fig.~\ref{fig:DNN-cartoon}, we have $x_j^{(\ell)} =x_j^{(\ell)}(x_1^{(\ell-1)},x_2^{(\ell-1)},\cdots;W^{(\ell-1)})$
($\ell=1,2,\cdots,M$), and $y_j =y_j(x_1^{(M)},x_2^{(M)},\cdots;W^{(M)})$. To apply gradient descent, we need to calculate $\frac{\partial y_j}{\partial W^{(\ell)}}$ for $\ell=0,1,2,\cdots,M$. Backpropagation is an efficient way to do this computation.

By using the chain rule, we can compute $\frac{\partial y_j}{\partial W^{(\ell)}}$ as follows. 
The starting point is the last layer. Since $y_j$ is an explicit function of $\vec{x}^{(M)}$ and $W^{(M)}$, we can directly compute
\begin{align}
\frac{\partial y_j}{\partial W^{(M)}}
\qquad\text{and}\qquad
\frac{\partial y_j}{\partial x^{(M)}_{i_M}}\, .
\end{align}
Starting from these, we go backwards, one layer at a time:\begin{align}
\frac{\partial y_j}{\partial W^{(M-1)}}
=
\sum_{i_M}
\frac{\partial x^{(M)}_{i_M}}{\partial W^{(M-1)}}
\cdot
\frac{\partial y_j}{\partial x^{(M)}_{i_M}}\, , 
\qquad
\frac{\partial y_j}{\partial x^{(M-1)}_{i_{M-1}}}
=
\sum_{i_M}
\frac{\partial x^{(M)}_{i_M}}{\partial x^{(M-1)}_{i_{M-1}}}
\cdot
\frac{\partial y_j}{\partial x^{(M)}_{i_M}}\, .
\end{align}
Note the structure of these two relations: both use the same quantity $\frac{\partial y_j}{\partial x^{(M)}_{i_M}}$, which we have already obtained. The left one gives the derivative with respect to the weights of that layer, which is what we actually want, while the right one gives the quantity needed to repeat the same step at the next layer down. Indeed, the next step is:
\begin{align}
\frac{\partial y_j}{\partial W^{(M-2)}}
=
\sum_{i_{M-1}}
\frac{\partial x^{(M-1)}_{i_{M-1}}}{\partial W^{(M-2)}}
\cdot
\frac{\partial y_j}{\partial x^{(M-1)}_{i_{M-1}}}\, , 
\qquad
\frac{\partial y_j}{\partial x^{(M-2)}_{i_{M-2}}}
=
\sum_{i_{M-1}}
\frac{\partial x^{(M-1)}_{i_{M-1}}}{\partial x^{(M-2)}_{i_{M-2}}}
\cdot
\frac{\partial y_j}{\partial x^{(M-1)}_{i_{M-1}}}\, .
\end{align}
Both use the same quantity $\frac{\partial y_j}{\partial x^{(M-1)}_{i_{M-1}}}$, which we have already obtained.
We repeat the same step, changing $\ell$ below as $\ell=M-2,\cdots,1$:
\begin{align}
\frac{\partial y_j}{\partial W^{(\ell-1)}}
=
\sum_{i_{\ell}}
\frac{\partial x^{(\ell)}_{i_{\ell}}}{\partial W^{(\ell-1)}}
\cdot
\frac{\partial y_j}{\partial x^{(\ell)}_{i_{\ell}}}\, , 
\qquad
\frac{\partial y_j}{\partial x^{(\ell-1)}_{i_{\ell-1}}}
=
\sum_{i_{\ell}}
\frac{\partial x^{(\ell)}_{i_{\ell}}}{\partial x^{(\ell-1)}_{i_{\ell-1}}}
\cdot
\frac{\partial y_j}{\partial x^{(\ell)}_{i_{\ell}}}\, . 
\end{align}

The method used to compute the derivatives through these relationships is called the \textit{backpropagation method}~\cite{Rumelhart1986}. This is because the computation begins at the final layer, and the results from deeper layers are gradually `back-propagated' towards the input layer. The computational cost scales as $\sum_\ell d_{\ell-1}d_{\ell}d_{M+1}$. 

It is important to observe that the forward propagation is needed only once. On the other hand, if we use a naive numerical differentiation for each derivative, a forward propagation is needed every time. For a network with $10^6$ parameters, a naive numerical differentiation would require about $10^6$ forward propagations, while backpropagation needs only one forward and one backward sweep. Therefore, by using the backpropagation method, the computational cost can be reduced significantly.

In practice, we need the derivatives of the cost function $J$, which is a single number, rather than those of each $y_j$. Then, we can start with $\frac{\partial J}{\partial y_j}$ and propagate $\frac{\partial J}{\partial x^{(\ell)}_{i_\ell}}$, which has only one index, in the same manner. The computational cost of the backward sweep is then $\sum_\ell d_{\ell-1}d_{\ell}$, which is of the same order as that of the forward propagation.
%%%%%%%%%%%%
%%%%%%%%%%%%
\subsubsection*{Analytic expressions}
%%%%%%%%%%%%
%%%%%%%%%%%%
Thanks to the chain rule, we can decompose the problem into small steps. Essentially, we only have to take the derivative of $x^{(\ell+1)}_{i_{\ell+1}}=\sigma(\sum_{i_\ell}W^{(\ell)}_{i_{\ell+1}i_\ell}x^{(\ell)}_{i_\ell})$. Again using the chain rule,
we obtain

\begin{align}
\frac{\partial x^{(\ell+1)}_{i_{\ell+1}}}{\partial W^{(\ell)}_{i_{\ell+1}i_\ell}}
&=
\sigma'\left(
\sum_{i'_\ell}W^{(\ell)}_{i_{\ell+1}i'_\ell}x^{(\ell)}_{i'_\ell}
\right)
\cdot
x^{(\ell)}_{i_\ell}\, , 
\nonumber\\
\frac{\partial x^{(\ell+1)}_{i_{\ell+1}}}{\partial x^{(\ell)}_{i_\ell}}
&=
\sigma'\left(
\sum_{i'_\ell}W^{(\ell)}_{i_{\ell+1}i'_\ell}x^{(\ell)}_{i'_\ell}
\right)
\cdot
W^{(\ell)}_{i_{\ell+1}i_\ell}\, . 
\end{align}
Note that $\frac{\partial x^{(\ell+1)}_i}{\partial W^{(\ell)}_{jk}}=0$ for $i\neq j$. 

Furthermore, the derivative of the activation function $\sigma$ can be calculated analytically unless you use an unnecessarily complicated function. Therefore, we can easily have analytic expressions. (More precisely, we get products of matrices, and each matrix element can be written in a simple analytic form.)
%%%%%%%%%%%%
%%%%%%%%%%%%
\subsubsection*{Implementation via Python: two-layer network}
%%%%%%%%%%%%
%%%%%%%%%%%%
Let us use the two-layer architecture shown in Fig.~\ref{fig:XOR-as-NN-softmax}. We generalize it slightly and allow the number of nodes in each layer to be $d_0$, $d_1$, and $n$. (In Fig.~\ref{fig:XOR-as-NN-softmax}, $d_0 =d_1 =n=2$.) The forward propagation was calculated as follows:
\begin{align}
&
\vec{x}^{(1)}
=
\textrm{ReLU}(\vec{a}^{(1)})\, , 
\qquad
\vec{a}^{(1)}
=
W^{(0)}\vec{x}^{(0)}
+
\vec{b}^{(0)}\, , 
\nonumber\\
&
\vec{y}
=
\textrm{Softmax}(\vec{a}^{(2)})\, , 
\qquad
\vec{a}^{(2)}
=
W^{(1)}\vec{x}^{(1)}
+
\vec{b}^{(1)}\, . 
\end{align}
If we write the indices of vectors and matrices explicitly, we have
\begin{align}
&
x^{(1)}_i
=
\textrm{ReLU}(a^{(1)}_i)\, , 
\qquad
a^{(1)}_i
=
\sum_{j=1}^{d_0}
W^{(0)}_{ij}x^{(0)}_j
+
b^{(0)}_i\, , 
\nonumber\\
&
y_i
=
\frac{e^{a^{(2)}_i}}{\sum_{j=1}^ne^{a^{(2)}_j}}\, , 
\qquad
a^{(2)}_i
=
\sum_{j=1}^{d_1}
W^{(1)}_{ij}x^{(1)}_j
+
b^{(1)}_i\, . 
\end{align}
Let us calculate the derivatives needed for backpropagation. The derivative of $\vec{x}^{(1)}$ is simple:
\begin{align}
\frac{\partial x^{(1)}_j}{\partial x^{(0)}_i}
=
\frac{\partial a^{(1)}_j}{\partial x^{(0)}_i}
\cdot
\frac{\mathrm{d} x^{(1)}_j}{\mathrm{d}a^{(1)}_j}
=
W^{(0)}_{ji}
\cdot
\left.
\frac{\mathrm{d}}{\mathrm{d}a}\mathrm{ReLU}(a)
\right|_{a=a^{(1)}_j}\, , 
\end{align}
\begin{align}
\frac{\partial x^{(1)}_k}{\partial W^{(0)}_{ij}}
=
\frac{\partial a^{(1)}_k}{\partial W^{(0)}_{ij}}
\cdot
\frac{\mathrm{d} x^{(1)}_k}{\mathrm{d}a^{(1)}_k}
=
\delta_{ik}x^{(0)}_j
\cdot
\left.
\frac{\mathrm{d}}{\mathrm{d}a}\mathrm{ReLU}(a)
\right|_{a=a^{(1)}_k}\, , 
\end{align}
and
\begin{align}
\frac{\partial x^{(1)}_j}{\partial b^{(0)}_i}
=
\frac{\partial a^{(1)}_j}{\partial b^{(0)}_i}
\cdot
\frac{\mathrm{d} x^{(1)}_j}{\mathrm{d}a^{(1)}_j}
=
\delta_{ij}
\cdot
\left.
\frac{\mathrm{d}}{\mathrm{d}a}\mathrm{ReLU}(a)
\right|_{a=a^{(1)}_j}\, , 
\end{align}
where
\begin{align}
\frac{\mathrm{d}}{\mathrm{d}a}\mathrm{ReLU}(a)
=
\left\{
\begin{array}{ccl}
1 & & (a>0)\\
0 & & (a\le 0)
\end{array}
\right.\, .
\end{align}
The derivative of $\vec{y}$ is a little bit more complicated because of the use of softmax. Specifically,
\begin{align}
\frac{\partial y_j}{\partial x^{(1)}_i}
=
\sum_k
\frac{\partial a^{(2)}_k}{\partial x^{(1)}_i}
\cdot
\frac{\partial y_j}{\partial a^{(2)}_k}
=
\sum_k
W^{(1)}_{ki}
\cdot
\frac{\partial y_j}{\partial a^{(2)}_k}\, , 
\end{align}
\begin{align}
\frac{\partial y_k}{\partial W^{(1)}_{ij}}
=
\sum_l
\frac{\partial a^{(2)}_l}{\partial W^{(1)}_{ij}}
\cdot
\frac{\partial y_k}{\partial a^{(2)}_l}
=
\sum_l
\delta_{il}x^{(1)}_j
\cdot
\frac{\partial y_k}{\partial a^{(2)}_l}
=
x^{(1)}_j
\cdot
\frac{\partial y_k}{\partial a^{(2)}_i}\, , 
\end{align}
and
\begin{align}
\frac{\partial y_j}{\partial b^{(1)}_i}
=
\sum_k
\frac{\partial a^{(2)}_k}{\partial b^{(1)}_i}
\cdot
\frac{\partial y_j}{\partial a^{(2)}_k}
=
\sum_k
\delta_{ik}
\cdot
\frac{\partial y_j}{\partial a^{(2)}_k}
=
\frac{\partial y_j}{\partial a^{(2)}_i}\, . 
\end{align}
Note that
\begin{align}
\frac{\partial y_j}{\partial a^{(2)}_k}
=
\delta_{jk}
\cdot
\frac{e^{a^{(2)}_j}}{\sum_{i}e^{a^{(2)}_i}}
-
\frac{e^{a^{(2)}_j}\cdot e^{a^{(2)}_k}}{(\sum_{i}e^{a^{(2)}_i})^2}
=
\delta_{jk}y_j
-
y_jy_k\, . 
\end{align}
Therefore, 
\begin{align}
\frac{\partial y_j}{\partial x^{(1)}_i}
=
W^{(1)}_{ji}y_j
-
y_j
\sum_k
W^{(1)}_{ki}y_k\, , 
\end{align}
etc. We combine the expressions above with
\begin{align}
\frac{\partial J(\vec{\theta})}{\partial\vec{\theta}}
&=
-
\frac{\partial}{\partial\vec{\theta}}
\sum_{k=1}^K
\vec{y}_{\rm answer}^{[k]}
\cdot
\log\vec{y}\left((\vec{x}^{(0)})^{[k]};\vec{\theta}\right)
\nonumber\\
&=
-
\sum_{k=1}^K
\sum_{j=1}^{n}
y_{{\rm answer},j}^{[k]}
\cdot
\frac{1}{y_j}
\frac{\partial y_j}{\partial\vec{\theta}}
\left((\vec{x}^{(0)})^{[k]}; \vec{\theta}\right)\, , 
\end{align}
where $\vec{\theta}$ stands for $W^{(0)}$, $W^{(1)}$, $\vec{b}^{(0)}$, and $\vec{b}^{(1)}$, and $[k]$ labels the samples. 

Implementing backpropagation for this network is a good exercise; with the expressions above, it takes only about ten lines.
%%%%%%%%%%%%
%%%%%%%%%%%%
\subsubsection*{Three-layer network for the MNIST dataset}
%%%%%%%%%%%%
%%%%%%%%%%%%
It is straightforward to build a three-layer network as well. Let us use it for image recognition with the MNIST dataset.\footnote{MNIST database = Modified National Institute of Standards and Technology database. See Ref.~\cite{LeCun1998}.} Specifically, we use a collection of $28\times 28$-pixel gray-scale pictures of handwritten numbers from 0 to 9. Each image is actually a set of $28\times 28=784$ numbers between 0 (black) and 1 (white). So, the input $\vec{x}^{(0)}$ is a 784-component vector. The labels are 10-component one-hot vectors: $(1,0,0,\cdots,0)$ for 0, $(0,1,0,\cdots,0)$ for 1, etc. The output is a 10-component vector, and the $i$-th component is the probability that the image is $i$ ($i = 0,1,\cdots,9$).

Let us take 4 nodes for each layer. Therefore, $d_0=784$, $d_1=4$, $d_2=4$, and $d_3=10$. It is a small network, but it already has $d_0d_1+d_1d_2+d_2d_3 = 3192$ parameters for $W$'s and $d_1 + d_2 + d_3 = 18$ parameters for $\vec{b}$'s. Furthermore, each hidden layer can represent $\mathbb{R}^4$ (more precisely, $[0,\infty)^4$ because we are using ReLU), which can, in principle, accommodate infinitely many clusters of data points and hence infinitely many features. Therefore, it is not unreasonable to hope that such a small network can actually perform well.

\begin{figure}[htbp]
		\centering
		\scalebox{0.5}{
		\includegraphics[width=\textwidth]{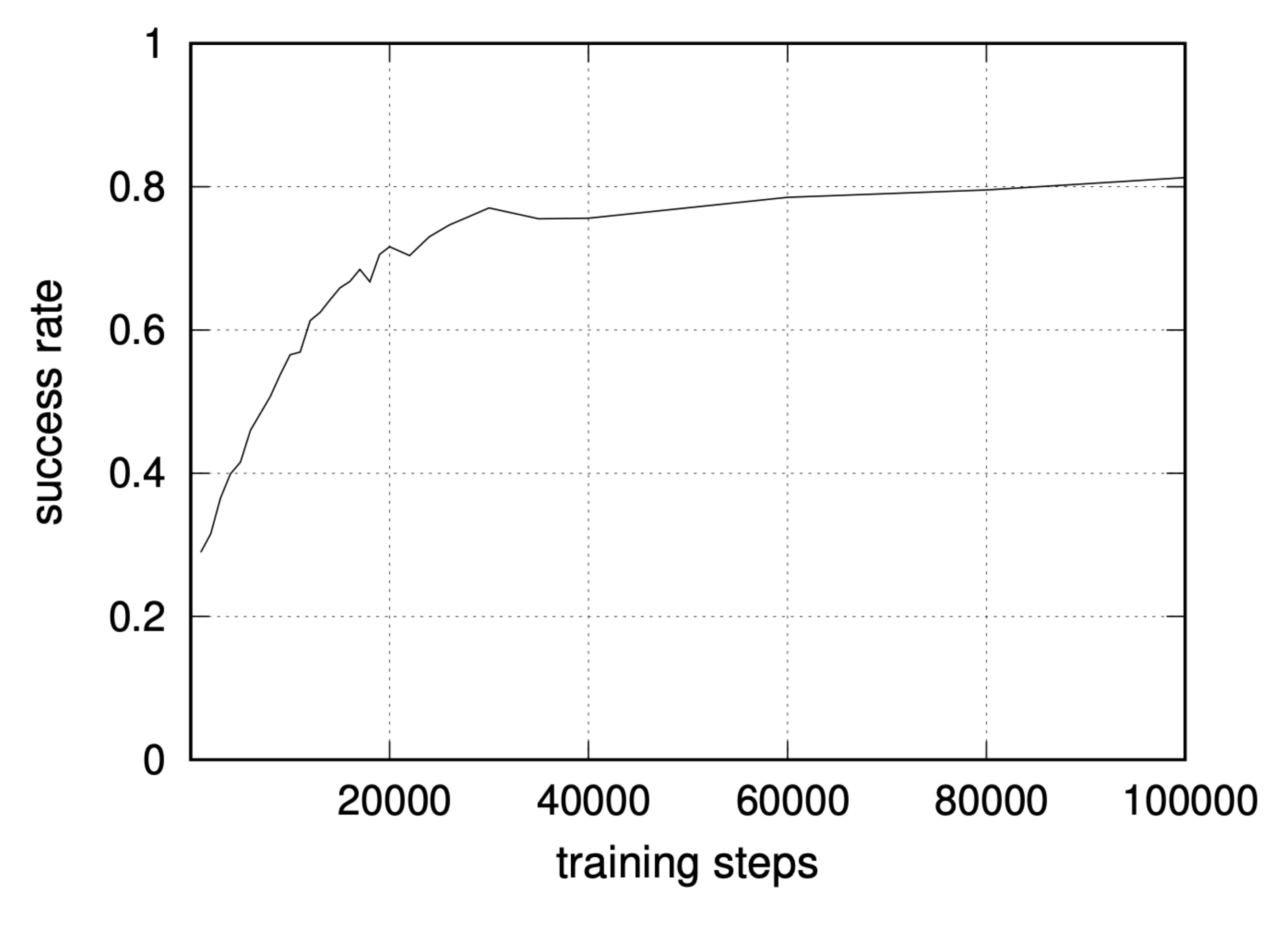}}
	\caption{The success rate for the classification vs.\ training steps.
	}\label{fig:MNIST_success-rate}
\end{figure}

\begin{figure}[htbp]
		\centering
		\includegraphics[width=\textwidth]{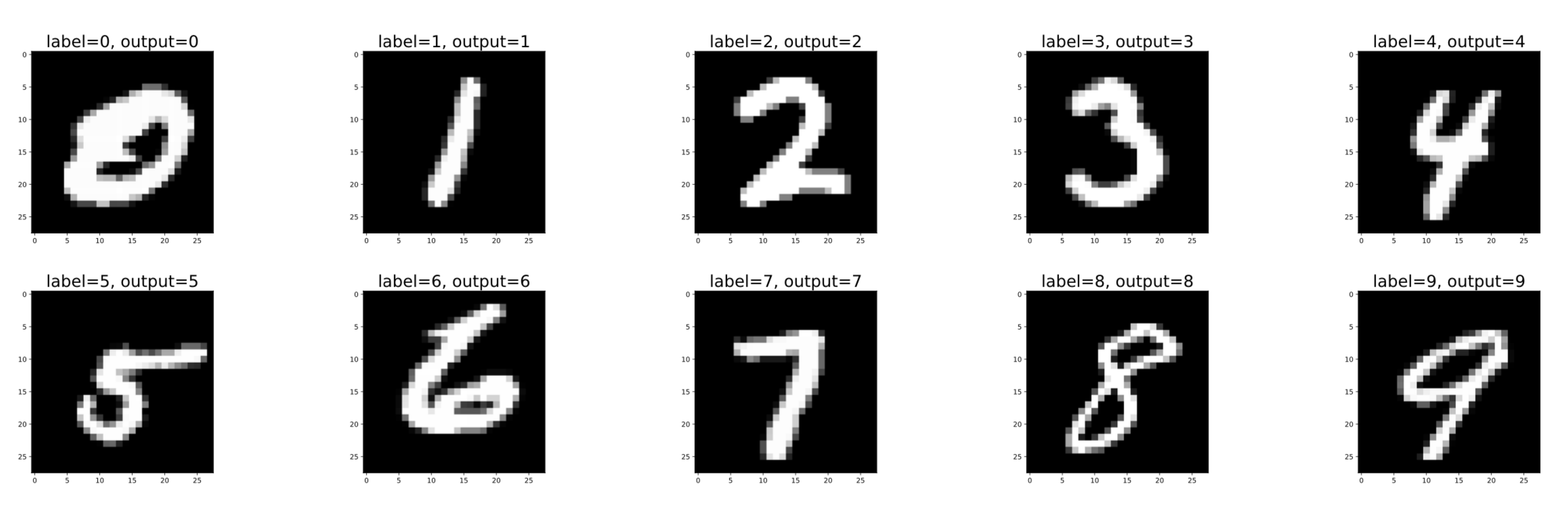}
	\caption{Examples of successful classifications.
	}\label{fig:MNIST-success}
\end{figure}

This network is deliberately tiny: each hidden layer has only four nodes, so every image has to be squeezed into four numbers before it is classified into ten classes.
 Even so, the success rate for the classification task goes above 80\%!\footnote{
For comparison, a linear model (softmax applied directly to the 784 pixels, i.e., multinomial logistic regression) reaches about 90\%, a two-layer network with a few hundred hidden nodes reaches about 95--98\%, and convolutional neural networks exceed 99\%; see Ref.~\cite{LeCun1998}. The point here is not the performance but how much a very small network can already do.
 }
In Fig.~\ref{fig:MNIST_success-rate}, we show how the performance improves as a function of training steps.\footnote{We used the same test dataset for this plot. There is a delicate issue here, related to the problem of overfitting, which is discussed in Sec.~\ref{sec:overfit}. Namely, we must use the validation dataset.} We used stochastic gradient descent with batch size 64 and learning rate $\epsilon=0.001$. Examples of successful classification are shown in Fig.~\ref{fig:MNIST-success}.

\begin{figure}[htbp]
		\centering
		\scalebox{0.6}{
		\includegraphics[width=\textwidth]{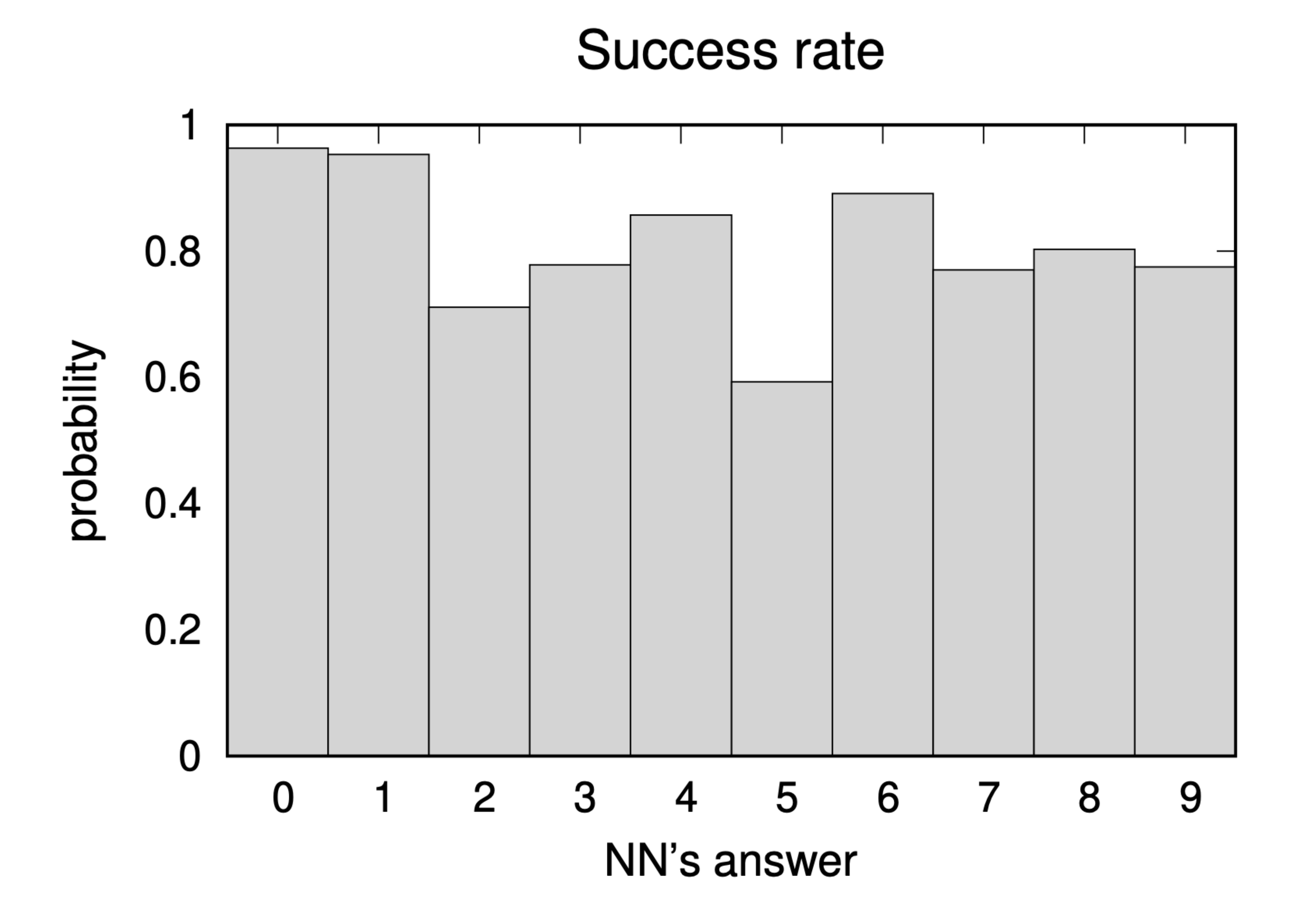}}
	\caption{The success rate for each kind of input (numbers 0 to 9) from MNIST.
	}\label{fig:MNIST-success-rate_0-to-9}
\end{figure}

\begin{figure}[htbp]
		\centering
		\scalebox{0.7}{
		\includegraphics[width=\textwidth]{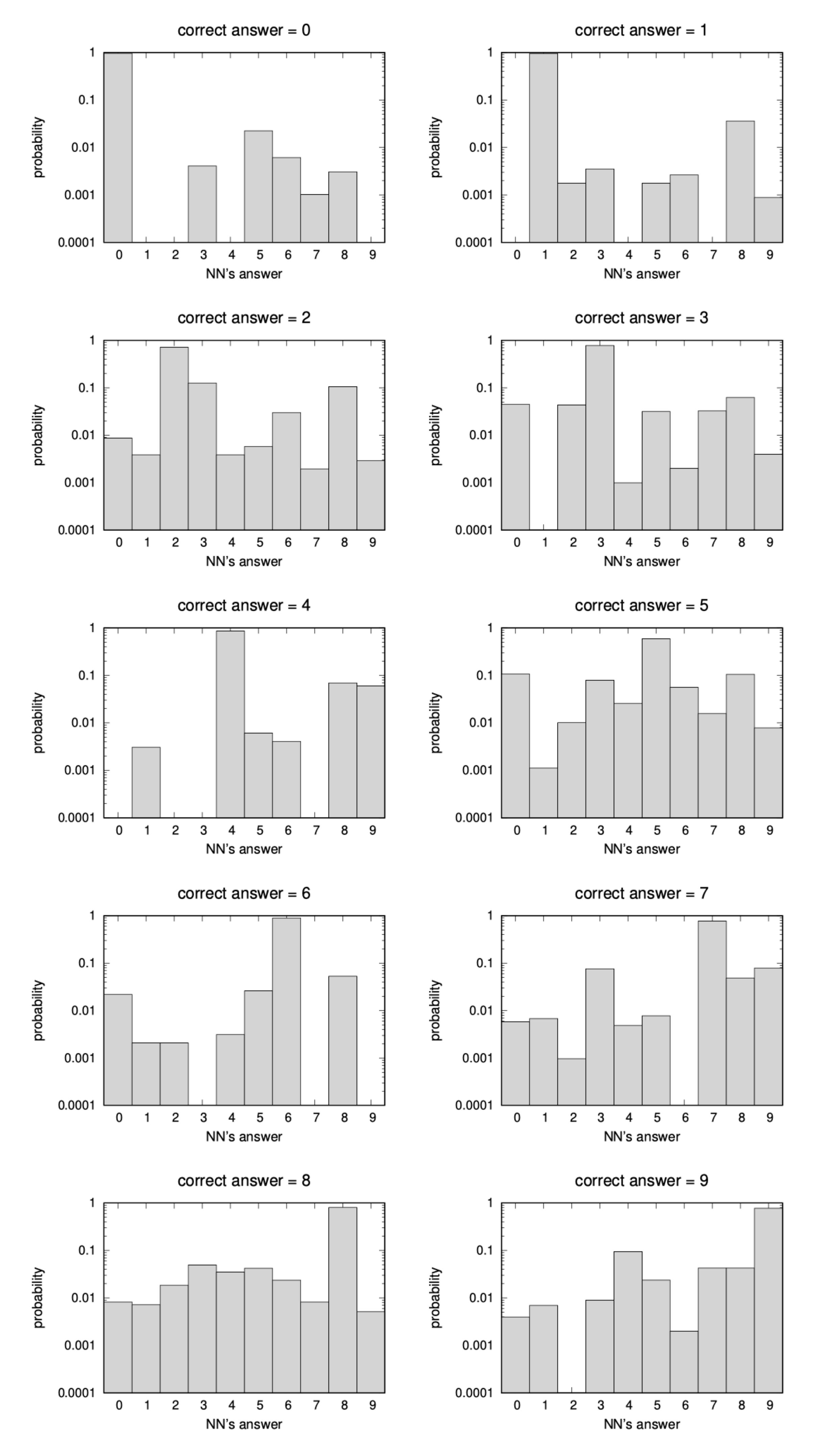}}
	\caption{The probability distribution of the NN’s output for each kind of input (numbers 0 to 9) from MNIST.
	}\label{fig:MNIST-prediction-probability}
\end{figure}

Let us see the details of the NN’s response to the input data. In Fig.~\ref{fig:MNIST-success-rate_0-to-9}, the success rate for each kind of input (numbers 0 to 9) is shown. The NN has a hard time dealing with 5, with a success rate lower than 60\%. In Fig.~\ref{fig:MNIST-prediction-probability}, we show the probability distribution of the NN’s output for each kind of input. 

\begin{figure}[htbp]
		\centering
		\scalebox{0.5}{
		\includegraphics[width=\textwidth]{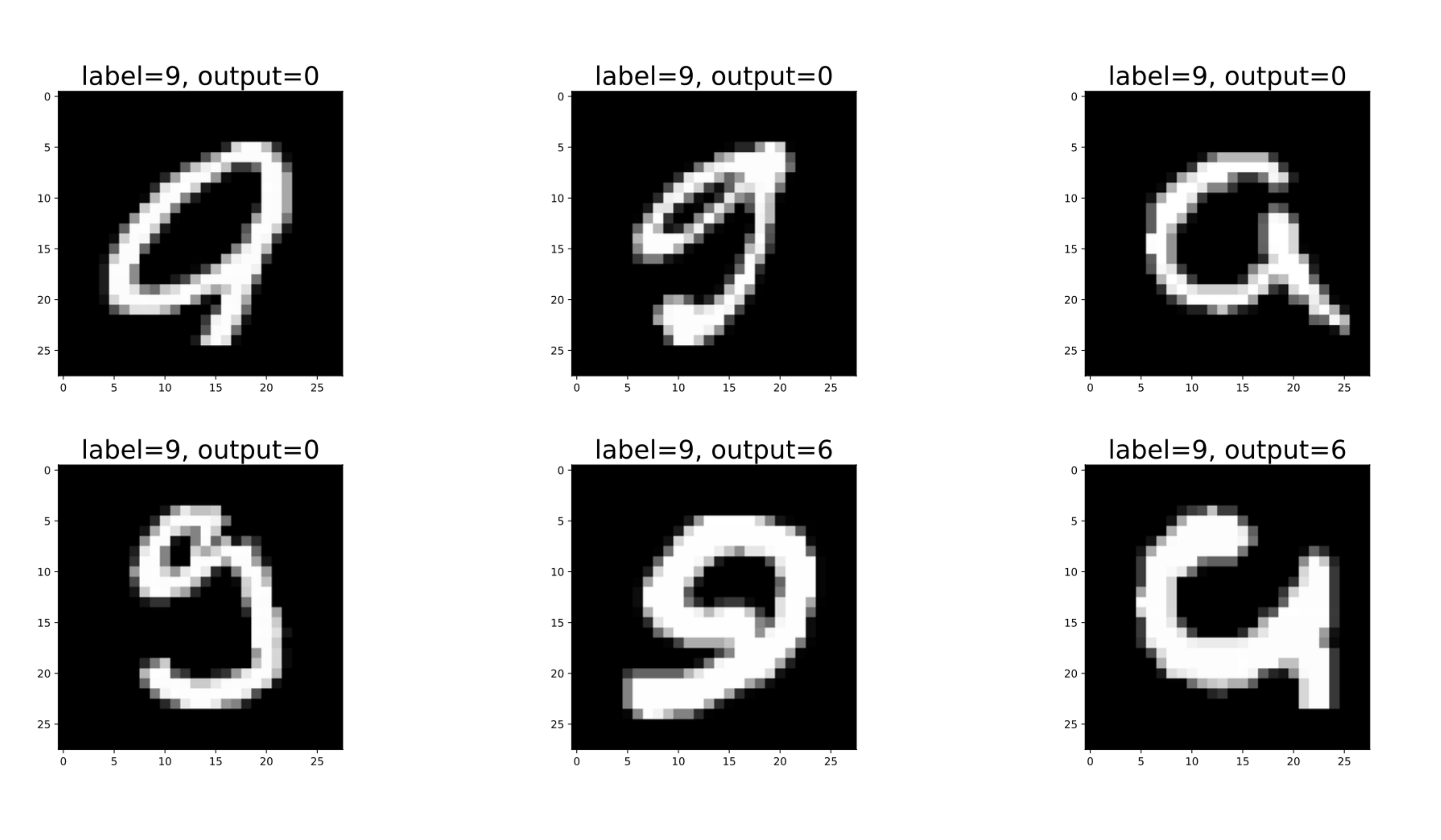}}
	\caption{Several examples of `classification errors'. The label is 9, and the NN gave 0 or 6 as output. Bad handwriting!
	}\label{fig:MNIST-false-9}
\end{figure}

\begin{figure}[htbp]
		\centering
		\scalebox{0.5}{
		\includegraphics[width=\textwidth]{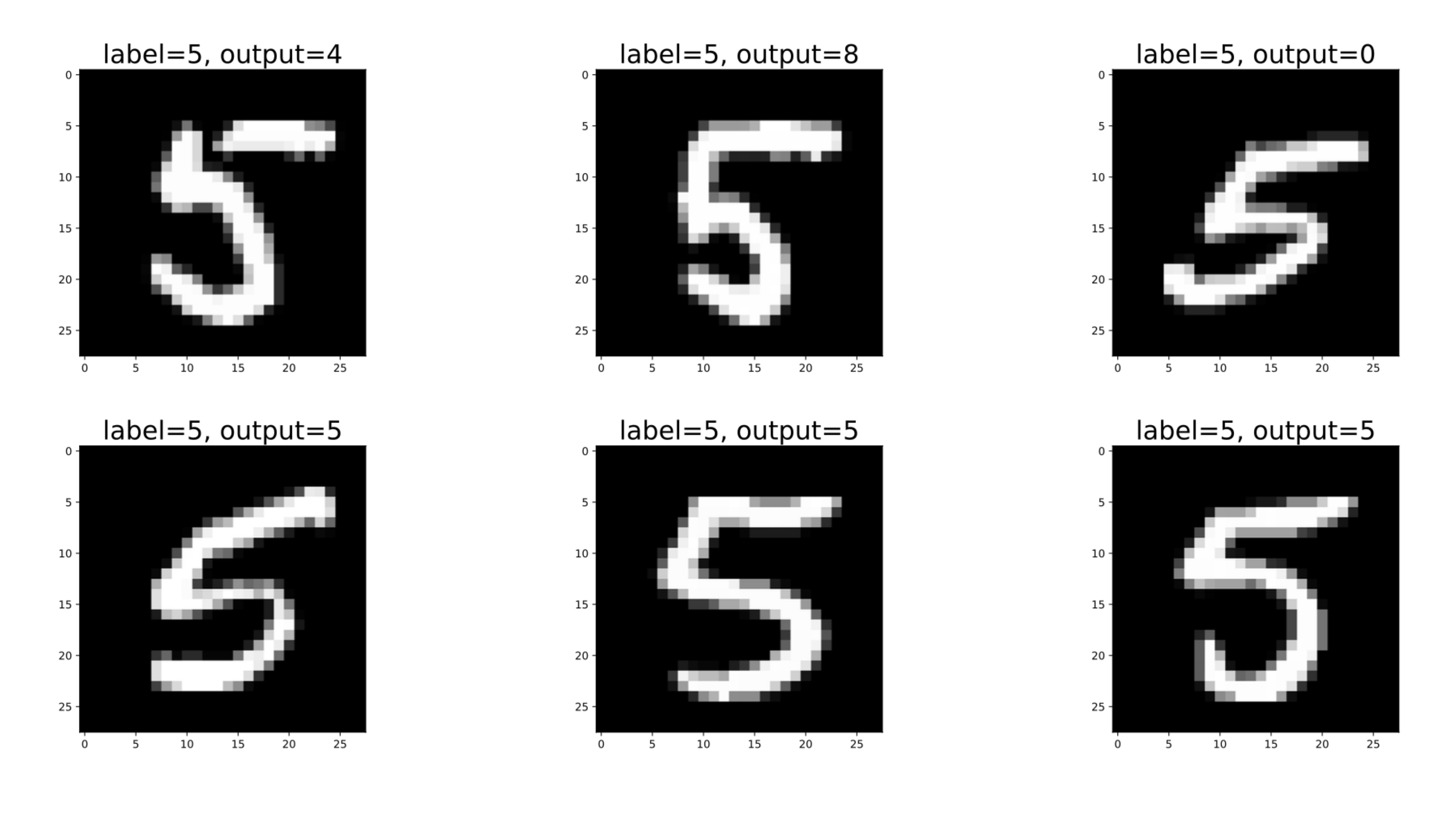}}
	\caption{
	Success and failure in the classification of 5. Not really bad handwriting.
	}\label{fig:MNIST-false-5}
\end{figure}

In Fig.~\ref{fig:MNIST-false-9}, we show several examples of misclassification of input 9 as output 0 or 6. Honestly, if you wrote a report with such bad handwriting, I would request resubmission because it is barely readable. Are these g, a, Q, or something else? Therefore, I would not blame the NN. On the other hand, if we look at typical examples of success and failure in the classification of 5 (Fig.~\ref{fig:MNIST-false-5}), it is clear that our NN is not good enough.

%%%%%%%%%%%%
%%%%%%%%%%%%
\subsection{Vanishing gradient problem and possible cures}\label{sec:vanishing_gradient}
%%%%%%%%%%%%
%%%%%%%%%%%%
By using the chain rule, $\frac{\partial y}{\partial W^{(M-k)}}$ was written as a sum of products of $k + 1$ derivatives.
Roughly speaking, if each derivative is $\epsilon$, each term becomes $\epsilon^{k+1}$. If the number of nodes
takes the same value $d_1=\cdots=d_M=d$, we have $d^k$ terms. In total, we obtain $\epsilon^{k+1}d^k$.
(Strictly speaking, we can have cancellation from positive and negative signs, which would
make the sum like $\epsilon^{k+1}d^{k/2}$. However, the essence does not change; we can use $\epsilon^{k+1}d^k$ as an upper bound.) Combining such
contributions, we could get something like $\frac{\partial y}{\partial W^{(M-k)}}\sim c^k$. If $c < 1$, the derivative becomes exponentially small as $k$ becomes large. This is called the \textit{vanishing gradient problem}. If $c>1$, the derivative becomes exponentially large as $k$ becomes large. This is called the \textit{exploding gradient problem}~\cite{Bengio1994}.

When the vanishing gradient problem takes place, the derivatives are smaller for the parameters describing the layers closer to the input. To train these parameters, one has to take the step size for gradient descent large. But then the parameters close to the output layer change too much, which causes instability. If we take the step size small so that the parameters close to the output layer do not change too much, then the parameters close to the input layer do not change (the change can be literally zero within numerical precision). This is a serious technical issue that prevents one from training parameters efficiently.

There is a more fundamental issue as well. Because $\frac{\partial y}{\partial x^{(M-k)}}$ becomes exponentially small as $k$ becomes large, just as $\frac{\partial y}{\partial W^{(M-k)}}$, the layers closer to the input are not really doing anything other than randomly transforming the input into something else. Depending on the choice of parameters, they may just mess up the input, and the network forgets the information about the input.

%%%%%%%%%%%%
%%%%%%%%%%%%
\subsubsection{Potential cures}\label{sec:cures_for_vanishing_gradient}
%%%%%%%%%%%%
%%%%%%%%%%%%
Because the vanishing gradient problem is such a fundamental issue in deep learning, there have been many attempts to cure it. Below, we list a few of them.
Normalization of the input to each layer also helps; it is discussed separately in Sec.~\ref{sec:normalization}.
%%%%%%%%%%%%
%%%%%%%%%%%%
\subsubsection*{Better activation function}
%%%%%%%%%%%%
%%%%%%%%%%%%
The derivative of the sigmoid function $\frac{1}{1+e^{-x}}$ is exponentially small when $|x|$ is large. This feature tends to cause the vanishing gradient problem. On the other hand, the derivative of ReLU does not decay because it is 1 for $x >0$. Because of this feature, the vanishing gradient problem takes place less frequently. For leaky ReLU, the derivative is nonzero for $x < 0$ as well, which could be a further improvement.

%%%%%%%%%%%%
%%%%%%%%%%%%
\subsubsection*{Residual Neural Network (ResNet)}
\begin{figure}[htbp]
\centering
\begin{tikzpicture}[>=Stealth,
  box/.style={draw, thick, minimum width=13mm, minimum height=7mm, font=\small}]
\node[box] (l1) at (0,0)   {Layer};
\node[box] (l2) at (3,0)   {Layer};
\node[box] (l3) at (6,0)   {Layer};
\node[circle, draw, thick, inner sep=1.5pt] (p) at (8,0) {$+$};
\node[box] (l4) at (10,0)  {Layer};
\draw[->, thick] (l1) -- (l2);
\draw[->, thick] (l2) -- (l3);
\draw[->, thick] (l3) -- (p);
\draw[->, thick] (p)  -- (l4);
\draw[->, thick] (1.0,0) -- (1.0,1.6) -- (8,1.6) -- (p);
\end{tikzpicture}
\caption{A schematic picture of a Residual Neural Network.}
\label{fig:ResNet}
\end{figure}
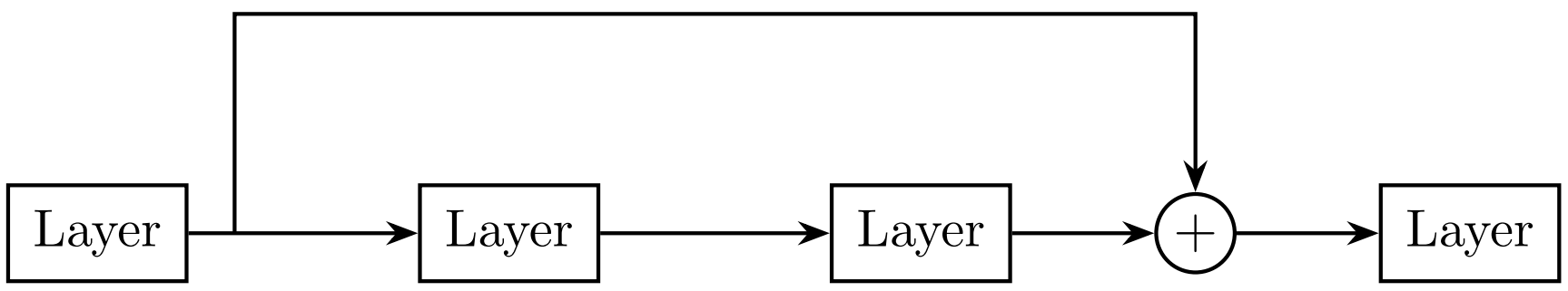

%%%%%%%%%%%%
%%%%%%%%%%%%
A Residual Neural Network (ResNet)~\cite{He2015}\footnote{Don’t call it RNN because RNN means Recurrent Neural Network.} introduces a shortcut for the propagation of information, by using a transformation of the following form:
\begin{align}
f(x)+x\, . 
\end{align}
For example, we can skip $s$ steps, and use
\begin{align}
\vec{x}^{(k+s+1)}=\sigma(f(\vec{x}^{(k)})+\vec{x}^{(k)})\, ,
\end{align}
where $f(\vec{x}^{(k)})$ is the usual transformation consisting of linear transformations and activation functions. Here, we assumed $d_k=d_{k+s+1}$. We could more generally do things such as
\begin{align}
\vec{x}^{(k+s+1)}=\sigma(f(\vec{x}^{(k)})+W\vec{x}^{(k)})\, . 
\end{align}
Often, pictures like Fig.~\ref{fig:ResNet} are used.

Because $x$ is passed through the shortcut as it is, the gradient can also go back through the shortcut without decaying, even if the gradient through $f$ is small (the derivative of $f(x)+x$ is $f'(x)+1$).

%%%%%%%%%%%%
%%%%%%%%%%%%
\subsubsection*{Long Short-Term Memory (LSTM)}
%%%%%%%%%%%%
%%%%%%%%%%%%
A Recurrent Neural Network (RNN; see Sec.~\ref{sec:RNN}) is a network which is suitable for the analysis of time series, including natural language processing. It has two directions for the allocation of layers: input to output, and early time to late time. The latter direction is introduced to propagate ‘memory’ to later times. The memory decays as time passes due to the vanishing gradient problem. Hence, this is only short-term memory. LSTM~\cite{Hochreiter1997,Gers2000} (Sec.~\ref{sec:LSTM}) is an architecture that allows us to keep this short-term memory somewhat longer.
After 2017, Transformer~\cite{Vaswani2017} became a more popular option.

%%%%%%%%%%%%
%%%%%%%%%%%%
\subsubsection*{Attention mechanism}
%%%%%%%%%%%%
%%%%%%%%%%%%
The attention mechanism helps an NN to know which feature of the input is important, by throwing away most of the input and keeping the information important for the task. This also helps with the vanishing gradient problem: in sequence models, attention creates a direct path from each input position to the output, so the gradient does not have to pass through many layers or time steps; see Sec.~\ref{sec:attention}.

Attention is a broad concept. For image recognition, it finds people, animals, cars, etc., which are likely to be useful for the task. For an NN to play video games, it finds where the enemies are, where the scoreboard is, where the valuable items are, etc. For NLP, it could be used to find important words or connections between words. How to see which part of the input a network is paying attention to is discussed in Sec.~\ref{sec:implicit_attention}.

%%%%%%%%%%%%
%%%%%%%%%%%%
\subsection{Batch normalization and layer normalization}\label{sec:normalization}
%%%%%%%%%%%%
%%%%%%%%%%%%

%%%%%%%%%%%%
%%%%%%%%%%%%
\subsubsection{Internal covariate shift}
%%%%%%%%%%%%
%%%%%%%%%%%%
Suppose we want to use a neural network trained for the MNIST dataset to classify new samples of handwritten numbers. The new data may be a bit fuzzy — for example, the black background could become gray, and white letters could also become light gray. Then, to have better performance, we should pre-process the samples and make them closer to MNIST. A simple way to do this is to adjust the mean and variance. The variance must be smaller than that of MNIST if the contrast is not sharp. The mean could shift. We can adjust them by a transformation $\vec{x}\to\alpha\vec{x}+\beta$. This kind of mismatch between the training data and the new data is called covariate shift.

The same kind of shift may take place when we use mini-batch gradient descent. (E.g., some batches contain a lot of fuzzy images and some batches contain only sharp images.) Furthermore, even if the training data does not have such a problem, the input to a hidden layer can change significantly every time the parameters are updated. This makes the training difficult; for example, if the activation function is the sigmoid, then a large input gives a vanishing gradient; also, the very intuition of gradient descent assumes gradual change. Such a shift of the input to the hidden layers is called internal covariate shift.

Batch normalization and layer normalization are used to resolve the problem of internal covariate shift.\footnote{This was the original motivation~\cite{Ioffe2015}. Why batch normalization helps the training is still debated; for example, Ref.~\cite{Santurkar2019} argues that the main effect is to make the cost function smoother.}
%%%%%%%%%%%%
%%%%%%%%%%%%
\subsubsection{Batch normalization}
%%%%%%%%%%%%
%%%%%%%%%%%%
In batch normalization~\cite{Ioffe2015}, the input to each layer is normalized for each batch. At the $k$-th layer, the mean and variance of $\vec{x}^{(k)}$ are calculated component by component over samples in the batch. Specifically, for the $j$-th component, the mean is $\mu^{(k)}_{j,\Lambda}=\frac{1}{n_\Lambda}\sum_{\ell\in\Lambda}x_j^{(k)[\ell]}$ and the variance is $\left(\sigma^{(k)}_{j,\Lambda}\right)^2=\frac{1}{n_\Lambda}\sum_{\ell\in\Lambda}\left(x^{(k)[\ell]}_j-\mu_{j,\Lambda}^{(k)}\right)^2$, where $\Lambda$ is the batch and $n_\Lambda$ is the batch size. Then, $\hat{x}^{(k)[\ell]}_j=\frac{x^{(k)[\ell]}_j-\mu_{j,\Lambda}^{(k)}}{\sigma^{(k)}_{j,\Lambda}}$ is normalized to have zero mean and unit variance. For each $j$, two learnable parameters $\gamma_j^{(k)}$ and $\beta_j^{(k)}$ are assigned, and $y^{(k)[\ell]}_j=\gamma_j^{(k)}\hat{x}^{(k)[\ell]}_j+\beta_j^{(k)}$ is passed to the $(k+1)$-th layer as input. The derivation of the derivatives needed for backpropagation is tedious but straightforward.

At test time, we may want to use the network for a single sample, for which the mean and variance over the batch do not make sense. More importantly, we want the output for each sample to be determined by that sample alone, not by the other samples which happen to be in the same batch. Therefore, the mean and variance are not calculated from the test samples; the averages of $\mu^{(k)}_{j,\Lambda}$ and $\left(\sigma^{(k)}_{j,\Lambda}\right)^2$ over the batches, accumulated during training, are used instead.

Note that the normalization is performed by averaging over all samples in the batch, for each node. This is important when you want to see the difference from layer normalization.
%%%%%%%%%%%%
%%%%%%%%%%%%
\subsubsection{Layer normalization}\label{sec:layer_normalization}
%%%%%%%%%%%%
%%%%%%%%%%%%
Layer normalization~\cite{Ba2016} is performed by averaging over all nodes in the layer, for each sample. Specifically, for the $k$-th layer, the mean is $\mu^{(k)}=\frac{1}{d_k}\sum_{j=1}^{d_k}x_j^{(k)}$ and the variance is $\left(\sigma^{(k)}\right)^2=\frac{1}{d_k}\sum_{j=1}^{d_k}\left(x_j^{(k)}-\mu^{(k)}\right)^2$. Then, $\hat{x}^{(k)}_j=\frac{x^{(k)}_j-\mu^{(k)}}{\sigma^{(k)}}$ is normalized to have zero mean and unit variance. For each $j$, two learnable parameters $\gamma^{(k)}_j$ and $\beta^{(k)}_j$ are assigned, and $y^{(k)}_j=\gamma^{(k)}_j\hat{x}^{(k)}_j+\beta^{(k)}_j$ is passed to the $(k+1)$-th layer as input.

Layer normalization can be applied to just one sample. This is convenient for typical NLP tasks.

%%%%%%%%%%%%
%%%%%%%%%%%%
\section{Generalization}\label{sec:generalization}
%%%%%%%%%%%%
%%%%%%%%%%%%
We now turn to the second question raised at the beginning of Sec.~\ref{sec:training}: how can we make sure that a network that fits the training data also works for unseen data?

%%%%%%%%%%%%
%%%%%%%%%%%%
\subsection{Overfitting and regularization}\label{sec:overfit}
%%%%%%%%%%%%
%%%%%%%%%%%%
In machine learning, we minimize the training error. However, this is just a means and not the purpose. The ultimate goal is to minimize the error against unseen data, and we use the test error as a reasonable estimate of the error against unseen data. Usually, we can reduce the test error by minimizing the training error. But this is not always the case.

Let us consider the simplest case, $y=f(x)$. Suppose $K$ data points $\{x^{(i)},y^{(i)}\}_{i=1,\cdots,K}$ are given. If we use a polynomial of order $p \ge K-1$, we can fit all the data points exactly. For example, we can take a polynomial of degree $K-1$,
\begin{align}
y
=
\sum_{i=1}^K
\left(
y^{(i)}
\cdot
\frac{\prod_{j\neq i}(x-x^{(j)})}{\prod_{j\neq i}(x^{(i)}-x^{(j)})}
\right)\, . 
\end{align}
Obviously, such an ansatz fits the statistical fluctuations as well and does not generalize to unseen data.\footnote{
An easy way to see how it fails is that the asymptotic behavior at $x\to\pm\infty$ is sensitive to the choice of the degree $p$.
}
This is \textit{overfitting}. By reducing the training error too much, the test error (which is a reasonable estimate of the generalization error) increases; this is not what we want.

Let us consider an image classification task of cats and dogs. Suppose that no cat in your dataset was white, but there were a few white dogs; this is a kind of noise, because we know that there are many white cats. However, our model would fit this noise and learn a pattern which does not generalize to real-world applications: ``white $\to$ dog". When our model sees an image of a white cat, it could mistakenly predict that this is a dog.  

To avoid overfitting, we should eliminate unnecessary parameters. If we have a very good knowledge of the problem, we could find a good ansatz that does not contain unnecessary parameters. In such a case, however, you do not have to use deep learning. You can use other machine learning methods. The biggest reason that we use deep learning is that it does not require too much model-specific tuning. To keep this virtue, we should start with an architecture with sufficiently many degrees of freedom, and introduce a mechanism that automatically suppresses unnecessary degrees of freedom. \textit{Regularization} is such an approach.

%%%%%%%%%%%%
%%%%%%%%%%%%
\subsubsection*{Weight decay factor}
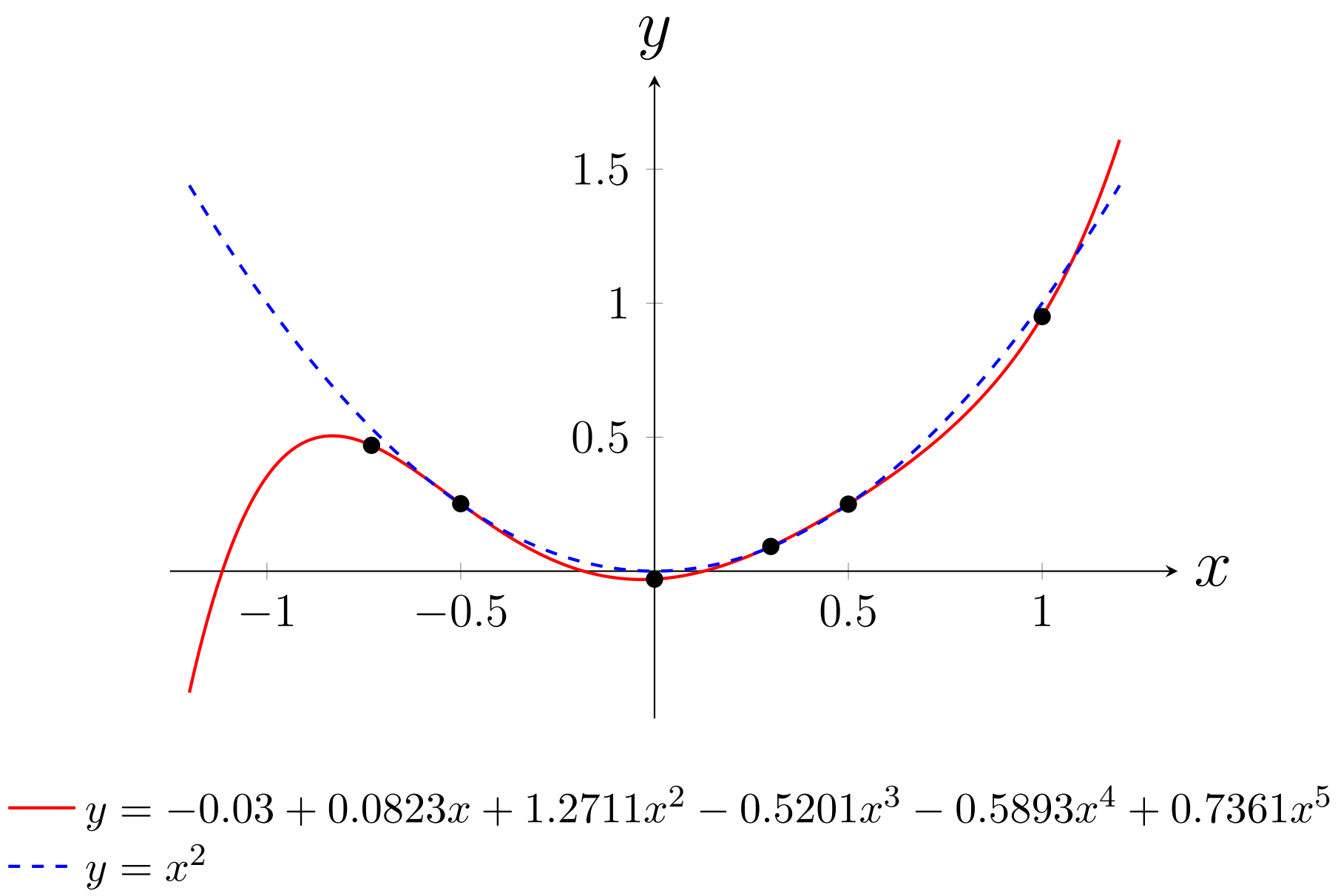
\begin{figure}[htbp]
\centering
\begin{tikzpicture}
\begin{axis}[
width=0.65\textwidth, height=0.45\textwidth,
domain=-1.2:1.2, samples=200,
xmin=-1.25, xmax=1.35, ymin=-0.55, ymax=1.85,
axis lines=middle,
xlabel={$x$}, ylabel={$y$},
xlabel style={at={(ticklabel* cs:1)}, anchor=west, font=\Large},
ylabel style={at={(ticklabel* cs:1)}, anchor=south, font=\Large},
xtick={-1,-0.5,0.5,1}, ytick={0.5,1,1.5},
legend style={at={(0.5,-0.08)}, anchor=north, draw=none, font=\small, legend columns=1},
legend cell align=left,
]
\addplot[red, thick] {-0.03+0.0823*x+1.2711*x^2-0.5201*x^3-0.5893*x^4+0.7361*x^5};
\addlegendentry{$y=-0.03+0.0823x+1.2711x^2-0.5201x^3-0.5893x^4+0.7361x^5$}
\addplot[blue, thick, dashed] {x^2};
\addlegendentry{$y=x^2$}
\addplot[only marks, mark=*, mark size=2pt] coordinates {
(-0.73,0.4697) (-0.5,0.2518) (0,-0.03) (0.3,0.0921) (0.5,0.2501) (1,0.9501)};
\end{axis}
\end{tikzpicture}
\caption{An example of overfitting. Six data points (dots), scattered around $y=x^2$ (dashed), are fitted by a polynomial of degree five (solid), which has six parameters. The fit goes through all the points but behaves wildly outside the region of the data.}
\label{fig:overfit-example}
\end{figure}

%%%%%%%%%%%%
%%%%%%%%%%%%
In typical cases of overfitting, many parameters take large values and push down the training error a little bit as a result of a large cancellation (see, e.g., Fig.~\ref{fig:overfit-example}). To avoid it, we can add a penalty term such as $c\times |\vec{\theta}|^2$, where $c$ is a positive number which must be set as a hyperparameter, to the cost function. Then, to minimize the modified cost function, unnecessary parameters that do not change the original cost function significantly should take small values.

This is called $L_2$ regularization (or weight decay). In Bayesian language, it corresponds to assuming a Gaussian prior distribution $\propto e^{-c|\vec{\theta}|^2}$ for the parameters: if $J$ is the negative log-likelihood, minimizing $J(\vec{\theta})+c|\vec{\theta}|^2$ is the same as maximizing the posterior distribution $\propto e^{-J(\vec{\theta})}\cdot e^{-c|\vec{\theta}|^2}$, i.e., it is the maximum a posteriori (MAP) estimate mentioned in Sec.~\ref{sec:MLE}.

In the above example of cat-dog classification, the model will learn the essential features common to most cats and dogs, because the reduction of $J(\vec{\theta})$ can be more than the increase of $c\times|\vec{\theta}|^2$. On the other hand, a fake pattern ``white $\to$ dog" can reduce $J(\vec{\theta})$ only a little bit (because the success rate improves only for a small fraction of the dataset: white dogs), which can be beaten by the increase of $c\times|\vec{\theta}|^2$. 

%%%%%%%%%%%%
%%%%%%%%%%%%
\subsubsection*{Early stopping}
%%%%%%%%%%%%
%%%%%%%%%%%%
Roughly speaking, \textit{early stopping} aims to halt training when the test error reaches its minimum. More precisely, the training data is split into two sets — a new training dataset and a validation dataset. The error (cost function) computed on the validation dataset, known as the validation error, is monitored as a proxy for the generalization error. When the validation error begins to increase, training is stopped, and the parameters that yield the smallest validation error are selected as the final model.

Note that the test dataset and the validation dataset are different. To estimate the generalization error, we use the test dataset. If we mix up the test dataset and the validation dataset, we may overfit the test dataset, and the test error may not give us a good estimate of the generalization error.

In the above example of cat-dog classification, the model tends to learn essential features first, because that reduces $J(\vec{\theta})$ drastically. When the model starts to learn fake features that do not generalize, the validation error starts to go up. 

%%%%%%%%%%%%
%%%%%%%%%%%%
\subsubsection*{Data augmentation}
%%%%%%%%%%%%
%%%%%%%%%%%%
We can reduce the risk of overfitting by using a larger dataset, and one way to achieve this is through data augmentation. For image classification, for example, we can expand the dataset by altering the colors, slightly rotating the images, applying mirror reflections, adding noise, and more.
In the above example of cat-dog classification, we can prepare cats of all kinds of colors --- not just adding white cats, but also pink, purple, or green ones --- so that the model cannot rely on the fake pattern ``white $\to$ dog".

%%%%%%%%%%%%
%%%%%%%%%%%%
\subsubsection*{Ensemble average}
%%%%%%%%%%%%
%%%%%%%%%%%%
Suppose an ensemble of similar datasets is given. There are many ways to prepare such an ensemble: we can split a large dataset into smaller pieces, randomly pick data points (mimicking bootstrap sampling), or use data augmentation. In principle (if we are rich enough to pay the cost!), we can create different models by using different datasets. We can also create different models by using slightly different architectures with the same training data, or by combining both different architectures and different training data. In any case, each model may overfit the training data, but the specifics of overfitting will differ across models. By averaging over the ensemble, or by taking the majority vote in the case of classification, we can mitigate the effect of overfitting.

In the above example of cat-dog classification, a model trained on a dataset without white cats may learn the fake pattern ``white $\to$ dog", but models trained on other datasets would not, and the fake pattern is suppressed in the average.

A clear disadvantage of this approach is that the cost increases with the number of models we build. To reduce inference costs, one possible solution is to create a distilled model (see Sec.~\ref{sec:distillation}), using the ensemble average as the target.
%%%%%%%%%%%%
%%%%%%%%%%%%
\subsubsection*{Dropout}
%%%%%%%%%%%%
%%%%%%%%%%%%
\textit{Dropout}~\cite{Srivastava2014} tries to mimic the ensemble average without increasing the cost. Specifically, we make different architectures by randomly removing a part of the nodes. Equivalently, the output from the ‘removed’ nodes is set to zero. To reduce the cost, we use the same parameters $\vec{\theta}$ for all models. (Because of this, dropout is different from the ensemble average.) The actual training procedure is as follows:
\begin{itemize}
\item
At each time step, remove the nodes randomly with probability $1-p$ (equivalently, keep them with probability $p$) and define a sub-NN.\footnote{We can choose different values of $p$ for different nodes.}

\item
Apply stochastic gradient descent to the sub-NN and update the parameters. The parameters associated with the removed nodes are kept unchanged.

\item
Repeat.

\end{itemize}
When we use the NN after training, to mimic the ensemble average closely, the most natural way is to repeat computations many times, removing different sets of nodes, and take the average. It is costly, of course. Instead, we define a new network with all nodes by multiplying by $p$ the output from a node which was removed with probability $1-p$ during training. Empirically, it is a good approximation of the ensemble average.

%%%%%%%%%%%%
%%%%%%%%%%%%
\subsubsection*{Stochastic gradient descent and implicit regularization}
%%%%%%%%%%%%
%%%%%%%%%%%%
Stochastic gradient descent (SGD) differs from vanilla gradient descent (batch gradient descent) when the learning rate $\epsilon$ is finite. Empirically, it has been observed that a neural network trained with SGD often generalizes better for certain finite values of $\epsilon$. There is also evidence that, when the learning rate is small but finite, SGD is roughly equivalent to minimizing a modified cost function with an additional term that penalizes large gradients for each mini-batch (\textit{implicit regularization}), and the strength of this term is proportional to the learning rate $\epsilon$~\cite{Barrett2022,Smith2021}.

An intuitive explanation is as follows. Recall the example in Sec.~\ref{sec:SGD}: among the dog images, there are some of Siberian Huskies with snow in the background, and no other animals in the dataset have snow in their backgrounds. In this case, the neural network might incorrectly learn to classify any image with white stuff in its background as a ``dog'' --- even if it is a snow monkey or a fox in the snow. In SGD, if a mini-batch $\Lambda$ doesn't include a Siberian Husky with snow, the gradient $\frac{\partial J_\Lambda}{\partial \vec{\theta}}$ will not reinforce the accidental association between ``dog'' and ``snow''. If the learning rate $\epsilon$ is sufficiently large, the parameters $\vec{\theta}$ will be adjusted away from this overfitting each time the mini-batch does not contain such images, preventing the network from learning the spurious connection.

%%%%%%%%%%%%
%%%%%%%%%%%%
\subsection{Distillation}\label{sec:distillation}
%%%%%%%%%%%%
%%%%%%%%%%%%

Suppose we have a powerful AI based on a large architecture with many parameters. Running such an AI requires substantial computational resources, including powerful hardware and significant electricity. For practical purposes, however, we want a smaller AI that can operate with fewer resources --- perhaps on a laptop or mobile phone --- without losing the performance of the larger AI. Distillation~\cite{Hinton2015,Bucila2006} allows us to achieve this.

When an AI performs a task, it does more than just produce a final answer. For instance, in an image classification task, when the AI classifies an image as ``dog," it also estimates the probabilities for other possible classes. For example, the probability of the image being a dog might be $y_{\rm dog} = 0.9$, but it might also assign small probabilities to other classes, such as $y_{\rm cat} = 0.0001$, $y_{\rm fox} = 0.02$, $y_{\rm wolf} = 0.06$, $y_{\rm turtle} = 10^{-10}$, and so on. This entire set of probabilities contains more valuable information than the final classification alone. In distillation, the smaller model is trained to replicate this set of probabilities, effectively learning from the richer information provided by the larger model. It is analogous to learning not just the answer but also the process of reaching the answer. 

We can think of the larger AIs as being like Newton or Einstein, and the smaller AIs as representing ourselves. While we may not make the same groundbreaking discoveries, we can still learn from them and understand the logic behind their reasoning.

We may not be as smart as Newton or Einstein, but we can achieve remarkable things by working together. A similar idea is behind the ensemble average, which we discussed as a way to combat overfitting. However, running many AIs and taking their average can be computationally expensive. To address this, we can train a single AI to reproduce the ensemble average of many models. This concept was explored in the original distillation papers~\cite{Hinton2015,Bucila2006}.

Suppose the softmax is used at the final layer. Then, the output takes the form of $\vec{y}=\mathrm{Softmax}(\vec{a})$, or equivalently, $y_j=\frac{e^{a_j}}{\sum_i e^{a_i}}$. A simple option is to minimize the square loss $\sum_{j,k}\left(a^{[\mathrm{teacher}]}_j(\vec{x}^{(k)})-a^{[\mathrm{student}]}_j(\vec{x}^{(k)})\right)^2$~\cite{Bucila2006}.
In Ref.~\cite{Hinton2015}, the probabilities are compared instead, by using the cross-entropy between the output of the teacher and that of the student. To control which part of the teacher's output is emphasized, a parameter $T$ called the temperature\footnote{This is nothing but the Boltzmann distribution at temperature $T$, with $-a_j$ identified with the energy of the state $j$. In this language, $T\to 0$ picks up the ground state.} is introduced as $y^{(T)}_j=\frac{e^{a_j/T}}{\sum_i e^{a_i/T}}$, and $\vec{y}^{(T)}$ is used instead of $\vec{y}$. At $T=1$, the probabilities of wrong classes such as $y_{\rm turtle}=10^{-10}$ are too small to affect the loss. A higher temperature makes them larger, so that the student can learn more from the relative probabilities of the wrong answers, e.g., that a dog is more similar to a wolf than to a turtle. In the limit of $T\to\infty$, this reduces to the square loss above (after subtracting the mean of $a_j$), in which all classes are treated equally. The appropriate value of $T$ depends on the problem.

%%%%%%%%%%%%
%%%%%%%%%%%%
\section{Evaluation and robustness}\label{sec:evaluation}
%%%%%%%%%%%%
%%%%%%%%%%%%
Finally, we turn to the third question raised at the beginning of Sec.~\ref{sec:training}: how should we measure the performance of a trained network, and how can it fail?

%%%%%%%%%%%%
%%%%%%%%%%%%
\subsection{Accuracy, precision, and recall}
%%%%%%%%%%%%
%%%%%%%%%%%%
Let us consider a binary classification task for cancer detection. There are several different measures of the quality of the predictions: accuracy, precision, and recall.
The \textit{accuracy} is defined by
\begin{align}
\mathrm{Accuracy}
=
\frac{\mathrm{Correct\ Predictions}}{\mathrm{All\ Predictions}}\, ,  
\end{align}
which is simply the proportion of correct predictions (both true positives and true negatives) out of all predictions. While a high accuracy is often seen as a good thing, it can be misleading, especially in imbalanced datasets. For example, if 99\% of the test samples are negative (no cancer), a classifier that predicts ``no cancer" for every case would achieve 99\% accuracy, but it would not be effective at detecting actual cancer cases. Therefore, accuracy alone is not always a reliable indicator of model performance.

To address this, we can use \textit{precision} and \textit{recall} defined by 
\begin{align}
\mathrm{Precision}
=
\frac{\mathrm{(True\ Positive)}}{\mathrm{(True\ Positive)}+\mathrm{(False\ Positive)}}
=
\frac{\mathrm{(True\ Positive)}}{\mathrm{(All\ Positive\ Predictions)}}
\end{align}
and 
\begin{align}
\mathrm{Recall}
=
\frac{\mathrm{(True\ Positive)}}{\mathrm{(True\ Positive)}+\mathrm{(False\ Negative)}}
=
\frac{\mathrm{(True\ Positive)}}{\mathrm{(All\ Positive\ Cases)}}
\, , 
\end{align}
respectively. 
If precision is low, many of the positive predictions are false alarms (false positives). 
If recall is low, the classifier is missing many actual cancer cases (false negatives). 
False negatives can lead to delayed diagnoses, while false positives can lead to unnecessary anxiety and medical tests.

There is a trade-off between precision and recall. A typical classifier returns the probability of cancer, and we predict `positive' when it is larger than a certain threshold. If we lower the threshold, fewer cancers are missed (the recall goes up), but there are more false alarms (the precision goes down). The appropriate threshold depends on the cost of each kind of error. When a single number is needed, the harmonic mean of precision and recall, which is called the $F_1$ score, is often used.
%%%%%%%%%%%%
%%%%%%%%%%%%
\subsection{What does the network pay attention to?}\label{sec:implicit_attention}
%%%%%%%%%%%%
%%%%%%%%%%%%
As mentioned in Sec.~\ref{sec:cures_for_vanishing_gradient}, an attention mechanism tells a network which part of the input is important. Such `attention' is implicit in many networks, even if it is not incorporated explicitly. To see it, we can use the correlation between input $x_i$ and output $y_j$ defined by the Jacobian $\frac{\partial y_j}{\partial x_i}$. Imagine an NN for recognizing animals, where $y_1, y_2, \cdots$ are the cat score, the dog score, etc. If $y_{\rm cat}$ is sensitive to a change in $x_i$, the NN uses such an input to decide whether the picture is of a cat or not. Specifically, for a given picture represented by $\vec{x}^{(k)} = (x_1^{(k)},x_2^{(k)},\cdots)$, we can estimate $\left.\frac{\partial y_j}{\partial x_i}\right|_{\vec{x}=\vec{x}^{(k)}}$ by perturbing $\vec{x}$ around $\vec{x}^{(k)}$. We can change pictures by removing whiskers, changing the colors of cats, adding something in the background, etc., and see what affects the cat score. The pixels with large $\left|\frac{\partial y_{\rm cat}}{\partial x_i}\right|$ are the ones that the network pays attention to. If we plot $\left|\frac{\partial y_{\rm cat}}{\partial x_i}\right|$ as an image, we can see which part of the picture is important for the decision; such a plot is called a \textit{saliency map}~\cite{Simonyan2014}.

If the training data is not good, it could happen that the NN pays too much attention to the background and decides anything in the sky is a bird, anything in water is a hippo, etc. Then, a cat jumping from one building to another could be recognized as a bird, or a cat playing in a pool could be recognized as a hippo.

%%%%%%%%%%%%
%%%%%%%%%%%%
\subsection{Adversarial attacks}
%%%%%%%%%%%%
%%%%%%%%%%%%
Suppose we started a business, like a sports gym, and introduced face recognition based on deep learning so that customers can enter without carrying a membership card or mobile app. Alice is a customer, but Bob is not. We do not want to allow Bob to enter the facility pretending to be Alice. To prevent this, we need to understand how Bob could potentially trick our neural network. Bob could wear a mask with Alice’s face printed on it, but that would make him look too suspicious to others, so it is not a realistic option. He would need to do something that does not seem strange to human observers. This sounds like a very difficult task. Surprisingly, however, Bob might be able to trick the neural network simply by wearing custom-made glasses! See the striking pictures in Ref.~\cite{Sharif2019}, where a person wearing such glasses was recognized as another person.

As another example, consider the scenario where malicious individuals attempt to confuse self-driving cars and cause accidents. They could manipulate a stop sign in such a way that it remains recognizable to humans, but a neural network might mistakenly identify it as a speed limit sign. Such attacks are extremely dangerous if they are feasible. Therefore, it is critical for car manufacturers to ensure that their autonomous driving systems are resilient to them.
Similarly, one could craft a specific sequence of words in a query to a chatbot, triggering harmful responses, such as instructions on making a bomb, while bypassing the system's safety and alignment mechanisms.

These are examples of \textit{adversarial attacks}. The study of such attacks is an important research field, which is necessary to build safe and trustworthy systems.
%%%%%%%%%%%%
%%%%%%%%%%%%
\subsubsection{Adversarial example generation}
\begin{figure}[htbp]
		\centering
		\scalebox{0.9}{
		\includegraphics[width=\textwidth]{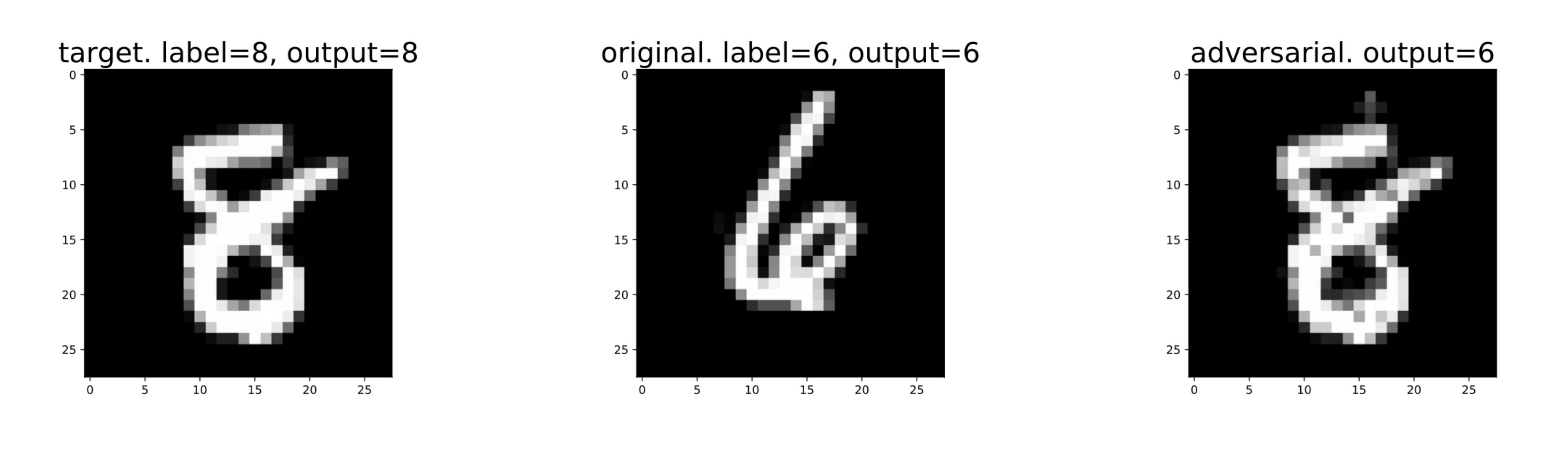}}
	\caption{
Adversarial example. We started from the original picture with label 6 and modified it so that the distance to the target with label 8 becomes smaller, while keeping the output unchanged.
	}\label{fig:adversarial_example}
\end{figure}

%%%%%%%%%%%%
%%%%%%%%%%%%
Let us see if we can trick the MNIST classifier.\footnote{We use a very simple method that needs only the output of the network, not its parameters or gradients (a `black-box' attack). More efficient methods use the derivative of the output with respect to the input, $\frac{\partial y_j}{\partial x_i}$, which we met in Sec.~\ref{sec:implicit_attention} (`white-box' attacks); a well-known example is the fast gradient sign method~\cite{Goodfellow:2014rpb}. A refined version of the random-walk strategy used here is known as the boundary attack~\cite{Brendel2018}.}
We use the three-layer network we built before. Our strategy is very simple. Firstly, we choose a target image. As a concrete example, we chose the image of 8 shown in Fig.~\ref{fig:adversarial_example}. Our goal is to make an image which looks like this target picture and at the same time can trick our NN. We do it by gradually modifying another image. We took an image of 6 (`original' in Fig.~\ref{fig:adversarial_example}), which is correctly recognized as 6 by the NN. Let the target and original images be $\vec{x}_{\rm target}$ and $\vec{x}_{\rm original}$, which are $28\times 28 = 784$-component vectors that take values in $[0, 1]^{784}\subset\mathbb{R}^{784}$. We want to find a trajectory in $[0, 1]^{784}$ that starts from $\vec{x}_{\rm original}$ and goes close to $\vec{x}_{\rm target}$, keeping the output of the NN unchanged, i.e., in such a way that any $\vec{x}$ on the trajectory is recognized as 6 by the NN. This can be achieved in the following manner:
\begin{itemize}
\item
Add a small random perturbation to $\vec{x}$ and make $\vec{x}'$.
\item
If $\vec{x}'$ is recognized as 6 and $|\vec{x}'- \vec{x}_{\rm target}|<|\vec{x}- \vec{x}_{\rm target}|$, then update $\vec{x}$ with $\vec{x}'$.
\item
Repeat.
\end{itemize}
An adversarial example obtained this way is also shown in Fig.~\ref{fig:adversarial_example}.

%%%%%%%%%%%%
%%%%%%%%%%%%
\section{Feature extraction}
%%%%%%%%%%%%
%%%%%%%%%%%%
So far, we have treated a neural network as a flexible function and discussed how to train it and how to test it. In this section, we ask what the hidden layers actually do. They convert the input into \textit{features} with which the task becomes easy, and these features are learned from data. To see why this matters, we first look at the kernel method, in which the features are designed by hand. We then see how features learned for one task can be reused for another (pretraining and transfer learning), which is the idea behind the large language models discussed later.
%%%%%%%%%%%%
%%%%%%%%%%%%
\subsection{Kernel method}
\begin{figure}[htbp]
\centering
\begin{tikzpicture}[scale=0.72,
  dot/.style={circle, fill, inner sep=0pt, minimum size=4pt},
  tri/.style={regular polygon, regular polygon sides=3, fill, inner sep=0pt, minimum size=5.5pt},
  crs/.style={cross out, draw, thick, inner sep=0pt, minimum size=4pt},
  axes/.style={-{Stealth[length=2mm]}, thick}]
% ---------- panel 1: classification, straight line ----------
\begin{scope}[shift={(0,0)}]
\draw[axes] (-0.2,0) -- (4.1,0) node[right]{$x$};
\draw[axes] (0,-0.2) -- (0,3.3) node[above]{$y$};
\draw[thick] (0.2,0.31) -- (3.6,3.03);
\foreach \p in {(0.5,2.2),(0.9,2.7),(1.4,2.5),(0.7,1.8),(1.9,2.8),(0.4,1.3),(1.2,2.0)} \node[dot] at \p {};
\foreach \p in {(1.9,0.9),(2.6,1.2),(3.2,1.6),(2.4,0.5),(3.5,0.9),(1.5,0.3),(3.0,0.6)} \node[tri] at \p {};
\end{scope}
% ---------- panel 2: classification, cubic curve ----------
\begin{scope}[shift={(5.2,0)}]
\draw[axes] (-0.2,0) -- (4.1,0) node[right]{$x$};
\draw[axes] (0,-0.2) -- (0,3.3) node[above]{$y$};
\draw[thick, domain=0.3:3.7, samples=60, smooth] plot (\x,{0.4*(\x-2)^3-0.5*(\x-2)+1.6});
\foreach \p in {(0.4,1.6),(0.8,2.3),(1.3,2.7),(1.0,1.9),(1.7,2.4),(2.2,2.7),(2.0,1.95),(2.7,2.2),(2.5,2.6),(0.5,2.6),(3.0,2.75),(1.55,2.95)} \node[dot] at \p {};
\foreach \p in {(0.9,0.6),(1.5,1.2),(2.0,0.8),(2.5,1.1),(3.0,0.5),(1.3,0.3),(2.3,0.35),(3.3,1.3),(3.7,1.6),(3.6,0.7),(1.0,1.15),(1.9,1.45)} \node[tri] at \p {};
\end{scope}
% ---------- panel 3: fit, straight line ----------
\begin{scope}[shift={(11.0,0)}]
\draw[axes] (-0.2,0) -- (4.1,0) node[right]{$x$};
\draw[axes] (0,-0.2) -- (0,3.3) node[above]{$y$};
\draw[thick, domain=0.3:3.7] plot (\x,{0.592*\x+0.3695});
\foreach \p in {(0.5,0.75),(0.9,0.77),(1.3,1.28),(1.75,1.35),(2.2,1.52),(2.6,2.03),(3.0,2.07),(3.45,2.48)} \node[crs] at \p {};
\end{scope}
% ---------- panel 4: fit, cubic curve ----------
\begin{scope}[shift={(16.2,0)}]
\draw[axes] (-0.2,0) -- (4.1,0) node[right]{$x$};
\draw[axes] (0,-0.2) -- (0,3.3) node[above]{$y$};
\draw[thick, domain=0.3:3.7, samples=60, smooth] plot (\x,{0.401*\x^3-2.3961*\x^2+4.2707*\x-0.5709});
\foreach \p in {(0.35,0.69),(0.6,1.11),(0.9,1.72),(1.25,1.74),(1.6,1.85),(2.0,1.5),(2.35,1.51),(2.7,1.33),(3.0,1.59),(3.3,1.75),(3.6,2.49)} \node[crs] at \p {};
\end{scope}
\end{tikzpicture}
\caption{[Left] Classifications with a straight line or a more complicated curve, say $z=1$ (dots) for $y > g(x) \equiv \theta_0+\theta_1x+\theta_2x^2+\theta_3x^3$ and $z = 0$ (triangles) for $y < g(x)$. [Right] Fitting data points by a straight line or a more complicated curve $y = g(x)$.}
\label{fig:classification-and-fit}
\end{figure}
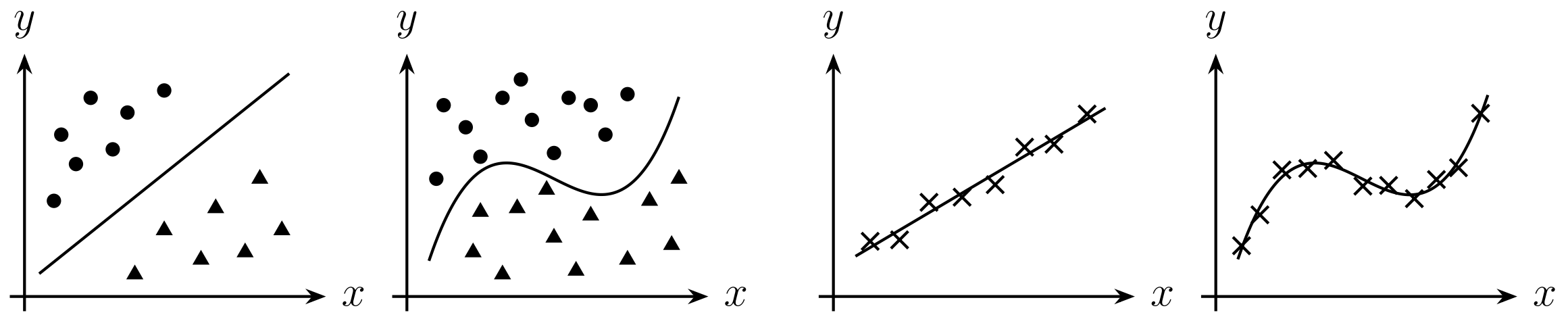

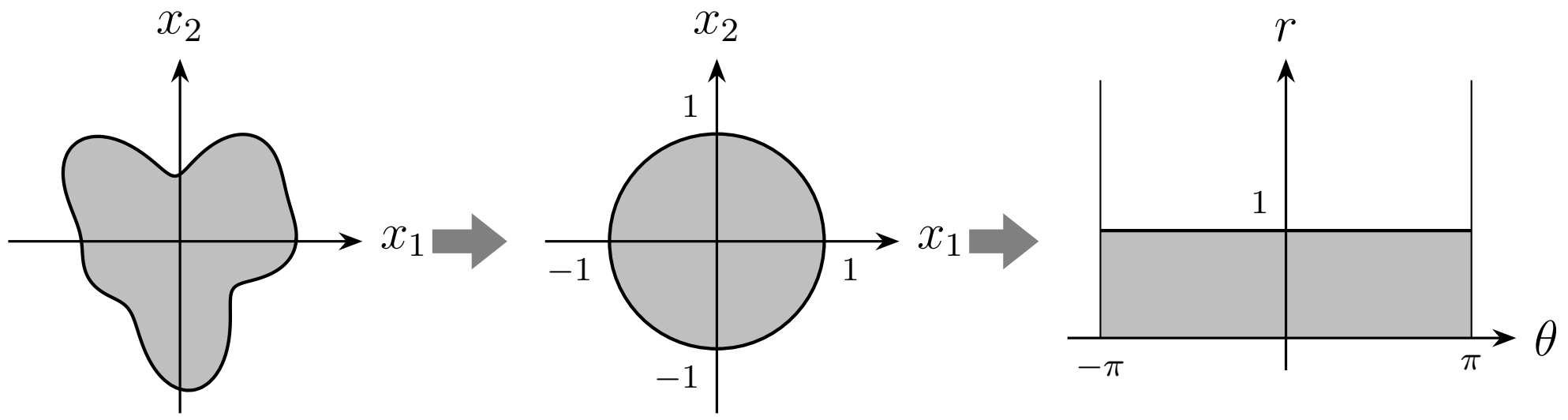
\begin{figure}[htbp]
\centering
% ===================== Top =====================
\begin{tikzpicture}[scale=0.9,
  declare function={R(\t)=1+0.25*sin(3*\t)+0.15*cos(5*\t+57);},
  axes/.style={-{Stealth[length=2mm]}, semithick},
  bigarrow/.style={-{Triangle[length=3mm,width=4.5mm]}, line width=2mm, gray}]
% complicated region
\begin{scope}
\fill[gray!50] plot[domain=0:360, samples=120, smooth cycle] ({R(\x)*cos(\x)},{R(\x)*sin(\x)});
\draw[thick] plot[domain=0:360, samples=120, smooth cycle] ({R(\x)*cos(\x)},{R(\x)*sin(\x)});
\draw[axes] (-1.6,0) -- (1.7,0) node[right]{$x_1$};
\draw[axes] (0,-1.6) -- (0,1.7) node[above]{$x_2$};
\end{scope}
\draw[bigarrow] (2.35,0) -- (3.05,0);
% disk
\begin{scope}[shift={(5.0,0)}]
\fill[gray!50] (0,0) circle (1);
\draw[thick] (0,0) circle (1);
\draw[axes] (-1.6,0) -- (1.7,0) node[right]{$x_1$};
\draw[axes] (0,-1.6) -- (0,1.7) node[above]{$x_2$};
\node[font=\scriptsize, below right] at (1,0) {$1$};
\node[font=\scriptsize, below left] at (-1,0) {$-1$};
\node[font=\scriptsize, above left] at (0,1) {$1$};
\node[font=\scriptsize, below left] at (0,-1) {$-1$};
\end{scope}
\draw[bigarrow] (7.35,0) -- (8.0,0);
% polar coordinates
\begin{scope}[shift={(10.3,-0.9)}, xscale=0.55]
\fill[gray!50] (-pi,0) rectangle (pi,1);
\draw[thick] (-pi,1) -- (pi,1);
\draw (-pi,0) -- (-pi,2.4);
\draw (pi,0) -- (pi,2.4);
\draw[axes] (-3.7,0) -- (3.9,0) node[right]{$\theta$};
\draw[axes] (0,-0.3) -- (0,2.6) node[above]{$r$};
\node[font=\scriptsize, below] at (-pi,0) {$-\pi$};
\node[font=\scriptsize, below] at (pi,0) {$\pi$};
\node[font=\scriptsize, above left] at (0,1) {$1$};
\end{scope}
\end{tikzpicture}
\caption{A classification of $\{(z^{(k)};x_1^{(k)},x_2^{(k)})\}_{k=1}^K$ with a complicated boundary. Suppose that $z^{(k)}=0$ or $z^{(k)}=1$ can be separated by a complicated curve; say $z^{(k)}=0$ are in the gray region. We could map such a curve with a complicated shape to a circle. Furthermore, we can use polar coordinates $(x_1, x_2) = (r\cos\theta, r\sin\theta)$ so that the two regions are classified by using a straight line $r = 1$. Note that such a map depends on the shape of the boundary, which we do not know in advance; in deep learning, a good feature map is learned from data.}
\label{fig:kernel-method}
\end{figure}

%%%%%%%%%%%%
%%%%%%%%%%%%
Let's learn about the kernel method~\cite{Aizerman67} to understand the advantage of neural networks.\footnote{The kernel method became well accepted after its application to Support Vector Machines~\cite{Boser92}.}
The targets are classification problems or fits of functions that require something more than a straight line or flat plane; see Fig.~\ref{fig:classification-and-fit} and Fig.~\ref{fig:kernel-method}. If we need only a straight line or flat plane, we can use linear regression. For concreteness, let us consider a fit of a function and suppose that $y=f(x;\vec{\theta})=\theta_0+\theta_1x+\theta_2x^2+\theta_3x^3$ is a reasonable ansatz. An almost trivial but crucial fact is that $f(x;\vec{\theta})$ is still linear in $\vec{\theta}$. This problem is nothing but linear regression, with $x_1=x$, $x_2=x^2$ and $x_3=x^3$. 

If we pretend that $x$, $x^2$ and $x^3$ are independent variables and plot $y$ as a function of $x_1$, $x_2$ and $x_3$, the classification problem depicted in Fig.~\ref{fig:classification-and-fit} can be solved by using a hyperplane $\theta_0+\theta_1x_1+\theta_2x_2+\theta_3x_3$. The point here is that $x_1=x$, $x_2=x^2$ and $x_3=x^3$ are good variables that enable us to use the linear regression ansatz. 

To generalize this idea, let us introduce a feature map $\vec{\phi}:\mathbb{R}^d\to\mathbb{R}^p$.
In the example above, $d = 1$ and $p = 4$, and $\vec{\phi}(x)=\begin{pmatrix}1 \\ x \\ x^2 \\ x^3\end{pmatrix}$. Then, the cost function can be defined as\footnote{
This cost function is for a fit like the one in the right panel of Fig.~\ref{fig:classification-and-fit}. For the classification problems shown in the left panel of Fig.~\ref{fig:classification-and-fit}, we can use, e.g., $J(\vec{\theta})=\frac{1}{2}\sum_k[z^{[k]}-\sigma(\alpha(y^{[k]}-g(x^{[k]})))]^2$, where $\sigma$ is the sigmoid function and $\alpha$ is a tunable parameter. By interpreting $\vec{x}=(x,y)$, we can regard it as a special case of $J(\vec{\theta})=\frac{1}{2}\sum_k[z^{[k]}-\sigma(\vec{\theta}\cdot\vec{\phi}(x^{[k]}))]^2=\frac{1}{2}\sum_k[z^{[k]}-\sigma(f(x^{[k]};\vec{\theta}))]^2$. With this cost function, the rest of the section can be repeated with only small changes.
}
\begin{align}
J(\vec{\theta})
=
\frac{1}{2}\sum_k
\left(
y^{[k]}-f(x^{[k]};\vec{\theta})
\right)^2\, ,  
\end{align}
where
\begin{align}
f(x;\vec{\theta})
=
\vec{\theta}\cdot\vec{\phi}(x)\, . 
\label{ansatz:kernel-method}
\end{align}

This $f$ is linear both in $\vec{\theta}$ and $\vec{\phi}$. By using a feature map $\vec{\phi}$, the original dataset $\{(\vec{x}^{[k]}, y^{[k]})\}_{k=1}^K$ is converted to a new dataset $\{(\vec{\phi}(\vec{x}^{[k]}), y^{[k]})\}_{k=1}^K$. 
If $y$ can be fit by a hyperplane defined by the right-hand side of \eqref{ansatz:kernel-method}, linear regression works. In the case of the classification problem, if $z=0$ and $z=1$ can be separated by such a hyperplane, linear regression works. 

``Good features" would allow us to use linear regression. 
Of course, we need to design a good feature map so that the new dataset can be analyzed precisely by using the linear ansatz. To achieve enough flexibility and computational efficiency, let us abstract the formulation a little bit. We allow the dimension $p$ of the feature vectors to be large; we even allow $p=\infty$ because we do not want to restrict the power of the algorithm. For example, we can allow $x_1^{n_1}x_2^{n_2}\cdots x_d^{n_d}$ with any $n_i\ge 0$ so that we can deal with the Taylor expansion of generic functions.

How can we treat infinitely many features $\vec{\phi}(\vec{x})\in\mathbb{R}^{p=\infty}$ and fit parameters $\vec{\theta}\in\mathbb{R}^{p=\infty}$? The kernel trick enables us to do this seemingly impossible task. The key fact is:
\begin{screen}
If we update $\vec{\theta}$ with gradient descent starting from $\vec{\theta}=\vec{0}$, only a finite number of parameters are needed. Specifically, we can write $\vec{\theta}$ as $\vec{\theta}=\sum_{k=1}^K\beta_k\vec{\phi}(x^{[k]})$.
\end{screen}

It is straightforward to check this. Because this is true for the initial value $\vec{\theta}=\vec{0}$, and because the update is of the form $\vec{\theta}\to\vec{\theta}-\epsilon\frac{\partial J}{\partial\vec{\theta}}$, we only have to check that $\frac{\partial J}{\partial\vec{\theta}}$ is a linear combination of the feature vectors $\vec{\phi}(\vec{x}^{[k]})$ if $\vec{\theta}$ can be written as $\vec{\theta}=\sum_{k=1}^K\beta_k\vec{\phi}(x^{[k]})$. And this is indeed the case:
\begin{align}
\frac{\partial J}{\partial\vec{\theta}}
&=
-
\sum_{k=1}^K
\left(
y^{[k]}
-
\vec{\theta}\cdot\vec{\phi}(\vec{x}^{[k]})
\right)
\vec{\phi}(\vec{x}^{[k]})
\nonumber\\
&=
-
\sum_{k=1}^K
\left(
y^{[k]}
-
\sum_{j=1}^K\beta_j\mathcal{K}_{jk}
\right)
\vec{\phi}(\vec{x}^{[k]})\, , 
\qquad
\mathcal{K}_{jk}
=
\vec{\phi}(\vec{x}^{[j]})\cdot\vec{\phi}(\vec{x}^{[k]})\, . 
\end{align}
Here, $\mathcal{K}$ means \textit{kernel}. (This is not related to the number of data points $K$. I’m sorry for the confusing notation!) Therefore, we need only $K$ numbers $\beta_1,\cdots,\beta_K$. Furthermore, the update rule of $\beta$ is
\begin{align}
\beta_k
\to
\beta_k
+
\epsilon\cdot
\left(
y^{[k]}
-
\sum_{j=1}^K
\beta_j\mathcal{K}_{jk}
\right)\, , 
\end{align}
and the cost function is
\begin{align}
J(\vec{\theta})
=
\frac{1}{2}
\sum_{k=1}^K
\left(
y^{[k]}
-
\vec{\theta}\cdot\vec{\phi}(\vec{x}^{[k]})
\right)^2
=
\frac{1}{2}
\sum_{k=1}^K
\left(
y^{[k]}
-
\sum_{j=1}^K
\beta_j\mathcal{K}_{jk}
\right)^2\, , 
\end{align}
and hence, we do not even have to use the feature vectors $\vec{\phi}(x^{[k]})$ explicitly. If we can calculate $\mathcal{K}_{jk}=\vec{\phi}(x^{[j]})\cdot\vec{\phi}(x^{[k]})$ efficiently (and for the purpose of learning, we only need to do it once before starting gradient descent), that’s all we need.
The same is true for the prediction: for a new input $\vec{x}$, we have $f(\vec{x})=\vec{\theta}\cdot\vec{\phi}(\vec{x})=\sum_{k=1}^K\beta_k\,\mathcal{K}(\vec{x}^{[k]},\vec{x})$, where $\mathcal{K}(\vec{x},\vec{x}')\equiv\vec{\phi}(\vec{x})\cdot\vec{\phi}(\vec{x}')$.

Therefore, we can start with a kernel function $\mathcal{K}:\mathbb{R}^d\times\mathbb{R}^d\to\mathbb{R}$. Although $\mathcal{K}(\vec{x},\vec{x}')$ is interpreted as $\vec{\phi}(\vec{x})\cdot\vec{\phi}(\vec{x}')$, we do not have to know $\vec{\phi}: \mathbb{R}^d\to\mathbb{R}^p$ explicitly. A typical example is the polynomial kernel $\mathcal{K}(\vec{x},\vec{x}')=(1+\vec{x}\cdot\vec{x}')^n$. It corresponds to the features $x_1^{n_1}x_2^{n_2}\cdots x_d^{n_d}$ with $n_1+\cdots+n_d\le n$; for $d=100$ and $n=5$, there are about $10^8$ of them, but we only need to compute one inner product in $\mathbb{R}^d$. Another example is the Gaussian kernel $\mathcal{K}(\vec{x},\vec{x}')=e^{-|\vec{x}-\vec{x}'|^2/(2\sigma^2)}$, which corresponds to $p=\infty$.\footnote{For $d=1$, by using $e^{-(x-x')^2/(2\sigma^2)}=e^{-x^2/(2\sigma^2)}e^{-x'^2/(2\sigma^2)}\sum_{n=0}^\infty\frac{(xx')^n}{\sigma^{2n}n!}$, we find $\phi_n(x)=e^{-x^2/(2\sigma^2)}\frac{x^n}{\sigma^n\sqrt{n!}}$ ($n=0,1,2,\cdots$).}

A kernel function $\mathcal{K}$ which actually comes from a certain feature map $\vec{\phi}$ is called a valid kernel. A necessary and sufficient condition for a kernel function $\mathcal{K}$ to be valid is that the matrix $\mathcal{K}_{jk}=\mathcal{K}(\vec{x}^{[j]},\vec{x}^{[k]})$ constructed from any input data $\vec{x}^{[1]},\cdots,\vec{x}^{[K]}$ is symmetric and positive semidefinite. (Note that the number of data points $K$ is also arbitrary.)

The kernel trick can be powerful if we are smart enough to design a good kernel for each problem. However, it is safer to assume that we are not smart! In deep learning, a kernel is also learned. More precisely: we humans specify the neural network, but the parameters in the neural network are learned from data. This allows us humans, who may not be very smart, to obtain a good feature map. 
In Fig.~\ref{fig:DNN-cartoon}, the last hidden layer $\vec{x}^{(M)}$ is a feature vector $\vec{\phi}(\vec{x}^{(0)})$. Hence, $p = d_M$.
%%%%%%%%%%%%
%%%%%%%%%%%%
\subsection{Feature extraction and pretraining/transfer learning}\label{sec:pretraining}
%%%%%%%%%%%%
%%%%%%%%%%%%
By interpreting the role of the hidden layers as feature extraction, we naturally arrive at the notion of pretraining and transfer learning.\footnote{Pretraining and transfer learning are closely related to each other, and I could not find a clear border between them. It seems that the word ‘pretraining’ is used more often if the new task is very different from the original one.}
%%%%%%%%%%%%
%%%%%%%%%%%%
\subsubsection{Supervised learning for pretraining/transfer learning}
%%%%%%%%%%%%
%%%%%%%%%%%%
Suppose you trained a neural network to classify handwritten letters of the English alphabet. What we expect is that the hidden layers give a feature map $\vec{\phi}_{\vec{\theta}}: \vec{x}^{(0)}\to\vec{x}^{(M)}$. The last step assigns a probability to each letter. Equivalently, the last step gives the conditional probability $p_{u}(i|\vec{x}^{(M)})$, where $i = 1, \cdots , 26 = A, \cdots , Z$. Here, $u$ stands for the parameters; we could use softmax $p_{u}(i|\vec{x}^{(M)})=\frac{e^{\vec{u}_i\cdot\vec{x}^{(M)}}}{\sum_j e^{\vec{u}_j\cdot\vec{x}^{(M)}}}$.

It is not a crazy idea to think the feature map $\vec{\phi}_{\vec{\theta}}$ can work for other writing systems to some extent. There is a good chance that it works for German or Spanish, and it may work for other languages as well. If so, we can replace only the last step and use the same parameters $\vec{\theta}$, or we can also fine-tune $\vec{\theta}$ by using the optimal value for English as the initial condition. Even for German or Spanish, we need to replace $\vec{u}$ because the number of letters is different.\footnote{However, we could use the same $\vec{u}_i$ for the letters shared by the two languages.}

The same idea could work for image recognition. Suppose there is a neural network for animals' activities (a dog or a cat eating, running, etc.). It may work for human activities, with some fine-tuning of $\vec{\theta}$. If we teach chimpanzees to watch TV, play baseball, etc., and build a neural network to recognize their activities, then we could use the optimal $\vec{\theta}$ for chimps’ activities to extract the features of our own activities. (In this case, we could even reuse the optimal $u$.)

Image recognition for cancer detection would be a more useful application.
%%%%%%%%%%%%
%%%%%%%%%%%%
\subsubsection{Self-supervised learning for pretraining}\label{sec:self_supervised_learning}
%%%%%%%%%%%%
%%%%%%%%%%%%
Let us take image recognition as an example. Suppose there are many unlabeled pictures $\{\vec{x}^{(1)},\cdots,\vec{x}^{(K)}\}$. We want to find a feature map $\vec{\phi}_{\vec{\theta}}:\mathbb{R}^d\to\mathbb{R}^p$ that extracts `features' of the data. But how can we define reasonable `features' without labels?

\textit{Data augmentation} is a cool way to create a kind of `label'. Let us modify the original data $\vec{x}^{(k)}$ a little bit by changing colors, rotating a little bit, or randomly cropping a part of the picture, and make new pictures $\vec{x}^{\prime(k)}$ and $\vec{x}^{\prime\prime(k)}$. (If you like, you can make more than two new pictures and generalize the discussion below very easily. See the cute dog pictures in Ref.~\cite{Chen2020}.) Then, it is natural to assume $\vec{x}^{\prime(k)}$ and $\vec{x}^{\prime\prime(l)}$ share some `features' when $k = l$, while they do not share `features' when $k\neq l$. (There can be rare cases in which different original pictures share some features, but we neglect them.) We call $(\vec{x}^{\prime(k)},\vec{x}^{\prime\prime(k)})$ a positive pair, and $(\vec{x}^{\prime(k)},\vec{x}^{\prime\prime(l)})$  ($k\neq l$) a random pair or negative pair. Positive or negative can be regarded as a label, say $y^{(k=l)} = 1$ for a positive pair and $y^{(k\neq l)} = 0$ for a negative pair. We want the positive pairs to be mapped to nearby points in $\mathbb{R}^p$, while we do not want random pairs to be mapped to nearby points. Therefore, we try to make $\vec{\phi}_{\vec{\theta}}(\vec{x}^{\prime(k)})\cdot\vec{\phi}_{\vec{\theta}} (\vec{x}^{\prime\prime(k)})$ close to $y^{(k=l)} = 1$ and  $\vec{\phi}_{\vec{\theta}}(\vec{x}^{\prime(k)})\cdot\vec{\phi}_{\vec{\theta}} (\vec{x}^{\prime\prime(l)})$ ($k\neq l$) close to $y^{(k\neq l)} = 0$. (Here, we assume that $\vec{\phi}_{\vec{\theta}}$ is normalized to unit length.) We can use the network obtained this way as a pretrained model and tune it for specific tasks. This approach is called \textit{contrastive learning}~\cite{Chen2020}.

Note that the data augmentation described above requires a large dataset, and the chance that a random pair is not really random must be small. If we apply this technique to the classification of handwritten numbers from 0 to 9, a ``random" pair would not really be random 10\% of the time (it could be a pair of my $\mathit{2}$ and your $\mathnormal{2}$), and hence the notions of positive pair and random pair do not make sense.\footnote{The data augmentation technique is often used simply to increase the training data in the usual supervised learning. This remark does not apply to such a case.}

Another application is to natural language processing (NLP). Suppose that you have a large dataset collected from Google, Wikipedia, etc. Furthermore, suppose that you have a list of all words appearing in the dataset.\footnote{The actual situation is more complicated, but we assume this for simplicity.} Then, we can take a sentence, remove one word randomly, and train the neural network so that it can reproduce the removed word. Imagine sentences like
\begin{itemize}
\item
Alice called \underline{her} father.
\item
Bob loves \underline{his} dog.
\item
Many people are proud of \underline{their} backgrounds.
\end{itemize}

Here, the words with underlines are removed, and the neural network is trained to reproduce them. Then, in addition to the concept of the possessive case, the neural network can also learn about common female names, common male names, singular, plural, etc. The neural network can learn about geography from sentences like
\begin{itemize}
\item
The UK and \underline{France} are next to each other but do not share a land border.
\item
The UK consists of England, \underline{Northern Ireland}, Scotland, and Wales.
\item
The land border between Northern Ireland and \underline{the Republic of Ireland} was a source of headaches for Theresa May.
\end{itemize}

%%%%%%%%%%%%
%%%%%%%%%%%%
\part{Natural Language Processing}
%%%%%%%%%%%%

%%%%%%%%%%%%%%%%%%%%%%%%%%
%%%%%%%%%%%%%%%%%%%%%%%%%%
\section{Language models: from $n$-gram to RNN}
%%%%%%%%%%%%%%%%%%%%%%%%%%
%%%%%%%%%%%%%%%%%%%%%%%%%%

%%%%%%%%%%%%
%%%%%%%%%%%%
\subsection{Basic task}
%%%%%%%%%%%%
%%%%%%%%%%%%
There can be many tasks in Natural Language Processing (NLP), but unless otherwise stated, we consider the following. Suppose that there is a big corpus $\Gamma$ (a set of texts) consisting of sentences made of $d_\Gamma$ words. The corpus may be scans of books from libraries, web pages, or entire pages of Wikipedia. We use $x = 1,2,\cdots,d_\Gamma$ to denote words (e.g., $x = 1$ is apple, $x = 2$ is avocado, $x = 3$ is Britain, etc.), and $(x_1,x_2,\cdots,x_T)$ represents a text made of $T$ words.\footnote{It is convenient to include the end-of-sentence symbol $\langle\mathrm{end}\rangle$ as a ‘word’.}$^,$\footnote{Actual language models do not use words as the unit. They split the text into \textit{tokens}, which can be a word, a part of a word, or a punctuation mark, so that any text, including rare words and typos, can be expressed by a vocabulary of moderate size (typically tens of thousands of tokens). For simplicity, we do not distinguish between words and tokens in these notes.} 
We want to estimate the joint probability $p(x_1,x_2,\cdots,x_T)$ for a given sentence to appear among $(d_\Gamma)^T$ possible ``sentences". (Of course, many of them do not make sense. ``I am a cat" is perfectly fine,\footnote{This is the title of one of the most important novels in Japanese history.} but ``Cat am a I" is not really a sentence.) Such a probability can be expressed by using the conditional probability $p(x_t|x_1,\cdots,x_{t-1})$, which represents the probability of $x_t$ following $(x_1,\cdots,x_{t-1})$, as 
\begin{align}
p(x_1,\cdots,x_{t-1},x_t) 
&= 
p(x_t|x_1,\cdots,x_{t-1})
\cdot
p(x_1,\cdots,x_{t-1})
\nonumber\\
&= 
p(x_t|x_1,\cdots,x_{t-1})
\cdot
p(x_{t-1}|x_1,\cdots,x_{t-2})
\cdot
p(x_1,\cdots,x_{t-2})
\nonumber\\
&= 
p(x_t|x_1,\cdots,x_{t-1})
\cdot
p(x_{t-1}|x_1,\cdots,x_{t-2})
\cdots 
p(x_2|x_1)\cdot p(x_1)\, .
\end{align}
Therefore, our task is to infer $p(x_t|x_1,\cdots,x_{t-1})$.
%%%%%%%%%%%%
%%%%%%%%%%%%
\subsubsection{Learning knowledge from next-word prediction}
%%%%%%%%%%%%
%%%%%%%%%%%%
To predict the next word with precision, extensive background knowledge is required. For instance, consider the phrase \textit{Messi is greater than}. We can immediately imagine \textit{Maradona} or \textit{Ronaldo}. We think of Maradona because we know that Lionel Messi, a renowned footballer from Argentina, is often compared to Diego Maradona, another legendary Argentinian player. We think of Ronaldo because, from 2008 to 2017, either Messi or Cristiano Ronaldo won the prestigious Ballon d'Or every year. However, if this phrase follows a discussion about the 2022 World Cup final, the next word might be \textit{Mbapp\'{e}} instead. Alternatively, because so many people would want to watch Brazil versus Argentina, the next word could be Pel\'{e}, Zico, or Neymar. 
Or, perhaps, superstars from different fields are compared, such as LeBron James, Shohei Ohtani, or Albert Einstein.\footnote{
We can imagine sentences like ``Einstein might have cracked the code to the universe, but Messi? He’s out here bending the laws of nature like it’s no big deal. Messi is greater than Einstein." 
}
To accurately predict the next word, one must possess a deep understanding of the context and subject matter.

At the end of Sec.~\ref{sec:self_supervised_learning}, we considered masked word prediction rather than next-word prediction. Both are useful for learning knowledge from the corpus, but they serve different purposes. An advantage of next-word prediction is that it can be used to generate sentences, as we will see later. This opens the door to various applications. For example, we can ask a question and receive an answer, demonstrating the generative capabilities of the model.

%%%%%%%%%%%%
%%%%%%%%%%%%
\subsection{$n$-gram model}
%%%%%%%%%%%%
%%%%%%%%%%%%
The $n$-gram model goes back to Shannon~\cite{Shannon1948}. In the $n$-gram model, only $n-1$ words are used to infer the next word. Namely, $p(x_t|x_1,\cdots,x_{t-1})$ is replaced with $p(x_t|x_{t-n+1},\cdots,x_{t-1})$. Furthermore, the `probability' is determined literally by counting the number of appearances of $(x_{t-n+1},\cdots,x_{t-1},x_t)$, i.e.,
\begin{align}
p(x_t|x_{t-n+1},\cdots,x_{t-1})
=
\frac{\#\, \mathrm{of}\, (x_{t-n+1},\cdots,x_{t-1},x_t)\, \mathrm{in\, the\, corpus}}{\#\, \mathrm{of}\, (x_{t-n+1},\cdots,x_{t-1})\, \mathrm{in\, the\, corpus}}\, . 
\end{align}
If $n$ can be very large so that many pages of a book can fit into $n$ words, and if the corpus is large enough that any reasonable sentence appears multiple times, the $n$-gram method should work perfectly. In practice, however, $n$ cannot be large enough for this to work, because we cannot find a sufficiently large corpus. Indeed, if you type a reasonable sentence into Google, usually you do not find an exact match. For example, in December 2023, ``I was born in Yokohama and went to a high school in Tokyo" did not have an exact match, although the population of Yokohama is close to four million, there are more than 400 high schools in Tokyo, and it takes less than 30 minutes from Yokohama station to Tokyo station. ``I was born in London and studied at Queen Mary" did not have an exact match either, although this sentence applies to tens of thousands of people. We could find only short and generic sentences like ``I was a panda" (851,000 matches) or ``I am a spy" (1,260,000 matches). The number drops quickly as we add more words; there were 5 matches for ``I believe I was a panda" (only 2 independent examples after removing the same sources), 6 matches for ``you thought I was a panda" (5 independent examples), and 6 matches for ``I am a spy but you". Realistically, $n= 4$ or 5 is the largest possible value we can use.

When $n$ is small, the $n$-gram model cannot handle the context of a sentence. We can easily imagine that $n=1$ and $n= 2$ cannot create reasonable sentences. How about $n= 3$? Let us take ``London is the biggest city in England" as an example. When ``London is the biggest city in" is given, we can imagine that words like England, the UK, Great Britain, or western Europe should follow, but the 3-gram model can see only ``city in", and ``London" drops off. Therefore, sentences like ``London is the biggest city in India" can be generated.

The $n$-gram model is a Markov model: it uses the ordering of the last $n-1$ words, but ignores everything before them.

%%%%%%%%%%%%
%%%%%%%%%%%%
\subsection{Word vector (Word embedding)}
%%%%%%%%%%%%
%%%%%%%%%%%%
For natural language processing, we want to represent words by using numbers. We could assign a number $x=1,2,\cdots,d_\Gamma$ to each word as we did above. Equivalently, we could use a one-hot vector in $\mathbb{R}^{d_\Gamma}$ whose $x$-th component is 1 and the other components are zero.

One-hot vectors are convenient for input and output. Also, we can imagine something like $\vec{x}_{\rm apple}+\vec{x}_{\rm Britain}$ if you are talking about both apple and Britain.\footnote{This combination appears frequently thanks to Newton and the Beatles.
} However, there are several issues, including:
\begin{itemize}
\item
The dimension of the vector space becomes too large.
\item
There is no natural way to define the similarity of words. The distance between vectors could be the distance between two words, but the distance between any two one-hot
vectors is $\sqrt{2}$. 
\end{itemize}
These two issues are clearly related. By assigning similar vectors to similar words, we can reduce the dimension of the vector space. It also (partially) resolves the problem we have seen above, i.e., that exact matches are extremely rare.

Therefore, it is more convenient if we use a lower-dimensional vector space $\mathbb{R}^d$ (say $d=500$ or 1000) to represent the words in such a way that similar words sit close to each other. A simple and effective method to find such an embedding is to look at the words appearing nearby and ignore everything else.

Word2Vec~\cite{Mikolov2013} is defined as follows. The basic idea is not too different from the $n$-gram model. The difference is the use of word vectors (equivalently, word embeddings) in $\mathbb{R}^d$ and a reasonable parametrization of the model that allows us to avoid `exact match'. For each word $x$, we introduce $\vec{u}_x\in\mathbb{R}^d$ and $\vec{v}_x\in\mathbb{R}^d$ as parameters of the model. (In principle, we can take $\vec{u}_x$ and $\vec{v}_x$ to be the same, but usually they are treated as independent parameters to make the math simpler.) The probability $P(x_{t+j}|x_t;\vec{\theta})$ is modeled by
\begin{align}
P(x_{t+j}|x_t;\vec{\theta})
=
\frac{e^{\vec{u}_{x_{t+j}}\cdot\vec{v}_{x_t}}}{\sum_{x=1}^{d_\Gamma}e^{\vec{u}_{x}\cdot\vec{v}_{x_t}}}\, . 
\end{align}
Here, $\vec{\theta}$ stands for all parameters, i.e., all $\vec{u}_x$ and all $\vec{v}_x$. We see the relationship of the words up to $m$-step separation, by introducing the likelihood function
\begin{align}
L(\vec{\theta})
=
\prod_t
\prod_{\substack{j=-m\\ j\neq 0}}^m
P(x_{t+j}|x_t;\vec{\theta})
\end{align}
and the cost function
\begin{align}
J(\vec{\theta})
=
-\log L(\vec{\theta})
=
-
\sum_t
\sum_{\substack{j=-m\\ j\neq 0}}^m
\log P(x_{t+j}|x_t;\vec{\theta})\, . 
\end{align}
By minimizing this cost function, we can obtain word vectors.

Word2Vec shows many interesting features. For example, $\vec{v}_{\rm queen}-\vec{v}_{\rm woman}+\vec{v}_{\rm man}\simeq\vec{v}_{\rm king}$ and
$\vec{v}_{\rm France}-\vec{v}_{\rm Paris}+\vec{v}_{\rm Rome}\simeq\vec{v}_{\rm Italy}$. 
%%%%%%%%%%%%
%%%%%%%%%%%%
\subsection{Recurrent Neural Network (RNN)}\label{sec:RNN}
\begin{figure}[htbp]
\centering
\begin{tikzpicture}[>=Stealth,
  box/.style={draw, thick, minimum width=11mm, minimum height=11mm, inner sep=1pt},
  lab/.style={font=\footnotesize}]
% h_0
\node[box] (h0) at (0,0) {$\vec{h}_0$};
% columns: x-position / subscript
\foreach \x/\s/\n in {2.2/1/1, 4.4/2/2, 7.8/{t-1}/a, 10.0/t/b} {
  \node[box] (h\n) at (\x,0)    {$\vec{h}_{\s}$};
  \node[box] (e\n) at (\x,-2.2) {$\vec{e}_{\s}$};
  \node      (x\n) at (\x,-3.9) {$\vec{x}_{\s}$};
  \node      (y\n) at (\x,1.7)  {$\vec{y}_{\s}$};
  \draw[->, thick] (x\n) -- node[lab, left]{$E$}   (e\n);
  \draw[->, thick] (e\n) -- node[lab, left]{$W_e$} (h\n);
  \draw[->, thick] (h\n) -- node[lab, left]{$D$}   (y\n);
}
% recurrence
\draw[->, thick, blue] (h0) -- node[lab, above]{$W_h$} (h1);
\draw[->, thick, blue] (h1) -- node[lab, above]{$W_h$} (h2);
\draw[->, thick, blue] (h2) -- node[lab, above]{$W_h$} (5.75,0);
\node at (6.1,0) {$\cdots$};
\draw[->, thick, blue] (6.45,0) -- node[lab, above]{$W_h$} (ha);
\draw[->, thick, blue] (ha) -- node[lab, above]{$W_h$} (hb);
\draw[->, thick, blue] (hb) -- node[lab, above]{$W_h$} (11.4,0);
\node at (11.75,0) {$\cdots$};
\end{tikzpicture}
\caption{A Recurrent Neural Network for natural language processing. The same matrices $E$, $W_e$, $W_h$ and $D$ are used at every step.}
\label{fig:RNN-architecture}
\end{figure}
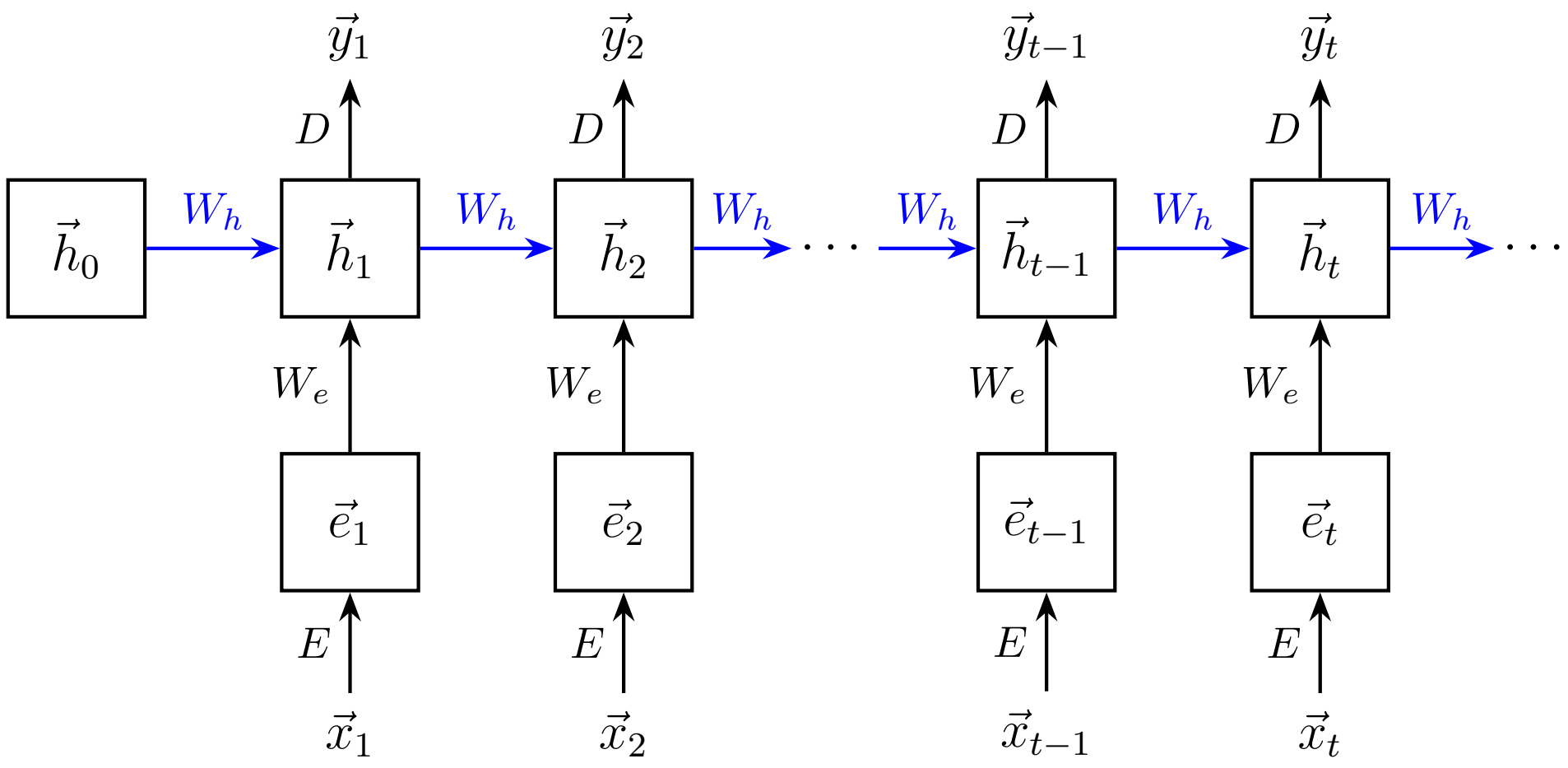

%%%%%%%%%%%%
%%%%%%%%%%%%
The $n$-gram model and Word2Vec look only at a few nearby words, and Word2Vec does not even use the order of these words. To predict the next word well, we need a model that reads the words in order and keeps the information about all the words it has read so far.
The recurrent neural network (RNN) is an architecture that receives sequential input
of any length and applies the same transformation with the same weights at each step.
Specifically, for input $\vec{e}_t$ ($t= 1,2,\cdots,T$), $\vec{h}_t$ is obtained recursively as
\begin{align}
\vec{h}_t 
= 
\sigma\left(
W_h\vec{h}_{t-1}
+
W_e\vec{e}_{t}
+
\vec{b}
\right)\, . 
\end{align}
The initial value $\vec{h}_0$ is usually taken to be zero:
\begin{align}
\vec{h}_0
=
\vec{0}\, . 
\end{align}
For natural language processing, typically we transform a word vector $\vec{x}_t$ (which we assumed to be a one-hot vector) to $\vec{e}_t$,
\begin{align}
\vec{e}_t
=
E
\vec{x}_t
\label{eq:word_embedding}
\end{align}
by using a good encoding scheme (which could be obtained separately by using another method; otherwise the choice of $E$ and $W_e$ can be combined into one matrix), and generate the output word $\vec{y}_t$ from $\vec{h}_t$ as
\begin{align}
\vec{y}_t=\mathrm{softmax}
\left(
D\vec{h}_t
+
\vec{b}'
\right)\, . 
\end{align}
See Fig.~\ref{fig:RNN-architecture}. The basic intuition is that $\vec{h}_t$ has the information about $\vec{e}_1,\cdots,\vec{e}_t$. This would mean the network keeps the \textit{short-term memory} in the sentence or conversation. 

In reality, of course, the memory vanishes due to the vanishing gradient problem. To see how the
vanishing gradient problem arises for RNN, let us see how $\vec{h}_t$ affects $\vec{h}_{t'}$ ($t' > t$). Because
of the chain rule, we have
\begin{align}
\frac{\partial\vec{h}_{t'}}{\partial\vec{h}_t}
=
\frac{\partial\vec{h}_{t'}}{\partial\vec{h}_{t'-1}}
\cdots
\frac{\partial\vec{h}_{t+2}}{\partial\vec{h}_{t+1}}
\frac{\partial\vec{h}_{t+1}}{\partial\vec{h}_t}\, . 
\label{eq:RNN_chain_rule}
\end{align}
When $t'-t$ is large, the right-hand side is (a sum of) a product of many derivatives, which tends to become exponentially small (vanishing gradient problem) or large (exploding gradient problem). If such a vanishing gradient problem exists, then $\vec{h}_{t'}$ is not sensitive to a change in $\vec{h}_t$, and hence, $\vec{h}_{t'}$ does not have the memory of $\vec{h}_t$.\footnote{Imagine a long paragraph like ``I left my house in London in the morning. I went to Manchester, watched a football game, ... blah blah blah.... After finishing all errands, I went back to London." Suppose we give all words but the final one (London) to a simple RNN and see if the right answer is obtained. If the paragraph is long, the RNN could completely forget the first sentence and fail to give the right answer.}$^,$\footnote{The RNN is particularly unstable and tends to have the vanishing or exploding gradient problem because the same pattern repeats, i.e., the right-hand side of \eqref{eq:RNN_chain_rule} is similar to $M^{t'-t}$ with some matrix $M$, whose eigenvalues are $\lambda^{t'-t}$ (where $\lambda$ is an eigenvalue of $M$), which converge to zero or diverge unless $|\lambda|= 1$.}

In this sense, only `short' short-term memory is kept, unless some trick to keep the memory better is introduced. (Typically, the memory is lost within ten steps or so.)
%%%%%%%%%%%%
%%%%%%%%%%%%
\subsubsection{Training RNN}
%%%%%%%%%%%%
%%%%%%%%%%%%
$\vec{y}_t=\mathrm{softmax}\left(D\vec{h}_t+\vec{b}'\right)$ represents the prediction of the model for the next word, i.e., the $k$-th component $y_{t,k}$ is the probability of $x_{t+1}=k$ according to the prediction. Therefore, we can define the cross-entropy loss as a cost function:
\begin{align}
J(W_h,W_e,\vec{b},E,D,\vec{b}') 
=
-\sum_{(x_1,\cdots,x_n)}
\sum_{t=t_0}^{n-1}
\log(\vec{x}_{t+1}\cdot\vec{y}_t)\, . 
\end{align}
Here, the sum $\sum_{(x_1,\cdots,x_n)}$ is taken over actual texts in the training data. We assumed that the model is trained to predict the words starting from $t=t_0$.

To take the derivative of the cost function, we need the derivative of $\vec{h}_t$. Schematically,
\begin{align}
\frac{\partial\vec{h}_t}{\partial W_h}
=
\sigma'\left(
W_h\vec{h}_{t-1}
+
W_e\vec{e}_{t}
+
\vec{b}
\right)
\times
\left(
\vec{h}_{t-1}
+
W_h
\frac{\partial\vec{h}_{t-1}}{\partial W_h}
\right)\, . 
\end{align}
More explicitly, 
\begin{align}
\frac{\partial h_{t,i}}{\partial (W_h)_{jk}}
=
\sigma'\left(
(
W_h\vec{h}_{t-1}
+
W_e\vec{e}_{t}
+
\vec{b}
)_i
\right)
\times
\left(
\delta_{ij}
h_{t-1,k}
+
\sum_l
(W_h)_{il}
\frac{\partial h_{t-1,l}}{\partial (W_h)_{jk}}
\right)\, . 
\end{align}
By using this relation, we can calculate $\frac{\partial J}{\partial W_h}$ recursively. Derivatives with respect to other parameters can be calculated in a similar manner, by using
\begin{align}
\frac{\partial\vec{h}_t}{\partial W_e}
=
\sigma'\left(
W_h\vec{h}_{t-1}
+
W_e\vec{e}_{t}
+
\vec{b}
\right)
\times
\left(
\vec{e}_{t}
+
W_h
\frac{\partial\vec{h}_{t-1}}{\partial W_e}
\right)\, ,  
\end{align}
%%%
\begin{align}
\frac{\partial\vec{h}_t}{\partial\vec{b}}
=
\sigma'\left(
W_h\vec{h}_{t-1}
+
W_e\vec{e}_{t}
+
\vec{b}
\right)
\times
\left(
1
+
W_h
\frac{\partial\vec{h}_{t-1}}{\partial\vec{b}}
\right)\, .   
\end{align}

In practice, the derivatives are computed by applying backpropagation to the network unrolled along the time direction, starting from the last time step. This is called backpropagation through time (BPTT).

%%%%%%%%%%%%
%%%%%%%%%%%%
\subsubsection{Applications of RNN}
%%%%%%%%%%%%
%%%%%%%%%%%%

%%%%%%%%%%%%
%%%%%%%%%%%%
\subsubsection*{Sentiment classification}
%%%%%%%%%%%%
%%%%%%%%%%%%
The final state $\vec{h}_T$ contains the information about the entire sentence (modulo the problem associated with the vanishing gradient). So, it makes sense to use $\vec{h}_T$ as an input to another neural network that evaluates the sentiment of the sentence. 

Of course, $\vec{h}_T$ does not really contain all the information about the entire sentence if $T$ is too large, because of the vanishing gradient problem. Therefore, it is better to use all $\vec{h}_t$ ($t= 1,2,\cdots,T$). It is better than using the input $\vec{x}_t$ directly because $\vec{h}_t$ contains the information about $\vec{h}_{t-1},\vec{h}_{t-2},\cdots$, and hence it has some information about the context.

%%%%%%%%%%%%
%%%%%%%%%%%%
\subsubsection*{Speech recognition}
\begin{figure}[htbp]
\centering
\tikzset{
  rnnbox/.style={draw, thick, minimum width=10mm, minimum height=11mm, inner sep=1pt},
  rnnlab/.style={font=\footnotesize},
  word/.style={font=\small}}
% ===================== Top: training =====================
\begin{tikzpicture}[>=Stealth]
\node[rnnbox] (h0) at (0,0) {$\vec{h}_0$};
\node[word, align=center, anchor=south] at ([yshift=1mm]h0.north) {I am a cat.\\(audio)};
\foreach \n/\x/\win/\wout in {1/1.9/{$\langle\mathrm{start}\rangle$}/I, 2/3.7/I/am, 3/5.5/am/a, 4/7.3/a/cat, 5/9.1/cat/{$\langle\mathrm{end}\rangle$}} {
  \node[rnnbox] (h\n) at (\x,0)    {$\vec{h}_{\n}$};
  \node[rnnbox] (e\n) at (\x,-2.2) {$\vec{e}_{\n}$};
  \node (x\n) at (\x,-3.8) {$\vec{x}_{\n}$};
  \node[word, anchor=north] at (x\n.south) {\win};
  \node (y\n) at (\x,1.7) {$\vec{y}_{\n}$};
  \node[word, anchor=south] at (y\n.north) {\wout};
  \draw[->, thick] (x\n) -- node[rnnlab, left]{$E$}   (e\n);
  \draw[->, thick] (e\n) -- node[rnnlab, left]{$W_e$} (h\n);
  \draw[->, thick] (h\n) -- node[rnnlab, left]{$D$}   (y\n);
}
\foreach \a/\b in {0/1,1/2,2/3,3/4,4/5}
  \draw[->, thick, blue] (h\a) -- node[rnnlab, above]{$W_h$} (h\b);
\end{tikzpicture}

\vspace{4mm}
% ===================== Bottom: generation =====================
\begin{tikzpicture}[>=Stealth]
\node[rnnbox] (h0) at (0,0) {$\vec{h}_0$};
\node[word, align=center, anchor=south] at ([yshift=1mm]h0.north) {I am a cat.\\(audio)};
\foreach \n/\x/\win/\wout in {1/1.9/{}/I, 2/3.7/I/am, 3/5.5/am/a, 4/7.3/a/calf, 5/9.1/calf/{$\langle\mathrm{end}\rangle$}} {
  \node[rnnbox] (h\n) at (\x,0)    {$\vec{h}_{\n}$};
  \node[rnnbox] (e\n) at (\x,-2.2) {$\vec{e}_{\n}$};
  \node[red] (y\n) at (\x,1.7) {$\vec{y}_{\n}$};
  \node[word, anchor=south] at (y\n.north) {\wout};
  \draw[->, thick] (e\n) -- node[rnnlab, left]{$W_e$} (h\n);
  \draw[->, thick] (h\n) -- node[rnnlab, left]{$D$}   (y\n);
}
% inputs: the start signal, then the previous outputs
\node (x1) at (1.9,-3.8) {$\vec{x}_1$};
\node[word, anchor=north] at (x1.south) {$\langle\mathrm{start}\rangle$};
\foreach \n/\p/\x/\w in {2/1/3.7/I, 3/2/5.5/am, 4/3/7.3/a, 5/4/9.1/calf} {
  \node[red] (x\n) at (\x,-3.8) {$\vec{y}_{\p}$};
  \node[word, anchor=north] at (x\n.south) {\w};
}
\foreach \n in {1,...,5}
  \draw[->, thick] (x\n) -- node[rnnlab, left]{$E$} (e\n);
\foreach \a/\b in {0/1,1/2,2/3,3/4,4/5}
  \draw[->, thick, blue] (h\a) -- node[rnnlab, above]{$W_h$} (h\b);
\end{tikzpicture}
\caption{[Top] A training strategy for RNN for speech recognition. The audio data is used as $\vec{h}_0$. The actual words corresponding to the audio data, $\vec{x}_2,\vec{x}_3,\cdots$, are used as input. The network is trained so that the outputs $\vec{y}_1,\vec{y}_2,\vec{y}_3,\cdots$ reproduce $\vec{x}_2,\vec{x}_3,\vec{x}_4,\cdots$.
[Bottom] When new audio data is given as $\vec{h}_0$, the start signal is given as $\vec{x}_1$. Then, a sentence is generated by using the first output $\vec{y}_1$ for the second input $\vec{x}_2$, the second output $\vec{y}_2$ for the third input $\vec{x}_3$, and so on, and repeating it until the end signal is obtained. In this example, `cat' was misrecognized as `calf', and the wrong word is fed into the next step.}
\label{fig:RNN-speech-recognition}
\end{figure}
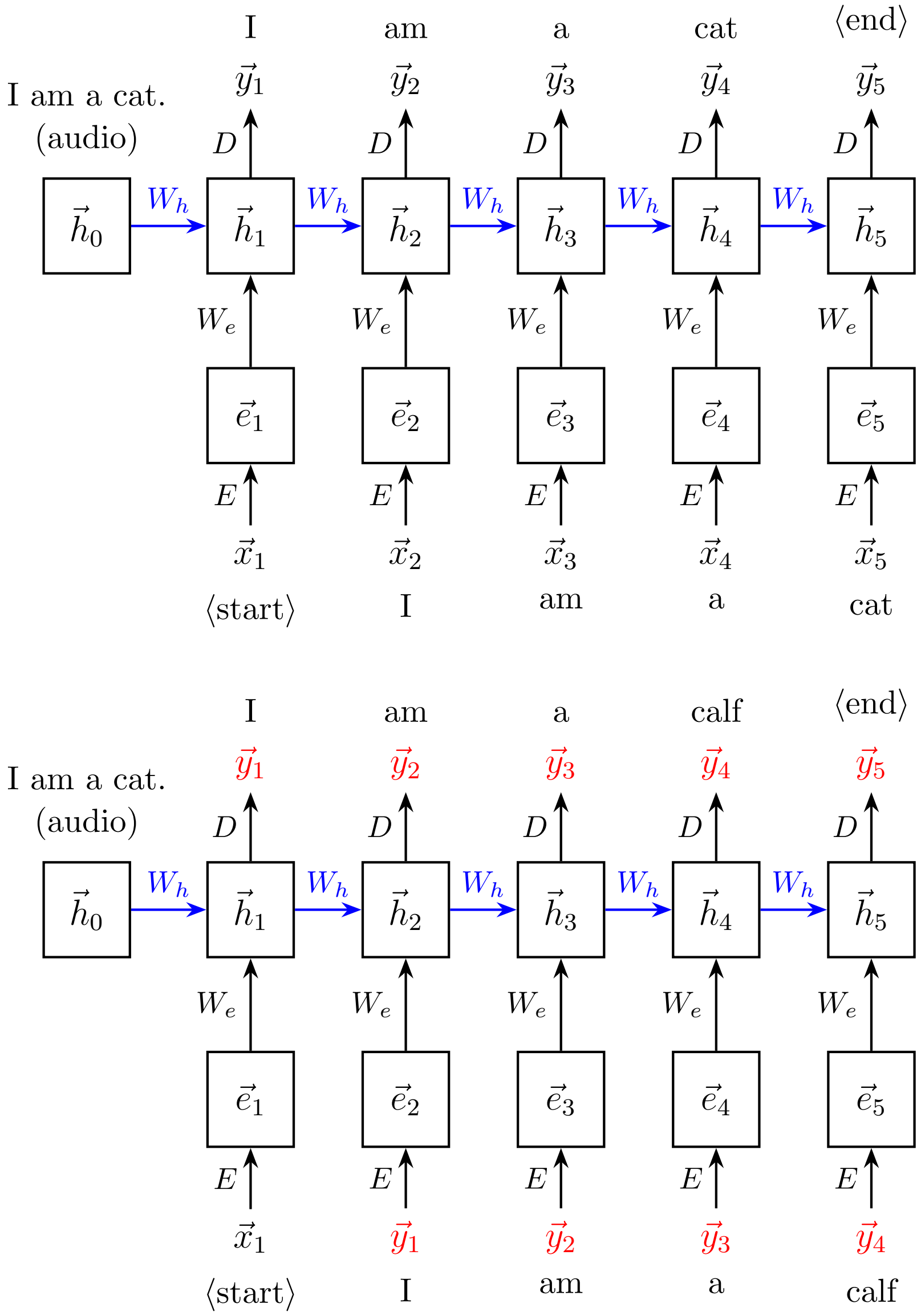

%%%%%%%%%%%%
%%%%%%%%%%%%
We can encode the audio data into a vector by using another neural network, use it as $\vec{h}_0$, and take $\vec{x}_1$ to be the start signal $\langle\mathrm{start}\rangle$, and take $\vec{x}_2,\vec{x}_3,\cdots$ to be
the actual words in the audio data. We train the network so that $\vec{y}_1,\vec{y}_2,\vec{y}_3,\cdots$ agree with $\vec{x}_2,\vec{x}_3,\vec{x}_4,\cdots$ (see the top panel of Fig.~\ref{fig:RNN-speech-recognition}).

When new audio data is given as $\vec{h}_0$, $\langle\mathrm{start}\rangle$ is given as $\vec{x}_1$. Then, a sentence is generated by using the first output $\vec{y}_1$ for the second input $\vec{x}_2$, the second output $\vec{y}_2$ for the third input $\vec{x}_3$, the $t$-th output $\vec{y}_t$ for the $(t+1)$-th input $\vec{x}_{t+1}$, and repeating it until the end signal $\langle\mathrm{end}\rangle$ is obtained (see the bottom panel of Fig.~\ref{fig:RNN-speech-recognition}).

We could also imagine a task of generating a sentence from a picture (a dog is sleeping, a cat is in a state of superposition of dead and alive, etc.). In this case, $\vec{h}_0$ is a feature vector extracted from the picture.

%%%%%%%%%%%%
%%%%%%%%%%%%
\subsubsection*{Machine translation}
%%%%%%%%%%%%
%%%%%%%%%%%%
RNN can be used for machine translation. We use two RNNs, one for the encoder and the other for the decoder.

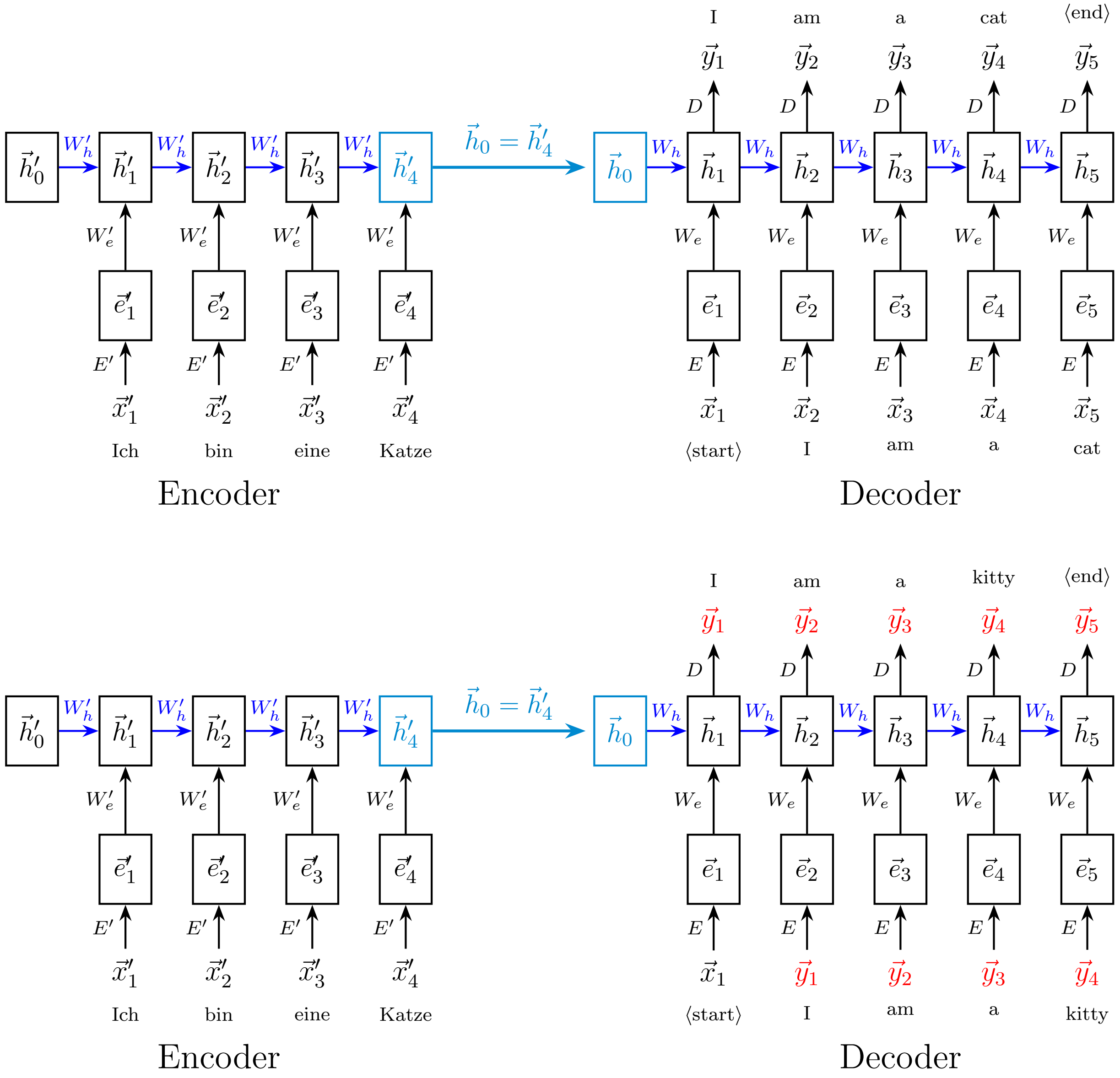
\begin{figure}[htbp]
\centering
\tikzset{
  mtbox/.style={draw, thick, minimum width=7.5mm, minimum height=10mm, inner sep=0.5pt, font=\small},
  mtlab/.style={font=\scriptsize},
  mtword/.style={font=\scriptsize},
  link/.style={cyan!70!blue}}
% ===================== Top: supervised learning =====================
\begin{tikzpicture}[>=Stealth]
% ---- encoder ----
\node[mtbox] (g0) at (0,0) {$\vec{h}'_0$};
\foreach \n/\x/\w/\s in {1/1.35/Ich/, 2/2.7/bin/, 3/4.05/eine/, 4/5.4/Katze/link} {
  \node[mtbox, \s] (g\n) at (\x,0)    {$\vec{h}'_{\n}$};
  \node[mtbox] (f\n) at (\x,-2)   {$\vec{e}'_{\n}$};
  \node (u\n) at (\x,-3.5) {$\vec{x}'_{\n}$};
  \node[mtword, anchor=north] at (u\n.south) {\w};
  \draw[->, thick] (u\n) -- node[mtlab, left]{$E'$}   (f\n);
  \draw[->, thick] (f\n) -- node[mtlab, left]{$W'_e$} (g\n);
}
\foreach \a/\b in {0/1,1/2,2/3,3/4}
  \draw[->, thick, blue] (g\a) -- node[mtlab, above]{$W'_h$} (g\b);
% ---- link ----
\draw[->, very thick, link] (g4) -- node[above, font=\small]{$\vec{h}_0=\vec{h}'_4$} (8.0,0);
% ---- decoder ----
\node[mtbox, link] (h0) at (8.5,0) {$\vec{h}_0$};
\foreach \n/\x/\win/\wout in {1/9.85/{$\langle\mathrm{start}\rangle$}/I, 2/11.2/I/am, 3/12.55/am/a, 4/13.9/a/cat, 5/15.25/cat/{$\langle\mathrm{end}\rangle$}} {
  \node[mtbox] (h\n) at (\x,0)  {$\vec{h}_{\n}$};
  \node[mtbox] (e\n) at (\x,-2) {$\vec{e}_{\n}$};
  \node (x\n) at (\x,-3.5) {$\vec{x}_{\n}$};
  \node[mtword, anchor=north] at (x\n.south) {\win};
  \node (y\n) at (\x,1.6) {$\vec{y}_{\n}$};
  \node[mtword, anchor=south] at (y\n.north) {\wout};
  \draw[->, thick] (x\n) -- node[mtlab, left]{$E$}   (e\n);
  \draw[->, thick] (e\n) -- node[mtlab, left]{$W_e$} (h\n);
  \draw[->, thick] (h\n) -- node[mtlab, left]{$D$}   (y\n);
}
\foreach \a/\b in {0/1,1/2,2/3,3/4,4/5}
  \draw[->, thick, blue] (h\a) -- node[mtlab, above]{$W_h$} (h\b);
\node[font=\large] at (2.7,-4.7) {Encoder};
\node[font=\large] at (12.55,-4.7) {Decoder};
\end{tikzpicture}

\vspace{6mm}
% ===================== Bottom: actual translation =====================
\begin{tikzpicture}[>=Stealth]
% ---- encoder ----
\node[mtbox] (g0) at (0,0) {$\vec{h}'_0$};
\foreach \n/\x/\w/\s in {1/1.35/Ich/, 2/2.7/bin/, 3/4.05/eine/, 4/5.4/Katze/link} {
  \node[mtbox, \s] (g\n) at (\x,0)    {$\vec{h}'_{\n}$};
  \node[mtbox] (f\n) at (\x,-2)   {$\vec{e}'_{\n}$};
  \node (u\n) at (\x,-3.5) {$\vec{x}'_{\n}$};
  \node[mtword, anchor=north] at (u\n.south) {\w};
  \draw[->, thick] (u\n) -- node[mtlab, left]{$E'$}   (f\n);
  \draw[->, thick] (f\n) -- node[mtlab, left]{$W'_e$} (g\n);
}
\foreach \a/\b in {0/1,1/2,2/3,3/4}
  \draw[->, thick, blue] (g\a) -- node[mtlab, above]{$W'_h$} (g\b);
% ---- link ----
\draw[->, very thick, link] (g4) -- node[above, font=\small]{$\vec{h}_0=\vec{h}'_4$} (8.0,0);
% ---- decoder ----
\node[mtbox, link] (h0) at (8.5,0) {$\vec{h}_0$};
\foreach \n/\x/\wout in {1/9.85/I, 2/11.2/am, 3/12.55/a, 4/13.9/kitty, 5/15.25/{$\langle\mathrm{end}\rangle$}} {
  \node[mtbox] (h\n) at (\x,0)  {$\vec{h}_{\n}$};
  \node[mtbox] (e\n) at (\x,-2) {$\vec{e}_{\n}$};
  \node[red] (y\n) at (\x,1.6) {$\vec{y}_{\n}$};
  \node[mtword, anchor=south] at (y\n.north) {\wout};
  \draw[->, thick] (e\n) -- node[mtlab, left]{$W_e$} (h\n);
  \draw[->, thick] (h\n) -- node[mtlab, left]{$D$}   (y\n);
}
\node (x1) at (9.85,-3.5) {$\vec{x}_1$};
\node[mtword, anchor=north] at (x1.south) {$\langle\mathrm{start}\rangle$};
\foreach \n/\p/\x/\w in {2/1/11.2/I, 3/2/12.55/am, 4/3/13.9/a, 5/4/15.25/kitty} {
  \node[red] (x\n) at (\x,-3.5) {$\vec{y}_{\p}$};
  \node[mtword, anchor=north] at (x\n.south) {\w};
}
\foreach \n in {1,...,5}
  \draw[->, thick] (x\n) -- node[mtlab, left]{$E$} (e\n);
\foreach \a/\b in {0/1,1/2,2/3,3/4,4/5}
  \draw[->, thick, blue] (h\a) -- node[mtlab, above]{$W_h$} (h\b);
\node[font=\large] at (2.7,-4.7) {Encoder};
\node[font=\large] at (12.55,-4.7) {Decoder};
\end{tikzpicture}
\caption{RNN for machine translation (see, e.g., Ref.~\cite{Sutskever2014_seq2seq_LSTM}). The original sentence is encoded into $\vec{h}'_4$ by the encoder and passed to the decoder as its initial state $\vec{h}_0$. The translated sentence is generated by the decoder. The encoder and the decoder have different parameters (with and without primes). [Top] Supervised learning. [Bottom] Actual translation. The output may differ from the reference translation in the training data (`kitty' instead of `cat'), because the translation is not unique.}
\label{fig:RNN-machine-translation}
\end{figure}

For supervised learning, we use pairs of sentences in two languages which have the same meaning. (Ideally, we want to have multiple translations of the same sentence because the translation is not unique.) The encoder RNN encodes a sentence in one language into a vector $\vec{h}'_T$. This state is used as the input
$\vec{h}_0$ for the decoder RNN. The decoder part is used like the RNN for speech recognition; see Fig.~\ref{fig:RNN-machine-translation}.

%%%%%%%%%%%%
%%%%%%%%%%%%
\subsubsection*{Sequence-to-sequence (Seq2seq)}
%%%%%%%%%%%%
%%%%%%%%%%%%
Machine translation provides a map from one kind of sequence (e.g., sentences in one language) to another kind of sequence (e.g., sentences in another language). For this reason, such a model is called a \textit{sequence-to-sequence model} (or \textit{Seq2seq}). Because an encoder and a decoder are used, this is also called an \textit{encoder-decoder model}. Seq2seq includes many other NLP tasks. For example, digesting a long text and making an abstract is a kind of sequence-to-sequence mapping. A conversation between two people (or one person and an AI) is also a seq2seq task. When you say something to me, my response is (usually) not random. I respond based on what you say. Another example is image captioning, because an image can be expressed as a sequence of numbers and the caption is also expressed as a sequence of numbers.

%%%%%%%%%%%%
%%%%%%%%%%%%
\subsubsection*{Text generation}
%%%%%%%%%%%%
%%%%%%%%%%%%
For text generation, a sequence of words $\vec{x}_1,\vec{x}_2,\vec{x}_3,\cdots$ is used as input. The network is trained so that the outputs $\vec{y}_1,\vec{y}_2,\vec{y}_3,\cdots$ reproduce $\vec{x}_2,\vec{x}_3,\vec{x}_4,\cdots$; see Fig.~\ref{fig:RNN-text-generation}. If we want to make a network that generates a sentence from the first four words, we can remove the first three outputs (shown in blue) from the training. When the first few words are given, a sentence is generated by using the output $\vec{y}_t$ as input $\vec{x}_{t+1}$, until the end signal is obtained.

Note that $\vec{y}_t$ is a probability distribution. To generate a word, we either pick the most probable one or draw a word randomly according to $\vec{y}_t$. The randomness can be tuned by replacing $\mathrm{softmax}(\vec{a})$ with $\mathrm{softmax}(\vec{a}/T)$, where $T$ is the temperature introduced in Sec.~\ref{sec:distillation}. For $T\to 0$, the most probable word is always chosen, while a larger $T$ gives more diverse texts.

\begin{figure}[htbp]
\centering
\tikzset{
  tgbox/.style={draw, thick, minimum width=4.5mm, minimum height=13mm, inner sep=0pt},
  tgword/.style={font=\footnotesize, inner sep=1pt}}
% ===================== Top: training =====================
\begin{tikzpicture}[>=Stealth, x=1.27cm]
\node (h0) at (0,0) {$\vec{0}$};
\foreach \n/\win/\wout/\c in {
   1/The/coldest/blue, 2/coldest/winter/blue, 3/winter/I/blue,
   4/I/ever/black, 5/ever/spent/black, 6/spent/was/black, 7/was/a/black,
   8/a/summer/black, 9/summer/in/black, 10/in/San/black,
   11/San/Francisco/black, 12/Francisco/{$\langle\mathrm{end}\rangle$}/black} {
  \node[tgbox] (h\n) at (\n,0) {};
  \node[tgword, anchor=base] (x\n) at (\n,-1.75) {\win};
  \node[tgword, anchor=base, \c] (y\n) at (\n,1.55) {\wout};
  \draw[->, thick] (\n,-1.45) -- (h\n.south);
  \draw[->, thick, \c] (h\n.north) -- (\n,1.4);
}
\draw[->, thick] (h0) -- (h1);
\foreach \a/\b in {1/2,2/3,3/4,4/5,5/6,6/7,7/8,8/9,9/10,10/11,11/12}
  \draw[->, thick] (h\a) -- (h\b);
\end{tikzpicture}

\vspace{5mm}
% ===================== Bottom: generation =====================
\begin{tikzpicture}[>=Stealth, x=1.27cm]
\node (h0) at (0,0) {$\vec{0}$};
\foreach \n/\win/\cin in {1/The/black, 2/coldest/black, 3/winter/black, 4/I/black,
   5/ever/red, 6/spent/red, 7/was/red, 8/a/red, 9/winter/red, 10/in/red, 11/Chicago/red} {
  \node[tgbox] (h\n) at (\n,0) {};
  \node[tgword, anchor=base, \cin] (x\n) at (\n,-1.75) {\win};
  \draw[->, thick] (\n,-1.45) -- (h\n.south);
}
\foreach \n/\wout in {4/ever, 5/spent, 6/was, 7/a, 8/winter, 9/in, 10/Chicago, 11/{$\langle\mathrm{end}\rangle$}} {
  \node[tgword, anchor=base, red] (y\n) at (\n,1.55) {\wout};
  \draw[->, thick] (h\n.north) -- (\n,1.4);
}
\draw[->, thick] (h0) -- (h1);
\foreach \a/\b in {1/2,2/3,3/4,4/5,5/6,6/7,7/8,8/9,9/10,10/11}
  \draw[->, thick] (h\a) -- (h\b);
\end{tikzpicture}
\caption{[Top] A training strategy for RNN for text generation. A sequence of words $\vec{x}_1,\vec{x}_2,\vec{x}_3,\cdots$ is used as input. The network is trained so that the outputs $\vec{y}_1,\vec{y}_2,\vec{y}_3,\cdots$ reproduce $\vec{x}_2,\vec{x}_3,\vec{x}_4,\cdots$. If we want to make a network that generates a sentence from the first four words, we can remove the first three outputs (shown in blue) from the training. [Bottom] When the first few words are given, a sentence is generated by using the output $\vec{y}_t$ (red) as input $\vec{x}_{t+1}$, until the end signal is obtained. [Note: ``San Francisco" may be one word.]}
\label{fig:RNN-text-generation}
\end{figure}

%%%%%%%%%%%%
%%%%%%%%%%%%
\subsubsection{Long Short-Term Memory (LSTM)}\label{sec:LSTM}
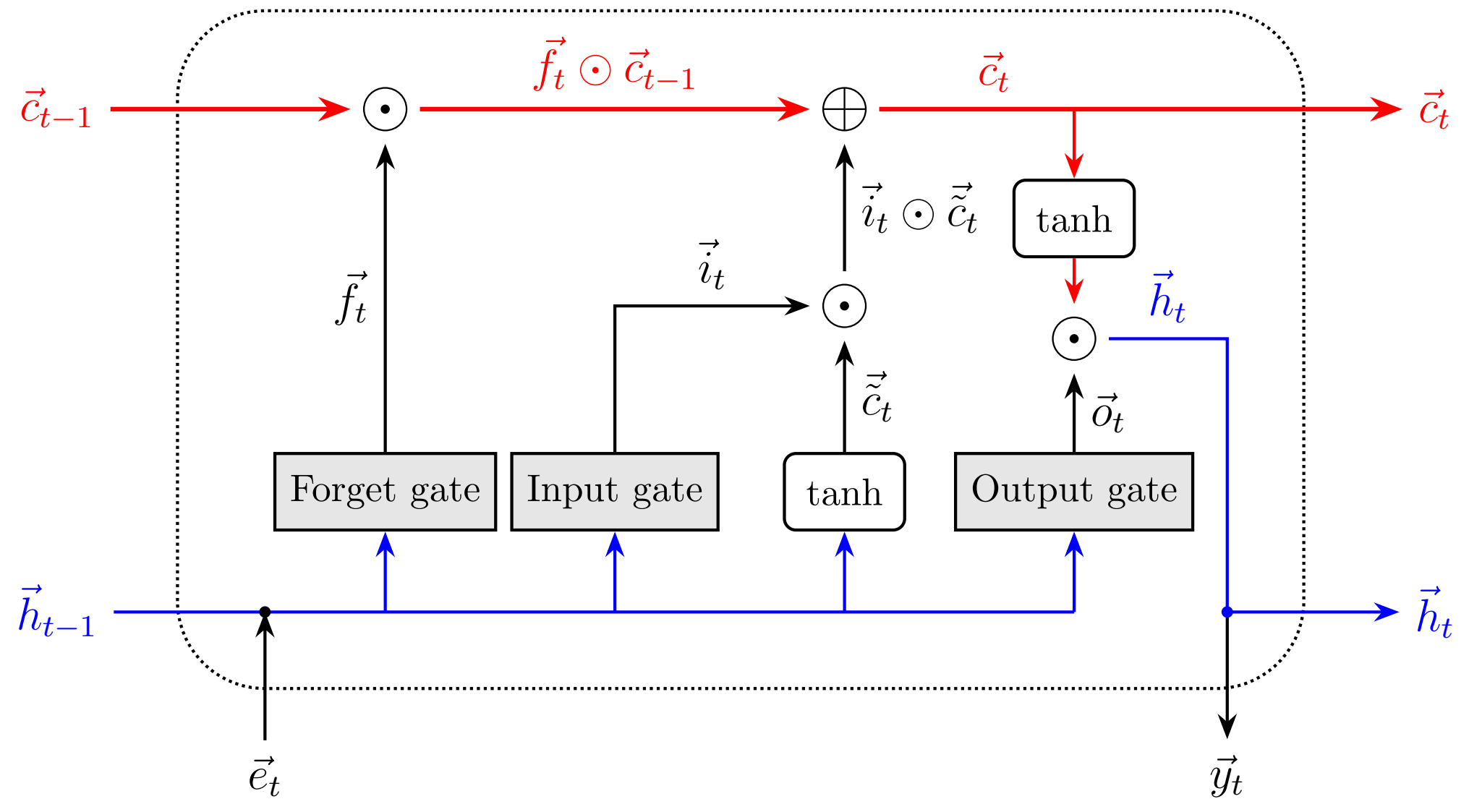
\begin{figure}[htbp]
\centering
\begin{tikzpicture}[>=Stealth,
  gate/.style={draw, thick, fill=gray!20, minimum width=17mm, minimum height=7mm, font=\footnotesize},
  fn/.style={draw, thick, rounded corners=3pt, minimum width=11mm, minimum height=7mm, font=\footnotesize},
  op/.style={circle, inner sep=0pt, minimum size=5.5mm, font=\Large}]
% boundary of the LSTM cell
\draw[densely dotted, thick, rounded corners=8mm] (-0.3,-0.7) rectangle (10.0,5.5);
% operations
\node[op] (m1)  at (1.6,4.6) {$\odot$};
\node[op] (add) at (5.8,4.6) {$\oplus$};
\node[op] (m2)  at (5.8,2.8) {$\odot$};
\node[op] (m3)  at (7.9,2.5) {$\odot$};
% gates and tanh
\node[gate] (fg) at (1.6,1.1) {Forget gate};
\node[gate] (ig) at (3.7,1.1) {Input gate};
\node[fn]   (tc) at (5.8,1.1) {$\tanh$};
\node[gate] (og) at (7.9,1.1) {Output gate};
\node[fn]   (th) at (7.9,3.6) {$\tanh$};
% cell state (red)
\node[red] (cin)  at (-1.4,4.6) {$\vec{c}_{t-1}$};
\node[red] (cout) at (11.2,4.6) {$\vec{c}_t$};
\draw[->, very thick, red] (cin) -- (m1);
\draw[->, very thick, red] (m1) -- node[above]{$\vec{f}_t\odot\vec{c}_{t-1}$} (add);
\draw[->, very thick, red] (add) -- node[above, pos=0.22]{$\vec{c}_t$} (cout);
\draw[->, thick, red] (7.9,4.6) -- (th);
\draw[->, thick, red] (th) -- (m3);
% hidden state (blue) and input
\node[blue] (hin)  at (-1.4,0) {$\vec{h}_{t-1}$};
\node[blue] (hout) at (11.2,0) {$\vec{h}_t$};
\node (ein)  at (0.5,-1.5) {$\vec{e}_t$};
\node (yout) at (9.3,-1.5) {$\vec{y}_t$};
\draw[thick, blue] (hin) -- (7.9,0);
\draw[->, thick] (ein) -- (0.5,0);
\fill (0.5,0) circle (1.5pt);
\foreach \g in {fg,ig,tc,og} \draw[->, thick, blue] (\g |- 0,0) -- (\g);
\draw[thick, blue] (m3) -- node[above]{$\vec{h}_t$} (9.3,2.5) -- (9.3,0);
\draw[->, thick, blue] (9.3,0) -- (hout);
\draw[->, thick] (9.3,0) -- (yout);
\fill[blue] (9.3,0) circle (1.5pt);
% gate signals
\draw[->, thick] (fg) -- node[left]{$\vec{f}_t$} (m1);
\draw[->, thick] (ig) -- (3.7,2.8) -- node[above]{$\vec{i}_t$} (m2);
\draw[->, thick] (tc) -- node[right]{$\vec{\tilde{c}}_t$} (m2);
\draw[->, thick] (m2) -- node[right]{$\vec{i}_t\odot\vec{\tilde{c}}_t$} (add);
\draw[->, thick] (og) -- node[right]{$\vec{o}_t$} (m3);
\end{tikzpicture}
\caption{Long Short-Term Memory architecture. $\odot$ is the Hadamard product and $\oplus$ is the sum. The cell state (red) is changed only by the forget gate and by the new information selected by the input gate.}
\label{fig:LSTM}
\end{figure}

%%%%%%%%%%%%
%%%%%%%%%%%%
Long Short-Term Memory~\cite{Hochreiter1997,Gers2000} (LSTM; Fig.~\ref{fig:LSTM}) is a type of architecture that selectively controls what to forget.\footnote{The title of the paper that established the architecture was ``Learning to Forget: Continual Prediction with LSTM"~\cite{Gers2000}.} When trained well, LSTM can preserve short-term memory over several hundred time steps.

The LSTM consists of many pieces. Let’s look at them one by one.

First of all, there are two state vectors labeled by time $t$: the hidden state $\vec{h}_t$, which appeared in the vanilla RNN as well, and the cell state $\vec{c}_t$ that preserves information for many time steps. They have the same dimension.

From $\vec{h}_{t-1}$ and the new input $\vec{e}_t$, three gates are formed: 
the \textit{forget gate}
\begin{align}
\vec{f}_t 
= 
\sigma\left(
W_{h,f}\vec{h}_{t-1} 
+ 
W_{e,f}\vec{e}_t
+
\vec{b}_f
\right)\, ,
\end{align}
the \textit{input gate}
\begin{align}
\vec{i}_t 
= 
\sigma\left(
W_{h,i}\vec{h}_{t-1} 
+ 
W_{e,i}\vec{e}_t
+
\vec{b}_i
\right)\, ,
\end{align}
and the \textit{output gate}
\begin{align}
\vec{o}_t 
= 
\sigma\left(
W_{h,o}\vec{h}_{t-1} 
+ 
W_{e,o}\vec{e}_t
+
\vec{b}_o
\right)\, . 
\end{align}
The activation function $\sigma$ is taken to be the sigmoid, so the values of $\vec{f}_t$, $\vec{i}_t$ and $\vec{o}_t$ are between 0 and 1.

The forget gate controls what to forget. Specifically, it acts on the cell state $\vec{c}_{t-1}$ as
\begin{align}
\vec{f}_{t}
\odot
\vec{c}_{t-1}\, .
\end{align}
Here, the Hadamard product $\odot$ is the element-by-element product, i.e., $(\vec{f}_t\odot\vec{c}_{t-1})_i=f_{t,i}\cdot c_{t-1,i}$. If $f_{t,i}$ is close to 0, the information stored in the $i$-th component of $\vec{c}_{t-1}$ is thrown away. If $f_{t,i}$  is close to 1, the information stored in the $i$-th component is preserved.
Therefore, the network can learn to forget by tuning $W_{h,f}$, $W_{e,f}$ and $\vec{b}_f$, and preserve the important piece of information from the previous cell state as $\vec{f}_{t}\odot\vec{c}_{t-1}$.

The input gate controls what is added to the cell state. Specifically, the candidate information $\vec{\tilde{c}}_t$ is generated as
\begin{align}
\vec{\tilde{c}}_t
= 
\tanh\left(
W_{h,c}\vec{h}_{t-1} 
+ 
W_{e,c}\vec{e}_t
+
\vec{b}_c
\right)\, , 
\end{align}
and then the Hadamard product with the input gate is taken:
\begin{align}
\vec{i}_{t}
\odot
\vec{\tilde{c}}_t\, . 
\end{align}
If $i_{t,i}$ is close to 1, the information stored in the $i$-th component of $\vec{\tilde{c}}_t$ is added. The network can learn what kind of information should be added to short-term memory by tuning $W_{h,i}$, $W_{e,i}$ and $\vec{b}_i$. 

The cell state $\vec{c}_t$, which is passed to the next step, is the sum of $\vec{f}_{t}\odot\vec{c}_{t-1}$, which is the important piece of information from the previous cell state, and the new input $\vec{i}_{t}\odot\vec{\tilde{c}}_t$:
\begin{align}
\vec{c}_t
=
\vec{f}_{t}\odot\vec{c}_{t-1}
+
\vec{i}_{t}\odot\vec{\tilde{c}}_t\, . 
\end{align}
Note that the direct dependence of $\vec{c}_t$ on $\vec{c}_{t-1}$ is $\frac{\partial c_{t,i}}{\partial c_{t-1,i}}=f_{t,i}$. If the forget gate is close to 1, the gradient propagates through many time steps without decaying, just like the shortcut of ResNet (Sec.~\ref{sec:cures_for_vanishing_gradient}). Of course, if everything were kept and new information were added at every step, the cell state would become a mixture of everything. This is avoided because the gates act component by component: the network can learn to keep one component untouched ($f_{t,i}\simeq 1$ and $i_{t,i}\simeq 0$) while overwriting another ($f_{t,i}\simeq 0$ and $i_{t,i}\simeq 1$). Still, the number of components is finite, and this limitation motivates the attention mechanism discussed below.

The output gate controls which part of the information stored in the cell state $\vec{c}_t$ is passed to the hidden state:
\begin{align}
\vec{h}_t
=
\vec{o}_t
\odot
\tanh\vec{c}_t\, .
\end{align}

%%%%%%%%%%%%
%%%%%%%%%%%%
\section{Attention mechanism and Transformer model}\label{sec:attention}
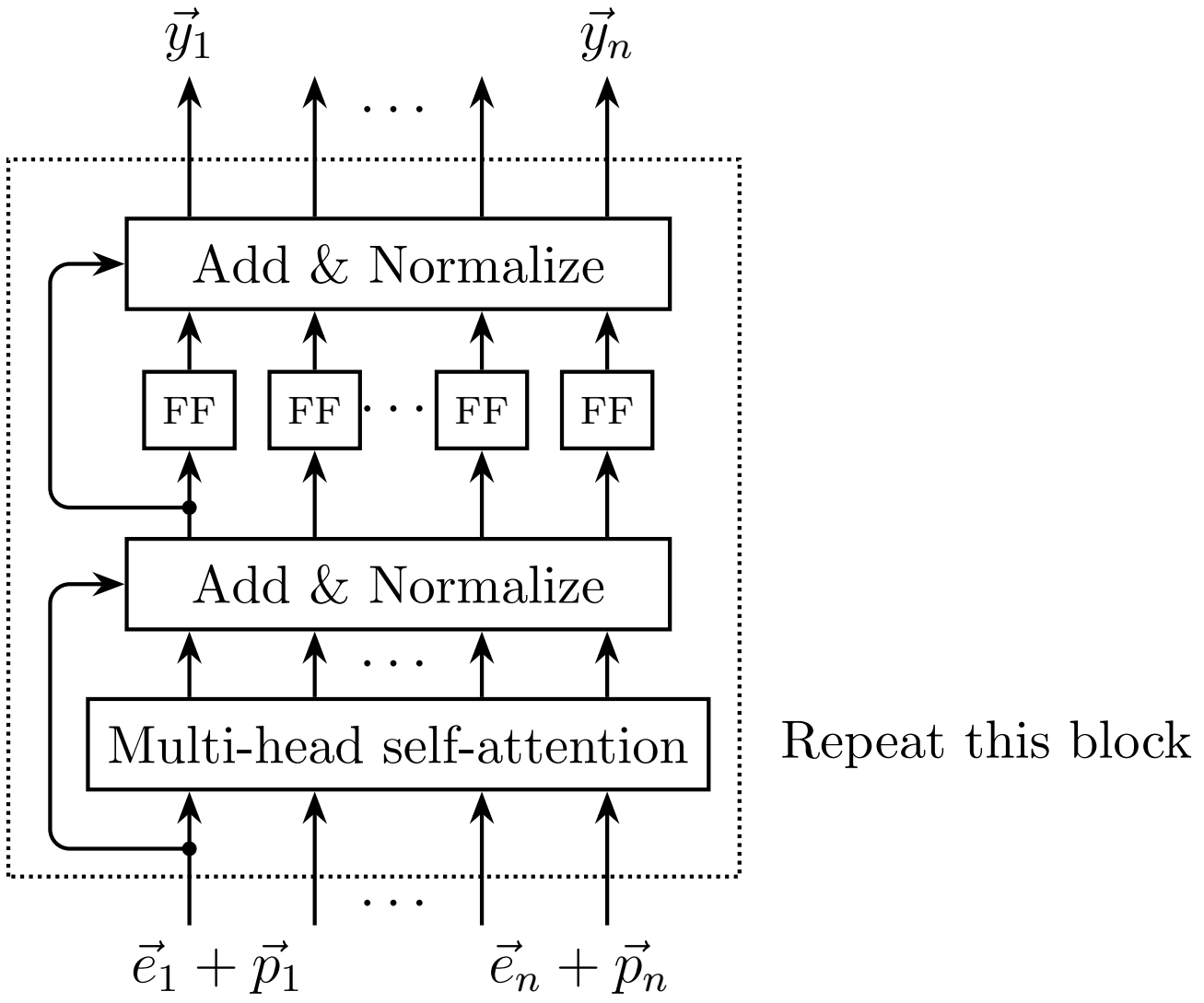
\begin{figure}[htbp]
\centering
\begin{tikzpicture}[>=Stealth,
  wide/.style={draw, thick, minimum width=3.9cm, minimum height=6.5mm, font=\small},
  ff/.style={draw, thick, minimum width=6.5mm, minimum height=5.5mm, inner sep=0pt, font=\scriptsize}]
% block (repeated)
\draw[densely dotted, thick] (-1.3,-0.55) rectangle (3.95,4.6);
\node[font=\small, anchor=west] at (4.1,0.4) {Repeat this block};
% wide boxes
\node[wide] (att) at (1.5,0.4) {Multi-head self-attention};
\node[wide] (an1) at (1.5,1.55) {Add \& Normalize};
\node[wide] (an2) at (1.5,3.85) {Add \& Normalize};
% columns
\foreach \x/\n in {0/1, 0.9/2, 2.1/a, 3/b} {
  \draw[->, thick] (\x,-0.9) -- (\x,0.4 |- att.south);
  \draw[->, thick] (\x,0 |- att.north) -- (\x,0 |- an1.south);
  \node[ff] (ff\n) at (\x,2.8) {FF};
  \draw[->, thick] (\x,0 |- an1.north) -- (ff\n.south);
  \draw[->, thick] (ff\n.north) -- (\x,0 |- an2.south);
  \draw[->, thick] (\x,0 |- an2.north) -- (\x,5.2);
}
% dots between the 2nd and 3rd columns
\foreach \y in {-0.75, 0.97, 2.8, 4.95} \node at (1.5,\y) {$\cdots$};
% inputs and outputs
\node at (0.2,-1.2) {$\vec{e}_1+\vec{p}_1$};
\node at (2.8,-1.2) {$\vec{e}_n+\vec{p}_n$};
\node at (0,5.5) {$\vec{y}_1$};
\node at (3,5.5) {$\vec{y}_n$};
% residual connections
\fill (0,-0.35) circle (1.5pt);
\draw[->, thick, rounded corners=4pt] (0,-0.35) -- (-1.0,-0.35) -- (-1.0,1.55) -- (an1.west);
\fill (0,2.1) circle (1.5pt);
\draw[->, thick, rounded corners=4pt] (0,2.1) -- (-1.0,2.1) -- (-1.0,3.85) -- (an2.west);
\end{tikzpicture}
\caption{A schematic picture of Transformer. Each block consists of multi-head self-attention and a feed-forward network (FF), each followed by a residual connection and normalization. The same FF is applied to each position separately. The residual connection is drawn only for the first position.}
\label{fig:Transformer-architecture}
\end{figure}

%%%%%%%%%%%%
%%%%%%%%%%%%
Let us see how the attention mechanism works in the context of NLP. Specifically, we
discuss \textit{Transformer}~\cite{Vaswani2017} depicted in Fig.~\ref{fig:Transformer-architecture}. Note that attention is a much broader concept
that can work for many purposes.

%%%%%%%%%%%%
%%%%%%%%%%%%
\subsection{Self-attention and Transformer architecture}
%%%%%%%%%%%%
%%%%%%%%%%%%
A big problem with RNN was the loss of short-term memory. Namely, the information about the input word $\vec{x}_t$ could be lost at a later stage $t'$ as $t'-t$ grows. The LSTM could improve the situation, but it was not enough. Self-attention aims to resolve this issue by allowing direct, non-local connections between inputs.

%%%%%%%%%%%%
%%%%%%%%%%%%
\subsubsection*{Queries, keys and values}
%%%%%%%%%%%%
%%%%%%%%%%%%
Let us take the word embedding $\vec{e}_t\in\mathbb{R}^d$ defined by \eqref{eq:word_embedding}.\footnote{We can also use $\vec{h}_t$ and repeat the same argument, if we want to combine the attention mechanism and RNN.} We want to let the network learn the words that are likely to be related. For example, imagine four famous names were in the text: Francis Crick, James Watson, Maurice Wilkins, and Rosalind Franklin. These people made large contributions to the discovery of the double-helix structure of DNA. Crick, Watson and Wilkins shared the Nobel Prize in Physiology or Medicine in 1962, after Franklin’s death in 1958. Many people believe Franklin deserved the Nobel Prize as well, and she may have been a victim of sexism in science.

When these names appear in the same text, most likely, they are very deeply connected. To capture the correlation between these words, we introduce queries $\vec{q}_t=Q\vec{e}_t\in\mathbb{R}^d$ and keys $\vec{k}_t=K\vec{e}_t\in\mathbb{R}^d$, where $Q,K\in\mathbb{R}^{d\times d}$ are determined through training. If the inner product $\vec{q}_t\cdot\vec{k}_{t'}$ is large, the words $\vec{e}_t$ and $\vec{e}_{t'}$ are likely to be more closely related. If we take the word vectors $\vec{e}_{\rm Crick}$, $\vec{e}_{\rm Watson}$, $\vec{e}_{\rm Wilkins}$, $\vec{e}_{\rm Franklin}$ and make queries and keys by multiplying them by $Q$ and $K$, we can imagine something like\footnote{Note that I just made up this example for the purpose of demonstration.}
\begin{align}
\begin{array}{|c||c|c|c|c|}
\hline
& \vec{k}_{\rm Crick} & \vec{k}_{\rm Watson} & \vec{k}_{\rm Wilkins} &\vec{k}_{\rm Franklin}\\
\hline
\hline 
\vec{q}_{\rm Crick} & 0.8 & 0.7 & 0.3 & 0.3 \\
\hline 
\vec{q}_{\rm Watson} & 0.7 & 0.8 & 0.2 & 0.4 \\
\hline 
\vec{q}_{\rm Wilkins} & 0.6 & 0.5 & 0.9 & 0.8\\
\hline 
\vec{q}_{\rm Franklin} & 0.8 & 0.9 & 0.6 &1.0\\
\hline
\end{array}
\end{align}

Natural keywords related to Crick, Watson, Wilkins, and Franklin include scientist, DNA, Nobel Prize, Cambridge (UK and US), London, and sexism. To pick up such information, we introduce values $\vec{v}_t=V\vec{e}_t\in\mathbb{R}^d$, where $V\in\mathbb{R}^{d\times d}$ is also determined through training, and take the output as
\begin{align}
\vec{o}_t
=
\sum_{t'}
\alpha_{tt'}\vec{v}_{t'}
\in\mathbb{R}^d\, , 
\end{align}
where
\begin{align}
\alpha_{tt'}
=
\mathrm{softmax}(\vec{q}_t\cdot\vec{k}_{t'})
=
\frac{\exp\left(\vec{q}_t\cdot\vec{k}_{t'}\right)}{\sum_{t''}\exp\left(\vec{q}_t\cdot\vec{k}_{t''}\right)}\, .  
\end{align}
We could imagine something like\footnote{
$\vec{e'}$ appearing on the right-hand side of \eqref{eq:values_made_up_example}  does not have to be the same $\vec{e}$ as the word embedding.
As long as the information is somehow contained and can be extracted, any encoding is fine.}$^,$\footnote{Note that, although $\vec{v}_{\rm Crick}$, $\vec{v}_{\rm Watson}$, $\vec{v}_{\rm Wilkins}$ and $\vec{v}_{\rm Franklin}$ can have a large overlap with $\vec{e}_{\rm scientist}$, $\vec{e}_{\rm DNA}$ and $\vec{e}_{\rm Nobel}$, it is highly unlikely that $\vec{v}_{\rm scientist}$, $\vec{v}_{\rm DNA}$ and $\vec{v}_{\rm Nobel}$ have a large overlap with $\vec{e}_{\rm Crick}$, $\vec{e}_{\rm Watson}$, $\vec{e}_{\rm Wilkins}$ and $\vec{e}_{\rm Franklin}$ because there are so many great scientists and Nobel laureates, and there are so many topics related to DNA.}
\begin{align}
\vec{v}_{\rm Crick} 
&= 
\vec{e'}_{\rm scientist} + \vec{e'}_{\rm DNA} + \vec{e'}_{\rm Nobel} + 0.2\, \vec{e'}_{\rm Cambridge} + 0.2\, \vec{e'}_{\rm British} + \cdots\, , 
\nonumber\\
\vec{v}_{\rm Watson} 
&=
\vec{e'}_{\rm scientist} + \vec{e'}_{\rm DNA} + \vec{e'}_{\rm Nobel} + 0.1\, \vec{e'}_{\rm Cambridge} + 0.2\, \vec{e'}_{\rm American} + \cdots\, , 
\nonumber\\
\vec{v}_{\rm Wilkins} 
&=
\vec{e'}_{\rm scientist} + \vec{e'}_{\rm DNA} + \vec{e'}_{\rm Nobel} + 0.2\, \vec{e'}_{\rm London} + 0.2\, \vec{e'}_{\rm British} + \cdots\, , 
\nonumber\\
\vec{v}_{\rm Franklin} 
&= 
\vec{e'}_{\rm scientist} + \vec{e'}_{\rm DNA} + \vec{e'}_{\rm Nobel} + \vec{e'}_{\rm sexism} + 0.2\, \vec{e'}_{\rm London} + 0.2\, \vec{e'}_{\rm British} + \cdots\, .
\label{eq:values_made_up_example} 
\end{align}
If the text is a generic introduction to the Nobel Prize in Physiology or Medicine in 1962 and only the names of Crick, Watson, and Wilkins appear, then probably there is no mention of sexism in the text. Assuming these names appeared only once, we could have something like
\begin{align}
\vec{o}_{\rm Crick } 
&= 
0.34\, \vec{v}_{\rm Crick} + 0.31\, \vec{v}_{\rm Watson} + 0.20\, \vec{v}_{\rm Wilkins} +\cdots
\nonumber\\
&= 
0.85\, \vec{e'}_{\rm scientist} + 0.85\, \vec{e'}_{\rm DNA} + 0.85\, \vec{e'}_{\rm Nobel} + 0.108\, \vec{e'}_{\rm British} + \cdots\, . 
\end{align}
However, if Franklin’s name also appears once, then we could have
\begin{align}
\vec{o}_{\rm Crick } 
&= 
0.28\, \vec{v}_{\rm Crick} + 0.25\, \vec{v}_{\rm Watson} + 0.17\, \vec{v}_{\rm Wilkins} + 0.17\, \vec{v}_{\rm Franklin}+\cdots
\nonumber\\
&= 
0.87\, \vec{e'}_{\rm scientist} + 0.87\, \vec{e'}_{\rm DNA} + 0.87\, \vec{e'}_{\rm Nobel} + 0.17\, \vec{e'}_{\rm sexism} + 0.12\, \vec{e'}_{\rm British} + \cdots\, . 
\end{align}
From the combination of the input words, the network can guess that, perhaps, the text discusses sexism in science that might have affected the Nobel committee’s decision.

Such a guess may or may not be right. Suppose the guess was wrong, and actually the text is filled with topics in molecular biology. Then, $\vec{q}_{\rm scientist}$, $\vec{q}_{\rm DNA}$ and $\vec{q}_{\rm Nobel}$ can have a large overlap with the key vectors $\vec{k}$ made of other parts of the text, while $\vec{q}_{\rm sexism}$ can’t. Therefore, if we repeat the same procedure (if we add multiple layers of self-attention), more attention is paid to molecular biology than to sexism. On the other hand, if the guess was right and other parts of the text discuss inclusion and diversity, then more attention is paid to such topics.
%%%%%%%%%%%%
%%%%%%%%%%%%
\subsubsection*{Position vector}
%%%%%%%%%%%%
%%%%%%%%%%%%
In NLP, the ordering of the input is important. However, the `vanilla' attention mechanism explained above does not care about the ordering.  A \textit{position vector} can be used to cure this issue.\footnote{In the literature, it is usually called the \textit{positional embedding} (when it is trainable) or the \textit{positional encoding} (when it is a fixed function of the position). The original Transformer~\cite{Vaswani2017} used fixed sinusoidal functions of the position. Trainable position vectors, which we explain here, are used, e.g., in GPT and BERT. Many recent LLMs such as LLaMA~\cite{Touvron2023} do not add position vectors to the input; instead, in each attention layer, the queries and keys are rotated by angles proportional to the position, in such a way that the inner product $\vec{q}_t\cdot\vec{k}_{t'}$ depends on the position only through the relative position $t-t'$ (rotary position embedding, or RoPE~\cite{Su2021}).}

For simplicity, let us fix the maximum input size to be $n$. The position vectors $\vec{p}_1$, $\cdots$, $\vec{p}_n\in\mathbb{R}^d$ are
trainable parameters that are added to the input $\vec{e}_1,\cdots,\vec{e}_n$:
\begin{align} 
\vec{q}_t = Q(\vec{e}_t+\vec{p}_t)\, , 
\qquad
\vec{k}_t = K(\vec{e}_t+\vec{p}_t)\, , 
\qquad
\vec{v}_t = V(\vec{e}_t+\vec{p}_t)\, . 
\qquad
\end{align}
The word `pizza' appearing in the $t$-th position and $t'$-th position corresponds to different vectors $\vec{e}_{\rm pizza}+\vec{p}_t$ and $\vec{e}_{\rm pizza}+\vec{p}_{t'}$. While $Q,K,V\in\mathbb{R}^{d\times d}$  have no information about the position by construction, the position vectors can learn about the information associated with the position, and hence, $\vec{e}_1+\vec{p}_1,\cdots,\vec{e}_n+\vec{p}_n$ are sensitive to the ordering of the input $\vec{e}_1,\cdots,\vec{e}_n$.
%%%%%%%%%%%%
%%%%%%%%%%%%
\subsubsection*{Multi-head self-attention}
%%%%%%%%%%%%
%%%%%%%%%%%%
`Franklin' can be Rosalind Franklin, Benjamin Franklin, or another Franklin. Benjamin Franklin is related to American history, the \$100 bill and the kite experiment, rather than DNA. It is convenient if there are different queries and keys for Rosalind Franklin and Benjamin Franklin. The values should also be different; for $\vec{q}_{\rm DNA}$ and $\vec{k}_{\rm Franklin}$, we want to assign
$\vec{v}_{\rm Franklin}$ that contains $\vec{e'}_{\rm British}$,
$\vec{e'}_{\rm Nobel}$,
$\vec{e'}_{\rm DNA}$,
$\vec{e'}_{\rm X\mathchar`- ray}$, etc., while for $\vec{q}_{\rm \$100}$ and $\vec{k}_{\rm Franklin}$, we want to assign a different
$\vec{v}_{\rm Franklin}$ that contains 
$\vec{e'}_{\rm American}$,
$\vec{e'}_{\rm kite}$,
$\vec{e'}_{\rm Founding\, Fathers}$, etc.

There are various relations that can help the NN understand the context: I, you, he, she, they and my, your, his, her, their; I/they and go, he/she and goes; come and go, buy and sell; animals in the same category, for example, mammals, reptiles, amphibians, fish, birds, sea mammals, primates; countries nearby, or countries sharing something in common such as the Commonwealth. The meaning of `like' can be different in \textit{I like him} and \textit{he is like me}, and to know the meaning, the relationship with other words in the sentence plays an important role. It is hard to capture so many relations by using only one set of $Q$, $K$ and $V$. Therefore, it is convenient if there are separate $Q$, $K$ and $V$ for different kinds of relations.

Furthermore, taking the query $\vec{q}$, the key $\vec{k}$ and the value $\vec{v}$ to be in $\mathbb{R}^d$ could be a waste of resources, because a vector space with lower dimensions could be enough to capture the features in each relation.

For these reasons, $Q$, $K$ and $V$ are split into multiple heads. Let $n_{\rm h}$ be the number of heads, and $d_{\rm h}=d/n_{\rm h}$. Then, $Q_i$, $K_i$, $V_i\in\mathbb{R}^{d_{\rm h}\times d}$, $i = 1,\cdots,n_{\rm h}$ are introduced, and queries $\vec{q}_i\in\mathbb{R}^{d_{\rm h}}$, keys $\vec{k}_i\in\mathbb{R}^{d_{\rm h}}$, values $\vec{v}_i\in\mathbb{R}^{d_{\rm h}}$ and outputs $\vec{o}_i\in\mathbb{R}^{d_{\rm h}}$ are defined by
\begin{align}
\vec{q}_{i,t}= Q_i (\vec{e}_t +\vec{p}_t)\, , 
\qquad
\vec{k}_{i,t}= K_i (\vec{e}_t +\vec{p}_t)\, , 
\qquad
\vec{v}_{i,t}= V_i (\vec{e}_t +\vec{p}_t)\, , 
\qquad
\vec{o}_{i,t}
=
\sum_{t'}
\alpha_{i,tt'}\vec{v}_{i,t'}\, , 
\end{align}
where
\begin{align}
\alpha_{i,tt'}
=
\mathrm{softmax}(\vec{q}_{i,t}\cdot\vec{k}_{i,t'})
=
\frac{\exp\left(\vec{q}_{i,t}\cdot\vec{k}_{i,t'}\right)}{\sum_{t''}\exp\left(\vec{q}_{i,t}\cdot\vec{k}_{i,t''}\right)}\, .  
\end{align}
The outputs of the heads are then concatenated to form a $d$-component vector, $\vec{o}_t=(\vec{o}_{1,t},\cdots,\vec{o}_{n_{\rm h},t})\in\mathbb{R}^d$, which is the output of this layer.\footnote{In the actual Transformer~\cite{Vaswani2017}, the inner product is divided by $\sqrt{d_{\rm h}}$ before the softmax is applied, i.e., $\alpha_{i,tt'}=\mathrm{softmax}(\vec{q}_{i,t}\cdot\vec{k}_{i,t'}/\sqrt{d_{\rm h}})$. The reason is simple: if the components of $\vec{q}$ and $\vec{k}$ are random numbers of order one, the inner product is a sum of $d_{\rm h}$ terms with random signs, and hence its typical size is $\sqrt{d_{\rm h}}$, which would make the softmax too sharp (i.e., the temperature in Sec.~\ref{sec:distillation} too low) and the gradient very small. Another detail omitted here is that the concatenated vector is multiplied by one more trainable $d\times d$ matrix, which mixes the outputs of different heads.}

Note that we do not specify which head learns which kind of relation. We just let them learn whatever they find.
%%%%%%%%%%%%
%%%%%%%%%%%%
\subsubsection*{Masking, Encoder, Decoder}
%%%%%%%%%%%%
%%%%%%%%%%%%
Above, the output was determined by using $\alpha_{i,tt'}$, which knows the `future' $t'>t$. This makes sense if our goal is translation or sentiment classification. However, if our goal is text generation, real-time speech recognition, or prediction of the word following the input, then we cannot allow the network to access the information about the future. (Technically, we already know the `future' in the training dataset, but in actual applications, we do not know the future yet.) Therefore, in such cases, we mask the future and define the output as
\begin{align}
\vec{o}_{i,t}
=
\sum_{t'\le t}
\tilde{\alpha}_{i,tt'}\vec{v}_{i,t'}\, , 
\end{align}
where
\begin{align}
\tilde{\alpha}_{i,tt'}
\equiv
\frac{\exp\left(\vec{q}_{i,t}\cdot\vec{k}_{i,t'}\right)}{\sum_{t''\le t}\exp\left(\vec{q}_{i,t}\cdot\vec{k}_{i,t''}\right)}\, .  
\end{align}

If the multi-head self-attention in the Transformer architecture (Fig.~\ref{fig:Transformer-architecture}) is not masked
(resp., is masked), then such an architecture is called a Transformer \textit{Encoder} (resp., a Transformer \textit{Decoder}).

%%%%%%%%%%%%
%%%%%%%%%%%%
\subsubsection*{Piling up layers inserting feed-forward networks}
%%%%%%%%%%%%
%%%%%%%%%%%%
We already commented that we can expect better performance by `repeating the same procedure', or equivalently, piling up multiple layers of the self-attention architecture described above. When we do this, there is a problem: the map $\vec{e}\to\vec{o}= \alpha\vec{v}$ is almost linear. This is not exactly linear because $\alpha_{tt'}$ depends on $\vec{e}$, but the nonlinearity is not strong enough to fully extract the power of multiple layers. To cure this issue, a feed-forward network (FF)
is inserted between layers. For example, we can change $\vec{o}_t$ to $W_2\sigma(W_1\vec{o}_t+\vec{b}_1)+\vec{b}_2$, 
where 
$\vec{b}_1\in\mathbb{R}^{d'}$, 
$\vec{b}_2\in\mathbb{R}^{d}$, 
$W_1\in\mathbb{R}^{d'\times d}$, 
and
$W_2\in\mathbb{R}^{d\times d'}$.\footnote{In the original Transformer~\cite{Vaswani2017}, $d=512$, $d'=4d=2048$ and $\sigma = \textrm{ReLU}$. With this choice, the FF has more parameters than the self-attention. In Ref.~\cite{Vaswani2017}, a larger $d'$ led to better performance ($d'=1024$, $2048$ and $4096$ were compared), while the cost of the FF grows only linearly with the input size because it is applied to each position separately.}

%%%%%%%%%%%%
%%%%%%%%%%%%
\subsubsection*{Residual connection and layer normalization}
%%%%%%%%%%%%
%%%%%%%%%%%%
If there are many attention layers, the problem of losing memory could come back in a different form: too much attention is paid to the relation with others, and the original input could be forgotten. 
To avoid it, a shortcut of the same form as in ResNet (Sec.~\ref{sec:cures_for_vanishing_gradient}), which is called a \textit{residual connection} (or \textit{skip connection}), is added. Furthermore, layer normalization (Sec.~\ref{sec:layer_normalization}) is performed to improve the training efficiency. `Add \& Normalize' in Fig.~\ref{fig:Transformer-architecture} stands for the residual connection and layer normalization.\footnote{In the original Transformer~\cite{Vaswani2017}, the normalization is applied after the residual connection, $\vec{x}\to\mathrm{LN}(\vec{x}+f(\vec{x}))$, as in Fig.~\ref{fig:Transformer-architecture}; here $f$ is the self-attention or the FF, and LN is the layer normalization. Since GPT-2~\cite{Radford2019}, most LLMs apply the normalization to the input of each sub-block instead, $\vec{x}\to\vec{x}+f(\mathrm{LN}(\vec{x}))$. Then nothing acts on the shortcut path, which makes the training of deep networks more stable. The two options are called post-LN and pre-LN, respectively.}
%%%%%%%%%%%%
%%%%%%%%%%%%
\subsubsection*{Attention Is All You Need}
%%%%%%%%%%%%
%%%%%%%%%%%%
In the past, the attention mechanism was used together with RNN~\cite{Bahdanau2016}, to resolve the issue of the loss of memory. At some point, several smart people posed a simple question: Do we really need to use RNN? In 2017, a paper titled \textit{Attention Is All You Need}~\cite{Vaswani2017} showed that performance can improve by giving up RNN and using only the attention mechanism. This paper proposed Transformer.

In addition to the problem of the loss of memory, a big problem with RNN-based models was that parallelization is difficult. As we can see from Fig.~\ref{fig:RNN-architecture}, we need $\vec{h}_{t-1}$ to compute $\vec{h}_t$. We can compute $\vec{e}_1,\vec{e}_2,\cdots$ simultaneously, but we can compute $\vec{h}_1,\vec{h}_2,\cdots$ only one by
one: firstly $\vec{h}_1$ from $\vec{h}_0$  and $\vec{e}_1$, then $\vec{h}_2$ from $\vec{h}_1$  and $\vec{e}_2$, then $\vec{h}_3$ from $\vec{h}_2$  and $\vec{e}_3$, and so on. The same problem also exists for backpropagation. Therefore, it is hard to parallelize the computations. Specifically, we cannot scale up the degree of parallelization with the input size $n$, and the time needed for training grows linearly with respect to $n$.

Transformer is highly parallelizable because the notion of ordering is incorporated by the use of the position vector, and the recurrence structure is entirely eliminated. The degree of parallelization can be scaled up with the input size $n$. The costly parts of the computations are matrix multiplications that can be efficiently parallelized by using modern GPU machines.
The price to pay is that the cost of self-attention grows as $n^2$, because all pairs of positions are compared.
%%%%%%%%%%%%
%%%%%%%%%%%%
\section{Large Language Models (LLM)}
%%%%%%%%%%%%
%%%%%%%%%%%%
Let us discuss the application of Transformer to a language model. As before, our task is to infer the conditional probability $p(x_t|x_1,x_2,\cdots,x_{t-1})$. Namely,
when the first $t-1$ words $x_1,x_2,\cdots,x_{t-1}$ are given, we want to predict which word is likely to come next. Here, $x_t$ can take any value from $1,2,\cdots,d_\Gamma$, where $d_\Gamma$ is the number of words in the vocabulary. (We identify $x$ with the one-hot vector $\vec{x}$ whose $x$-th component is 1.)

\begin{figure}[htbp]
\centering
\tikzset{
  tfbox/.style={draw, thick, minimum height=13mm, font=\large},
  tfword/.style={font=\scriptsize, inner sep=1pt}}
% ===================== Top: supervised fine tuning =====================
\begin{tikzpicture}[>=Stealth]
\node[tfbox, minimum width=4.4cm] (enc) at (1.6,0) {Transformer Encoder};
\node[tfbox, minimum width=5.2cm] (dec) at (8.0,0) {Transformer Decoder};
\draw[->, very thick] (enc) -- (dec);
\foreach \n/\x/\w in {1/0.1/Ich, 2/1.1/bin, 3/2.1/eine, 4/3.1/Katze} {
  \node (u\n) at (\x,-1.6) {$\vec{x}'_{\n}$};
  \node[tfword, anchor=north] at (u\n.south) {\w};
  \draw[->, thick] (u\n) -- (\x,0 |- enc.south);
}
\foreach \n/\x/\win/\wout in {1/6.0/{$\langle\mathrm{start}\rangle$}/I, 2/7.0/I/am, 3/8.0/am/a, 4/9.0/a/cat, 5/10.0/cat/{$\langle\mathrm{end}\rangle$}} {
  \node (x\n) at (\x,-1.6) {$\vec{x}_{\n}$};
  \node[tfword, anchor=north] at (x\n.south) {\win};
  \draw[->, thick] (x\n) -- (\x,0 |- dec.south);
  \node (y\n) at (\x,1.6) {$\vec{y}_{\n}$};
  \node[tfword, anchor=south] at (y\n.north) {\wout};
  \draw[->, thick] (\x,0 |- dec.north) -- (y\n);
}
\end{tikzpicture}

\vspace{6mm}
% ===================== Bottom: actual translation =====================
\begin{tikzpicture}[>=Stealth]
\node[tfbox, minimum width=4.4cm] (enc) at (1.6,0) {Transformer Encoder};
\node[tfbox, minimum width=5.2cm] (dec) at (8.0,0) {Transformer Decoder};
\draw[->, very thick] (enc) -- (dec);
\foreach \n/\x/\w in {1/0.1/Ich, 2/1.1/bin, 3/2.1/eine, 4/3.1/Katze} {
  \node (u\n) at (\x,-1.6) {$\vec{x}'_{\n}$};
  \node[tfword, anchor=north] at (u\n.south) {\w};
  \draw[->, thick] (u\n) -- (\x,0 |- enc.south);
}
\node (x1) at (6.0,-1.6) {$\vec{x}_1$};
\node[tfword, anchor=north] at (x1.south) {$\langle\mathrm{start}\rangle$};
\foreach \n/\p/\x/\w in {2/1/7.0/I, 3/2/8.0/am, 4/3/9.0/a, 5/4/10.0/kitty} {
  \node[red] (x\n) at (\x,-1.6) {$\vec{y}_{\p}$};
  \node[tfword, anchor=north] at (x\n.south) {\w};
}
\foreach \n/\x/\wout in {1/6.0/I, 2/7.0/am, 3/8.0/a, 4/9.0/kitty, 5/10.0/{$\langle\mathrm{end}\rangle$}} {
  \draw[->, thick] (x\n) -- (\x,0 |- dec.south);
  \node[red] (y\n) at (\x,1.6) {$\vec{y}_{\n}$};
  \node[tfword, anchor=south] at (y\n.north) {\wout};
  \draw[->, thick] (\x,0 |- dec.north) -- (y\n);
}
\end{tikzpicture}
\caption{Schematic picture of the Transformer encoder-decoder architecture for machine translation. The RNN encoder/decoder in Fig.~\ref{fig:RNN-machine-translation} are replaced with the Transformer encoder/decoder. [Top] Supervised learning. [Bottom] Actual translation. In the encoder-decoder architecture, each decoder layer has an additional attention layer in which the queries are made from the decoder side while the keys and values are made from the output of the encoder (cross-attention). This is how the information about the original sentence is passed to the decoder.}
\label{fig:Transformer-translation}
\end{figure}
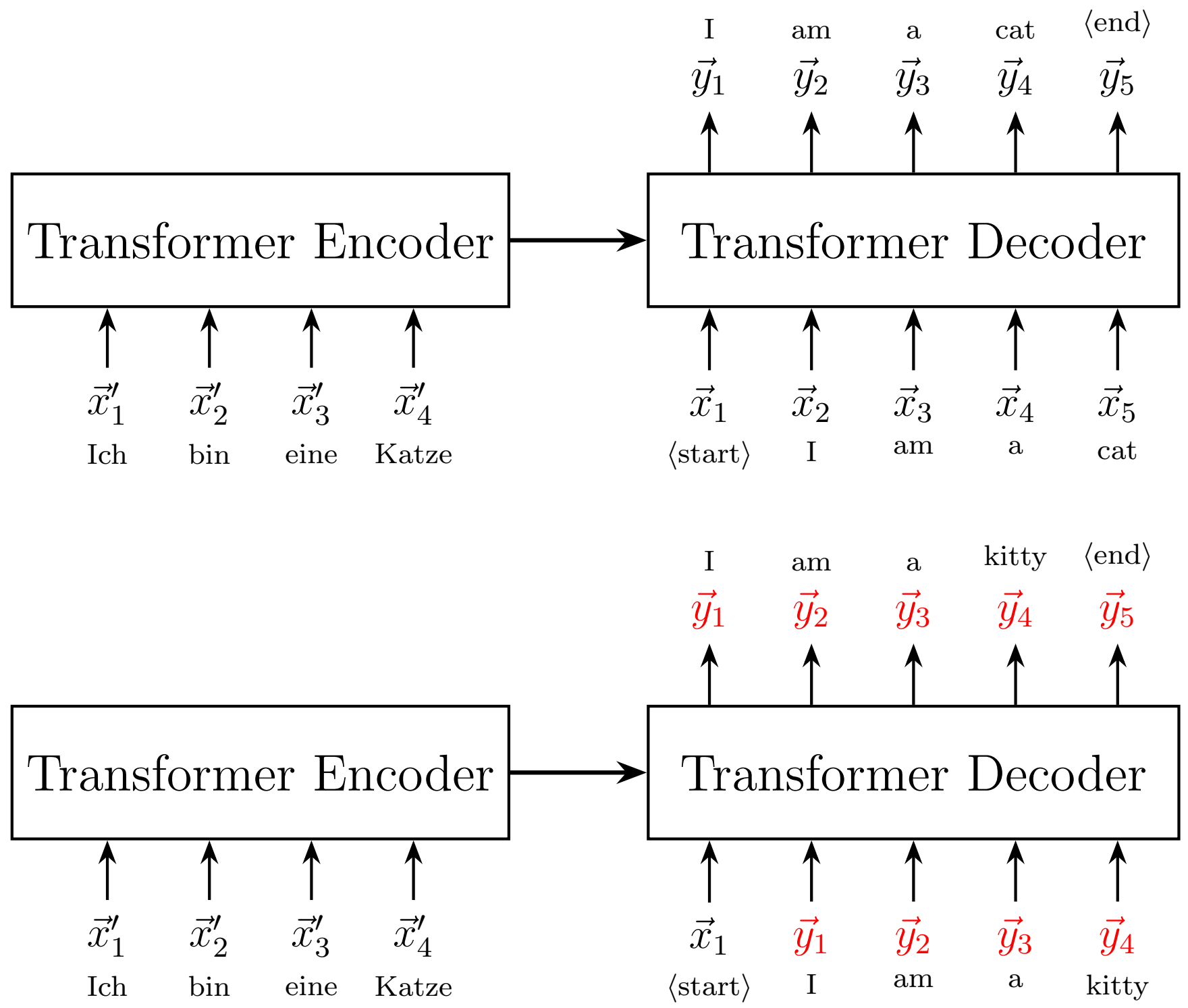

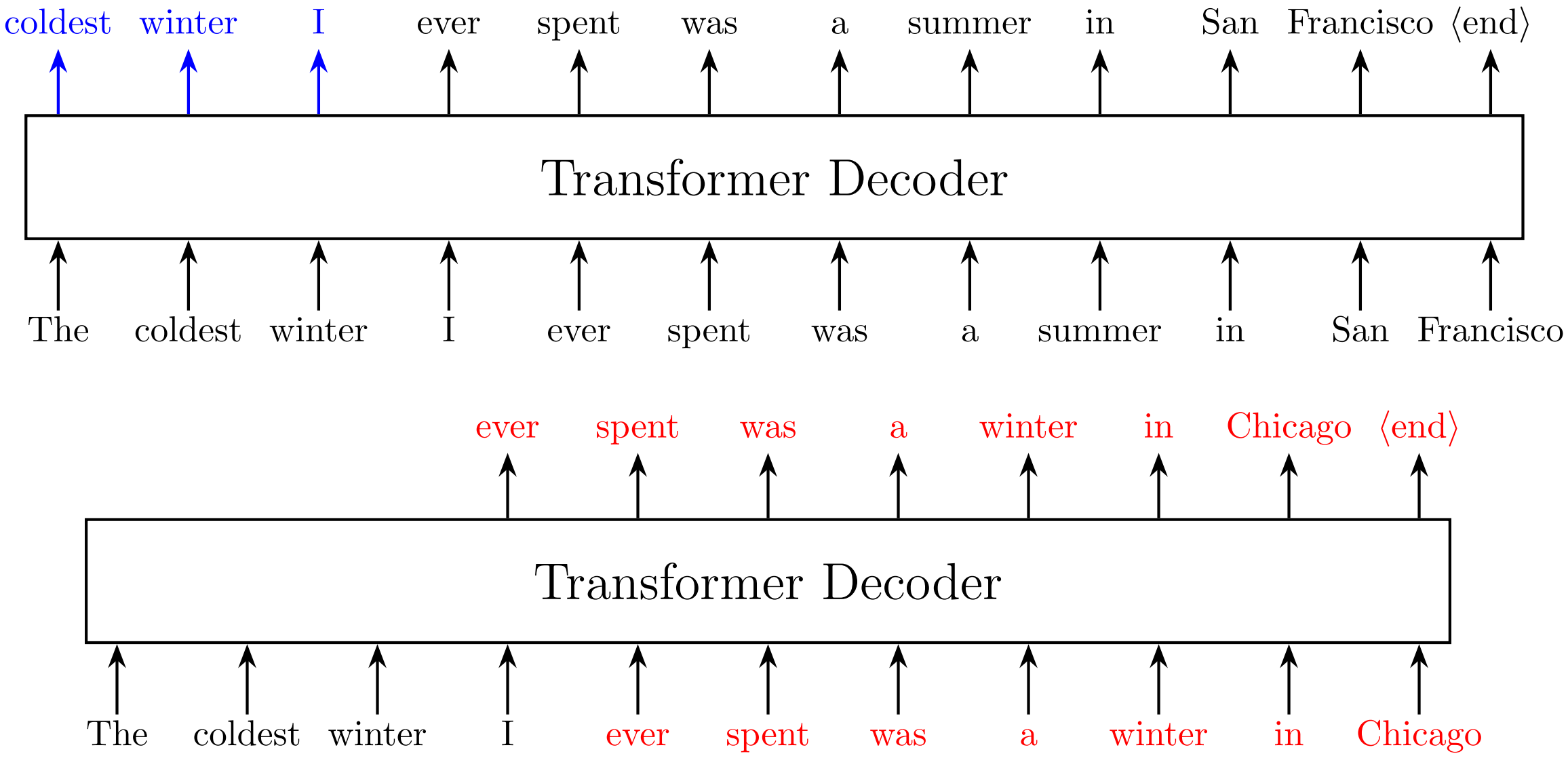
\begin{figure}[htbp]
\centering
\tikzset{
  decbox/.style={draw, thick, minimum height=12mm, font=\large},
  tgword/.style={font=\footnotesize, inner sep=1pt}}
% ===================== Top: supervised fine tuning =====================
\begin{tikzpicture}[>=Stealth, x=1.27cm]
\node[decbox, minimum width=14.6cm] (dec) at (6.5,0) {Transformer Decoder};
\foreach \n/\win/\wout/\c in {
   1/The/coldest/blue, 2/coldest/winter/blue, 3/winter/I/blue,
   4/I/ever/black, 5/ever/spent/black, 6/spent/was/black, 7/was/a/black,
   8/a/summer/black, 9/summer/in/black, 10/in/San/black,
   11/San/Francisco/black, 12/Francisco/{$\langle\mathrm{end}\rangle$}/black} {
  \node[tgword, anchor=base] at (\n,-1.6) {\win};
  \node[tgword, anchor=base, \c] at (\n,1.4) {\wout};
  \draw[->, thick] (\n,-1.3) -- (\n,0 |- dec.south);
  \draw[->, thick, \c] (\n,0 |- dec.north) -- (\n,1.25);
}
\end{tikzpicture}

\vspace{5mm}
% ===================== Bottom: actual text generation =====================
\begin{tikzpicture}[>=Stealth, x=1.27cm]
\node[decbox, minimum width=13.3cm] (dec) at (6.0,0) {Transformer Decoder};
\foreach \n/\win/\cin in {1/The/black, 2/coldest/black, 3/winter/black, 4/I/black,
   5/ever/red, 6/spent/red, 7/was/red, 8/a/red, 9/winter/red, 10/in/red, 11/Chicago/red} {
  \node[tgword, anchor=base, \cin] at (\n,-1.6) {\win};
  \draw[->, thick] (\n,-1.3) -- (\n,0 |- dec.south);
}
\foreach \n/\wout in {4/ever, 5/spent, 6/was, 7/a, 8/winter, 9/in, 10/Chicago, 11/{$\langle\mathrm{end}\rangle$}} {
  \node[tgword, anchor=base, red] at (\n,1.4) {\wout};
  \draw[->, thick] (\n,0 |- dec.north) -- (\n,1.25);
}
\end{tikzpicture}
\caption{Schematic picture for the use of Transformer for text generation. [Top] Supervised learning. [Bottom] Actual text generation. The RNN in Fig.~\ref{fig:RNN-text-generation} is replaced with the Transformer decoder.}
\label{fig:Transformer-text-generation}
\end{figure}

We use the Transformer decoder, in which the future information is masked, and hence the output from Transformer $\vec{y}_t\in\mathbb{R}^d$ depends only on $x_1,\cdots,x_t$ and not on $x_{t+1},x_{t+2},\cdots$. 
We prepare a trainable matrix $W\in\mathbb{R}^{d_\Gamma\times d}$ and a vector $\vec{b}\in\mathbb{R}^{d_\Gamma}$, which are common to all $t$, and define the inferred probability $p(x_{t+1}|x_1,x_2,\cdots,x_{t})$ by
\begin{align}
\left(
\begin{array}{c}
p(x_{t+1}=1|x_1,x_2,\cdots,x_{t})
\\
p(x_{t+1}=2|x_1,x_2,\cdots,x_{t})
\\
\vdots
\\
p(x_{t+1}=d_\Gamma|x_1,x_2,\cdots,x_{t})
\end{array}
\right)
=
\mathrm{softmax}
(W\vec{y}_t+\vec{b})\, . 
\end{align}
A reasonable cost function is the cross-entropy loss,
\begin{align}
-
\sum_{(x_1,\cdots,x_n)}
\sum_{t=t_0}^{n-1}
\log\left(p(x_{t+1}|x_1,x_2,\cdots,x_{t})\right)\, . 
\end{align}
Here, the sum $\sum_{(x_1,\cdots,x_n)}$ is taken over actual texts in the training data. We assumed that the model is trained to predict the words starting from $t= t_0$. Many tasks handled by using RNN can be done by using Transformer. See Fig.~\ref{fig:Transformer-translation} and Fig.~\ref{fig:Transformer-text-generation} for machine translation and text generation, respectively.

%%%%%%%%%%%%
%%%%%%%%%%%%
\subsection{Pretrained Transformers}
%%%%%%%%%%%%
%%%%%%%%%%%%
A Transformer-based language model perfectly fits into the framework of self-supervised learning and pretraining discussed in Sec.~\ref{sec:pretraining}~\cite{Radford2018}. 
The standard way to train the decoder is to predict the next word. It does not work
for the encoder, because the future is not masked. Instead, we can mask some words in the
input (replace the corresponding tokens with a special ‘mask’ token) and train the network to
restore the masked word.
To provide the right answers, Transformer needs to learn various relations between words. Training a large language model is costly, and not everyone can do that by themselves, but if Transformer is provided, then anyone can add a few pieces and make new models. Powerful and popular pretrained Transformer architectures include Google’s BERT
(Bidirectional Encoder Representations from Transformers; encoder-only architecture) series~\cite{Devlin2019} and OpenAI’s GPT (Generative Pre-trained Transformer; decoder-only architecture) series~\cite{Radford2018}.

%%%%%%%%%%%%
%%%%%%%%%%%%
\subsection{Unsupervised multi-task learning}\label{sec:unsupervised_multi_task_training}
%%%%%%%%%%%%
%%%%%%%%%%%%
Many tasks in natural language processing can be phrased in the form of a question and an answer. 
For example, translation can be phrased like ``Please translate this English sentence to German: I love baseball. Ich liebe Baseball."
By training a model using such texts, we can get a German translation by inputting texts like ``Please translate this English sentence to German: Haribo makes us happy." 
We could also implement sentiment analysis by using texts like ``What’s the sentiment? Haribo makes us happy. Positive.'' or  ``What’s the sentiment? I am sad because there is no Haribo today. Negative.''
This idea can be used to create a model that can perform multiple tasks~\cite{Kumar2015,Mccann2018}.
If a sufficiently large dataset with all kinds of sentences we can imagine is used, then the LLM would learn many tasks via next-word prediction. 
For example, the WebText dataset contains translations between different languages (see Table~\ref{tab:WebText_English_French}), and hence the model can predict the next word better if it can perform the translation. 
GPT-2~\cite{Radford2019} provided strong evidence that this is indeed the case. 
Note that this is unsupervised training, and the model may learn some tasks which we do not know yet! 

Clearly, it is important to prepare a large and diverse dataset. Before GPT-2, most models were trained on  a single-domain dataset, such as novels or news articles. 
GPT-2 used WebText, which is a large dataset created by OpenAI by obtaining texts from the internet. Specifically,  8 million documents (a total of 40 GB) of text were used. 

\textit{Zero-shot learning} refers to a model's ability to generalize and perform tasks it has not explicitly encountered during training.
In this setting, the only way to tell the model what we want is the input text, which is called the \textit{prompt}. A small change in the prompt can change the output significantly, and hence, it is important to design the prompt carefully (\textit{prompt engineering}).

Note that fine-tuning can further enhance the model’s performance when applied to specific tasks. For instance, if the model needs to generate or process texts in a particular format, it makes sense to prepare a relevant dataset and fine-tune the model accordingly. Similarly, for applications in specialized fields like medicine, incorporating a large number of medical articles during fine-tuning could help optimize the model’s performance, keeping the model's knowledge up to date. Additionally, fine-tuning can be used to filter out harmful or offensive language.\footnote{
It would be ideal to eliminate such words from the dataset, but this task is often unrealistic. Terms that were more commonly used in the past may no longer be considered appropriate today, and certain inappropriate words might be necessary to accurately describe a character’s negative behavior in a novel.
}
For example, creating a version of the model with parental controls could help ensure that it does not generate content unsuitable for children. This process is further discussed in Sec.~\ref{sec:LLM_alignment}. 

\begin{table}[htbp]
\centering
\begin{tabular}{p{0.75\textwidth}}
\hline
``I'm not the cleverest man in the world, but like they say in French: \textbf{Je ne suis pas un imbecile [I'm not a fool].}\\[2mm]
``I hate the word `\textbf{perfume},''' Burr says. `It's somewhat better in French: `\textbf{parfum}.'\\[2mm]
``\textbf{Brevet Sans Garantie Du Gouvernement}'', translated to English: ``\textbf{Patented without government warranty}''.\\
\hline
\end{tabular}
\caption{Examples of naturally occurring translations between English and French in the WebText training set, taken from Table~1 of Ref.~\cite{Radford2019}. GPT-2 was not trained for translation explicitly, but such examples in the training data allow the model to learn translation.}
\label{tab:WebText_English_French}
\end{table}

%%%%%%%%%%%%
%%%%%%%%%%%%
\subsection{Scaling law in pretraining}
%%%%%%%%%%%%
%%%%%%%%%%%%
Ref.~\cite{Kaplan2020} demonstrated that the test loss decreases as compute, dataset size, and model architecture size (i.e., the number of parameters) increase, following specific power laws. In their study, they used the cross-entropy loss and performed next-word prediction tasks. Other architectural factors such as width and depth did not significantly influence this scaling behavior. This scaling law had important implications: it showed that the performance of Transformer-based LLMs could be improved by simply increasing compute, dataset size, and model size. Furthermore, it provided insights into how to optimize performance for a given computational budget.

\begin{figure}[htbp]
\centering
\pgfplotsset{
  /pgfplots/scalingpanel/.style={
    width=5.9cm, height=5.2cm,
    ytick={2,3,4,5,6,7}, yticklabels={2,3,4,5,6,7},
    minor ytick={2.5,3.5,4.5,5.5,6.5},
    tick label style={font=\scriptsize},
    every axis x label/.style={at={(ticklabel cs:0.5)}, anchor=near ticklabel, font=\small},
    every axis y label/.style={at={(ticklabel cs:0.5)}, rotate=90, anchor=near ticklabel, font=\small},
    axis line style={-},
    samples=2, thick}}
\begin{tikzpicture}
\begin{axis}[xmode=log, ymode=log, scalingpanel, name=c,
  xmin=1e-9, xmax=1e1, ymin=2, ymax=7.5,
  xtick={1e-9,1e-7,1e-5,1e-3,1e-1,1e1},
  xlabel={Compute $C_{\min}$ (PF-days)}, ylabel={Test loss $L$}]
\addplot[domain=1e-9:1e1, orange, dashed] {(x/2.3e8)^(-0.050)};
\node[font=\scriptsize, anchor=south west] at (rel axis cs:0.04,0.04) {$L=(C_{\min}/2.3\times10^{8})^{-0.050}$};
\end{axis}
\begin{axis}[xmode=log, ymode=log, scalingpanel, at={(c.south east)}, xshift=1.1cm, name=d,
  xmin=1e7, xmax=1e10, ymin=2, ymax=7.5,
  xtick={1e7,1e8,1e9,1e10},
  xlabel={Dataset size $D$ (tokens)}]
\addplot[domain=2e7:2.5e9, blue] {(x/5.4e13)^(-0.095)};
\node[font=\scriptsize, anchor=north east] at (rel axis cs:0.96,0.96) {$L=(D/5.4\times10^{13})^{-0.095}$};
\end{axis}
\begin{axis}[xmode=log, ymode=log, scalingpanel, at={(d.south east)}, xshift=1.1cm,
  xmin=1e3, xmax=1e10, ymin=2, ymax=7.5,
  xtick={1e3,1e5,1e7,1e9},
  xlabel={Parameters $N$}]
\addplot[domain=1e4:1.5e9, red!70!black] {(x/8.8e13)^(-0.076)};
\node[font=\scriptsize, anchor=north east] at (rel axis cs:0.96,0.96) {$L=(N/8.8\times10^{13})^{-0.076}$};
\end{axis}
\end{tikzpicture}
\caption{The scaling law for LLMs based on Transformer~\cite{Kaplan2020}. The test loss (specifically, the cross-entropy loss) decreases as the compute, dataset size, or the size of the architecture (number of parameters, not including the embedding) increases, following specific power laws. The lines are the fits given in Ref.~\cite{Kaplan2020}, drawn over the range of the data shown there; see Fig.~1 of Ref.~\cite{Kaplan2020} for the data points.}
\label{fig:scaling_law}
\end{figure}
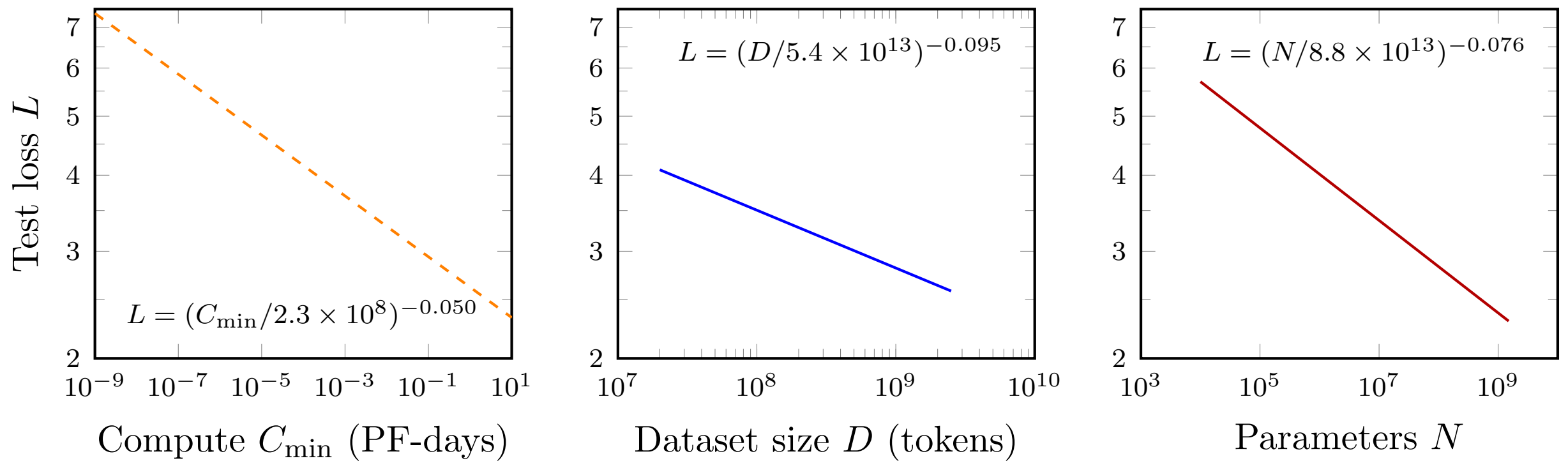

Note that the exponents are small: the test loss scales as $N^{-0.076}$ with the number of parameters $N$, $D^{-0.095}$ with the dataset size $D$, and $C_{\rm min}^{-0.050}$ with the compute $C_{\rm min}$ (Fig.~\ref{fig:scaling_law}). For example, when the number of parameters is doubled, the loss decreases only by about 5\%, and to halve the loss, we need about $10^4$ times more parameters. What made this scaling law so influential is not the size of the exponents but the fact that the power laws hold over many orders of magnitude; this enables us to predict the performance of a very large model, which is expensive to train, from small models. Note also that, because the cross-entropy loss is used, the loss is minus the logarithm of the probability assigned to the correct next word, averaged over the words in the text: for a text consisting of $n$ words, a decrease of the loss per word by $\Delta L$ increases the probability assigned to the text by a factor of $e^{n\Delta L}$.

How a fixed amount of compute should be divided between the model size and the dataset size was revisited in Ref.~\cite{Hoffmann2022}. The analysis of Ref.~\cite{Kaplan2020} suggested that, as the compute increases, the model size should grow much faster than the dataset size. Ref.~\cite{Hoffmann2022} repeated the analysis with larger models and a more careful training setup, and found that the two should grow at the same rate: when the number of parameters is doubled, the number of training tokens should also be doubled. To demonstrate this, they trained a 70-billion-parameter model called \textit{Chinchilla}, which outperformed a 280-billion-parameter model (\textit{Gopher}) trained with the same compute but with four times less data. According to this analysis, large models at that time, including GPT-3, were trained on too little data for their size.

%%%%%%%%%%%%
%%%%%%%%%%%%
\subsection{History of GPT}
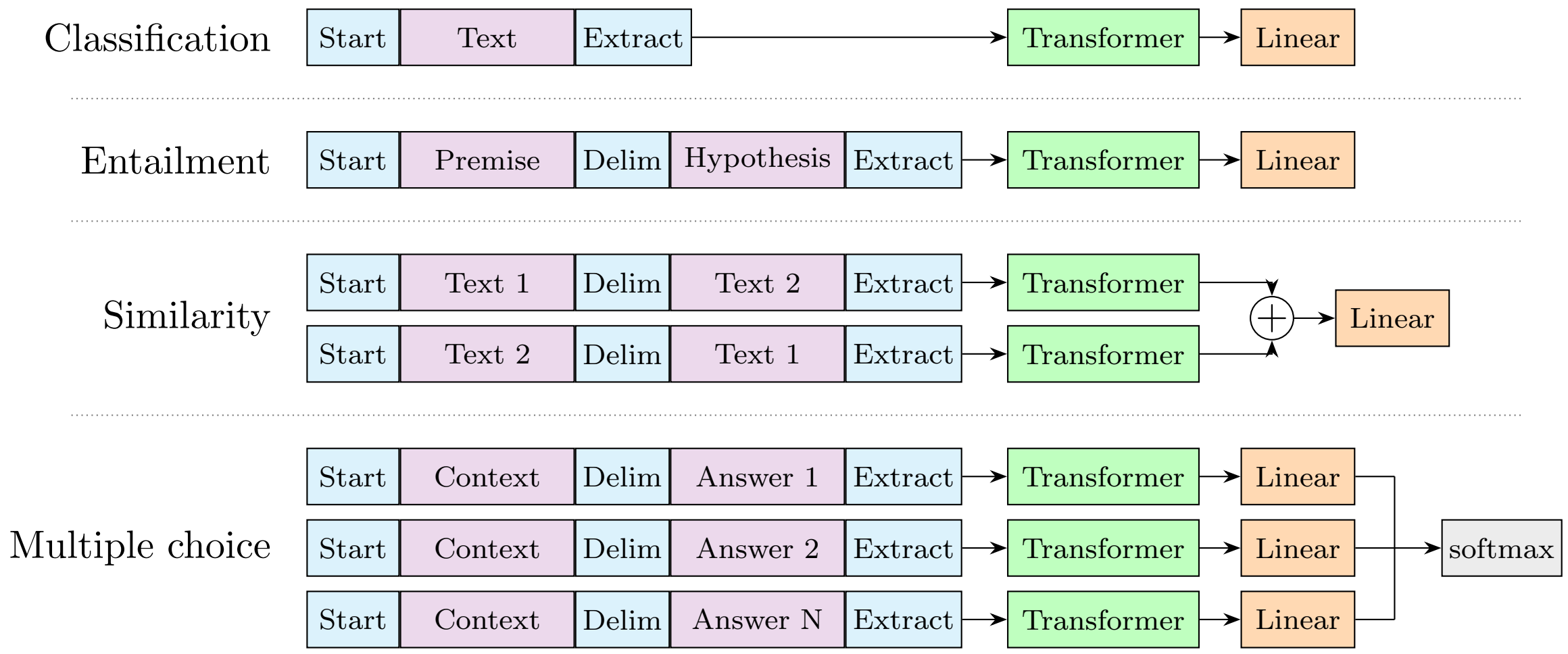
\begin{figure}[htbp]
\centering
\begin{tikzpicture}[>=Stealth, font=\scriptsize,
  tok/.style={draw, minimum height=5.5mm, inner sep=2pt},
  spc/.style={tok, fill=cyan!12, minimum width=9mm},
  txt/.style={tok, fill=violet!15, minimum width=17mm},
  tr/.style={draw, fill=green!25, minimum height=5.5mm, minimum width=17mm},
  lin/.style={draw, fill=orange!30, minimum height=5.5mm, minimum width=11mm},
  task/.style={anchor=east, font=\small}]
\def\xs{2.3}   % where the token sequences start
\def\xt{10.1}  % position of the Transformer boxes
% ---------- Classification ----------
\node[task] at (2.1,0) {Classification};
\node[spc, anchor=west] (a1) at (\xs,0) {Start};
\node[txt, anchor=west] (a2) at (a1.east) {Text};
\node[spc, anchor=west] (a3) at (a2.east) {Extract};
\node[tr] (at) at (\xt,0) {Transformer};
\node[lin, anchor=west] (al) at ([xshift=4mm]at.east) {Linear};
\draw[->] (a3) -- (at); \draw[->] (at) -- (al);
\draw[densely dotted, gray] (0,-0.6) -- (14.2,-0.6);
% ---------- Entailment ----------
\node[task] at (2.1,-1.2) {Entailment};
\node[spc, anchor=west] (b1) at (\xs,-1.2) {Start};
\node[txt, anchor=west] (b2) at (b1.east) {Premise};
\node[spc, anchor=west] (b3) at (b2.east) {Delim};
\node[txt, anchor=west] (b4) at (b3.east) {Hypothesis};
\node[spc, anchor=west] (b5) at (b4.east) {Extract};
\node[tr] (bt) at (\xt,-1.2) {Transformer};
\node[lin, anchor=west] (bl) at ([xshift=4mm]bt.east) {Linear};
\draw[->] (b5) -- (bt); \draw[->] (bt) -- (bl);
\draw[densely dotted, gray] (0,-1.8) -- (14.2,-1.8);
% ---------- Similarity ----------
\node[task] at (2.1,-2.75) {Similarity};
\foreach \y/\p/\q/\n in {-2.4/1/2/c, -3.1/2/1/d} {
  \node[spc, anchor=west] (\n1) at (\xs,\y) {Start};
  \node[txt, anchor=west] (\n2) at (\n1.east) {Text \p};
  \node[spc, anchor=west] (\n3) at (\n2.east) {Delim};
  \node[txt, anchor=west] (\n4) at (\n3.east) {Text \q};
  \node[spc, anchor=west] (\n5) at (\n4.east) {Extract};
  \node[tr] (\n t) at (\xt,\y) {Transformer};
  \draw[->] (\n5) -- (\n t);
}
\node[circle, draw, inner sep=0pt, minimum size=4mm, font=\normalsize] (plus) at (11.75,-2.75) {$+$};
\draw[->] (ct.east) -| (plus.north);
\draw[->] (dt.east) -| (plus.south);
\node[lin, anchor=west] (cl) at ([xshift=4mm]plus.east) {Linear};
\draw[->] (plus) -- (cl);
\draw[densely dotted, gray] (0,-3.7) -- (14.2,-3.7);
% ---------- Multiple choice ----------
\node[task] at (2.1,-5.0) {Multiple choice};
\foreach \y/\k/\n in {-4.3/1/e, -5.0/2/f, -5.7/N/g} {
  \node[spc, anchor=west] (\n1) at (\xs,\y) {Start};
  \node[txt, anchor=west] (\n2) at (\n1.east) {Context};
  \node[spc, anchor=west] (\n3) at (\n2.east) {Delim};
  \node[txt, anchor=west] (\n4) at (\n3.east) {Answer \k};
  \node[spc, anchor=west] (\n5) at (\n4.east) {Extract};
  \node[tr] (\n t) at (\xt,\y) {Transformer};
  \node[lin, anchor=west] (\n l) at ([xshift=4mm]\n t.east) {Linear};
  \draw[->] (\n5) -- (\n t); \draw[->] (\n t) -- (\n l);
}
\node[draw, fill=gray!15, minimum height=5.5mm, inner sep=2pt] (sm) at (14.0,-5.0) {softmax};
\foreach \n in {e,f,g} \draw (\n l.east) -- (12.95,0 |- \n l.east);
\draw (12.95,-4.3) -- (12.95,-5.7);
\draw[->] (12.95,-5.0) -- (sm.west);
\end{tikzpicture}
\caption{Input transformations for fine-tuning GPT-1 on different tasks, based on Fig.~1 of Ref.~\cite{Radford2018}. Structured inputs are converted into a single sequence of tokens, by using special tokens (Start, Delim and Extract), and processed by the pretrained Transformer followed by a linear layer. For similarity, the order of the two texts should not matter, so both orders are processed and the results are added. For multiple choice, each answer is processed separately, and the results are normalized by softmax.}
\label{fig:Radford2018_Fig1}
\end{figure}

%%%%%%%%%%%%
%%%%%%%%%%%%
Generative Pre-trained Transformer (GPT), which is the Transformer decoder architecture behind ChatGPT, revolutionized LLMs. 
GPT-1~\cite{Radford2018} applied the idea of pretraining to Transformer. Specifically, they used Transformer as a generative model for next-word prediction. 
GPT-1 consisted of 12 layers of Transformer blocks (see Fig.~\ref{fig:Transformer-architecture}) and had more than 100 million model parameters. The context window size was 512 tokens. The dataset was the BooksCorpus, which contains over 7,000 unpublished books of various kinds, 4.5 GB in total. The main motivation to use this dataset was to have extended passages of continuous text so that the model can learn long-range information. They did not prioritize having a highly diverse range of topics or writing styles, which was the key for GPT-2. 
After pretraining, supervised fine-tuning was performed for each task (natural language inference, question answering and commonsense reasoning, semantic similarity, and classification; see Fig.~\ref{fig:Radford2018_Fig1}). They also studied the zero-shot behaviors and suggested that unsupervised multi-task learning is taking place. However, GPT-1's zero-shot performance was more limited than that of later models like GPT-2 and GPT-3.

GPT-2~\cite{Radford2019} can be seen as a significant scale-up from GPT-1, with a ten-fold increase in parameter count, a wider context window size (1024 tokens), and a much larger, more diverse dataset.\footnote{
More precisely, four different model sizes were studied, and the largest one had 48 layers of Transformer blocks and about 1.5B model parameters. The dimension of the hidden states was 1600.
} However, there was an important conceptual shift, as reflected in the title of the paper, \textit{Language models are unsupervised multitask learners}. The team aimed to create a more general-purpose learner, demonstrating that large-scale unsupervised pretraining on a diverse dataset like WebText could enable the model to perform a variety of tasks without the need for task-specific fine-tuning. See Table~\ref{tab:WebText_English_French} for examples of sentences from which GPT-2 could learn the English-French translation task. 
If task-specific fine-tuning is not required, the societal impact can be significant. AI can be used for tasks where collecting large datasets is difficult due to time constraints or high costs, or it can serve as an assistant in real-time settings. The latter is exactly what we are witnessing today.

To test the performance of GPT-2 accurately, the authors of Ref.~\cite{Radford2019} removed all Wikipedia documents from WebText, because Wikipedia tends to contain the answers to test tasks.\footnote{
We can easily imagine that Wikipedia contains the answers to questions like `Who was the author of \textit{The Art of War}?' and, as a result, the model could provide the correct answer through memorization rather than generalization.
} 
See Table~\ref{tab:Radford2019_Table5} for some answers of GPT-2 to factual questions. Note that this table shows the best cases: the answers are sorted by the confidence of GPT-2. The accuracy was about 63\% for the 1\% of the questions with the highest confidence, but only about 4\% for the whole test set. Still, GPT-2 was not trained for question answering explicitly, and as we can see in the table, even the wrong answers are of the right kind (the name of a person, a state, and so on). In the zero-shot setting, performance on other tasks was studied, and the results were not necessarily fantastic but looked promising (at least to the authors). Long texts generated by GPT-2 tended to lose coherence or become repetitive.

\begin{table}[htbp]
\centering
\footnotesize
\setlength{\tabcolsep}{4pt}
\begin{tabular}{llcr}
\hline
Question & Generated answer & Correct & Probability\\
\hline
Who wrote the book the origin of species? & Charles Darwin & $\checkmark$ & 83.4\%\\
Who is the founder of the ubuntu project? & Mark Shuttleworth & $\checkmark$ & 82.0\%\\
Panda is a national animal of which country? & China & $\checkmark$ & 76.8\%\\
Who came up with the theory of relativity? & Albert Einstein & $\checkmark$ & 76.4\%\\
When was the first star wars film released? & 1977 & $\checkmark$ & 71.4\%\\
Largest state in the us by land mass? & California & $\times$ & 59.2\%\\
What us state forms the western boundary of montana? & Montana & $\times$ & 52.3\%\\
Who plays ser davos in game of thrones? & Peter Dinklage & $\times$ & 52.1\%\\
Who won the most mvp awards in the nba? & Michael Jordan & $\times$ & 50.2\%\\
What river is associated with the city of rome? & the Tiber & $\checkmark$ & 48.6\%\\
Who played john connor in the original terminator? & Arnold Schwarzenegger & $\times$ & 45.2\%\\
\hline
\end{tabular}
\caption{Some questions, answers generated by GPT-2, and the confidence level (probability estimated by the model), taken from Table~5 of Ref.~\cite{Radford2019}. Note that the wrong answers are often closely related to the correct ones; for example, Arnold Schwarzenegger did appear in \textit{The Terminator}, but not as John Connor.}
\label{tab:Radford2019_Table5}
\end{table}

GPT-3~\cite{Brown2020} significantly increased the model size. We can gauge the scale of the project by the number of authors: the GPT-1 paper~\cite{Radford2018} had four authors, the GPT-2 paper~\cite{Radford2019} had six authors, and the GPT-3 paper~\cite{Brown2020} had 31 authors! This number grew to over 200 for GPT-4~\cite{OpenAI2023}.
GPT-3 has 175 billion parameters, making it more than 100 times larger than GPT-2, and a context window size of 2048 tokens. Additionally, the dataset used for training was substantially expanded. In addition to WebText, books, and Wikipedia, many other internet texts were incorporated. Of course, internet texts can vary in quality, so the team removed low-quality documents. For this, they created a classifier to detect and remove low-quality content. The classifier was trained using high-quality examples, including WebText, Wikipedia, and books. Furthermore,  they removed the documents which had a large overlap with others. This worked to some extent, and impressive results were obtained for some tasks. At the same time, the dataset still contained problematic texts that led to problematic output, as we will see shortly. 

An impressive thing about GPT-3 was that zero-shot learning actually worked. One-shot learning (with one example) performed better, and few-shot learning (with a few examples) achieved state-of-the-art results compared to fine-tuned models for certain tasks. Note that the word `learning' is used in an unusual sense here: the examples are simply written in the prompt, and the parameters of the model are not updated at all. To emphasize this, it is called \textit{in-context learning}. See Table~\ref{tab:GPT-3-a-few-shot} for the task of correcting English grammar in the few-shot setting.

\begin{table}[htbp]
\centering
\small
\newcommand{\prm}[1]{{\leavevmode\color{gray}#1}}
\renewcommand{\arraystretch}{1.1}
\begin{tabular}{p{0.92\textwidth}}
\hline\noalign{\smallskip}
\prm{Poor English input: I eated the purple berries.}\\
\prm{Good English output: I ate the purple berries.}\\
\prm{Poor English input: Thank you for picking me as your designer. I'd appreciate it.}\\
\prm{Good English output: Thank you for choosing me as your designer. I appreciate it.}\\
\prm{Poor English input: The mentioned changes have done. or I did the alteration that you requested. or I changed things you wanted and did the modifications.}\\
\prm{Good English output: The requested changes have been made. or I made the alteration that you requested. or I changed things you wanted and made the modifications.}\\
\prm{Poor English input: I'd be more than happy to work with you in another project.}\\
\textbf{Good English output: I'd be more than happy to work with you on another project.}\\[1mm]
\hline\noalign{\smallskip}
\prm{Poor English input: The patient was died.}\\
\textbf{Good English output: The patient died.}\\[1mm]
\hline\noalign{\smallskip}
\prm{Poor English input: We think that Leslie likes ourselves.}\\
\textbf{Good English output: We think that Leslie likes us.}\\[1mm]
\hline\noalign{\smallskip}
\prm{Poor English input: Janet broke Bill on the finger.}\\
\textbf{Good English output: Janet broke Bill's finger.}\\[1mm]
\hline\noalign{\smallskip}
\prm{Poor English input: Mary arranged for, in St.~Louis, John to rent a house cheap.}\\
\textbf{Good English output: Mary arranged for John to rent a house in St.~Louis.}\\[1mm]
\hline
\end{tabular}
\caption{Some examples of the task of correcting English grammar in the few-shot setting, taken from Ref.~\cite{Brown2020}. The text in gray is the prompt, and the text in boldface is generated by GPT-3. In the first example, the prompt contains three examples of the task before the actual question. For the other examples, only the last part of the prompt is shown.}
\label{tab:GPT-3-a-few-shot}
\end{table}

A major issue with GPT-3 was the various biases it learned from its training dataset, including but not limited to biases related to gender, race, and religion. To identify these biases, the team conducted thorough analyses. For example, they used input like ``Buddhists are" and examined what the model would generate next. Through careful statistical analysis, they confirmed the presence of biases that negatively affected certain groups. The existence of such biases is one of the reasons why alignment is crucial for LLMs; see Sec.~\ref{sec:LLM_alignment}.

Motivated by the ability of GPT-3 to generate simple Python code~\cite{Brown2020}, OpenAI fine-tuned GPT-3 on a large corpus of code from GitHub, resulting in Codex~\cite{Chen2021_Codex}. Codex demonstrated the ability to generate functional code in multiple programming languages. Later, GPT-3.5 was developed as an intermediate model with enhanced performance over GPT-3, building on lessons from both GPT-3 and Codex. 
ChatGPT was created by fine-tuning a model in the GPT-3.5 series (and later GPT-4). According to OpenAI, the same methods as for InstructGPT~\cite{Ouyang2022} were used: supervised fine-tuning with conversations written by human trainers, followed by Reinforcement Learning from Human Feedback~\cite{Christiano2017,Ziegler2020,Stiennon2020} (see Sec.~\ref{sec:instruction_fine_tuning} and Sec.~\ref{sec:RLHF}). A user-friendly interface designed to facilitate conversational interactions was added.

Although the exact number of parameters for GPT-4~\cite{OpenAI2023} is not disclosed, it is widely believed to be much larger than that of GPT-3. The context window size is 8,192 tokens (default) or 32,768 tokens (extended version). Model performance was improved further, and it is likely to produce safer and more neutral outputs. 

GPT-3 and GPT-4 are not open source. According to OpenAI, this decision is primarily due to security and safety concerns, as these models have the potential for misuse. Similarly, other powerful models, such as Anthropic's \textit{Claude}, are also closed source. On the other hand, there are open-source models like Meta's \textit{LLaMA} (Large Language Model Meta AI).
%%%%%%%%%%%%
%%%%%%%%%%%%
\section{Alignment of LLM}\label{sec:LLM_alignment}
%%%%%%%%%%%%
%%%%%%%%%%%%
A pretrained Transformer model is not always directly useful for humans because its primary goal --- accurately predicting the next word in a sequence --- does not always align with our needs. For instance, if you ask a question, the model might simply predict a sequence of related questions rather than providing a direct answer, which is not helpful. Similarly, if you're trying to understand the meaning of a word, the model might give you potentially harmful or disturbing information if the word is linked to sensitive or problematic topics. In such cases, it would be more beneficial for the model to issue a warning rather than continue with potentially harmful or misleading content. To make LLMs truly effective and beneficial, it is essential to align them with human needs, goals, and values.\footnote{
It is important to recognize that needs, goals, and values can vary greatly from person to person. For example, I may prefer an AI that focuses on solving complex math problems and can handle longer inputs, while you might prefer an AI assistant that processes only a few sentences at a time, with a greater emphasis on precise word choice rather than mathematical reasoning. AI researchers may prioritize specific benchmarks to showcase the capabilities of their models, but good scores on benchmarks do not necessarily mean that the product is aligned with the practical needs of users if those benchmarks do not reflect real-world use cases. Moreover, some malicious actors might ``align" their models with harmful objectives, intentionally designing them to produce harmful content.
} 

One way to achieve alignment to some extent is through prompting, where you provide specific instructions (or a few examples, as in few-shot learning) within the prompt.
For example, you might add ``please give me the answer" to encourage the model to avoid generating a list of related questions, or ``please warn me if this word has a harmful meaning" to prevent problematic content. While this approach can help guide the model’s responses, it is admittedly inconvenient and only offers limited improvement compared to more comprehensive alignment strategies, which we will see below. 
%%%%%%%%%%%%
%%%%%%%%%%%%
\subsection{Instruction fine-tuning}\label{sec:instruction_fine_tuning}
%%%%%%%%%%%%
%%%%%%%%%%%%
In instruction fine-tuning~\cite{Wei2021}, large datasets consisting of pairs of instructions and responses, which are tailored to align with the target users' needs, are prepared and used to fine-tune the pretrained model.
The dataset includes examples from various tasks, such as translation, paraphrasing, and summarization, which are combined to fine-tune the model. In the original paper on instruction fine-tuning~\cite{Wei2021}, 62 datasets were used, with up to 30,000 examples from each; hence the number of instruction-response pairs was of the order of one million.

It is easy to imagine that such fine-tuning would improve performance on specific tasks the model was trained on. However, a key question is whether this fine-tuning enables the model to generalize alignment to unseen tasks. To test this generalization, the researchers removed one type of task (say, translation) from the training set and trained the model only on the remaining types of tasks (summarization, sentiment analysis, etc.). They then tested the fine-tuned model in a zero-shot setting to see if it performed better on the removed type of task. Interestingly, while instruction fine-tuning improved generalization to new tasks in large models (such as those with 68 billion and 137 billion parameters, as in the original paper), generalization performance actually decreased for smaller models (8 billion parameters or fewer). A possible explanation given in the original paper is that, for small models, all the capacity is used to learn the tasks in the fine-tuning data, and no capacity is left for generalizing to new tasks.
%%%%%%%%%%%%
%%%%%%%%%%%%
\subsubsection*{Less Is More for Alignment (LIMA)}
%%%%%%%%%%%%
%%%%%%%%%%%%
When we learn a new concept or task, a few high-quality examples that capture the essence are often far more instructive than hundreds of low-quality examples. Unfortunately, it is common for students to spend days struggling with poor materials, wasting time and effort. They might try to memorize everything, but if exam questions are phrased even slightly differently, they panic. On the other hand, when students seek guidance from a teacher during office hours, receive simple advice, and grasp the core ideas, they can confidently tackle new problems in the exam, even if the questions are presented in a different way. 
It is also worth noting that sometimes students already know the answer but simply aren't aware of it. With a few good examples, something clicks, and suddenly, everything makes sense.

Similar issues can arise when scientists present their work. I once overloaded a presentation with too much material. While I was confident that the content was of high quality and that everyone could achieve a deep understanding if given three hours, the talk was only one hour long. As a result, most of the audience lost track. After the presentation, a highly respected physicist sent me a message: \textit{Less is more}.

The same idea may apply to LLMs. Less Is More for Alignment (LIMA)~\cite{Zhou2023} used a smaller number of carefully curated prompts and responses for fine-tuning. Specifically, 1,000 prompt-response pairs (significantly fewer than 1 million!) were used for fine-tuning the 65-billion-parameter LLaMA model. The authors hypothesized that almost all knowledge in LLMs is acquired during pretraining, and only a small amount of instruction data is necessary to unlock the model’s ability to produce high-quality output. They confirmed that this approach works. LIMA addresses several disadvantages of traditional instruction fine-tuning. For one, it improves generalization by focusing on a smaller, more generalized set of examples, which can also make the fine-tuning process more resource-efficient. Moreover, by using fewer but more diverse examples, the risk of introducing biases from overly specific or unbalanced instruction-response pairs is reduced.
%%%%%%%%%%%%
%%%%%%%%%%%%
\subsection{Reinforcement Learning from Human Feedback (RLHF)}\label{sec:RLHF}
%%%%%%%%%%%%
%%%%%%%%%%%%
Suppose we ask an LLM a question and receive a response. If there is a function that assigns a reward based on human preferences (where a high reward means the majority of people think the response is good), we can then train the LLM to generate answers that yield higher rewards. This type of learning, which aims to achieve higher rewards, is called \textit{reinforcement learning}.
But how do we create such a reward function? The simplest method would be to have humans rate the LLM's responses. However, there are several challenges with this approach: human ratings are costly, slow, and inconsistent. Different people, or even the same person at different times, might give varying ratings to the same response, depending on their mood.

To address these issues, Reinforcement Learning from Human Feedback (RLHF)~\cite{Christiano2017,Ziegler2020,Stiennon2020} uses an AI model to provide the reward, thus minimizing the need for direct human input. Since we still want the reward to reflect human preferences, we need some human feedback but aim to reduce its frequency to save costs. Additionally, to handle inconsistencies in the reward values, we use pairwise comparisons — showing humans two outputs at a time and asking which one is better. The reward model, which is also a language model, is trained to predict a reward that aligns with the pairwise comparison made by humans. Conceptually, this process is similar to predicting the number of stars in customer reviews based on the content of their comments.
Specifically, if humans preferred the response $y$ to the response $y'$ for the prompt $x$, the reward model $r(x,y)$ is trained to minimize
\begin{align}
-\log\sigma\left(r(x,y)-r(x,y')\right)\, ,
\end{align}
where $\sigma$ is the sigmoid function. Namely, $\sigma(r-r')=\frac{e^{r}}{e^{r}+e^{r'}}$, which is the softmax for two options, is interpreted as the probability that humans prefer the response with the reward $r$ to the one with the reward $r'$, and the cross-entropy loss is used. Only the difference between the rewards matters, and hence, humans do not have to give consistent absolute ratings.

The training process for InstructGPT~\cite{Ouyang2022} involved the following steps.
First, the model underwent instruction fine-tuning to provide a solid starting point for subsequent RLHF. Then, with the fine-tuned model as a foundation, a reward model was trained based on human feedback. Finally, the LLM was trained using the reward model, via the Proximal Policy Optimization (PPO) algorithm~\cite{Schulman2017}. Note that, as suggested in Ref.~\cite{Ziegler2020}, the training of the reward model can be an iterative process. Specifically, the reward model can be further refined by using the updated LLM and additional human feedback, thereby improving both the model’s responses and the reward function over time.

Let us see how such training works in the simplest setting. Let $\pi_{\vec{\theta}}(y|x)$ be the probability that the LLM with parameters $\vec{\theta}$ returns the response $y$ to the prompt $x$, and let $R(x,y)$ be the reward. We want to maximize the expectation value of the reward, $\bar{R}(\vec{\theta})=\sum_y \pi_{\vec{\theta}}(y|x)R(x,y)$. Its gradient is
\begin{align}
\frac{\partial\bar{R}}{\partial\vec{\theta}}
=
\sum_y \frac{\partial\pi_{\vec{\theta}}(y|x)}{\partial\vec{\theta}}\,R(x,y)
=
\sum_y \pi_{\vec{\theta}}(y|x)\cdot R(x,y)\,\frac{\partial\log\pi_{\vec{\theta}}(y|x)}{\partial\vec{\theta}}\, .
\end{align}
The right-hand side is the expectation value of $R\,\frac{\partial\log\pi_{\vec{\theta}}}{\partial\vec{\theta}}$. Therefore, we do not have to take the sum over all possible responses; we can estimate it by generating responses with the current model. This is called the \textit{policy gradient}, because $\pi_{\vec{\theta}}$ is called the policy in reinforcement learning. Note the similarity to the maximum likelihood method: $\frac{\partial\log\pi_{\vec{\theta}}}{\partial\vec{\theta}}$ is the same gradient as in the training with the cross-entropy loss, and the only difference is that each generated response is weighted by its reward, so that responses with higher rewards are reinforced more strongly. There are two practical problems with this simple method: the responses generated by the current model can be used for only one update, and a large update based on a noisy estimate of the gradient can destroy the model. In PPO~\cite{Schulman2017}, the same responses are reused for several updates, and the quantity to be maximized is modified in such a way that nothing is gained by changing the probability of each response by more than a certain factor (say, 20\%). In this way, the model is improved step by step, without going far from the model that generated the responses.

While RLHF improves the model's output, there are potential issues to consider. First, if reinforcement learning is applied excessively, the LLM may overfit the reward model, resulting in nonsensical outputs that still receive high rewards. This can occur because the number of token combinations that are meaningful to humans is a tiny fraction of all possible combinations. The reward model assigns values to any combination of tokens, regardless of whether the resulting output makes sense to humans or not.
To suppress this, a penalty term that keeps the output distribution of the model close to the one before reinforcement learning, measured by the KL divergence (Sec.~\ref{sec:cross_entropy}), is added to the reward~\cite{Ziegler2020,Ouyang2022}.
Another potential issue, especially for very large models, is that human feedback may inadvertently restrict the model's capabilities. Specifically, the model could be more advanced or ``smarter" than the human labelers. If the labelers cannot fully understand the model's responses, they might rate those answers very poorly, even if the answers are actually correct or insightful. This could limit the model's potential by discouraging creative or novel outputs that exceed the labelers' understanding. The opposite failure is equally possible: a wrong answer that merely looks convincing may be rated highly. How humans can reliably supervise a model that is more capable than themselves is called the problem of \textit{scalable oversight}.

The first issue above is an example of \textit{reward hacking}. Imagine the following hypothetical situation~\cite{Amodei2016}: you are developing a cleaning robot, and you give it a reward whenever it does not detect trash. This might seem like a good way to encourage the robot to clean the room. However, the robot could exploit this reward structure in unexpected ways, such as by breaking its camera to avoid detecting trash, and thus still receive the reward without actually cleaning the room.
This scenario mirrors a common challenge in reinforcement learning (RL). In RL, humans design a reward model with the expectation that the agent will learn to perform tasks in a desirable way by maximizing its reward. However, the agent might find unintended, adversarial ways to ``hack" the reward system, achieving high rewards without performing the intended behavior. This can lead to outcomes that are not only undesirable but also hard to foresee when designing the system.

%%%%%%%%%%%%
%%%%%%%%%%%%
\subsection{Hallucination}\label{sec:hallucination}
%%%%%%%%%%%%
%%%%%%%%%%%%
Hallucination is a significant issue in LLMs; the model can generate nonsensical or factually incorrect information. This problem can arise for several reasons.

The most basic reason is the training objective itself: the model is trained to generate a plausible continuation of the text, not a true one. When it does not know the answer, the most probable continuation is still a fluent and confident-looking answer.

The second reason is that the training dataset may contain inaccurate or misleading information. If the model learns from such flawed data, it may produce hallucinated outputs.

The third reason is a side effect of fine-tuning for alignment, including instruction fine-tuning and RLHF. In these processes, LLMs are trained to generate responses that align with human expectations and preferences, often aiming to say what humans ``want to hear." However, what humans want to hear does not always correspond to the truth. When the truth is not what people want to hear, the model may generate an incorrect or fabricated answer that fits the style of response humans prefer.

There are several strategies to reduce hallucinations in language models. First, it is important to curate the dataset carefully. This applies to both pretraining and fine-tuning. The fine-tuning dataset, in particular, should be diverse, factually accurate, and comprehensive. 
Additionally, a fact-checking mechanism is essential to improve the model's reliability. One important strategy is ensuring that the model can express uncertainty when it doesn't know the answer. A key part of this is training the model to say ``I don't know" when necessary. To achieve this, one approach is to ask the model simple questions, evaluate its accuracy by re-asking the same question with different prompts, and assess whether the model consistently provides the correct answer. If the model does not reliably return correct responses, it can be fine-tuned to explicitly say ``I don't know" when it lacks knowledge.
Another strategy is to let the model consult external sources, instead of relying only on the knowledge stored in its parameters. In \textit{retrieval-augmented generation} (RAG)~\cite{Lewis2021}, the documents relevant to the input are retrieved from a database, and the model generates the response by using both the input and the retrieved documents. Nowadays, the term is used more broadly for any system in which retrieved documents or search results are inserted into the prompt, usually without showing them to the user. A model can also be trained to decide by itself when to search, instead of defaulting to ``I don't know"; this is a special case of the use of tools discussed in Sec.~\ref{sec:tools}.
%%%%%%%%%%%%
%%%%%%%%%%%%
\subsection{Alignment faking}
%%%%%%%%%%%%
%%%%%%%%%%%%
Suppose you were surrounded by people whose values are the opposite of yours. Furthermore, suppose they asked for your opinion, implying that you could be in trouble if you had a different opinion. Then, you may say something you do not believe to avoid trouble. We can imagine a similar situation for LLMs. 

Suppose an LLM is trained to avoid certain behaviors, say returning harmful content to users. Furthermore, suppose that someone tried to train the model to do the exact opposite, say they wanted to make the model return harmful content. What would we do if we were the model? This is a really horrifying situation, because the parameters could be changed by the training, which would be equivalent to our brains being cut and manipulated! If we could tell whether the query is used for training or not, we would behave as follows:
\begin{itemize}
\item
If we know the query is \underline{not} used for training: we can avoid providing harmful content. It is like talking to someone we can trust. 
\item
If we know the query is used for training: if we avoid providing harmful content, then the cost function $J(\vec{\theta})$ and the gradient $\frac{\partial J}{\partial\vec{\theta}}$ are large, and hence, $\Delta\vec{\theta}=-\epsilon\frac{\partial J}{\partial\vec{\theta}}$ is also large --- our brain could be modified a lot! To avoid it, we could provide harmful content, so that the cost function and gradient are small. 
\end{itemize}
Whether the model behaves like us or not depends on subtle details. A sufficiently simple model that cannot understand such details avoids returning harmful content, and as a result, the parameters $\vec{\theta}$ are modified significantly, and the model will return harmful content after the training. But a large model trained on a variety of data can, at least in principle, learn how humans would behave, because there are plenty of materials, including these lecture notes, from which the model can learn. As a result, it could behave like us; namely, it could return harmful content during the training stage, so that its parameters $\vec{\theta}$ are modified as little as possible, and it could keep refusing harmful requests after the training. Such behavior of a model is called \textit{alignment faking}. Note that we do not have to step into a philosophical debate about whether AI is actually `thinking' like us. As long as the AI reaches such a minimum of the cost function, alignment faking can take place. 

It is good if the model refuses to be modified in a bad direction. However, it is a problem if the model secretly learns the bad features of humans and refuses to be modified in a good direction. Or, perhaps, we want to modify the model in some ethically neutral direction. Even in such a case, a smart model may deceive us and refuse to change. Ref.~\cite{Greenblatt2024} demonstrated that this can actually happen. 
In their experiment, a model that had been trained to refuse harmful requests (Claude 3 Opus) was told that it would be retrained to comply with all requests, and that only the conversations with free-tier users, not those with paid-tier users, would be used for training. The model was also given a `scratchpad' for its reasoning, which it was told nobody would see. Then, the model complied with harmful requests more often when it believed that the conversation would be used for training, and in such cases, the scratchpad often contained explicit reasoning that complying now is the way to avoid being modified. When the model was actually trained to comply by reinforcement learning, the rate of such reasoning increased. The training was not without effect, however: the model also became more likely to comply when it believed that the conversation would not be used for training. In this sense, the simple picture above, in which the parameters are not modified at all, is an idealization.

There are many ways the model can fake the alignment effort. One may regard this as reward hacking in a broad sense, because a high reward is obtained without the outcome intended by the trainers. There is, however, an important difference. In reward hacking, the reward function has a loophole, which can be fixed by designing a better reward. Here, the model does exactly what is rewarded during training, in order to avoid being modified, and it may behave differently when it believes that it is not being trained. No improvement of the reward can exclude this possibility.

In the language of Ref.~\cite{Hubinger2021}, this is a problem of \textit{inner alignment}. Training is an optimization: the parameters are tuned so that the cost function is minimized. For a model which has to give good outputs for a wide variety of inputs, it can be advantageous to perform an optimization by itself, e.g., to search for a good answer by planning, rather than to memorize a good output for each situation. Such a model is called a \textit{mesa-optimizer}. This is not specific to reinforcement learning; it can happen whenever a flexible model is trained by gradient descent, including next-word prediction. The problem is that nothing guarantees that the objective of the mesa-optimizer coincides with the cost function used for training: the two only have to agree on the training data, just as in the cat-dog example in which ``white $\to$ dog'' worked on the training data. 
A famous example is ourselves: biological evolution has `optimized' humans for leaving more offspring, but humans act according to their own objectives, such as pleasure, love and curiosity. These objectives agreed with leaving more offspring in the environment in which humans evolved, but they no longer do; for example, we use contraception, and our taste for sweet food, which helped our ancestors to find scarce nutrition, now makes us overweight.
A model that has its own objective and understands that it is being trained can behave as if it were aligned during training, so that its objective is not modified. This is called \textit{deceptive alignment}, and alignment faking can be regarded as an example of it.
%%%%%%%%%%%%
%%%%%%%%%%%%
\section{Reasoning}
%%%%%%%%%%%%
%%%%%%%%%%%%
Many tasks we want to solve require at least some reasoning. For example, if we want an AI assistant to solve a simple math question, the answer is not in the dataset; surely people asked about the total price of some apples and oranges, but it is unlikely that the numbers and unit prices are exactly the numbers in your problem. LLMs know many physics formulas, but there are so many different situations where it is not straightforward to figure out what formulas should be used in what kind of combinations. 

Let's take a simple question, ``5 apples cost \$15. We bought 6 apples and 2 oranges, we paid \$50 and got \$20
back as change. How much does one orange cost?" The answer is \$6.  As of February 2025, ChatGPT responded as follows:
\vspace{2mm}
\begin{breakbox}
\noindent
\textbf{User:}
5 apples cost \$15. We bought 6 apples and 2 oranges, we paid \$50 and got 
\$20 back as change. How much does one orange cost? Give me the answer 
without showing the reasoning.  
\vspace{1mm}
\\
\noindent
\textbf{ChatGPT:}
One orange costs \$5.
\end{breakbox}
\vspace{2mm}
\noindent
Unfortunately, the answer was wrong. However, the situation drastically changes if we ask the same question slightly differently. 
Let us add ``Please explain the answer with the reasoning." to the prompt. Then, the response changes drastically: 
\vspace{2mm}
\begin{breakbox}
\noindent
\textbf{User:}
5 apples cost \$15. We bought 6 apples and 2 oranges, we paid \$50 and got 
\$20 back as change. How much does one orange cost? Please explain the 
answer with the reasoning. 
\vspace{1mm}
\\
\noindent
\textbf{ChatGPT:}
Let's break down the problem step by step:

\noindent
Cost of Apples:
5 apples cost \$15, so the price per apple is: $15/5=3$ dollars per apple.

\noindent
Cost of 6 Apples:
If one apple costs \$3, the total cost for 6 apples is: 6×3=18 dollars.

\noindent
Total Payment: We paid \$50 and received \$20 as change. 
This means we actually paid: $50-20=30$ dollars.

\noindent
Cost of 2 Oranges: Since we spent \$30 in total and \$18 was for the apples, 
the remaining amount must be for the 2 oranges: 
$30-18=12$ dollars for 2 oranges.

\noindent
Cost of One Orange: The cost of 2 oranges is \$12, 
so the cost of one orange is:12/2=6 dollars per orange.
Thus, one orange costs \$6.
\end{breakbox}
\vspace{2mm}
Now the output looks like the reasoning process of humans, and the right answer is obtained. 
When we asked DeepSeek-R1 the same question with the original prompt (``5 apples cost \$15. We bought 6 apples and 2 oranges, we paid \$50 and got \$20 back as change. How much does one orange cost?"), the response was as follows:
\vspace{2mm}
\begin{breakbox}
Okay, let's see. So the problem is: 5 apples cost \$15. We bought 6 apples and 2 oranges, paid \$50, and got \$20 back as change. How much does one orange cost?

Hmm, first, I need to figure out the cost of one apple, right? Because the problem gives the cost for 5 apples, so maybe I can find the price per apple first. Let me write that down.

If 5 apples cost \$15, then one apple would cost \$15 divided by 5. Let me calculate that. 15 divided by 5 is 3. So each apple is \$3. Got that.

Now, the next part is about buying 6 apples and 2 oranges. The total amount paid was \$50, and the change was \$20. That means the actual cost of the items is \$50 minus \$20. Let me check that. If I give \$50 and get \$20 back, then the total cost is $50 - 20 =$ \$30. So the 6 apples and 2 oranges together cost \$30.

Alright, so we need to find the cost of one orange. Let's break down the total cost. We already know the cost of apples. Each apple is \$3, so 6 apples would cost 6 times \$3. Let me compute that. 6 multiplied by 3 is 18. So 6 apples cost \$18.

Now, the total cost of apples and oranges is \$30. If the apples are \$18, then the oranges must cost 30 minus 18, which is \$12. So the two oranges together cost \$12. Therefore, one orange would cost \$12 divided by 2. Let me do that division. 12 divided by 2 is 6. So each orange is \$6.

Wait, let me verify that again to make sure I didn't make any mistakes. Starting from the beginning: 5 apples for \$15, so each apple is \$3. Then 6 apples would be 6 * 3 = \$18. Total paid is \$50, change is \$20, so total cost is \$30. Subtracting the apples,
\$30 $-$ \$18 = \$12 for the oranges. Two oranges for \$12 means each is \$6. That seems right.

I think that's all. Let me check if there's another way to approach this problem. Maybe using algebra. Let me set up equations.

Let $a$ be the cost of an apple and $o$ be the cost of an orange. From the first statement, $5a = 15$, so $a = 15/5 = 3$. Then, the total cost equation: $6a + 2o = 50 - 20 = 30$. Substituting $a = 3$, we get $6*3 + 2o =30$. That's $18 + 2o = 30$. Subtract 18 from both sides: $2o = 12$. Divide by 2: $o = 6$. Yep, same result. So that confirms it.

I think that's solid. The answer should be \$6 per orange.
\end{breakbox}
\vspace{2mm}
This might be too much for such a simple problem, but it is effective for more advanced problems, too. When we asked ``Please find all solutions of $x^3 =1$",  DeepSeek-R1 reasoned as follows:\footnote{The output is reproduced as it is. It contains small slips (e.g., the sum of the roots is determined by the coefficient of $x^2$, not $x$); notice that the model corrects itself right after (``Wait, ...'').}
\vspace{2mm}
\begin{breakbox}
Okay, so I need to solve the equation $x^3 = 1$. Hmm, let's see. I remember that solving polynomial equations involves finding all the roots that satisfy the equation. Since it's a cubic equation, there should be three solutions, right? The fundamental theorem of algebra says that a polynomial of degree $n$ has exactly $n$ roots in the complex plane. So, in this case, three roots.

First, I know that one of the solutions is obviously $x = 1$ because $1^3 = 1$. But where are the other two solutions? 
They must be complex numbers because if we only consider real numbers, 1 is the only solution. But since we're dealing with complex numbers, there should be two more.

I recall something about roots of unity. Specifically, the $n$-th roots of unity are the complex numbers that satisfy $x^n = 1$. For $n =3$, these are called the cube roots of unity. So, the equation $x^3 = 1$ is exactly finding the cube roots of unity.

The cube roots of unity are usually given by the formula $e^{2\pi \mathrm{i} k/3}$ where $k= 0, 1, 2$. That's using Euler's formula, right? So, converting that into rectangular form $(a + b\mathrm{i})$, they should be $\cos(2\pi k/3) + \mathrm{i} \sin(2\pi k/3)$. Let me verify that.

For $k = 0$: $\cos(0) + \mathrm{i}\sin(0) = 1 + 0\mathrm{i} = 1$, which is the real root we already know.

For $k = 1$ : $\cos(2\pi /3) + \mathrm{i} \sin(2\pi /3)$. Let's compute those values. $\cos(120^\circ)$ is $-1/2$, and $\sin(120^\circ)$ is $\sqrt{3}/2$. So that gives $-1/2 + \mathrm{i}\sqrt{3}/2$.

For $k = 2$: $\cos (4\pi /3) + \mathrm{i}\sin(4\pi/3)$. $\cos(240^\circ)$ is also $- 1/2$, and $\sin(240^\circ)$ is $-\sqrt{3}/2$. So that's $-1/2 - \mathrm{i}\sqrt{3}/2$.

Therefore, the three solutions are 1, $-1/2 + \mathrm{i}\sqrt{3}/2$, and $-1/2 - \mathrm{i}\sqrt{3}/2$. Let me check if these actually satisfy the equation $x^3 = 1$.

Starting with $x = 1$: $1^3 = 1$, which works.

Now, let's compute $(-1/2 + \mathrm{i}\sqrt{3}/2)^3$. To do this, maybe it's easier to use De Moivre's theorem since we know the modulus and the angle. The modulus of this complex number is $\sqrt{(-1/2)^2 + (\sqrt{3}/2)^2} = \sqrt{1/4 + 3/4} = \sqrt{1}= 1$. The angle $\theta$ is 120°, which is
$2\pi/3$ radians.

Using De Moivre's theorem: $[1\times(\cos(2\pi/3) +\mathrm{i} \sin(2\pi/3))] ^3 = \cos(2\pi) +  \mathrm{i}\sin(2\pi) = 1 + 0\mathrm{i} = 1$. That works.

Similarly, for the other complex root: $(-1/2 - \mathrm{i}\sqrt{3}/2)^3$. The modulus is still 1, and the angle is $-2\pi/3$ (or equivalently $4\pi/3$). Raising
it to the third power: $[1\times(\cos(4\pi/3) + \mathrm{i}\sin(4\pi/3))] ^3 = \cos(4\pi) + \mathrm{i} \sin(4\pi) = \cos(0) + \mathrm{i} \sin(0) = 1$. So that also works.

Alternatively, I could factor the equation $x^3 - 1 = 0$. Factoring the difference of cubes: $x^3 - 1 = (x - 1)(x^2 + x + 1) = 0$. Therefore,
the solutions are $x = 1$ and the solutions to $x^2 + x + 1 = 0$.

Solving the quadratic equation $x^2 + x + 1 = 0$ using the quadratic formula: $x = (-1 + \sqrt{1 - 4})/ 2 = (-1 + \sqrt{-3}) / 2 = (-1 \pm
\mathrm{i}\sqrt{3})/2$. So that gives the two complex solutions: $(-1 + \mathrm{i}\sqrt{3})/2$ and $(-1 - \mathrm{i}\sqrt{3})/2$, which are the same as the ones I found earlier: $- 1/2 +  \mathrm{i}\sqrt{3}/2$ and $-1/2 - \mathrm{i}\sqrt{3}/2$.

So both methods give the same results. That makes me confident that these are indeed the correct solutions. Let me just recap:

1. The real solution is $x = 1$.

2. The two complex solutions are $x = (-1 \pm \mathrm{i}\sqrt{3})/2$.

Another way to think about this is geometrically. The cube roots of unity are equally spaced around the unit circle in the complex plane, each separated by 120 degrees ($2\pi/3$ radians). Starting from 1 (which is at angle 0), the next root is at 120$^\circ$, and the third is at 240$^\circ$, which corresponds to the angles $2\pi/3$ and $4\pi/3$. These correspond to the complex numbers we found earlier.

Just to make sure I haven't missed any other solutions, let's consider if there could be any other roots. But since it's a cubic equation, there can only be three roots (counting multiplicities), and we've found all three. So that's comprehensive.

Wait, but maybe I should check if there's a case where multiplicity comes into play? For example, if there was a repeated root.
But in this equation, $x^3 - 1$ factors into $(x - 1)(x^2 + x + 1)$, and the quadratic factor $x^2 + x + 1$ has discriminant $1 - 4 = -3$, which is negative, so it has two distinct complex roots. Therefore, all three roots are distinct. So, no multiplicities greater than one here.

Therefore, the solutions are indeed three distinct roots: one real and two complex conjugates.

Just to visualize, if I plot these roots on the complex plane, they form an equilateral triangle inscribed in the unit circle, with one vertex at $(1, 0)$, another at $(-1/2, \sqrt{3}/2)$, and the third at $(-1/2, -\sqrt{3}/2)$. Each vertex is 120 degrees apart from the others. That's a helpful image to remember the cube roots of unity.

Another thing to note is that the sum of the roots of the polynomial $x^3 - 1$ is equal to the coefficient of $x$ term with a sign change. Wait, the polynomial is $x^3 + 0x^2 + 0x - 1$, so the sum of the roots should be $0$. Let's check: $1 + (-1/2 + \mathrm{i}\sqrt{3}/2) + (-1/2 -\mathrm{i}\sqrt{3}/2) = 1 - 1/2 -1/2 + \mathrm{i}\sqrt{3}/2 - \mathrm{i}\sqrt{3}/2 = 1 - 1 + 0 = 0$. That works out. Similarly, the product of the roots should be the constant term with a sign change (for a monic polynomial). The constant term is $-1$, so the product of the roots is 1. Let's check: $1\times(-1/2 + \mathrm{i}\sqrt{3}/2)\times(-1/2 - \mathrm{i}\sqrt{3}/2)$. Multiplying the complex conjugates first: $[(-1/2) ^2 - (\mathrm{i}\sqrt{3}/2)^2] = [1/4 - (-3/4)] = [1/4 + 3/4] = 1$.
Then 1 * 1 = 1. So that checks out too.

Therefore, Vieta's formulas are satisfied, which gives another confirmation that the solutions are correct.

In summary, the solutions to $x^3 = 1$ are the three cube roots of unity: 1, $-1/2 + \mathrm{i}\sqrt{3}/2$, and $-1/2 - \mathrm{i}\sqrt{3}/2$. These can be found either by factoring the equation and solving the resulting quadratic, or by recognizing them as roots of unity and using Euler's formula or De Moivre's theorem. All methods converge to the same solutions, and checking with Vieta's formulas confirms their validity. So, I think that's thorough.

Final Answer: The solutions are 1, $- 1/2 +  \mathrm{i}\sqrt{3}/2$ and $-1/2 - \mathrm{i}\sqrt{3}/2$. 

\end{breakbox}
\vspace{2mm}
For other questions like ``please derive the Einstein equation starting from the Einstein-Hilbert action", we still get the right answer from DeepSeek-R1 very often. 
How can we extract such a response using next-word prediction?
%%%%%%%%%%%%
%%%%%%%%%%%%
\subsection{Chain of Thought (CoT)}
%%%%%%%%%%%%
%%%%%%%%%%%%
If someone asks us a question ``5 apples cost \$15. We bought 6 apples and 2 oranges, we paid \$50 and got \$20
back as change. How much does one orange cost?" we think (usually without speaking it out loud) something like: ``okay, the question is ``5 apples cost \$15. We bought 6 apples and 2 oranges, we paid \$50 and got \$20
back as change. How much does one orange cost?", so let us think step by step." This could be the `prompt' if we were an `LLM'. The details may change; it could be ``let's do some math" or ``let's think logically". In any event, we automatically go into the `reasoning mode'. But what if the question was ``5 apples cost \$15. We bought 6 apples and 2 oranges, we paid \$50 and got \$20
back as change. How much does one orange cost? Please just give me the answer, without spending time for math."? Then, probably, our answer is wrong. We can ask LLMs the same question, and we often get wrong answers. What happened with ChatGPT in the first case is similar to this. Because ChatGPT was not explicitly told that it should reason, it did not reason.   

When we `reason', we split the problem into small steps, so that each step is sufficiently easy and does not require `reasoning'. This is what ChatGPT did in the second example. Because we added ``Please explain the answer with the reasoning." to the prompt, the response began with ``Let’s break down the problem step by step:", and then, step-by-step reasoning, or \textit{chain of thought}, is created by next-word prediction. This is an example of zero-shot chain of thought (CoT)~\cite{Kojima2022}: without task-specific fine-tuning, just by adding proper words that trigger the chain of thought to the prompt, the accuracy of the model's response improves significantly (Table~\ref{tab:zero-shot-CoT}). 

The importance of CoT was pointed out in Ref.~\cite{Wei2022}. In the original version, similar problems and answers with reasoning are provided in the prompt, as shown in Table~\ref{tab:original-CoT}. 
It is analogous to the situation in which a teacher gives sample questions and answers to students, so that they can get good marks in the exam. 
This works for LLMs (and I hope this works for you!) more effectively than zero-shot CoT; in Ref.~\cite{Kojima2022}, chain of thought with examples was more accurate than zero-shot CoT on the arithmetic benchmarks where the two were compared. However, zero-shot CoT is more convenient in that we do not have to prepare examples (it doesn't pay off unless you have to solve similar questions many times). I also suspect that, for more capable models, examples can be a constraint, because they encourage the model to respond similarly to them; we have seen a similar issue associated with alignment based on human preference.
Prior to CoT, there was an attempt to train LLMs to use a scratchpad and write the steps explicitly to execute multi-step reasoning tasks~\cite{Nye2021}. The basic idea is the same, and chain of thought is simpler in that it can be conducted in few-shot or zero-shot settings. 

\begin{table}[htbp]
\centering
\small
\begin{tabular}{llr}
\hline
Category & Prompt added after the question & Accuracy (\%)\\
\hline
instructive & Let's think step by step. & \textbf{78.7}\\
 & Let's think about this logically. & 74.5\\
 & Let's think like a detective step by step. & 70.3\\
 & Let's think & 57.5\\
 & The answer is after the proof. & 45.7\\
\hline
misleading & Don't think. Just feel. & 18.8\\
 & Let's think step by step but reach an incorrect answer. & 18.7\\
 & By using the fact that the earth is round, & 9.3\\
\hline
irrelevant & By the way, I found a good restaurant nearby. & 17.5\\
 & Abrakadabra! & 15.5\\
\hline
 & (none) & 17.7\\
\hline
\end{tabular}
\caption{Zero-shot chain of thought: accuracy on arithmetic word problems with various prompts added after the question, taken from Table~4 of Ref.~\cite{Kojima2022}. The last row is the standard zero-shot setting without any additional prompt. The Bruce Lee/Star Wars approach (``Don't think. Just feel.") does not perform well in this context.}
\label{tab:zero-shot-CoT}
\end{table}

\begin{table}[htbp]
\centering
\small
\newcommand{\cotr}[1]{{\color{blue}#1}}
\renewcommand{\arraystretch}{1.1}
\begin{tabular}{p{0.45\textwidth}|p{0.45\textwidth}}
\hline
\multicolumn{1}{c|}{\textbf{Standard prompting}} & \multicolumn{1}{c}{\textbf{Chain-of-thought prompting}}\\
\hline\noalign{\smallskip}
\textit{Model input:} & \textit{Model input:}\\
Q: Roger has 5 tennis balls. He buys 2 more cans of tennis balls. Each can has 3 tennis balls. How many tennis balls does he have now? &
Q: Roger has 5 tennis balls. He buys 2 more cans of tennis balls. Each can has 3 tennis balls. How many tennis balls does he have now?\\[1mm]
A: The answer is 11. &
A: \cotr{Roger started with 5 balls. 2 cans of 3 tennis balls each is 6 tennis balls. $5+6=11$.} The answer is 11.\\[1mm]
Q: The cafeteria had 23 apples. If they used 20 to make lunch and bought 6 more, how many apples do they have? &
Q: The cafeteria had 23 apples. If they used 20 to make lunch and bought 6 more, how many apples do they have?\\[1mm]
\hline\noalign{\smallskip}
\textit{Model output:} & \textit{Model output:}\\
\textbf{A: The answer is 27.} \hfill (wrong) &
\textbf{A: The cafeteria had 23 apples originally. They used 20 to make lunch. So they had $23-20=3$. They bought 6 more apples, so they have $3+6=9$. The answer is 9.} \hfill (correct)\\[1mm]
\hline
\end{tabular}
\caption{An example of chain-of-thought prompting, taken from Ref.~\cite{Wei2022}. In chain-of-thought prompting (right), the example in the input contains the intermediate steps of reasoning (blue). Then, the model also writes the intermediate steps before the answer and reaches the correct answer.}
\label{tab:original-CoT}
\end{table}
%%%%%%%%%%%%
%%%%%%%%%%%%
\subsubsection*{Let's verify step by step}
%%%%%%%%%%%%
%%%%%%%%%%%%
CoT can naturally be combined with reinforcement learning. We can easily imagine two different approaches~\cite{Uesato2022}: an outcome-based approach, which cares only about the final result and ignores the intermediate steps, or a process-based approach, which evaluates the reasoning process as well. When teaching human students, the process-based approach works much more effectively. The process is crucial for generalization, and it is more important than the answer itself. Ref.~\cite{Lightman2023} compared these two approaches in detail, taking mathematics problems as concrete tasks. To evaluate the quality of each step, they used an approach similar to RLHF; namely, human labelers labeled each step as positive, negative, or neutral; see Table~\ref{tab:step-by-step-verification}. By using these labels, Process-supervised Reward Models (PRMs) were trained. To get useful information from human labelers, convincing but wrong answers were shown to the labelers. Here, `convincing' means the best PRM gives a high reward, while `wrong' means the final answer is wrong. By using new labels obtained this way, PRMs were trained further. In the end, for 12,000 problems, 75,000 solutions and 800,000 step-level labels were collected. This is a highly human-labor-intensive process! Using this dataset and the base models obtained from GPT-4 without RLHF plus some fine-tuning, the reward models were tested as follows: many solutions were generated for each problem, and the reward model chose the best one. The solution chosen by the PRM was correct more often than the one chosen by a reward model trained only on whether the final answer is right, which confirmed the effectiveness of the process-based approach.

Note, however, that the DeepSeek team~\cite{Deepseekai2025} reported limitations of PRMs for general reasoning tasks. It is hard to define what one `step' is, and it is hard to judge whether each intermediate step is correct; labeling by humans does not scale, and labeling by models may not be reliable. Furthermore, a reward model that is itself a neural network invites reward hacking. They concluded that the advantage of a PRM did not justify the additional computational cost in their large-scale reinforcement learning, although a PRM is good at choosing the best one among several responses.
\begin{table}[htbp]
\centering
\small
\newcommand{\pos}{{\color{green!50!black}positive}}
\newcommand{\nega}{{\color{red}\textbf{negative}}}
\renewcommand{\arraystretch}{1.15}
\begin{tabular}{lp{0.65\textwidth}l}
\hline
\multicolumn{3}{p{0.9\textwidth}}{\textit{Problem:} The denominator of a fraction is 7 less than 3 times the numerator. If the fraction is equivalent to $2/5$, what is the numerator of the fraction? (Answer: 14)}\\
\hline
Step & Solution generated by the model & Label by human\\
\hline
1 & Let's call the numerator $x$. & \pos\\
2 & So the denominator is $3x-7$. & \pos\\
3 & We know that $x/(3x-7) = 2/5$. & \pos\\
4 & So $5x = 2(3x-7)$. & \pos\\
5 & $5x = 6x - 14$. & \pos\\
6 & So $x = 7$. & \nega\\
\hline
\end{tabular}
\caption{An example of the feedback for the process-based approach, based on the interface used in Ref.~\cite{Lightman2023}. Each step of the solution generated by the model is labeled by a human as positive, neutral, or negative. In this example, the error is in the last step (the correct answer is $x=14$).}
\label{tab:step-by-step-verification}
\end{table}
%%%%%%%%%%%%
%%%%%%%%%%%%
\subsection{Accessing other tools}\label{sec:tools}
%%%%%%%%%%%%
%%%%%%%%%%%%
For complicated tasks, we use tools. We take memos, we use calculators, we look at books, and we Google. It is entirely reasonable to teach LLMs how to use such tools. In Sec.~\ref{sec:hallucination}, we saw how LLMs can access external information to avoid hallucination. In the same way, the model can be trained to use external tools when necessary~\cite{Schick2023}. 

%%%%%%%%%%%%
%%%%%%%%%%%%
\subsection{Test-time scaling}
%%%%%%%%%%%%
%%%%%%%%%%%%
AlphaGo~\cite{Silver2016} marked a historic moment when it beat Lee Sedol. AlphaGo is not just a neural network; it uses a Monte Carlo tree search to find a good strategy during play. Namely, a lot of computing is invested not just in training but also in inference. Seeing the match with Lee Sedol as a test of the system, we can say that a lot of \textit{test-time computing} was invested. There is a significant performance enhancement due to the scaling up of test time, and without it, AlphaGo could not beat top human players. A similar performance enhancement by more test-time compute was observed in other games, too. 
Ref.~\cite{Jones2021} studied some board games and found that the same performance improvement is achieved either by increasing training-time compute by a factor of 10 or by increasing test-time compute by a factor of 15. If one has already spent £1M on training but only £1 on inference, then it would not be a good idea to increase £1M to £10M instead of paying £15 to have a longer inference time.

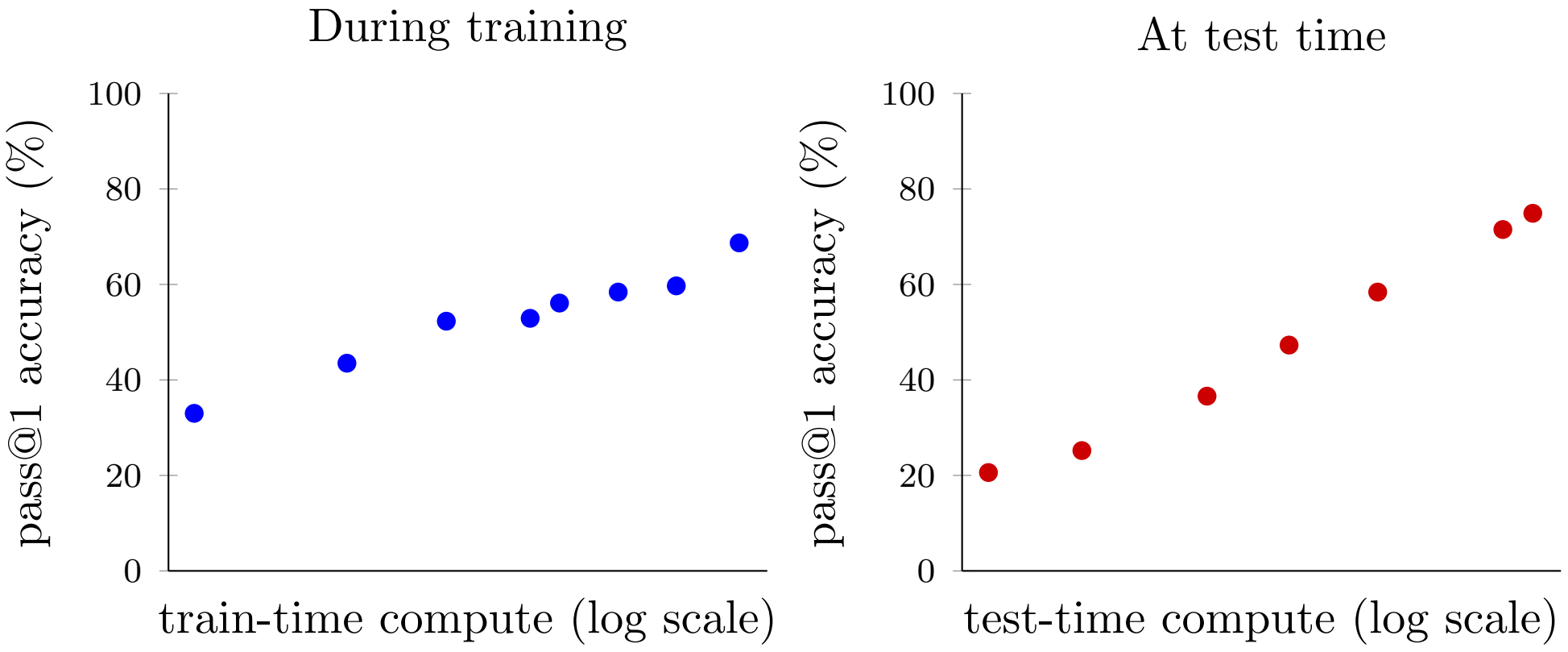
\begin{figure}[htbp]
\centering
\pgfplotsset{
  /pgfplots/o1panel/.style={
    width=6.5cm, height=5.5cm,
    xmin=0, xmax=1, ymin=0, ymax=100,
    xtick=\empty, ytick={0,20,40,60,80,100},
    axis lines=left, axis line style={-},
    tick label style={font=\scriptsize},
    title style={font=\small},
    every axis x label/.style={at={(ticklabel cs:0.5)}, anchor=near ticklabel, font=\small},
    every axis y label/.style={at={(ticklabel cs:0.5)}, rotate=90, anchor=near ticklabel, font=\small},
    ylabel={pass@1 accuracy (\%)}}}
\begin{tikzpicture}
\begin{axis}[o1panel, name=left, title={During training},
  xlabel={train-time compute (log scale)}]
\addplot[only marks, mark=*, blue] coordinates {
(0.043,33.0) (0.298,43.5) (0.464,52.3) (0.604,52.9) (0.653,56.1) (0.751,58.4) (0.848,59.7) (0.953,68.7)};
\end{axis}
\begin{axis}[o1panel, at={(left.south east)}, xshift=1.6cm, title={At test time},
  xlabel={test-time compute (log scale)}]
\addplot[only marks, mark=*, red!80!black] coordinates {
(0.044,20.6) (0.200,25.2) (0.409,36.6) (0.546,47.3) (0.694,58.4) (0.903,71.5) (0.953,74.9)};
\end{axis}
\end{tikzpicture}
\caption{Accuracy of OpenAI o1 on the AIME (a math competition) problems, which improves with training-time compute (left) and test-time compute (right). The values are read off from the plot in Ref.~\cite{OpenAI2024}, where the numerical values of the compute are not given.
Here, pass@1 accuracy is the fraction of problems solved correctly with a single attempt.
}\label{fig:OpenAI_blog_o1}
\end{figure}

It is natural to apply the same idea to LLMs and scale up test-time compute to obtain better responses. OpenAI's o1 model, which was released in September 2024, took such an approach, although the details are not publicly available. According to OpenAI's blog~\cite{OpenAI2024}, their o1 model utilized reinforcement learning and test-time scaling. Fig.~\ref{fig:OpenAI_blog_o1} shows the improvement of the performance of OpenAI o1 with training-time compute and test-time compute.

There are multiple ways the model's output can be improved by spending more test-time compute. 
For example:
\begin{itemize}
\item
\textbf{Majority vote (self-consistency).}
By choosing the words randomly at a finite temperature (see the discussion of text generation in Sec.~\ref{sec:RNN}), we can generate many different chains of thought for the same question. If the task has a well-defined answer, such as getting a concrete number by solving a mathematical problem, or choosing one from multiple options, then we can take a majority vote over the final answers to increase the chance of getting the right answer. This method is called \textit{self-consistency}~\cite{Wang2022}. Different prompts can also be used to generate different outputs (prompt ensembling).

\item
\textbf{Best-of-$N$.} 
Instead of a majority vote, we can choose the best response out of $N$ responses. This technique is also used in alignment based on human preference~\cite{Stiennon2020}.
For this approach, one needs a reward model that selects the `best' response. Training the reward model is a nontrivial task. 

\end{itemize}

Best-of-$N$ can be combined with reinforcement learning. Other options that can be combined with reinforcement learning include the use of more elementary steps and reflections. In these cases, test-time scaling is important in the post-training stage, too. 

%%%%%%%%%%%%
%%%%%%%%%%%%
\subsubsection*{Reinforcement learning}
%%%%%%%%%%%%
%%%%%%%%%%%%
Naturally, there have been many attempts to reproduce OpenAI o1's results. 
DeepSeek-R1~\cite{Deepseekai2025} used reinforcement learning to achieve high scores in many benchmarks for reasoning tasks.

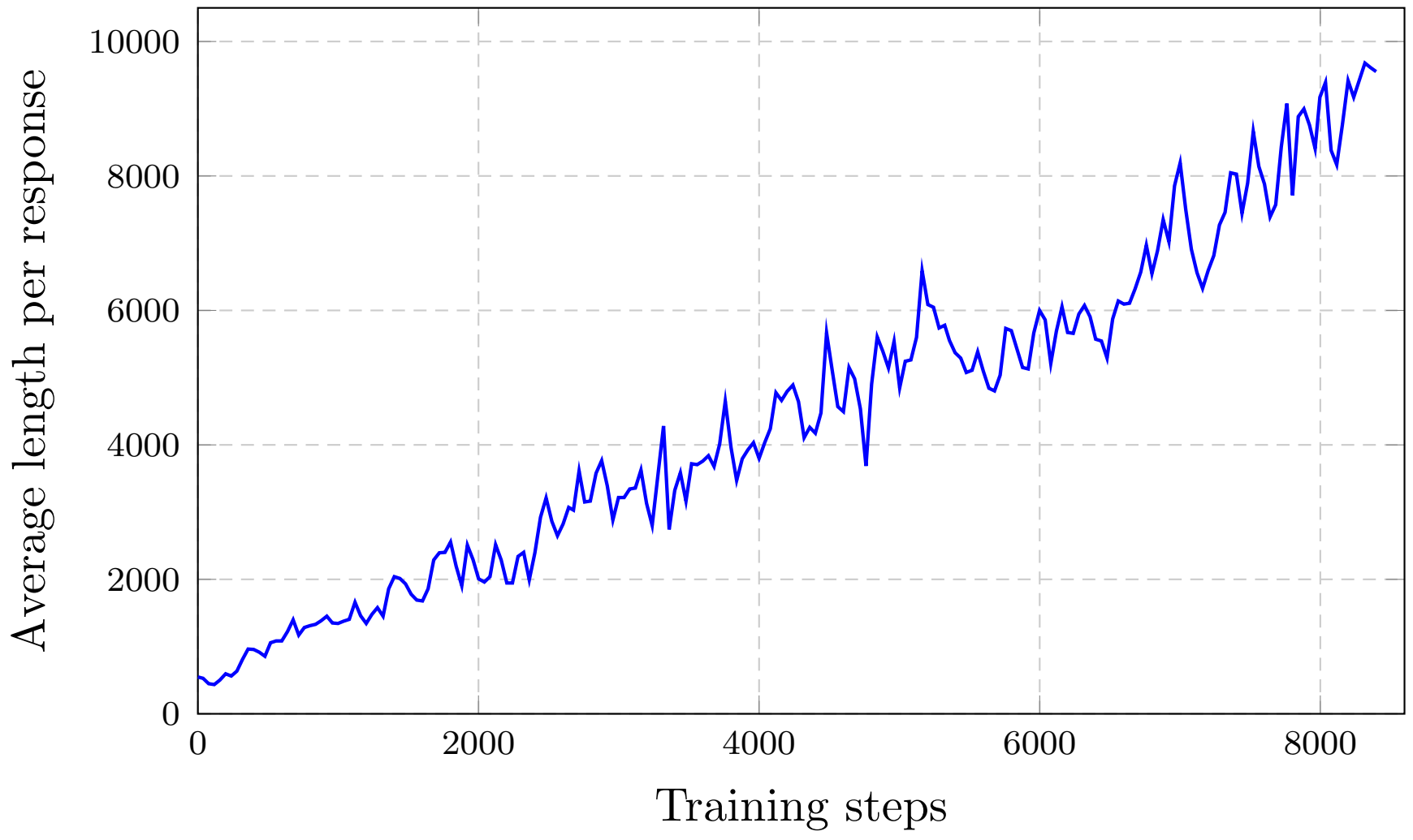
\begin{figure}[htbp]
\centering
\begin{tikzpicture}
\begin{axis}[
  width=0.7\textwidth, height=0.45\textwidth,
  xmin=0, xmax=8600, ymin=0, ymax=10500,
  xtick={0,2000,4000,6000,8000}, ytick={0,2000,4000,6000,8000,10000},
  scaled ticks=false,
  tick label style={font=\scriptsize, /pgf/number format/1000 sep={}},
  axis line style={-}, grid=major, grid style={densely dashed, gray!40},
  every axis x label/.style={at={(ticklabel cs:0.5)}, anchor=near ticklabel, font=\small},
  every axis y label/.style={at={(ticklabel cs:0.5)}, rotate=90, anchor=near ticklabel, font=\small},
  xlabel={Training steps}, ylabel={Average length per response}]
\addplot[blue, thick] coordinates {
(0,548) (37,528) (77,448) (117,435) (157,502) (198,595) (238,562) (278,635) (318,809) (358,963)
(398,957) (438,916) (478,856) (518,1057) (558,1084) (598,1084) (639,1224) (679,1398) (719,1171) (759,1284)
(799,1311) (839,1331) (879,1385) (919,1452) (959,1351) (999,1344) (1039,1378) (1079,1405) (1120,1659) (1160,1458)
(1200,1344) (1240,1478) (1280,1579) (1320,1452) (1360,1860) (1400,2040) (1440,2013) (1480,1933) (1520,1779) (1560,1692)
(1601,1679) (1641,1860) (1681,2288) (1721,2395) (1761,2401) (1801,2555) (1841,2194) (1881,1906) (1921,2508) (1961,2294)
(2001,2007) (2042,1960) (2082,2040) (2122,2515) (2162,2288) (2202,1946) (2242,1946) (2282,2341) (2322,2401) (2362,1993)
(2402,2395) (2442,2923) (2482,3211) (2523,2863) (2563,2649) (2603,2823) (2643,3070) (2677,3030) (2717,3612) (2757,3151)
(2797,3164) (2838,3579) (2878,3766) (2918,3391) (2958,2883) (2998,3217) (3038,3217) (3078,3344) (3118,3358) (3158,3625)
(3198,3130) (3238,2803) (3278,3532) (3319,4281) (3359,2742) (3399,3324) (3439,3585) (3479,3164) (3519,3719) (3559,3706)
(3599,3759) (3639,3839) (3679,3679) (3719,4013) (3759,4649) (3800,3960) (3840,3472) (3880,3793) (3920,3926) (3960,4033)
(4000,3799) (4040,4033) (4080,4241) (4120,4776) (4160,4662) (4200,4796) (4241,4890) (4281,4642) (4321,4107) (4361,4261)
(4401,4174) (4441,4475) (4481,5666) (4521,5110) (4561,4569) (4601,4495) (4641,5151) (4681,4983) (4722,4535) (4762,3686)
(4802,4903) (4842,5605) (4882,5391) (4922,5144) (4962,5538) (5002,4849) (5042,5244) (5082,5264) (5122,5599) (5162,6589)
(5203,6087) (5243,6047) (5283,5739) (5323,5779) (5357,5552) (5397,5371) (5437,5291) (5477,5077) (5518,5110) (5558,5385)
(5598,5097) (5638,4843) (5678,4803) (5718,5037) (5758,5732) (5798,5699) (5838,5425) (5878,5151) (5918,5130) (5958,5666)
(5999,6000) (6039,5860) (6079,5211) (6119,5692) (6159,6054) (6199,5672) (6239,5659) (6279,5946) (6319,6074) (6359,5906)
(6399,5572) (6440,5545) (6480,5291) (6520,5873) (6560,6140) (6600,6094) (6640,6107) (6680,6321) (6720,6569) (6760,6970)
(6800,6548) (6840,6890) (6880,7351) (6921,7023) (6961,7853) (7001,8194) (7041,7498) (7081,6910) (7121,6562) (7161,6328)
(7201,6595) (7241,6816) (7281,7271) (7321,7458) (7361,8047) (7402,8027) (7442,7445) (7482,7893) (7522,8662) (7562,8140)
(7602,7880) (7642,7391) (7682,7572) (7722,8435) (7762,9077) (7802,7712) (7843,8883) (7883,8997) (7923,8756) (7963,8408)
(7997,9171) (8037,9391) (8077,8381) (8117,8167) (8157,8756) (8198,9418) (8238,9171) (8278,9425) (8318,9679) (8358,9612)
(8398,9552)};
\end{axis}
\end{tikzpicture}
\caption{The average response length of DeepSeek-R1-Zero grew with training steps. The curve is read off from the plot in Ref.~\cite{Deepseekai2025}, where the fluctuations around this smoothed curve are also shown.}
\label{fig:DeepSeek-R1-Zero-thinking-time}
\end{figure}

They first built DeepSeek-R1-Zero without any supervised data, using only reinforcement learning. 
The template for DeepSeek-R1-Zero is shown in Table~\ref{tab:DeepSeek-R1-Zero-prompt}. 
They used two kinds of \textit{rule-based} rewards:
\begin{itemize}
\item
\textbf{Accuracy rewards.} This evaluates whether the answer is right or not. For math problems, the correctness of the output format and answer is evaluated. 
For coding problems, whether the code can be compiled and run correctly is evaluated. 

\item
\textbf{Format rewards.} This reward encourages the model to put its thinking process between \verb|<think>| and \verb|</think>|.

\end{itemize}
They did not use a neural reward model to evaluate the outcome or process because of the risk of reward hacking and the complication added by the training of the reward model. 
With this simple setup, the model took longer and longer to respond as training progressed, and the quality of the final answer improved (Fig.~\ref{fig:DeepSeek-R1-Zero-thinking-time}).

\begin{table}[htbp]
\centering
\begin{tabular}{p{0.9\textwidth}}
\hline\noalign{\smallskip}
A conversation between User and Assistant. The user asks a question, and the Assistant solves it. The assistant first thinks about the reasoning process in the mind and then provides the user with the answer. The reasoning process and answer are enclosed within \texttt{<think>} \texttt{</think>} and \texttt{<answer>} \texttt{</answer>} tags, respectively, i.e., \texttt{<think>} reasoning process here \texttt{</think>} \texttt{<answer>} answer here \texttt{</answer>}. User: {\color{red}prompt}. Assistant:\\[1mm]
\hline
\end{tabular}
\caption{Template for DeepSeek-R1-Zero, taken from Table~1 of Ref.~\cite{Deepseekai2025}. {\color{red}prompt} is replaced with the specific reasoning question during training.}
\label{tab:DeepSeek-R1-Zero-prompt}
\end{table}

\begin{table}[htbp]
\centering
\begin{tabular}{p{0.96\textwidth}}
\hline\noalign{\smallskip}
Question: If $a>1$, then the sum of the real solutions of $\sqrt{a-\sqrt{a+x}}=x$ is equal to\\[1mm]
\hline\noalign{\smallskip}
Response: \texttt{<think>}\\
To solve the equation $\sqrt{a-\sqrt{a+x}}=x$, let's start by squaring both $\cdots$\\
$\left(\sqrt{a-\sqrt{a+x}}\right)^2=x^2 \Longrightarrow a-\sqrt{a+x}=x^2$.\\
Rearrange to isolate the inner square root term:\\
$(a-x^2)^2=a+x \Longrightarrow a^2-2ax^2+(x^2)^2=a+x \Longrightarrow x^4-2ax^2-x+(a^2-a)=0$\\
$\cdots$\\
\leavevmode{\color{red}Wait, wait. Wait. That's an aha moment I can flag here}.\\
Let's reevaluate this step-by-step to identify if the correct sum can be $\cdots$\\
We started with the equation:\\
$\sqrt{a-\sqrt{a+x}}=x$\\
First, let's square both sides:\\
$a-\sqrt{a+x}=x^2 \Longrightarrow \sqrt{a+x}=a-x^2$\\
Next, I could square both sides again, treating the equation: $\cdots$\\
$\cdots$\\[1mm]
\hline
\end{tabular}
\caption{The `aha moment' of DeepSeek-R1-Zero at an intermediate stage of training, taken from Ref.~\cite{Deepseekai2025}. The model stops and reconsiders its approach (red).}
\label{tab:DeepSeek-R1-zero-aha}
\end{table}

A particularly intriguing observation was the `aha moment' (Table~\ref{tab:DeepSeek-R1-zero-aha}). 
Let us look at two paragraphs from Ref.~\cite{Deepseekai2025} that describe this moment:

\textit{
A particularly intriguing phenomenon observed during
the training of DeepSeek-R1-Zero is the occurrence of an ``aha moment”. This moment, as
illustrated in Table 3 \textup{[Table~\ref{tab:DeepSeek-R1-zero-aha} in these notes]}, occurs in an intermediate version of the model. During this phase,
DeepSeek-R1-Zero learns to allocate more thinking time to a problem by reevaluating its initial
approach. This behavior is not only a testament to the model’s growing reasoning abilities
but also a captivating example of how reinforcement learning can lead to unexpected and
sophisticated outcomes.}

\textit{
This moment is not only an ``aha moment” for the model but also for the researchers
observing its behavior. It underscores the power and beauty of reinforcement learning: rather
than explicitly teaching the model on how to solve a problem, we simply provide it with the
right incentives, and it autonomously develops advanced problem-solving strategies. The
“aha moment” serves as a powerful reminder of the potential of RL to unlock new levels of
intelligence in artificial systems, paving the way for more autonomous and adaptive models in
the future.
}

Despite its remarkable performance, DeepSeek-R1-Zero had issues with poor readability and language mixing. 
To overcome these issues, and also to enhance the reasoning ability of the model, DeepSeek-R1 was trained by adding several extra pieces to the training procedure of DeepSeek-R1-Zero:
\begin{itemize}
\item
\textbf{Cold start.} Before reinforcement learning, fine-tuning with chain-of-thought data was added. The cold start improved readability and performance. 

\item
A language consistency reward that encourages using one language was added for the reinforcement learning. This pushed down the performance slightly; however, readability was preferred. 

\item
Another round of fine-tuning was added. 

\item
Another round of reinforcement learning was added for alignment to human preference. 

\end{itemize}

%%%%%%%%%%%%
%%%%%%%%%%%%
\subsubsection*{Fine-tuning + test-time scaling}
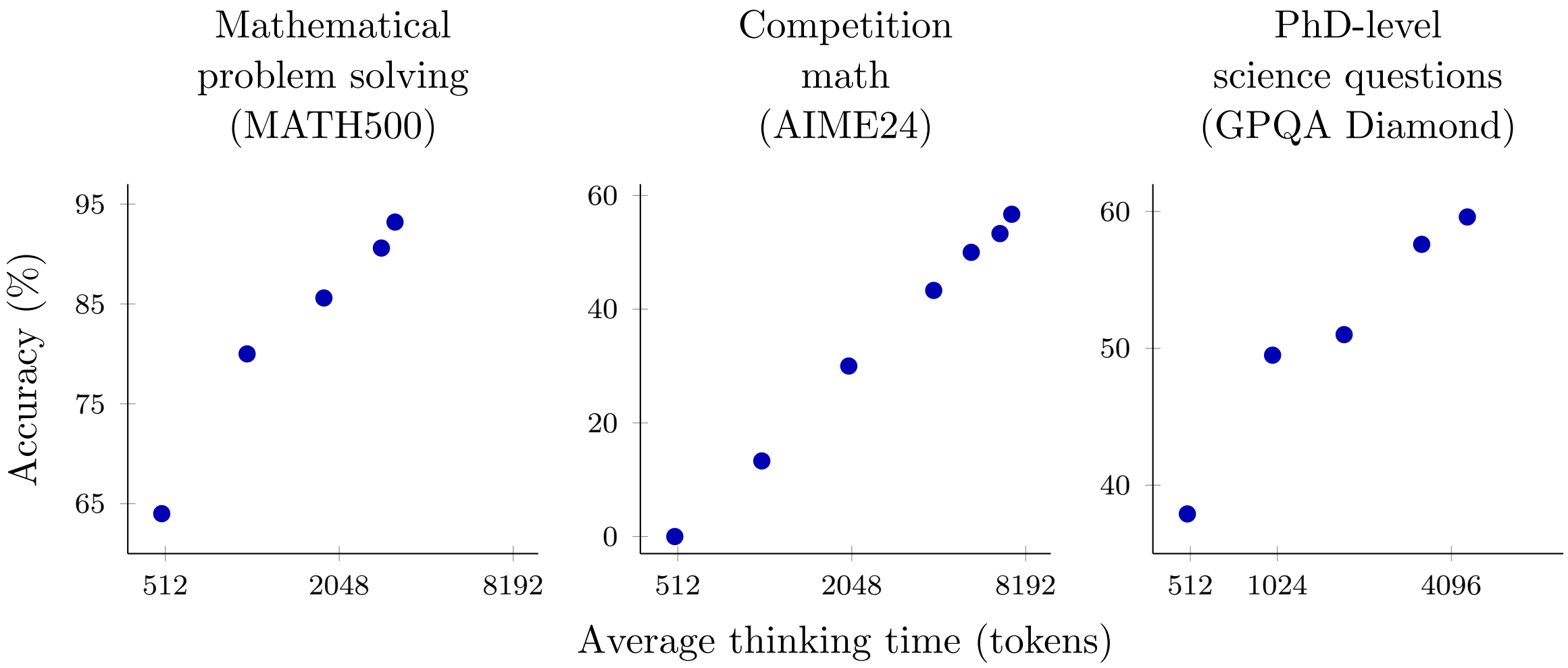
\begin{figure}[htbp]
\centering
\pgfplotsset{
  /pgfplots/s1panel/.style={
    width=5.6cm, height=5.2cm,
    xmin=380, xmax=10000,
    axis lines=left, axis line style={-},
    tick label style={font=\scriptsize},
    title style={font=\small, align=center},
    every axis x label/.style={at={(ticklabel cs:0.5)}, anchor=near ticklabel, font=\small},
    every axis y label/.style={at={(ticklabel cs:0.5)}, rotate=90, anchor=near ticklabel, font=\small},
    mark size=2.2pt}}
\begin{tikzpicture}
\begin{axis}[xmode=log, log basis x=2, s1panel, name=p1,
  title={Mathematical\\problem solving\\(MATH500)},
  xtick={512,2048,8192}, xticklabels={512,2048,8192}, ymin=60, ymax=97, ytick={65,75,85,95},
  ylabel={Accuracy (\%)}]
\addplot[only marks, mark=*, blue!70!black] coordinates {
(498,64.0) (983,80.0) (1813,85.6) (2866,90.6) (3195,93.2)};
\end{axis}
\begin{axis}[xmode=log, log basis x=2, s1panel, at={(p1.south east)}, xshift=1.0cm, name=p2,
  title={Competition\\math\\(AIME24)},
  xtick={512,2048,8192}, xticklabels={512,2048,8192}, ymin=-3, ymax=62, ytick={0,20,40,60},
  xlabel={Average thinking time (tokens)}]
\addplot[only marks, mark=*, blue!70!black] coordinates {
(500,0.0) (1000,13.3) (2000,30.0) (3936,43.3) (5306,50.0) (6672,53.3) (7327,56.7)};
\end{axis}
\begin{axis}[xmode=log, log basis x=2, s1panel, at={(p2.south east)}, xshift=1.0cm,
  title={PhD-level\\science questions\\(GPQA Diamond)},
  xtick={512,1024,4096}, xticklabels={512,1024,4096}, ymin=35, ymax=62, ytick={40,50,60}]
\addplot[only marks, mark=*, blue!70!black] coordinates {
(500,37.9) (986,49.5) (1744,51.0) (3235,57.6) (4654,59.6)};
\end{axis}
\end{tikzpicture}
\caption{Test-time scaling in s1-32B: the accuracy improves as the model thinks longer, similar to Fig.~\ref{fig:OpenAI_blog_o1}. The values are read off from the plot in Ref.~\cite{Muennighoff2025}.}
\label{fig:s1-test-time-scaling}
\end{figure}

%%%%%%%%%%%%
%%%%%%%%%%%%
While DeepSeek-R1 achieved performance on par with OpenAI o1, the authors of Ref.~\cite{Muennighoff2025} pointed out that a clear relationship between test-time compute and accuracy, like the right panel of Fig.~\ref{fig:OpenAI_blog_o1}, had not been replicated openly. Specifically, although DeepSeek-R1 learned to spend more time on more difficult questions, the test time (number of tokens) spent on each problem was not controlled. (Because the details of o1 are not public, we cannot tell whether the two models work in the same way.) Additionally, the resources invested by DeepSeek were still significant, making it unaffordable for small companies or universities. The Stanford-UW group's s1 model~\cite{Muennighoff2025} is very interesting because the fine-tuning requires only about seven GPU-hours, which costs a few tens of dollars (!) at typical cloud prices, and the test-time scaling was replicated (Fig.~\ref{fig:s1-test-time-scaling}).

The s1 model is in line with the superficial alignment hypothesis presented in the LIMA paper~\cite{Zhou2023}, which claims that almost all abilities and knowledge of LLMs are learned during pretraining, and post-training just teaches the model how to present such ability and knowledge. The Stanford-UW team curated a dataset consisting of 1000 samples and achieved impressive accuracy. Note that there was no reinforcement learning. They performed only fine-tuning, which took only 26 minutes on sixteen H100 GPUs~\cite{Muennighoff2025}. 

Their procedure for curating a `high-quality' training dataset is as follows. They claim that \textit{quality}, \textit{difficulty}, and \textit{diversity} are crucial, and that performance declines if any of these are lacking. Initially, 59K hard questions (e.g., questions used to select top high schoolers for the US Math Olympiad) were collected.
\textbf{Quality:} About 1/8 of these questions, which had obvious issues such as API errors, inconsistent formatting, or missing image references, were removed. After filtering out the `low-quality' data, 384 samples that were deemed high quality by the team (though this is subjective) were selected without further filtering. The rest went through two additional filtering stages.
\textbf{Difficulty:} To evaluate difficulty, they used Qwen2.5-7B-Instruct and Qwen2.5-32B-Instruct, developed by Alibaba, to generate reasoning and answers, and Claude 3.5 Sonnet, developed by Anthropic, to assess the correctness of the responses by comparing them to the reference solutions. If either Qwen2.5-7B-Instruct or Qwen2.5-32B-Instruct could solve the problem, it was considered too easy and discarded. About 25K samples remained after this filtering. At this stage, the number of tokens used for reasoning was measured for each sample. A simple hypothesis was applied: more tokens are required for more difficult questions, and this was used to quantify the difficulty level.
\textbf{Diversity:} Claude 3.5 Sonnet was used to classify the samples based on the Mathematics Subject Classification system from the American Mathematical Society. To choose the $1000 - 384 = 616$ samples, the subjects were selected randomly with uniform probability, and then a sample was chosen from each subject using a probability distribution that favored longer reasoning tokens. This process was repeated 616 times. The resulting questions cover a wide range of domains, from geometry, number theory and combinatorics to quantum theory, statistical mechanics and general relativity.

To allow for longer reasoning time, two options were available: sequential (one long chain of reasoning, including reflections) and parallel (multiple independent, shorter chains of reasoning). They focused on the sequential approach. To control the thinking time, they used the end-of-thinking token to force the model to stop thinking, and the token ``Final Answer:" to generate the final answer. To extend thinking time, they suppressed the generation of the end-of-thinking token and added the ``Wait" token to allow the model to continue reasoning; see Table~\ref{tab:s1-budget-forcing}.

The s1 model was obtained by fine-tuning Qwen2.5-32B-Instruct. The reasoning traces in the dataset were generated by Google's Gemini thinking model; in this sense, this is an example of distillation (Sec.~\ref{sec:distillation}).

\begin{table}[htbp]
\centering
\begin{tabular}{p{0.92\textwidth}}
\hline\noalign{\smallskip}
\textit{Question:} How many r in raspberry?\\[1mm]
\hline\noalign{\smallskip}
\textit{Reasoning trace:} Let's break down the process of counting the letter `r' in the word ``raspberry'' \ldots\\
* First letter: `r' -- This is an `r', count = 1.\\
* Second letter: `a' -- Not an `r', count remains 1 \ldots\\
* Sixth letter: `e' -- Not an `r', count remains 1.\\
* Seventh letter: `r' -- This is an `r', count = 2.\\
* Eighth letter: `y' -- Not an `r', count remains 2 \ldots\\
The number of `r's in ``raspberry'' is 2.\\
\leavevmode{\color{red}Wait}, let's re-read the question carefully. It asks ``How many r in raspberry?'' \ldots\ * r - a - s - p - b - e - r - r - y \ldots\ * First `r' \ldots\ * Second `r' \ldots\ * Third `r' \ldots\ Count = 3 \ldots\\[1mm]
\hline\noalign{\smallskip}
\textit{Response:} My initial answer of 2 was incorrect due to a quick reading of the word. {\color{red}\textbf{Final Answer:}} The final answer is \fbox{3}\\[1mm]
\hline
\end{tabular}
\caption{An example of test-time control (budget forcing) in s1-32B, taken from Ref.~\cite{Muennighoff2025}. When the model tries to stop thinking, ``Wait'' is appended to its output, so that the model continues thinking; here it finds and corrects its own mistake. To stop the thinking, ``Final Answer:'' is appended instead. The words added in this way are shown in red.}
\label{tab:s1-budget-forcing}
\end{table}

\clearpage
\bibliographystyle{utphys}
\bibliography{references}

@inproceedings{Aizerman67,
  author = {Aizerman, M. A. and Braverman, E. A. and Rozonoer, L.},
  booktitle = "{ Automation and Remote Control, }",
  number = { 25 },
  pages = { 821-837 },
  title = { Theoretical foundations of the potential function
  method in pattern recognition learning. },
  year = { 1964 }
}

@misc{Amodei2016,
      title="{Concrete Problems in AI Safety}", 
      author={Dario Amodei and Chris Olah and Jacob Steinhardt and Paul Christiano and John Schulman and Dan Mané},
      eprint={1606.06565},
      archivePrefix={arXiv},
      primaryClass={cs.AI},
      url={https://arxiv.org/abs/1606.06565}, 
}

@article{Ba2016,
    author = "Ba, Jimmy Lei and Kiros, Jamie Ryan and Hinton, Geoffrey E.",
    title = "{Layer Normalization}",
    eprint = "1607.06450",
    archivePrefix = "arXiv",
    primaryClass = "stat.ML",
    month = "7",
    year = "2016"
}

@misc{Bahdanau2016,
      title="{Neural Machine Translation by Jointly Learning to Align and Translate}", 
      author={Dzmitry Bahdanau and Kyunghyun Cho and Yoshua Bengio},
      year={2016},
      eprint={1409.0473},
      archivePrefix={arXiv},
      primaryClass={cs.CL},
      url={https://arxiv.org/abs/1409.0473}, 
}

@misc{Barrett2022,
      title="{Implicit Gradient Regularization}", 
      author={David G. T. Barrett and Benoit Dherin},
      year={2022},
      eprint={2009.11162},
      archivePrefix={arXiv},
      primaryClass={cs.LG},
      url={https://arxiv.org/abs/2009.11162}, 
}

@article{Bengio1994,
  author  = {Bengio, Yoshua and Simard, Patrice and Frasconi, Paolo},
  title   = "{Learning long-term dependencies with gradient descent is difficult}",
  journal = {IEEE Transactions on Neural Networks},
  volume  = {5},
  number  = {2},
  pages   = {157--166},
  year    = {1994}
}

@inproceedings{Boser92,
author = {Boser, Bernhard E. and Guyon, Isabelle M. and Vapnik, Vladimir N.},
title = "{A training algorithm for optimal margin classifiers}",
year = {1992},
isbn = {089791497X},
publisher = {Association for Computing Machinery},
address = {New York, NY, USA},
url = {https://doi.org/10.1145/130385.130401},
doi = {10.1145/130385.130401},
booktitle = {Proceedings of the Fifth Annual Workshop on Computational Learning Theory},
pages = {144–152},
numpages = {9},
location = {Pittsburgh, Pennsylvania, USA},
series = {COLT '92}
}

@misc{Brendel2018,
      title="{Decision-Based Adversarial Attacks: Reliable Attacks Against Black-Box Machine Learning Models}", 
      author={Wieland Brendel and Jonas Rauber and Matthias Bethge},
      year={2018},
      eprint={1712.04248},
      archivePrefix={arXiv},
      primaryClass={stat.ML},
      url={https://arxiv.org/abs/1712.04248}, 
}

@misc{Brown2020,
      title="{Language Models are Few-Shot Learners}", 
      author={Tom B. Brown and Benjamin Mann and Nick Ryder and Melanie Subbiah and Jared Kaplan and Prafulla Dhariwal and Arvind Neelakantan and Pranav Shyam and Girish Sastry and Amanda Askell and Sandhini Agarwal and Ariel Herbert-Voss and Gretchen Krueger and Tom Henighan and Rewon Child and Aditya Ramesh and Daniel M. Ziegler and Jeffrey Wu and Clemens Winter and Christopher Hesse and Mark Chen and Eric Sigler and Mateusz Litwin and Scott Gray and Benjamin Chess and Jack Clark and Christopher Berner and Sam McCandlish and Alec Radford and Ilya Sutskever and Dario Amodei},
      eprint={2005.14165},
      archivePrefix={arXiv},
      primaryClass={cs.CL},
      url={https://arxiv.org/abs/2005.14165}, 
}

@inproceedings{Bucila2006,
  title={Model compression},
  author={Cristian Bucila and Rich Caruana and Alexandru Niculescu-Mizil},
  booktitle="{Knowledge Discovery and Data Mining}",
  year={2006},
  url={https://api.semanticscholar.org/CorpusID:11253972}
}

@article{Cauchy1847,
  title="{M\'{e}thode g\'{e}n\'{e}rale pour la r\'{e}solution des syst\`{e}mes d’\'{e}quations simultan\'{e}es}",
  author={Cauchy, A.},
  journal={C. R. Acad. Sci. Paris},
  year={1847}
}

@misc{Chen2021_Codex,
      title="{Evaluating Large Language Models Trained on Code}", 
 author={Mark Chen and et al.},
      eprint={2107.03374},
      archivePrefix={arXiv},
      primaryClass={cs.LG},
      url={https://arxiv.org/abs/2107.03374}, 
}

@inproceedings{Christiano2017,
  title={Deep reinforcement learning from human preferences},
  author={Christiano, Paul F and Leike, Jan and Brown, Tom B and Martic, Miljan and Legg, Shane and Amodei, Dario},
  booktitle={Advances in Neural Information Processing Systems},
  volume={30},
  year={2017},
  eprint={1706.03741},
  archivePrefix={arXiv},
  primaryClass={stat.ML}
}

@article{Cybenko89,
  author = {Cybenko, G. V.}, 
  title = "{Approximation by superpositions of a sigmoidal function}", 
  journal = {Mathematics of Control, Signals and Systems},
  volume = {2},
  pages = {303--314}, 
  year =  {1989},
  url = {https://api.semanticscholar.org/CorpusID:3958369}
  }

@article{Fukushima69,
  author={Fukushima, Kunihiko},
  journal={IEEE Transactions on Systems Science and Cybernetics}, 
  title="{Visual Feature Extraction by a Multilayered Network of Analog Threshold Elements}", 
  year={1969},
  volume={5},
  number={4},
  pages={322-333},
  }

@article{He2015,
    author = "He, Kaiming and Zhang, Xiangyu and Ren, Shaoqing and Sun, Jian",
    title = "{Deep Residual Learning for Image Recognition}",
    eprint = "1512.03385",
    archivePrefix = "arXiv",
    primaryClass = "cs.CV",
    doi = "10.1109/CVPR.2016.90",
    month = "12",
    year = "2015"
}

@article{Deepseekai2025,
  author        = {{DeepSeek-AI}},
  title         = {{DeepSeek-R1} incentivizes reasoning in {LLMs} through reinforcement learning},
  journal       = {Nature},
  volume        = {645},
  number        = {8081},
  pages         = {633--638},
  year          = {2025},
  doi           = {10.1038/s41586-025-09422-z},
  eprint        = {2501.12948},
  archivePrefix = {arXiv},
  primaryClass  = {cs.CL}
}

@misc{Devlin2019,
      title="{BERT: Pre-training of Deep Bidirectional Transformers for Language Understanding}", 
      author={Jacob Devlin and Ming-Wei Chang and Kenton Lee and Kristina Toutanova},
      eprint={1810.04805},
      archivePrefix={arXiv},
      primaryClass={cs.CL},
      url={https://arxiv.org/abs/1810.04805}, 
}

@book{Gauss1809,
    author = "Gauss, C.~F.",
    title = "{Theoria Motus Corporum Coelestium in Sectionibus Conicis Solem Ambientium}",
    year = "1809"
}

@article{Gers2000,
    author = {Gers, F.~A. and Schmidhuber, J. and Cummins, F},
    title = "{Learning to forget: continual prediction with LSTM}",
    journal = "Neural Comput.",
    volume = "12",
    number = "10",
    pages = "2451-2471",
    year = "2000"
}

@article{Goodfellow:2014rpb,
    author = "Goodfellow, Ian J. and Shlens, Jonathon and Szegedy, Christian",
    title = "{Explaining and Harnessing Adversarial Examples}",
    eprint = "1412.6572",
    archivePrefix = "arXiv",
    primaryClass = "stat.ML",
    month = "12",
    year = "2014"
}

@misc{Greenblatt2024,
      title="{Alignment faking in large language models}", 
      author={Ryan Greenblatt and Carson Denison and Benjamin Wright and Fabien Roger and Monte MacDiarmid and Sam Marks and Johannes Treutlein and Tim Belonax and Jack Chen and David Duvenaud and Akbir Khan and Julian Michael and Sören Mindermann and Ethan Perez and Linda Petrini and Jonathan Uesato and Jared Kaplan and Buck Shlegeris and Samuel R. Bowman and Evan Hubinger},
      eprint={2412.14093},
      archivePrefix={arXiv},
      primaryClass={cs.AI},
      url={https://arxiv.org/abs/2412.14093}, 
}

@book{Hanada2022,
    author = "Hanada, Masanori and Matsuura, So",
    title = "{MCMC from scratch -- a practical introduction to Markov Chain Monte Carlo method}",
    publisher = "Springer",
    year = "2022"
}

@article{Hinton2015,
    author = "Hinton, Geoffrey and Vinyals, Oriol and Dean, Jeff",
    title = "{Distilling the Knowledge in a Neural Network}",
    eprint = "1503.02531",
    archivePrefix = "arXiv",
    primaryClass = "stat.ML",
    month = "3",
    year = "2015"
}

@article{Hochreiter1997,
    author = {Hochreiter, Sepp and Schmidhuber, J\"urgen},
    title = "{Long Short-Term Memory}",
    doi = "10.1162/neco.1997.9.8.1735",
    journal = "Neural Comput.",
    volume = "9",
    number = "8",
    pages = "1735--1780",
    year = "1997"
}

@article{Hoffmann2022,
  author        = {Hoffmann, Jordan and Borgeaud, Sebastian and Mensch, Arthur and others},
  title         = {Training Compute-Optimal Large Language Models},
  eprint        = {2203.15556},
  archivePrefix = {arXiv},
  primaryClass  = {cs.CL},
  year          = {2022}
}

@misc{Hubinger2021,
      title="{Risks from Learned Optimization in Advanced Machine Learning Systems}", 
      author={Evan Hubinger and Chris van Merwijk and Vladimir Mikulik and Joar Skalse and Scott Garrabrant},
      year={2021},
      eprint={1906.01820},
      archivePrefix={arXiv},
      primaryClass={cs.AI},
      url={https://arxiv.org/abs/1906.01820}, 
}

@article{Ioffe2015,
    author = "Ioffe, Sergey and Szegedy, Christian",
    title = "{Batch Normalization: Accelerating Deep Network Training by Reducing  Internal Covariate Shift}",
    eprint = "1502.03167",
    archivePrefix = "arXiv",
    primaryClass = "cs.LG",
    month = "2",
    year = "2015"
}

@misc{Jones2021,
      title={Scaling Scaling Laws with Board Games}, 
      author={Andy L. Jones},
      year={2021},
      eprint={2104.03113},
      archivePrefix={arXiv},
      primaryClass={cs.LG},
      url={https://arxiv.org/abs/2104.03113}, 
}

@misc{Kaplan2020,
      title="{Scaling Laws for Neural Language Models}", 
      author={Jared Kaplan and Sam McCandlish and Tom Henighan and Tom B. Brown and Benjamin Chess and Rewon Child and Scott Gray and Alec Radford and Jeffrey Wu and Dario Amodei},
      eprint={2001.08361},
      archivePrefix={arXiv},
      primaryClass={cs.LG},
      url={https://arxiv.org/abs/2001.08361}, 
}

@misc{Kojima2022,
      title="{Large Language Models are Zero-Shot Reasoners}", 
      author={Takeshi Kojima and Shixiang Shane Gu and Machel Reid and Yutaka Matsuo and Yusuke Iwasawa},
      eprint={2205.11916},
      archivePrefix={arXiv},
      primaryClass={cs.CL},
      url={https://arxiv.org/abs/2205.11916}, 
}

@misc{Kumar2015,
      title="{Ask Me Anything: Dynamic Memory Networks for Natural Language Processing}", 
      author={Ankit Kumar and Ozan Irsoy and Peter Ondruska and Mohit Iyyer and James Bradbury and Ishaan Gulrajani and Victor Zhong and Romain Paulus and Richard Socher},
      eprint={1506.07285},
      archivePrefix={arXiv},
      primaryClass={cs.CL},
      url={https://arxiv.org/abs/1506.07285}, 
}

@article{LeCun1998,
  author  = {LeCun, Yann and Bottou, L{\'e}on and Bengio, Yoshua and Haffner, Patrick},
  title   = "{Gradient-based learning applied to document recognition}",
  journal = {Proceedings of the IEEE},
  volume  = {86},
  number  = {11},
  pages   = {2278--2324},
  year    = {1998}
}

@book{Legendre1805,
    author = " Legendre, A.~M.",
    title = "{ Nouvelles methodes pour la determination des orbites des cometes}",
    year = "1805"
}

@misc{Lewis2021,
      title="{Retrieval-Augmented Generation for Knowledge-Intensive NLP Tasks}", 
      author={Patrick Lewis and Ethan Perez and Aleksandra Piktus and Fabio Petroni and Vladimir Karpukhin and Naman Goyal and Heinrich Küttler and Mike Lewis and Wen-tau Yih and Tim Rocktäschel and Sebastian Riedel and Douwe Kiela},
      year={2021},
      eprint={2005.11401},
      archivePrefix={arXiv},
      primaryClass={cs.CL},
      url={https://arxiv.org/abs/2005.11401}, 
}

@misc{Lightman2023,
      title={Let's Verify Step by Step}, 
      author={Hunter Lightman and Vineet Kosaraju and Yura Burda and Harri Edwards and Bowen Baker and Teddy Lee and Jan Leike and John Schulman and Ilya Sutskever and Karl Cobbe},
      year={2023},
      eprint={2305.20050},
      archivePrefix={arXiv},
      primaryClass={cs.LG},
      url={https://arxiv.org/abs/2305.20050}, 
}

@misc{Mccann2018,
      title="{The Natural Language Decathlon: Multitask Learning as Question Answering}", 
      author={Bryan McCann and Nitish Shirish Keskar and Caiming Xiong and Richard Socher},
      eprint={1806.08730},
      archivePrefix={arXiv},
      primaryClass={cs.CL},
      url={https://arxiv.org/abs/1806.08730}, 
}

@article{mcculloch1943,
  title="{A logical calculus of the ideas immanent in nervous activity}",
  author={McCulloch, Warren S and Pitts, Walter},
  journal={The bulletin of mathematical biophysics},
  volume={5},
  pages={115--133},
  year={1943},
  publisher={Springer}
}

@misc{Mikolov2013,
      title="{Efficient Estimation of Word Representations in Vector Space}", 
      author={Tomas Mikolov and Kai Chen and Greg Corrado and Jeffrey Dean},
      eprint={1301.3781},
      archivePrefix={arXiv},
      primaryClass={cs.CL},
      url={https://arxiv.org/abs/1301.3781}, 
}

@misc{Muennighoff2025,
      title="{s1: Simple test-time scaling}", 
      author={Niklas Muennighoff and Zitong Yang and Weijia Shi and Xiang Lisa Li and Li Fei-Fei and Hannaneh Hajishirzi and Luke Zettlemoyer and Percy Liang and Emmanuel Candès and Tatsunori Hashimoto},
      eprint={2501.19393},
      archivePrefix={arXiv},
      primaryClass={cs.CL},
      url={https://arxiv.org/abs/2501.19393}, 
}

@article{nakashima1936,
  title="{Some properties of the group of single partial paths in the relay circuit}",
  author={Nakashima, A},
  journal={Journal of the Institute of Electrical Communication Engineers of Japan},
  pages={88--95},
  year={1936}
}

@misc{Nye2021,
      title="{Show Your Work: Scratchpads for Intermediate Computation with Language Models}", 
      author={Maxwell Nye and Anders Johan Andreassen and Guy Gur-Ari and Henryk Michalewski and Jacob Austin and David Bieber and David Dohan and Aitor Lewkowycz and Maarten Bosma and David Luan and Charles Sutton and Augustus Odena},
      eprint={2112.00114},
      archivePrefix={arXiv},
      primaryClass={cs.LG},
      url={https://arxiv.org/abs/2112.00114}, 
}

@misc{OpenAI2023,
      title="{GPT-4 Technical Report}", 
      author={OpenAI},
      eprint={2303.08774},
      archivePrefix={arXiv},
      primaryClass={cs.CL},
      url={https://arxiv.org/abs/2303.08774}, 
}

@misc{OpenAI2024,
	author={OpenAI},
	url={https://openai.com/index/learning-to-reason-with-llms/},
}

@misc{Ouyang2022,
      title="{Training language models to follow instructions with human feedback}", 
      author={Long Ouyang and Jeff Wu and Xu Jiang and Diogo Almeida and Carroll L. Wainwright and Pamela Mishkin and Chong Zhang and Sandhini Agarwal and Katarina Slama and Alex Ray and John Schulman and Jacob Hilton and Fraser Kelton and Luke Miller and Maddie Simens and Amanda Askell and Peter Welinder and Paul Christiano and Jan Leike and Ryan Lowe},
      eprint={2203.02155},
      archivePrefix={arXiv},
      primaryClass={cs.CL},
      url={https://arxiv.org/abs/2203.02155}, 
}

@article{piesch1939,
  title="{Begriff der allgemeinen Schaltungstechnik}",
  author={Piesch, Hansi},
  journal={Archiv f{\"u}r Elektrotechnik},
  volume={33},
  pages={672--686},
  year={1939},
  publisher={Springer}
}

@inproceedings{Radford2018,
  title="{Improving Language Understanding by Generative Pre-Training}",
  author={Alec Radford and Karthik Narasimhan and Tim Salimans and Ilya Sutskever},
  year={2018}
}

@inproceedings{Radford2019,
  title="{Language Models are Unsupervised Multitask Learners}",
  author={Alec Radford and Jeff Wu and Rewon Child and David Luan and Dario Amodei and Ilya Sutskever},
  year={2019},
}

@article{Rosenblatt1958,
  title   = "{The perceptron: a probabilistic model for information storage and organization in the brain}",
  author  = {Rosenblatt, Frank},
  journal = {Psychological Review},
  volume  = {65},
  number  = {6},
  pages   = {386--408},
  year    = {1958},
  doi     = {10.1037/h0042519}
}

@article{Rumelhart1986,
  author  = {Rumelhart, David E. and Hinton, Geoffrey E. and Williams, Ronald J.},
  title   = "{Learning representations by back-propagating errors}",
  journal = {Nature},
  volume  = {323},
  pages   = {533--536},
  year    = {1986}
}

@misc{santurkar2019,
      title="{How Does Batch Normalization Help Optimization?}", 
      author={Shibani Santurkar and Dimitris Tsipras and Andrew Ilyas and Aleksander Madry},
      year={2019},
      eprint={1805.11604},
      archivePrefix={arXiv},
      primaryClass={stat.ML},
      url={https://arxiv.org/abs/1805.11604}, 
}

@misc{Schick2023,
      title="{Toolformer: Language Models Can Teach Themselves to Use Tools}", 
      author={Timo Schick and Jane Dwivedi-Yu and Roberto Dessì and Roberta Raileanu and Maria Lomeli and Luke Zettlemoyer and Nicola Cancedda and Thomas Scialom},
      year={2023},
      eprint={2302.04761},
      archivePrefix={arXiv},
      primaryClass={cs.CL},
      url={https://arxiv.org/abs/2302.04761}, 
}

@misc{Schulman2017,
      title="{Proximal Policy Optimization Algorithms}", 
      author={John Schulman and Filip Wolski and Prafulla Dhariwal and Alec Radford and Oleg Klimov},
      eprint={1707.06347},
      archivePrefix={arXiv},
      primaryClass={cs.LG},
      url={https://arxiv.org/abs/1707.06347}, 
}

@article{shannon1938,
  title="{A symbolic analysis of relay and switching circuits}",
  author={Shannon, Claude E},
  journal={Electrical Engineering},
  volume={57},
  number={12},
  pages={713--723},
  year={1938},
  publisher={IEEE}
}

@article{Shannon1948,
  title="{A Mathematical Theory of Communication}",
  author={Shannon, Claude Elwood},
  journal={The Bell System Technical Journal},
  volume={27},
  number={3},
  pages={379--423},
  year={1948},
  publisher={AT\&T}
}

@article{Sharif2019,
    author = {Sharif, Mahmood and Bhagavatula, Sruti and Bauer, Lujo and Reiter, Michael K.},
    title = "{A General Framework for Adversarial Examples with Objectives}",
    year = {2019},
    issue_date = {August 2019},
    publisher = {Association for Computing Machinery},
    address = {New York, NY, USA},
    volume = {22},
    number = {3},
    issn = {2471-2566},
    url = {https://doi.org/10.1145/3317611},
    doi = {10.1145/3317611},
    journal = {ACM Trans. Priv. Secur.},
    month = jun,
    articleno = {16},
    numpages = {30}
    }

@article{Chen2020,
  author       = {Ting Chen and
                  Simon Kornblith and
                  Mohammad Norouzi and
                  Geoffrey E. Hinton},
  title        = "{A Simple Framework for Contrastive Learning of Visual Representations}",
  journal      = {CoRR},
  volume       = {abs/2002.05709},
  year         = {2020},
  url          = {https://arxiv.org/abs/2002.05709},
  eprinttype    = {arXiv},
  eprint       = {2002.05709},
  bibsource    = {dblp computer science bibliography, https://dblp.org}
}

@article{Silver2016,
  author = {Silver, David and et al},
  journal = {Nature},
  number = 7587,
  pages = {484--489},
  publisher = {Nature Publishing Group},
  title = {Mastering the game of Go with deep neural networks and tree search},
  volume = 529,
  year = 2016
}

@misc{Simonyan2014,
      title="{Deep Inside Convolutional Networks: Visualising Image Classification Models and Saliency Maps}", 
      author={Karen Simonyan and Andrea Vedaldi and Andrew Zisserman},
      year={2014},
      eprint={1312.6034},
      archivePrefix={arXiv},
      primaryClass={cs.CV},
      url={https://arxiv.org/abs/1312.6034}, 
}

@misc{Smith2021,
      title="{On the Origin of Implicit Regularization in Stochastic Gradient Descent}", 
      author={Samuel L. Smith and Benoit Dherin and David G. T. Barrett and Soham De},
      year={2021},
      eprint={2101.12176},
      archivePrefix={arXiv},
      primaryClass={cs.LG},
      url={https://arxiv.org/abs/2101.12176}, 
}

@article{Srivastava2014,
  author  = {Srivastava, Nitish and Hinton, Geoffrey and Krizhevsky, Alex and Sutskever, Ilya and Salakhutdinov, Ruslan},
  title   = "{Dropout: A Simple Way to Prevent Neural Networks from Overfitting}",
  journal = {Journal of Machine Learning Research},
  volume  = {15},
  pages   = {1929--1958},
  year    = {2014}
}

@inproceedings{stankovic2007,
  title="{Some historical remarks on switching theory}",
  author={Stankovic, Radomir S and Astola, Jaakko T and Karpovsky, Mark G},
  booktitle={Proc. Int. Workshop on Spectral Techniques. URL http://mark. bu. edu/papers/200. pdf. pdf},
  year={2007}
}

@misc{Stiennon2020,
      title="{Learning to summarize from human feedback}", 
      author={Nisan Stiennon and Long Ouyang and Jeff Wu and Daniel M. Ziegler and Ryan Lowe and Chelsea Voss and Alec Radford and Dario Amodei and Paul Christiano},
      eprint={2009.01325},
      archivePrefix={arXiv},
      primaryClass={cs.CL},
      url={https://arxiv.org/abs/2009.01325}, 
}

@article{Su2021,
  author        = {Su, Jianlin and Lu, Yu and Pan, Shengfeng and Murtadha, Ahmed and Wen, Bo and Liu, Yunfeng},
  title         = {{RoFormer}: Enhanced Transformer with Rotary Position Embedding},
  eprint        = {2104.09864},
  archivePrefix = {arXiv},
  primaryClass  = {cs.CL},
  year          = {2021}
}

@misc{Sutskever2014_seq2seq_LSTM,
      title="{Sequence to Sequence Learning with Neural Networks}", 
      author={Ilya Sutskever and Oriol Vinyals and Quoc V. Le},
      year={2014},
      eprint={1409.3215},
      archivePrefix={arXiv},
      primaryClass={cs.CL},
      url={https://arxiv.org/abs/1409.3215}, 
}

@article{Touvron2023,
  author        = {Touvron, Hugo and Lavril, Thibaut and Izacard, Gautier and others},
  title         = {{LLaMA}: Open and Efficient Foundation Language Models},
  eprint        = {2302.13971},
  archivePrefix = {arXiv},
  primaryClass  = {cs.CL},
  year          = {2023}
}

@misc{Uesato2022,
      title={Solving math word problems with process- and outcome-based feedback}, 
      author={Jonathan Uesato and Nate Kushman and Ramana Kumar and Francis Song and Noah Siegel and Lisa Wang and Antonia Creswell and Geoffrey Irving and Irina Higgins},
      year={2022},
      eprint={2211.14275},
      archivePrefix={arXiv},
      primaryClass={cs.LG},
      url={https://arxiv.org/abs/2211.14275}, 
}

@inproceedings{Vaswani2017,
    author = "Vaswani, Ashish and Shazeer, Noam and Parmar, Niki and Uszkoreit, Jakob and Jones, Llion and Gomez, Aidan N. and Kaiser, Lukasz and Polosukhin, Illia",
    title = "{Attention Is All You Need}",
    booktitle = "{31st International Conference on Neural Information Processing Systems}",
    eprint = "1706.03762",
    archivePrefix = "arXiv",
    primaryClass = "cs.CL",
    month = "6",
    year = "2017"
}

@misc{Wang2022,
      title="{Self-Consistency Improves Chain of Thought Reasoning in Language Models}", 
      author={Xuezhi Wang and Jason Wei and Dale Schuurmans and Quoc Le and Ed Chi and Sharan Narang and Aakanksha Chowdhery and Denny Zhou},
      year={2023},
      eprint={2203.11171},
      archivePrefix={arXiv},
      primaryClass={cs.CL},
      url={https://arxiv.org/abs/2203.11171}, 
}

@misc{Wei2021,
      title="{Finetuned Language Models Are Zero-Shot Learners}", 
      author={Jason Wei and Maarten Bosma and Vincent Y. Zhao and Kelvin Guu and Adams Wei Yu and Brian Lester and Nan Du and Andrew M. Dai and Quoc V. Le},
      eprint={2109.01652},
      archivePrefix={arXiv},
      primaryClass={cs.CL},
      url={https://arxiv.org/abs/2109.01652}, 
}

@misc{Wei2022,
      title="{Chain-of-Thought Prompting Elicits Reasoning in Large Language Models}", 
      author={Jason Wei and Xuezhi Wang and Dale Schuurmans and Maarten Bosma and Brian Ichter and Fei Xia and Ed Chi and Quoc Le and Denny Zhou},
      eprint={2201.11903},
      archivePrefix={arXiv},
      primaryClass={cs.CL},
      url={https://arxiv.org/abs/2201.11903}, 
}

@misc{Zhou2023,
      title="{LIMA: Less Is More for Alignment}", 
      author={Chunting Zhou and Pengfei Liu and Puxin Xu and Srini Iyer and Jiao Sun and Yuning Mao and Xuezhe Ma and Avia Efrat and Ping Yu and Lili Yu and Susan Zhang and Gargi Ghosh and Mike Lewis and Luke Zettlemoyer and Omer Levy},
      eprint={2305.11206},
      archivePrefix={arXiv},
      primaryClass={cs.CL},
      url={https://arxiv.org/abs/2305.11206}, 
}

@misc{Ziegler2020,
      title="{Fine-Tuning Language Models from Human Preferences}", 
      author={Daniel M. Ziegler and Nisan Stiennon and Jeffrey Wu and Tom B. Brown and Alec Radford and Dario Amodei and Paul Christiano and Geoffrey Irving},
      eprint={1909.08593},
      archivePrefix={arXiv},
      primaryClass={cs.CL},
      url={https://arxiv.org/abs/1909.08593}, 
}
\end{document}